\documentclass[lettersize,journal]{IEEEtran}
\usepackage{amssymb}
\usepackage{amsmath}
\usepackage{times}
\usepackage[ruled,vlined]{algorithm2e}
\usepackage{bm}
\usepackage{hyperref}
\usepackage{xspace}
\usepackage[capitalise]{cleveref}
\usepackage{siunitx}
\usepackage{graphicx}
\usepackage{subcaption}
\usepackage{overpic}
\usepackage{color}
\usepackage{multirow}
\usepackage[table,x11names,dvipsnames]{xcolor}
\usepackage{alphalph}
\usepackage{comment}

\newcommand{\lsj}[1]{{\color{black}#1}}

\makeatletter
\DeclareRobustCommand\onedot{\futurelet\@let@token\@onedot}
\def\@onedot{\ifx\@let@token.\else.\null\fi\xspace}

\def\eg{\emph{e.g}\onedot} 
\def\ie{\emph{i.e}\onedot} 
 
\def\etc{\emph{etc}\onedot} \def\vs{\emph{vs}\onedot}
 
\def\etal{\emph{et al}\onedot}

\begin{document}
%
\title{Neuromorphic Synergy for Video Binarization
}
%
%
%
%

\author{Shijie Lin, Xiang Zhang, Lei Yang, Lei Yu, Bin Zhou, Xiaowei Luo, Wenping Wang, and Jia Pan

\IEEEcompsocitemizethanks{
\IEEEcompsocthanksitem Shijie Lin and Jia Pan (corresponding author) are with the Department of Computer Science, The University of Hong Kong and Centre for Transformative Garment Production, Hong Kong SAR, China. E-mail: lsj2048@connect.hku.hk, jpan@cs.hku.hk
\IEEEcompsocthanksitem Lei Yang is with the Centre for Transformative Garment Production, Hong Kong, China. E-mail: l.yang@transgp.hk
\IEEEcompsocthanksitem Xiang Zhang and Lei Yu are with the School of Electronic and Information, Wuhan University, Wuhan, China. E-mail: \{xiangz,ly.wd\}@whu.edu.cn.
\IEEEcompsocthanksitem Bin Zhou is with the School of Computer Science and Engineering, Beihang University, Beijing, China. E-mail:
zhoubin@buaa.edu.cn 
\IEEEcompsocthanksitem Xiaowei Luo is with the Department of Architecture and Civil Engineering, City University of Hong Kong, Hong Kong SAR, China. E-mail: xiaowluo@cityu.edu.hk
\IEEEcompsocthanksitem Wenping Wang is with the Department of Computer Science and Engineering, Texas A\&M University, Texas, USA. E-mail: wenping@tamu.edu
\IEEEcompsocthanksitem This project is supported by the Innovation and Technology Commission of the HKSAR Government under the InnoHK initiative, ITF GHP/126/21GD and HKU's CRF seed grant. 
}

}

%
%

\markboth{}%
{Shell \MakeLowercase{\textit{et al.}}: Bare Demo of IEEEtran.cls for Computer Society Journals}



\IEEEtitleabstractindextext{%
\begin{abstract}
\lsj{Bimodal objects, such as the checkerboard pattern used in camera calibration, markers for object tracking, and text on road signs, to name a few, are prevalent in our daily lives
and serve as a visual form to embed information that can be easily recognized by vision systems.
While binarization from intensity images is crucial for extracting the embedded information in the bimodal objects, few previous works consider the task of binarization of blurry images due to the relative motion between the vision sensor and the environment.
The blurry images can result in a loss in the binarization quality and thus degrade the downstream applications where the vision system is in motion.
Recently, neuromorphic cameras offer new capabilities for alleviating motion blur, 
but it is non-trivial to first deblur and then binarize the images in a real-time manner.
In this work, we propose an event-based binary reconstruction method that leverages the prior knowledge of the bimodal target's properties to perform inference independently in both event space and image space and merge the results from both domains to generate a sharp binary image. We also develop an efficient integration method to propagate this binary image to high frame rate binary video. Finally, we develop a novel method to naturally fuse events and images for unsupervised threshold identification. The proposed method is evaluated in publicly available and our collected data sequence, and shows the proposed method can outperform the SOTA methods to generate high frame rate binary video in real-time on CPU-only devices.}

\end{abstract}

\begin{IEEEkeywords}
Image Binarization, Neuromorphic Event Camera, Motion Deblurring, High Frame-rate Video Restoration
\end{IEEEkeywords}}

\maketitle

\IEEEdisplaynontitleabstractindextext

%
\IEEEpeerreviewmaketitle

\section{Introduction}\label{sec:introduction}
\begin{figure}[t]
\begin{subfigure}{0.49\linewidth}
  \includegraphics[width=\textwidth]{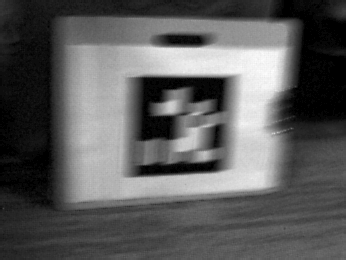}
  \caption{The blurred image}    \label{fig:blur_ori_enhanced}
\end{subfigure}
\hfill
\begin{subfigure}{0.49\linewidth}
  \includegraphics[width=\textwidth]{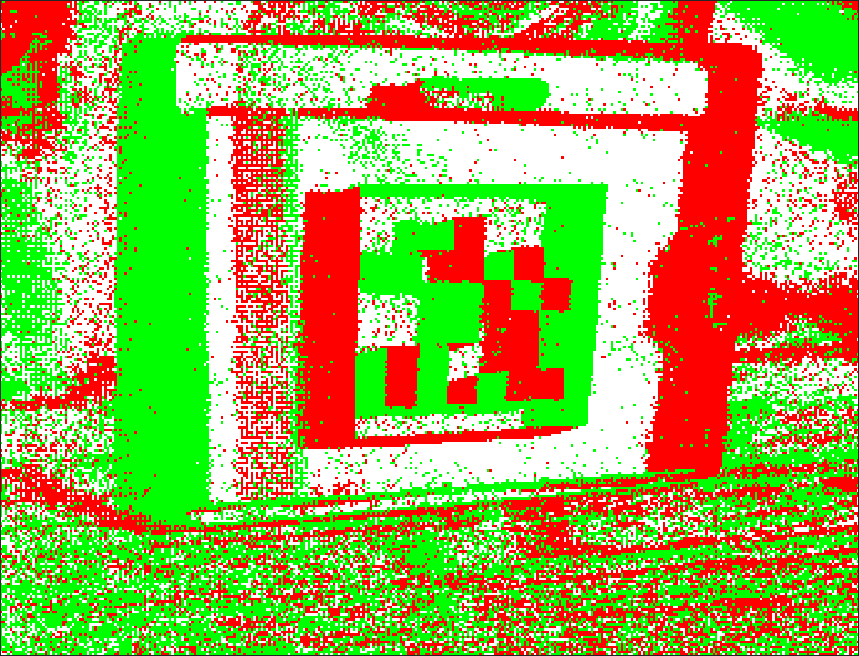}
  \caption{Events}    
  \label{fig:tag_events}
\end{subfigure}
\hfill
\begin{subfigure}{0.325\linewidth}
  \includegraphics[width=\textwidth]{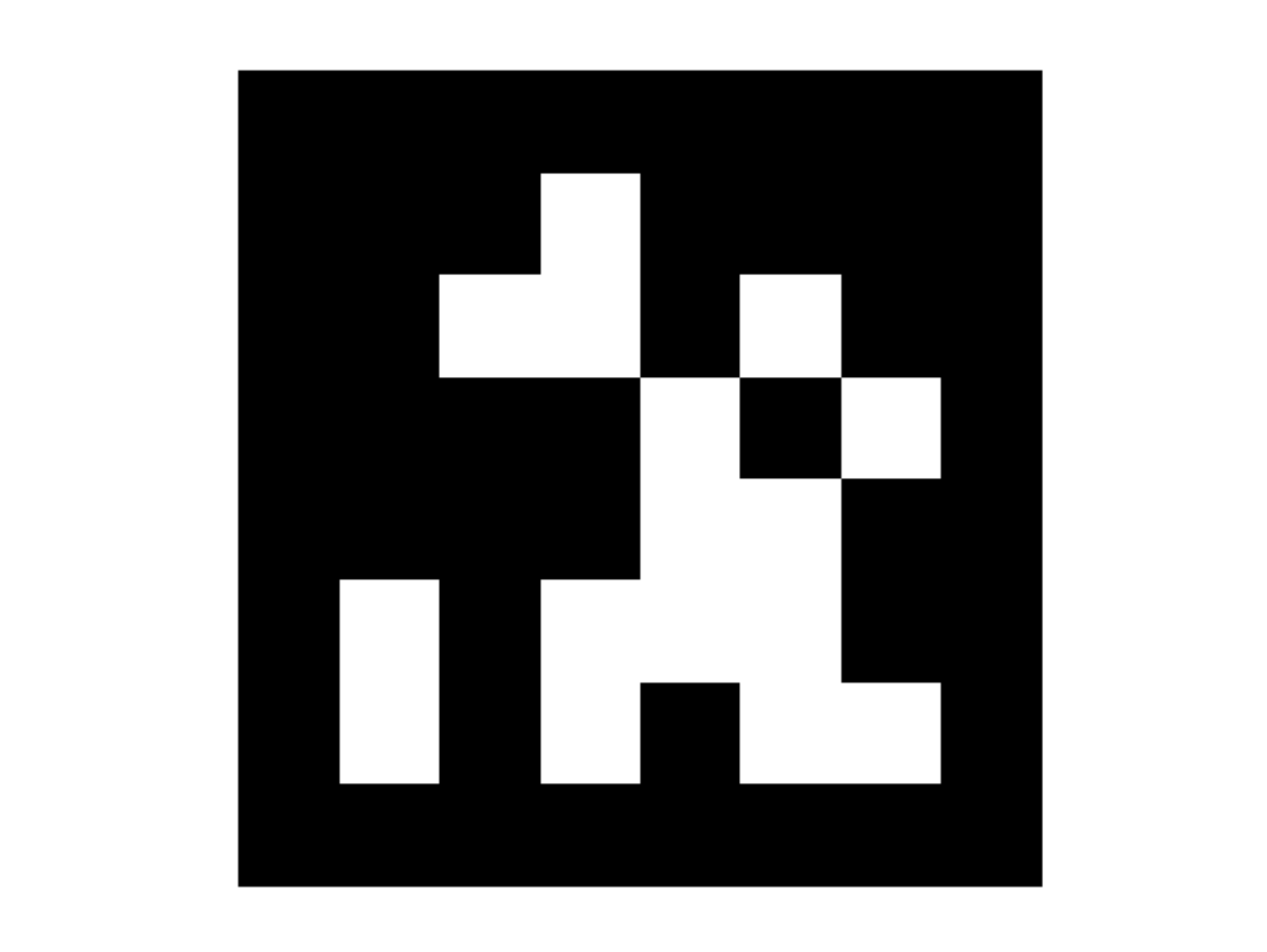}
  \caption{The Apriltag \cite{olson2011apriltag}}    
  \label{fig:id0}
\end{subfigure}
\hfill
\begin{subfigure}{0.325\linewidth}
  \begin{overpic}[width=\textwidth]{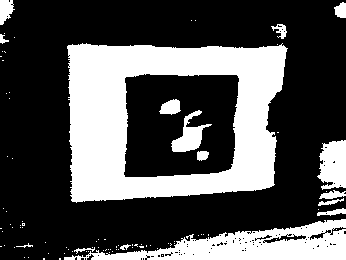}
    \put(0,6){ \textcolor{red}{\bf \small Detection Failed}}
  \end{overpic}  
  \caption{Binarization \cite{mustafa2018binarization}}    \label{fig:blur_det}
\end{subfigure}
\hfill
\begin{subfigure}{0.325\linewidth}
  \begin{overpic}[width=\textwidth]{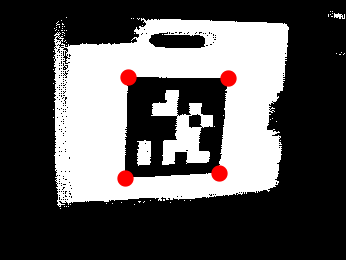}
    \put(0,6){ \textcolor{green}{\bf \small Detection Success}}
  \end{overpic}  
  \caption{Ours}    \label{fig:erb_tag}
\end{subfigure}
\caption{\lsj{Examples of video binarization in the presence of motion blur. Given (a) the blurry images and (b) the corresponding events (red: positive events, green: negative events) of (c) the Apriltag \cite{olson2011apriltag}, our method can produce (e) sharp binary videos, enabling accurate tag detection (marked by the red dots) which is hardly achieved using (d) conventional image binarization \cite{mustafa2018binarization}. Best viewed in color.}
} 
\end{figure}

\IEEEPARstart{B}{imodal} objects are ubiquitous in our daily lives, such as checkerboard patterns, visual markers/tags, and texts on road signs. They are designed to encode information for various downstream applications like camera calibration, object recognition and tracking, and navigation in complex scenes~\cite{wang2016apriltag,pfrommer2019tagslam,rufli2008automatic,mufti2021automatic}.
\lsj{While it is trivial to binarize an intensity image containing a bimodal pattern with sharp contrast, it is known to be challenging to accurately and efficiently recover bimodal patterns from images captured in motion, for example from a flying drone, due to the motion blur. 
Therefore, current robotic systems need to slow down their speed or even stop to recognize the information embedded in the bimodal objects.
Otherwise, the loss in the binarization quality due to motion blur will degrade performances in the downstream tasks, such as the tag detection as shown in \cref{fig:blur_det,fig:erb_tag,fig:blur_det}. 
Hence, enabling efficient and high-quality binarization of images captured in motion is of paramount importance to allow \textit{moving robots} to decode the information embedded in the bimodal objects in a complex environment, offering a key enabler for a variety of robotic applications in dynamic scenes.}

\lsj{Existing methods for binarization, e.g.,~\cite{calvo2019selectional,su2012robust,mustafa2018binarization}, are designed to process intensity images with sharp contrast. While they are lightweight and efficient (\ie, can achieve real-time performance), they are unable to handle images with motion blur. 
Efficient restoration of the clarity of such blurry images is difficult. 
This is because motion cues required to remove the motion blur are not readily available for conventional intensity-based camera systems. Therefore, a complex optimization problem to solve the motion cues is needed to restore the sharp contrast in the bimodal pattern for binarization. However, it is challenging to solve such an optimization problem under limited time and resource budgets for applications on robots where on-board computing is often required.}

Recently, neuromorphic vision sensors (or event cameras) have received considerable attention due to their high temporal resolution and low latency \cite{dvs128,DAVIS} (less than \SI{1}{ms}). Event cameras can attain promising performance in applications such as autofocus~\cite{lin2022autofocus}, visual tracking~\cite{zheng2022spike,li2022asynchronous}, and optical flow estimation~\cite{wan2022learning}. \lsj{Among these, a relevant application of event cameras is to reconstruct intensity images captured in motion with events~\cite{yu2023learning,jiang2020learning,edipami,cadena2021spade,wang2021asynchronous,haoyu2020learning,chen2022residual}. The events from a neuromorphic sensor indicate the intensity changes at event spikes and hence provide intra-frame motion cues of dynamic scenes, showing its potential to restore the sharp contrast of bimodal patterns in binarization tasks.}

\lsj{However, na\"ive combination of event-based image reconstruction and image binarization is not applicable to tasks requiring real-time, robust video binarization.} The reason is two-fold. First, event-based reconstruction methods typically rely on computationally intensive techniques like deep learning \cite{jiang2020learning,xu2021motion,yu2023learning} or convex optimization \cite{xu2010two,edipami}. Using images and events in image reconstruction involves event-wise double integrals that impose a theoretical limit on reconstruction efficiency~\cite{edipami}. 
Second, event cameras are subject to inherent limitations, such as refractory periods \cite{delbruck2021feedback} and transmission dropping \cite{berner20075}, making it challenging to quantify large intensity variations in high contrast areas of the bimodal objects. As a result, most event-based reconstructions will generate halo artifacts and distort subsequent binarization. 
Thus far, no previous works allow the neuromorphic synergy for motion-invariant binarization, resulting in a significant gap between the video binarization and the neuromorphic cameras.

In this study,  we develop a novel event-based binary reconstruction (EBR) framework for real-time, robust video binarization under complex motion. We demonstrate that by leveraging the inherent properties of bimodality and its correlation with blurry images and events, the proposed framework enables simultaneous motion deblurring and binarization, generating high frame-rate binary videos in real-time (\cref{fig:erb_tag}). 
Our event-based binary image reconstruction can be naturally integrated with the simple thresholding process to achieve an unsupervised threshold estimation.
The estimation process also relaxes the need for accurate contrast determination, thus enhancing performance and robustness. The overall pipeline is asynchronous and linear in time complexity, ensuring its applicability to on-board computing devices with limited computational resources.

In summary, our contributions are:
\begin{itemize}
    \item \lsj{We develop the dual-stage binarization to produce latent binary images, providing a novel way to directly produce the binary images in time complexity linear to event number and avoid solving the intricate problem of event-wise doubly integral.}
    \item \lsj{We propose an efficient method to produce high frame-rate sharp binary videos under complex motions.}
    \item\lsj{ We develop a novel threshold estimation method that naturally fuses events and blurry images to produce an optimal and motion-invariant threshold.}
    \item We extensively evaluate our proposed method on various datasets, demonstrating state-of-the-art performance and efficiency with CPU-only devices. \\
    \textbf{Project page}: \url{https://github.com/eleboss/EBR}.
\end{itemize}

\section{Related Work}
Event-based cameras, such as the dynamic and active pixel vision sensor (DAVIS) \cite{DAVIS} and the dynamic vision sensor (DVS) \cite{dvs128}, are designed to emulate the human visual system by detecting logarithmic changes in luminance. This generates a continuous, asynchronous stream of events that encode unique information about variations in brightness. 
Despite many studies that have examined the benefits of high temporal resolution enabled by event-based cameras, the optimal utilization of such cameras for binary images remains unclear. Previous works by Adam \etal \cite{loch2021event} involved high-rate tracking of fiducial markers using events, while Sarmadi \etal \cite{sarmadi2021detection} proposed fitting the line directly in the time surface of events to detect fiducial markers. Nagata \etal \cite{nagata2020qr} suggested estimating and initializing the marker's motion and affine transformation and then optimizing these two parameters to decode the QR code. However, previous attempts have only addressed specific use cases of binary images, whereas the general problem regarding generating sharp binary videos has not been discussed.

\noindent{\textbf{Image Binarization.}}
Image binarization usually serves as a preprocessing for various applications~\cite{khurshid2009comparison,bradley2007adaptive,wolf2004extraction,mustafa2018binarization}. The earliest attempts to perform image binarization relied on global thresholding methods based on image statistics \cite{otsu1979threshold}. Subsequent works aimed to improve binarization performance by estimating multiple thresholds locally through techniques such as adaptive image contrast \cite{su2012robust}, foreground concavities \cite{howe2013document}, among others. However, these approaches often fail to generalize to scenes with degraded textures. Recent learning-based methods have leveraged convolutional neural networks to extract pixel-wise \cite{calvo2017pixel}, or patch-wise \cite{calvo2019selectional} semantic information. Nevertheless, these methods still heavily depend on high-quality inputs and suffer from performance degradation facing motion blur in real-world scenarios \cite{sulaiman2019degraded}. 

\noindent{\textbf{Motion Deblurring.}} The general motion deblurring tasks are designed for recovering sharp intensity images and removing the blur. The field of single image deblurring has made significant progress using various gradient-based regularizers such as Gaussian scale mixture \cite{fergus2006removing}, and $L_0$-norm regularizer \cite{xu2013unnatural}. In addition, non-gradient-based priors such as the extreme channel (dark/bright channel) prior \cite{darkchannel} has also been explored. Given the difficulty in estimating blur parameters and latent images from a single image, recent research attention has gradually shifted to leveraging powerful deep neural networks (CNN). Sun \etal \cite{sun2015learning} introduced a convolutional neural network (CNN) for estimating locally linear blur kernels. Nah \etal \cite{nah2017deep} proposed a multi-scale CNN that can restore latent images in an end-to-end learning process without assuming any constrained blur kernel model. However, deep deblurring methods usually require large datasets to train the model, and sharp images are typically needed as supervisory input, which is not always available in practice for blurry images. What's worse, using images only, the task of motion deblurring is severely ill-posed~\cite{purohit2019bringing}, which greatly limits the performance of previous solutions in terms of efficiency and accuracy under complex motions.

\noindent{\textbf{Event-based Motion Deblurring.}} 
Event cameras report asynchronous events that potentially embed motion information, alleviating the difficulty of motion deblurring and providing a novel direction to address the motion deblurring more effectively. Pan \etal \cite{edipami} first reveal the physical relation of events and images using the event double integral (EDI) model and develop a multi-frame optimization framework to estimate the event contrast. Such physical relation is further adopted in various learning-based methods \cite{yu2023learning,lin2020learning,jiang2020learning} to achieve better deblurring performance. 
\lsj{Lin \etal \cite{lin2020learning} leverage the convolutional neural network to implement the EDI model, which achieved significant improvement regarding motion deblurring. Yu \etal \cite{yu2023learning} jointly consider image resolution, sensor noise, and event double integral model and apply an event-enhanced sparse learning network to recover the sharp image under various resolutions.} 
However, both the optimization~\cite{edipami} and the learning-based approaches~\cite{yu2023learning,lin2020learning} require high computational costs for inference. Besides, these works are mostly designed for the recovery of intensity images, which cannot satisfy the efficiency bound for image binarization and cannot be directly adopted for motion deblurring for image binarization.

\noindent{\textbf{Motion Deblurring for Binary Image.}} 
Deblurring binary images, including textual or barcode representations, represents a specialized subset within the broader field of image deblurring, garnering sustained research interest due to the widespread application of binary-coded markers. Cho and Wang \cite{cho2012text} introduced the employment of stroke width transformation to discern sharp edges for kernel estimation. Jiang \etal. \cite{jiang2017text} employed double-well potential to achieve text image deblurring, an approach later refined by Li \etal. \cite{lv2018binary} via adaptive foreground and background value calculations. Additional research endeavors have focused on QR-code recovery through methods such as corner position recognition \cite{van2015regularization}, Kullback-Leibler divergence application \cite{rioux2019blind}, and linear motion assumptions \cite{shi2020fast}. Nonetheless, extant solutions grapple with intricate motion and the inherent ill-posedness of the problem, often yielding suboptimal performance or marked inefficiencies. No established solutions have capitalized on event and image data to reconstruct binary images.

\section{Problem Definition} \label{sec:ps}
Human-designed bimodal objects, such as visual markers and texts in road signs, are prevalent in our daily environment. These objects utilize high-contrast bimodal patterns (\eg, black and white) to encode information for efficient detection by vision systems. 
\lsj{However, existing image binarization approaches for processing bimodal objects are limited to sharp images, preventing their use in a wider range of applications in robotic and mobile applications where the images taken may contain motion blur.}
\lsj{To allow vision systems to recognize bimodal patterns in motion, we propose event-based binary reconstruction (EBR) to incorporate events into the binarization process to achieve high frame-rate and blur-free binary videos from degraded video input due to motion blur.}

\noindent\textbf{Event-based Binary Reconstruction (EBR):} Given the potentially blurred intensity image $I(\mathbf{x}) \in [0,255]$ captured within exposure period $\mathcal{T}$ and its corresponding event stream $\mathcal{E}$, the EBR task aims to generate sharp binary images directly from blurry inputs, i.e.,
\begin{align}
B(t,\mathbf{x}) = \text{Event-BR}(t; I(\mathbf{x}), \mathcal{E}), \quad \forall t\in\mathcal{T}, \label{eq:EBR}
\end{align}
where $B(t,\mathbf{x})\in \{0, 1\}$ denotes the binary image at an arbitrary time $t\in\mathcal{T}$, $\mathbf{x}=(x,y)$ represents the pixel position, and $\text{Event-BR}(\cdot)$ is an operator enabling the generation of sharp, high-rate binary outputs. 

\noindent\textbf{EBR \vs Image Binarization (IB):} The IB task generates a binary image $B_{I}(\mathbf{x})$ from an intensity image $I(\mathbf{x})$:
\begin{align}
B_{I}(\mathbf{x}) = \text{IB}( I(\mathbf{x})), 
\label{eq:IB}
\end{align}
where $\text{IB}(\cdot)$ is an IB operator. Most IB methods~\cite{otsu1979threshold,bradley2007adaptive,moghaddam2010multi,saddami2019effective} are designed to leverage information (e.g., texture, semantic, statistics) within a sharp intensity image but often fail to handle images with motion blur~\cite{sulaiman2019degraded}. In contrast, our EBR utilizes motion information furnished by the events, enabling motion-invariant generation of high frame-rate binary videos.

\noindent\textbf{EBR \vs Event-based Image Reconstruction (EIR):} The EIR task recovers sharp latent images $L(t,\mathbf{x})$ from the intensity image $I(\mathbf{x})$ and the corresponding events $\mathcal{E}$:
\begin{align}
L(t,\mathbf{x}) = \text{Event-IR}(t; I(\mathbf{x}), \mathcal{E}), \quad \forall t\in\mathcal{T}, \label{eq:md}
\end{align}
where $\text{Event-IR}(\cdot)$ represents an EIR operator. Existing EIR techniques primarily concentrate on the generation of latent intensity images $L(t,\mathbf{x}) \in [0,255]$ (for 8-bit images), utilizing resource-intensive approaches such as deep learning \cite{bishan2020event} or optimization \cite{edipami}. In contrast to EIR, our EBR aims to produce binary output efficiently, 
focusing on the robust decoding of the embedded information in bimodal patterns.


As summarized above, our EBR task presents unique demands different from IB or EIR tasks. 
\lsj{It shall be able to handle motion blur generated by complex arbitrary motions, manage events with high emission rates, and cope with random noises in the events. Furthermore, it shall be able to operate on various computing devices, meeting strict time constraints.}
Consequently, the development and implementation of EBR methods must address these distinct challenges to provide efficient, real-time binary video reconstruction while maintaining high-quality outputs for various applications.

\begin{figure}[t]
  \centering
  \includegraphics[width=\linewidth]{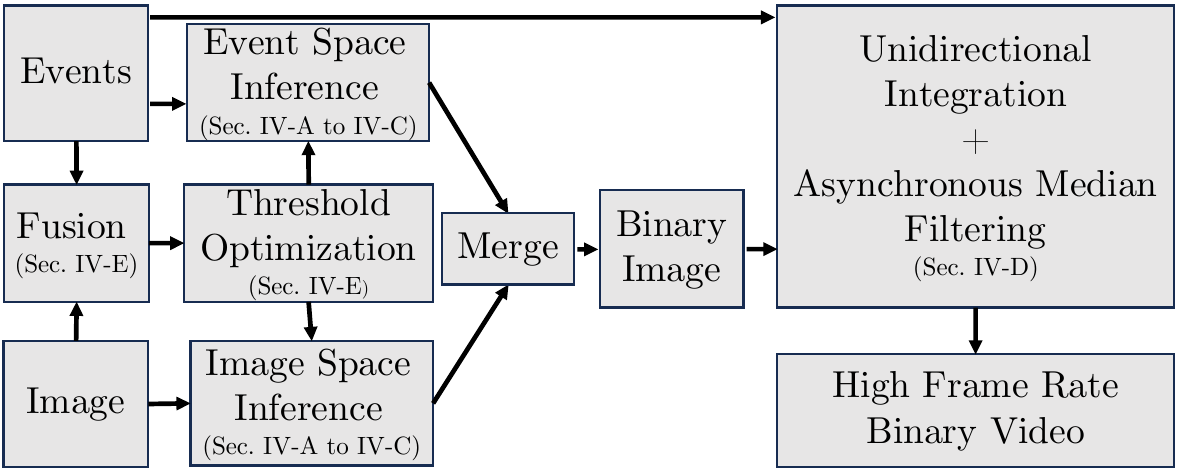}
  \caption{\lsj{Overall pipeline of the proposed method.}}
  \label{fig:pipeline}
\end{figure}
\begin{figure}[t]
\begin{subfigure}{\linewidth}
  \includegraphics[width=0.485\textwidth]{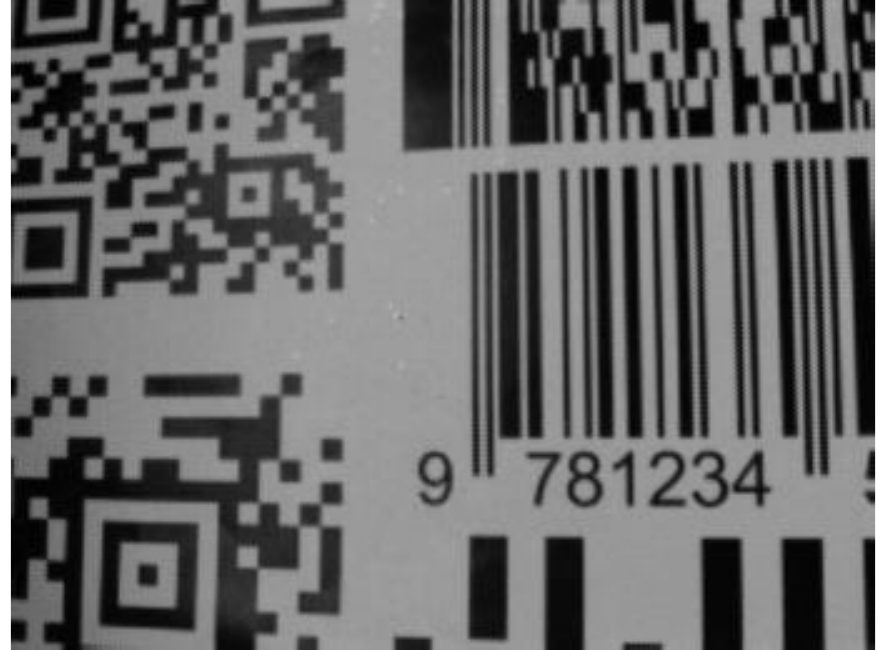}
  \includegraphics[width=0.485\textwidth]{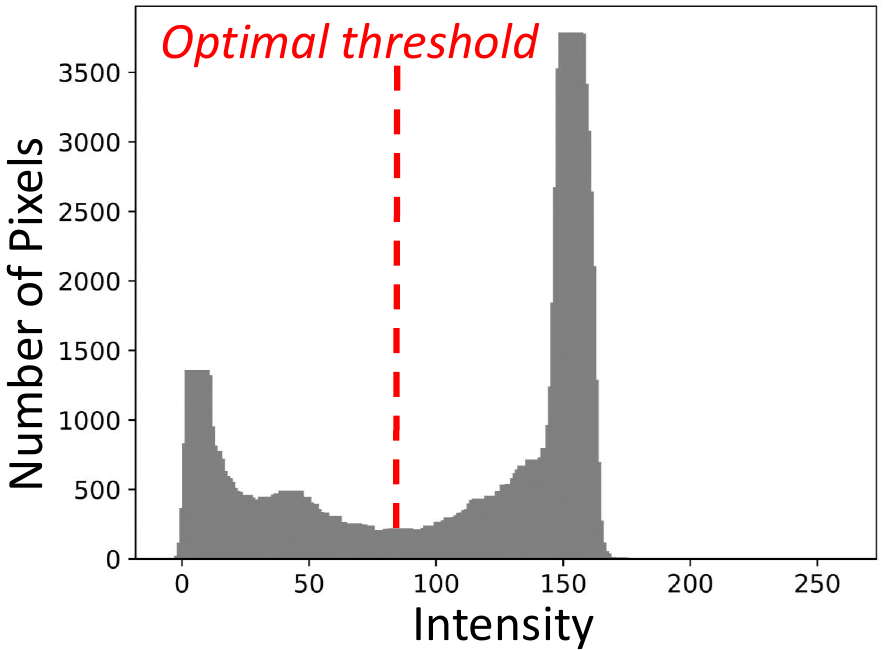}
  \caption{An image with sharp contrast and its corresponding intensity histogram.}    
  \label{fig:bimodality_1}
\end{subfigure}
\vfill
\begin{subfigure}{\linewidth}
  \includegraphics[width=0.485\textwidth]{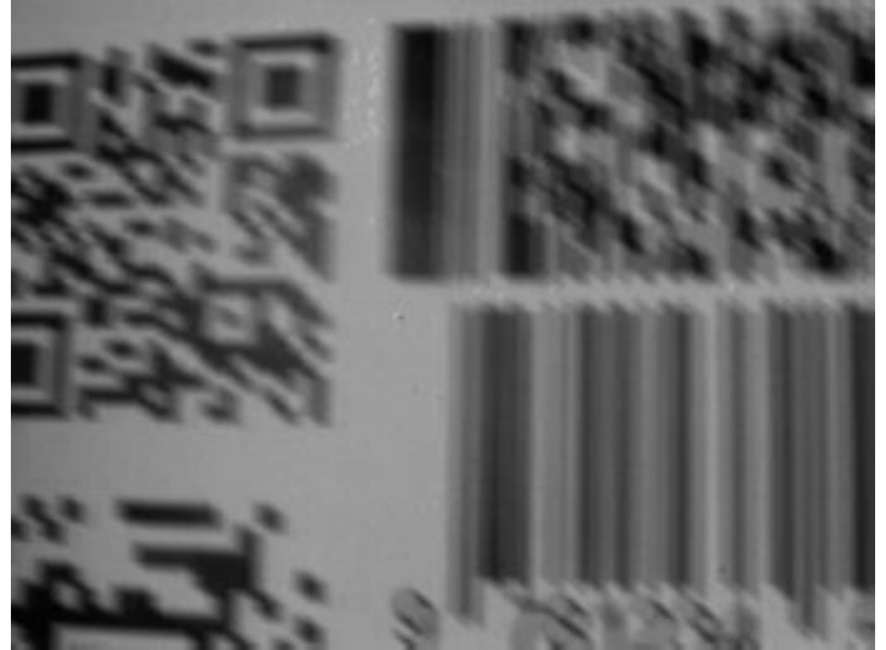}
  \includegraphics[width=0.485\textwidth]{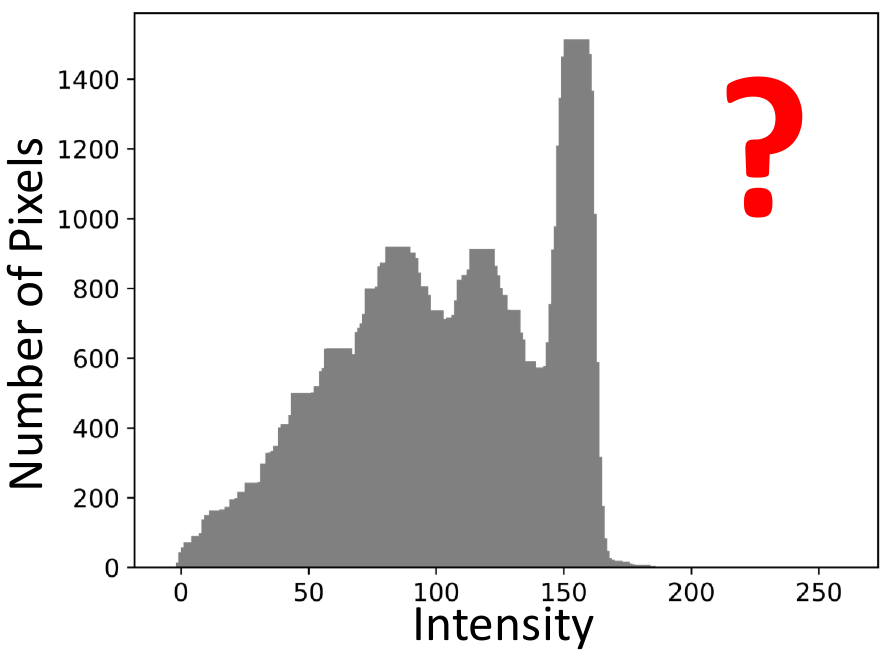}
  \caption{A blurred image and its corresponding intensity histogram.}    
  \label{fig:bimodality_2}
\end{subfigure}
\caption{ 
 \lsj{Illustration of how motion blur obfuscates bimodality in images. (a) The histogram of a static image shows a clear bimodal pattern, allowing effective identification of the optimal threshold. (b) However, motion blur may cause an averaging of pixel intensities, leading to the disappearance of bimodality in the histogram.}
} 
\end{figure}
\section{Approach}
\lsj{
Our goal is to incorporate events into the binarization process, allowing the efficient, blur-free generation of high frame-rate binary videos. The overall pipeline of the proposed method is shown in \cref{fig:pipeline}, consisting of three major steps. First, the image and its corresponding events are fused for the unsupervised threshold estimation (\cref{sec:thresh}). Second, we infer the binary status of each pixel in event space and image space respectively and merge the results from both spaces to generate a binary image (\cref{sec:bimodality,sec:sep,sec:dual}). Third, our method generates high-rate binary videos based on the binary image and events using unidirectional integration and asynchronous median filtering (\cref{sec:video}).}

\subsection{Motion Blur and Bimodality}
\label{sec:bimodality}

Bimodal objects like barcodes and texts have high-contrast bimodal patterns allowing easy identification. As a result, images of these objects have a bimodal pixel intensity distribution; 
\lsj{see \cref{fig:bimodality_1} for example. In ideal conditions, image binarization techniques can well segment out the bimodal object, showing clear bimodal patterns. However, relative motion between the sensor and the object can cause blurring, 
distorting the bimodal distribution,
making accurate thresholding of the observed distribution difficult in practical situations (\eg, \cref{fig:bimodality_2}).}

\lsj{
Using the event camera \cite{DAVIS} and following the event double integral model \cite{edipami}, one can reconstruct the latent image $L(t,\mathbf{x})$ from a blurry image $I(\mathbf{x}) = \frac{1}{T} \int_{t \in \mathcal{T}} L(t,\mathbf{x}) dt$, where $\mathcal{T} = [t_s, t_s + T]$ represents the exposure time interval starting at time $t_s$, and the corresponding events $\mathcal{E} = \{ \mathbf{e}_k\}_{k=1}^{N_e}$, with $N_e$ denoting the number of events. The $k$-th event $\mathbf{e}_k = (\mathbf{x}_k, t_k, p_k)$ is triggered whenever the log-scale intensity variations surpass the event contrast\footnote{Event contrast $c$ and another two thresholds $\theta_e$ and $\theta_I$ will be discussed in \cref{sec:thresh}.} $c>0$ at time $t_k$, resulting in polarity $p\in\{+1,-1\}$ in $\mathbf{e}_k$, indicating an increase or decrease in intensity.
}

After reconstruction, the bimodal histogram can be retrieved from the latent image for binarization. However, event double integral increases computational complexity, making real-time processing infeasible. Furthermore, artifacts for the event camera like refractory period and transmission dropping result in artifacts in the results of reconstruction methods \cite{edipami}, which affect binarization results, leading to undesirable outcomes (\eg, \cref{fig:deblur_bin_ebt_halfscale_5,fig:deblur_bin_hqf_halfscale_5,fig:deblur_bin_reblur_halfscale_5}).



\begin{figure*}[htb]
\centering
\begin{subfigure}{0.24\linewidth}
  \includegraphics[width=\textwidth]{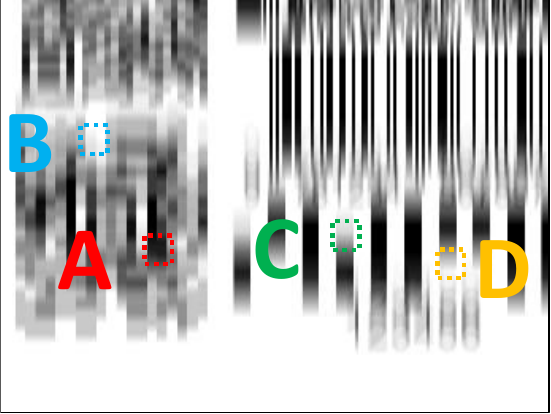}
  \caption{Intensity image $I(\mathbf{x})$}    \label{fig:four_1}
\end{subfigure}
\hfill
\begin{subfigure}{0.24\linewidth}
  \includegraphics[width=\textwidth]{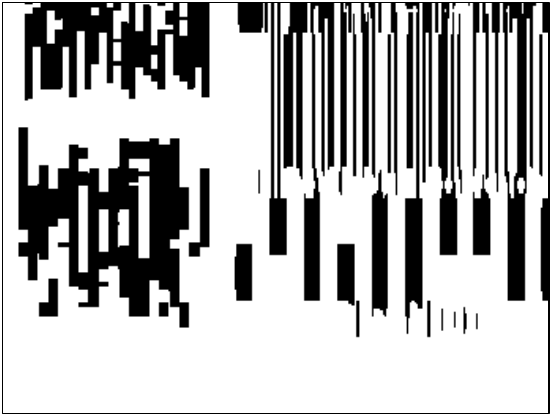}
  \caption{Binary result of (a) $B_{I}(\mathbf{x})$}    \label{fig:four_2}
\end{subfigure}
\hfill
\begin{subfigure}{0.24\linewidth}
  \includegraphics[width=\textwidth]{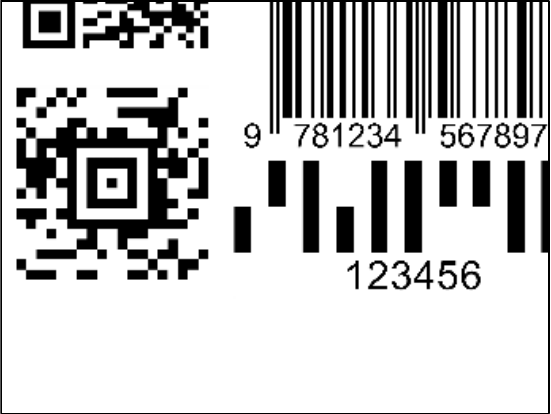}
  \caption{Ground-truth}    \label{fig:four_3}
\end{subfigure}
\hfill
\begin{subfigure}{0.24\linewidth}
  \includegraphics[width=\textwidth]{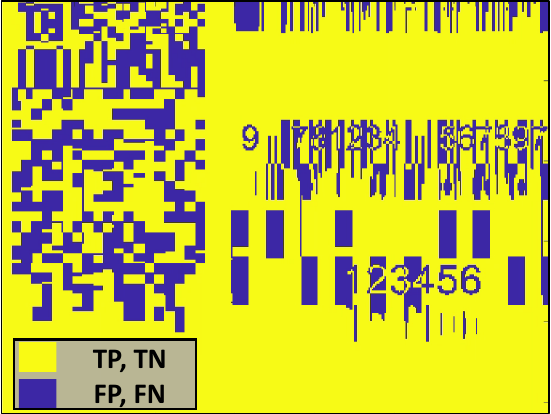}
  \caption{Results comparing with GT}    \label{fig:four_4}
\end{subfigure}
\vfill
\begin{subfigure}{0.24\linewidth}
  \includegraphics[width=\textwidth]{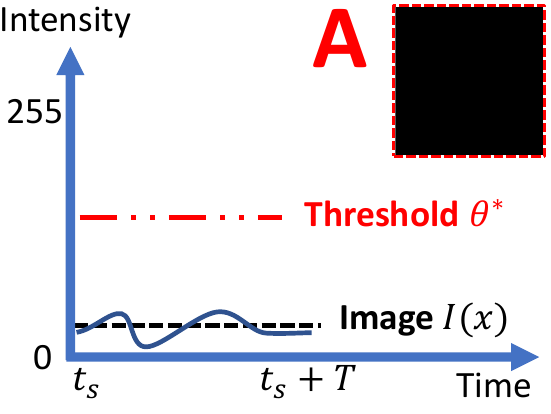}
  \caption{True Negative}    \label{fig:four_5}
\end{subfigure}
\hfill
\begin{subfigure}{0.24\linewidth}
\includegraphics[width=\textwidth]{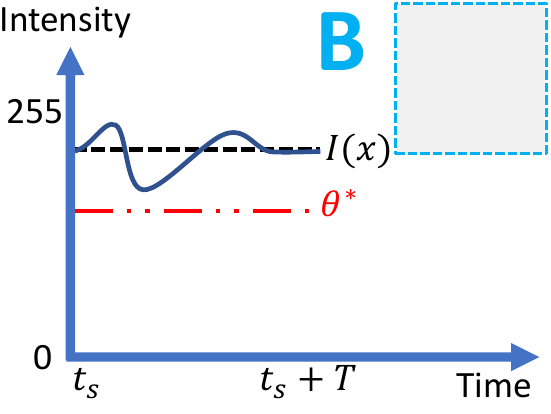}
  \caption{True Positive}    \label{fig:four_6}
\end{subfigure}
\hfill
\begin{subfigure}{0.24\linewidth}
\includegraphics[width=\textwidth]{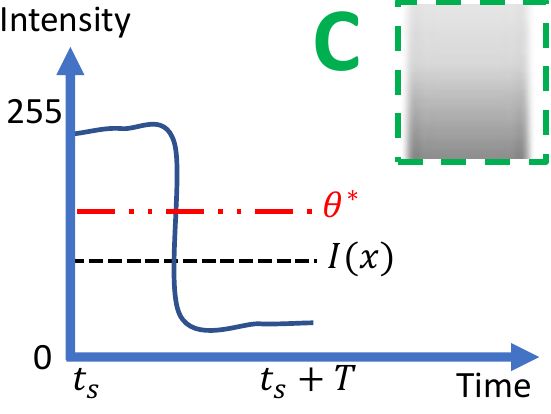}
  \caption{False Negative}    \label{fig:four_7}
\end{subfigure}
\hfill
\begin{subfigure}{0.24\linewidth}
\includegraphics[width=\textwidth]{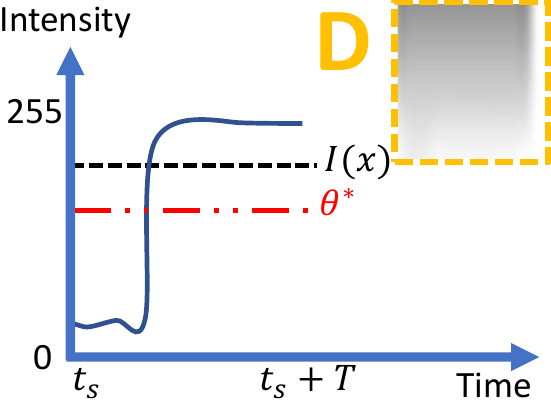}
  \caption{False Positive}    \label{fig:four_8}
\end{subfigure}
  \caption{
  Illustration of pixel classification. In comparing (b) the binary image estimated from (a) the blurry intensity image with (c) the ground truth, we discovered that not all pixels are misclassified, resulting in false positive and false negative classifications, as shown in (d). By examining the pixel areas in (a) the blurry image, we found that the intensity values of pixels in areas like (e) A and (f) B experienced little change during blurring, resulting in accurate binary classification. However, the intensity values of pixels in areas like (g) C and (h) D exceeded the threshold, leading to incorrect binary classification. Best viewed in color.}
\end{figure*}
\subsection{Separating Pixels into Two Subsets}
\label{sec:sep}
To overcome these challenges, rather than solving the intensity reconstruction problem, we propose utilizing the bimodality present in bimodal objects to predict the binary class of each pixel, treating binarization under motion blur as a binary classification problem.



In motion-distorted images, applying an optimal threshold $\theta^*$ estimated using the latent image $L(t,\mathbf{x})$ may produce four possibilities: True Positive (TP), True Negative (TN), False Positive (FP), and False Negative (FN).

\begin{itemize}
\item True Positive (TP) and True Negative (TN). Thresholding results in the motion-blurred images are equivalent to the ground truth, implying that the threshold correctly predicts even when motion blur occurs.
\item False Positive (FP) and False Negative (FN). Thresholding in the motion-blurred images leads to 
incorrect results compared to the ground truth, indicating that motion blur distorts the results of these pixels.
\end{itemize}

The output may contain all four conditions. On the one hand, in the presence of motion blur (\eg, \cref{fig:four_1}), the optimal threshold generates numerous FP and FN and degrades the results (\cref{fig:four_2}). On the other hand, comparing \cref{fig:four_2} with \cref{fig:four_3}, we might still observe that TP and TN persist in the output, as marked in \cref{fig:four_4}. Since TP and TN do not require restoration, we concentrate on recovering FP and FN by leveraging events to analyze their generation due to motion and then 
adopt pixel-wise bimodality to recover them. After that, the remaining pixels could be treated as TP and TN and directly estimated from the blurry image. Therefore, we can classify all pixels into two subsets $\mathbb{X} = \mathbb{T} \cup \mathbb{F}$:
\begin{itemize}
  \itemsep0em
  \item \textbf{True pixels} $\mathbf{x}^t \in \mathbb{T}$. We define the true pixel where the binary output generated by the optimal threshold $B_{I}(\mathbf{x}^t)$ is not affected by motion, \ie, $B_L(t_\text{s}, \mathbf{x}^t) = B_{I}(\mathbf{x}^t)$, where $B_L(t_\text{s}, \mathbf{x}^t)$ is the latent binary image at time $t_s$.
  \item \textbf{False pixels} $\mathbf{x}^f \in \mathbb{F}$. We define the false pixel where the intensity is distorted by motion, leading to the misclassification of binary output, \ie, $B_L(t_\text{s}, \mathbf{x}^f) \neq  B_{I}(\mathbf{x}^f)$.
\end{itemize}

Then, we develop the dual-stage binarization approach to process each class of pixel in one of the stages to recover $B_L(t_\text{s}, \mathbf{x}^f)$ and $B_L(t_\text{s}, \mathbf{x}^t)$, and ultimately combine the two results to restore the final $B_L(t_\text{s}, \mathbf{x})$. Based on $B_L(t_\text{s}, \mathbf{x})$ and the corresponding events, we can propagate the latent binary image to derive the high frame-rate binary video.

\begin{figure*}[htb]
\centering
\begin{subfigure}{0.245\linewidth}
    \includegraphics[width=\textwidth]{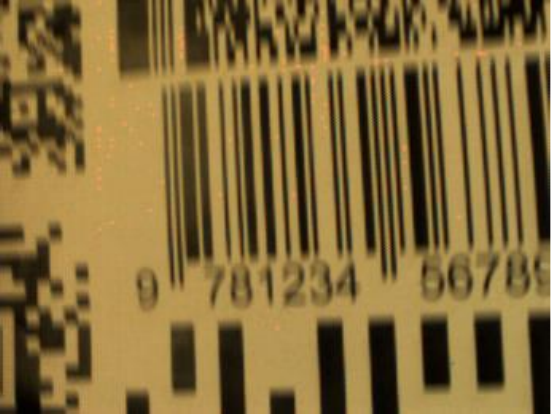}
    \caption{Intensity Image $I(\mathbf{x})$} 
    \label{fig:stage_1}
\end{subfigure}
\hfill
\begin{subfigure}{0.245\linewidth}
    \includegraphics[width=\textwidth]{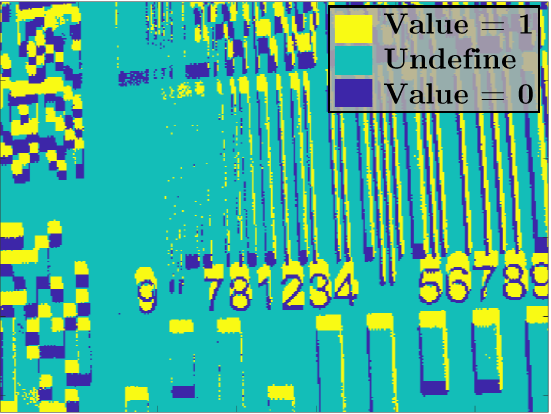}
  \caption{From events $B_L(t_\text{s}, \mathbf{x}^f)$}  
  \label{fig:stage_2}  
\end{subfigure}
\hfill
\begin{subfigure}{0.245\linewidth}
   \includegraphics[width=\textwidth]{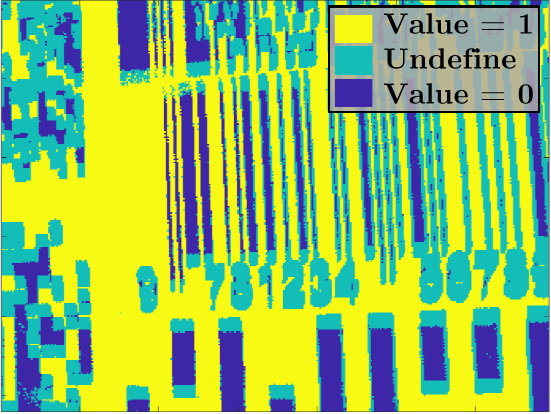}
  \caption{From image $B_L(t_\text{s}, \mathbf{x}^t)$} 
   \label{fig:stage_3}
\end{subfigure}
\hfill
\begin{subfigure}{0.245\linewidth}
    \includegraphics[width=\textwidth]{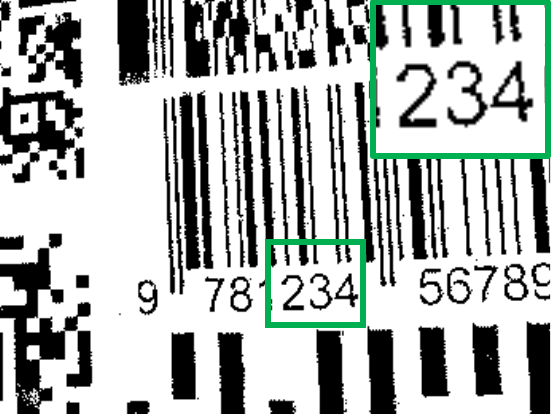}
  \caption{Binary Output $B_{L}(t_\text{s},\mathbf{x})$}
  \label{fig:stage_4}
\end{subfigure}
\vfill
\begin{subfigure}{0.135\linewidth}
    \includegraphics[width=\textwidth]{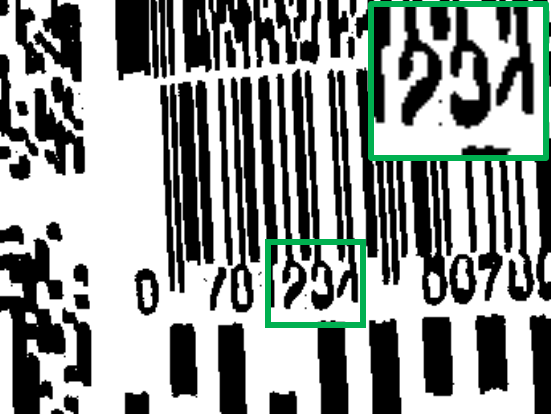}
  \caption{Howe \cite{howe2013document}}
  \label{fig:stage_5}
\end{subfigure}
  \hfill
\begin{subfigure}{0.135\linewidth}
    \includegraphics[width=\textwidth]{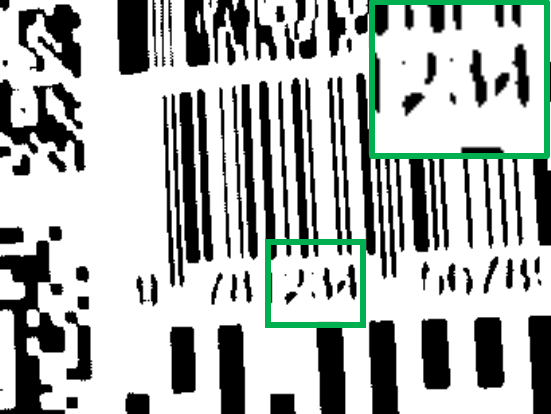}
  \caption{Nick \cite{khurshid2009comparison}}
  \label{fig:stage_6}
\end{subfigure}
  \hfill
\begin{subfigure}{0.135\linewidth}
    \includegraphics[width=\textwidth]{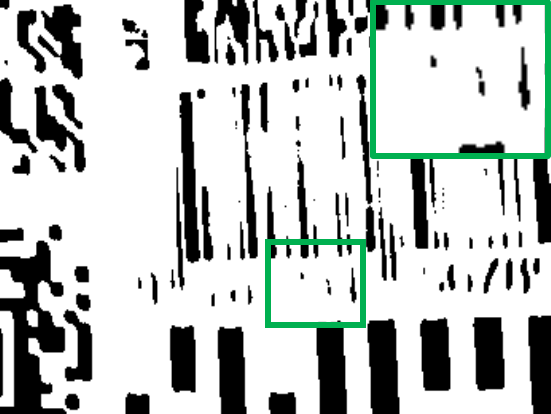}
  \caption{Dplink-Net \cite{xiong2021dp}}
  \label{fig:stage_7}
\end{subfigure}
  \hfill
\begin{subfigure}{0.135\linewidth}
    \includegraphics[width=\textwidth]{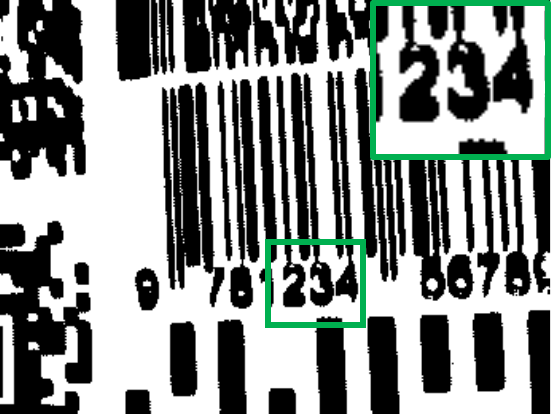}
  \caption{Wan \cite{mustafa2018binarization}}
  \label{fig:stage_8}
  \end{subfigure}
  \hfill
  \begin{subfigure}{0.135\linewidth}
    \includegraphics[width=\textwidth]{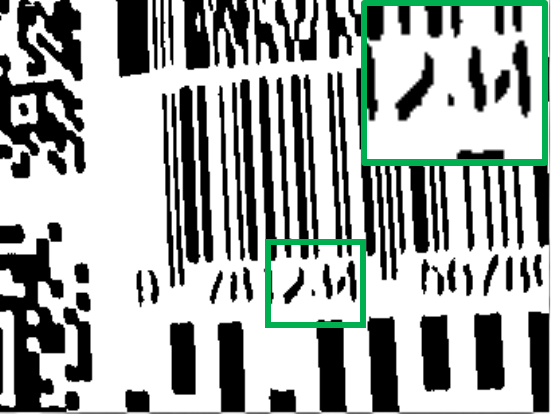}
  \caption{AE \cite{calvo2019selectional}}
  \label{fig:stage_9}
\end{subfigure}
  \hfill
\begin{subfigure}{0.135\linewidth}
    \includegraphics[width=\textwidth]{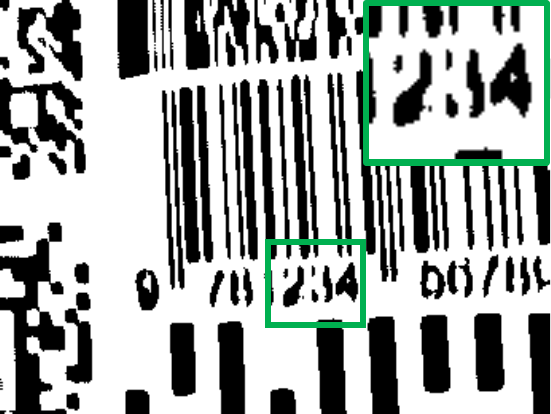}
  \caption{Adaptive \cite{bradley2007adaptive}}
  \label{fig:stage_10}
\end{subfigure}
\begin{subfigure}{0.135\linewidth}
    \includegraphics[width=\textwidth]{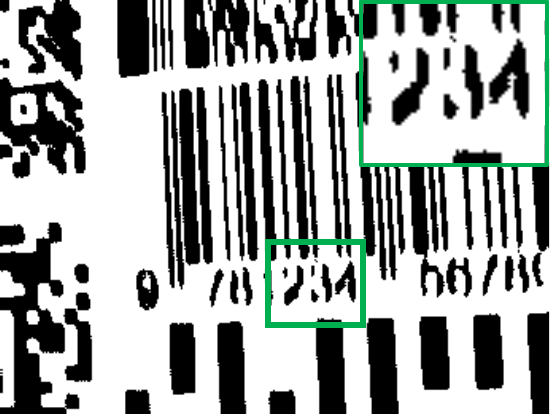}
  \caption{Wolf \cite{wolf2004extraction}}
  \label{fig:stage_11}
\end{subfigure}

  \caption{Demonstration of the dual-stage binarization and its comparisons with conventional image binarization. Given the events and (a) the blurred image captured in the exposure duration, our method first deducts (b) the binary results of false pixels $B_L(t_\text{s}, \mathbf{x}^f)$ and then merges it with (c) the binary result of true pixels $B_L(t_\text{s}, \mathbf{x}^t)$ estimated from the intensity image to recover (d) the complete sharp binary image $B_{L}(t_\text{s},\mathbf{x})$. (e) - (k) The results of conventional image binarization. Best viewed in color.
  }
  \label{fig:stage}
\end{figure*}

\subsection{Dual-stage Binarization}
\label{sec:dual}
We aim to directly predict the latent binary image $B_L(t_\text{s}, \mathbf{x})$ of latent images $L(t_\text{s}, \mathbf{x})$ using a potentially blurred intensity image $I(\mathbf{x})$ captured over the exposure period $\mathcal{T} = [t_\text{s},t_\text{s}+T]$ and the concurrent events $\mathcal{E}$ at $t_k \in \mathcal{T}$. 
To this end, we first recover binary output of false pixels $B_L(t_\text{s}, \mathbf{x}^f)$ using events (\cref{fig:stage_2}), and then merge it with the output $B_L(t_\text{s}, \mathbf{x}^t)$ (\cref{fig:stage_3}) estimated from images to produce latent binary image $B_L(t_s,\mathbf{x})$, which is immune from motion blur (\cref{fig:stage_4}):
\begin{align} 
B_L(t_s,\mathbf{x}) =  B_L(t_\text{s}, \mathbf{x}^f) \cup  B_L(t_\text{s}, \mathbf{x}^t), \label{eq:eib}
\end{align}
where $\cup$ is the Logical-OR operator. 

\subsubsection{Inference in the event space}

In the initial stage, we recover the binary image of false pixels $\mathbf{x}^f$, which belong to set $\mathbb{F}$. Typically, bimodal objects such as barcodes and markers exhibit high contrast patterns and have strongly bimodal intensity distributions. That is to say, to distort the binarization output and result in false negatives (FN) and false positives (FP), there needs to be a strong intensity variation during motion blurring (\cref{fig:four_1}), causing the intensity to vary from small to large or vice versa. 

For instance, under motion blur, pixels in regions A and B (\cref{fig:four_5} and \cref{fig:four_6}), which are affected by their neighboring pixels and share similar intensity values, undergo slight changes in their ultimate average intensity. Consequently, these pixels are correctly classified by the threshold, resulting in true positives (TP) and true negatives (TN). However, for FN and FP cases, large intensity variations occur in pixels in regions C and D (\cref{fig:four_7} and \cref{fig:four_8}), leading to shifts in the ultimate average intensity and resulting in classification errors. 
Nevertheless, these variations are robustly recorded by events, which are triggered based on the initial log intensity level at time $t_s$. Due to the bimodal nature of the object, the initial intensity level is set to either very high (\cref{fig:four_7}) or very low (\cref{fig:four_8}) values. Thus, there is only one way for these pixels to vary, which is either to decrease and generate massive negative events or to increase and generate massive positive events. 
Consequently, we can recover the binary results by detecting the first large rising or falling edge using the following logic: 
\begin{align}\label{eq:logic}
  B_L(t_\text{s},\mathbf{x}^f)=\left\{\begin{array}{ll}
    0, & \text { if the first large edge is rising} , \\
     1, & \text { if the first large edge is falling}.
     \end{array}\right.
\end{align}

To obtain the logical expression described by \cref{eq:logic}, we propose a bi-directional integration that effectively combines the intensity variations from positive and negative events. The integration terminates once the variation along one direction exceeds the edge threshold $\theta_{e} > 0$, at which point the rising or falling edge is detected, \ie,
\begin{align}\label{eq:altered}
  B_L(t_\text{s},\mathbf{x}^f)=\left\{\begin{array}{ll}
    0, & \text { if} \quad \Delta I(t_e,\mathbf{x}^f,+1) > \theta_{e}, \\
     1, & \text { if} \quad \Delta I(t_e,\mathbf{x}^f,-1) > \theta_{e},
     \end{array}\right.
\end{align}
where 
\lsj{
\begin{align}
    \Delta I(t,\mathbf{x},p) = c\sum_{k=1}^{N_e} \sigma(p-p_k) \sigma(\mathbf{x} - \mathbf{x}_k) H(t_k - t_e),
\end{align}
$t_e$ is the end time of the interval $[t_s, t_e]$ in which the integration in positive or negative direction reaches the threshold. $\sigma(\cdot)$ indicates the Kronecker delta \cite{gehrig2020eklt}, and $H(\cdot)$ is a step function defined as:
\begin{align}
    H(\Delta t) =
\begin{cases}
    0, & \text{for } \Delta t \geq 0, \\
    1, & \text{for } \Delta t < 0.
\end{cases}
\end{align}
Since we use the first edge that is large enough to exceed the threshold, small noisy edges should not exceed the threshold, thus avoiding potential misclassification and ensuring the overall validity of the results.
}
Therefore, we can use the integration of events to infer the binary image of false pixels $B_L(t_s,\mathbf{x}^f)$ (\cref{fig:stage_2}). The pixels that are not classified by \cref{eq:altered} are left undefined and will be processed in the next stage. 

\subsubsection{Inference in the image space}
In the first stage, we treat pixels that trigger large edges as false pixels and deduct their true binary status at time $t_s$. But to complete the whole binary image $B_L(t_\text{s}, \mathbf{x})$, we still lack the result of true pixels, \ie, $B_L(t_\text{s}, \mathbf{x}^t)$. For the true pixels $\mathbf{x}^t$, the intensity variations are generally small (\eg, \cref{fig:four_5} and \cref{fig:four_6}), generating fewer events than the false ones. Thus, we treat the pixels without sufficient events that pass the large edge detection in \cref{eq:altered} as the true pixels, \ie, pixels that belong to the subset $\mathbb{T} = \mathbb{X} \backslash \mathbb{F}$. 

Given that the intensity fluctuations are small, we can directly apply a threshold $\theta_{I}$ to obtain the correct classification result:
\begin{align}\label{eq:true}
  B_L(t_\text{s},\mathbf{x}^t)=\left\{\begin{array}{ll}
    0, & \text { if} \quad I(\mathbf{x}^t) \leq \theta_{I}, \\
    1, & \text { if} \quad I(\mathbf{x}^t) > \theta_{I}.
     \end{array}\right.
\end{align}
The result of estimated $B_L(t_\text{s}, \mathbf{x}^t)$ is shown in \cref{fig:stage_3}. Finally, we merge results from the first stage (\cref{fig:stage_2}) and the second stage (\cref{fig:stage_3}) to get the clear binary image $B_L( t_\text{s},\mathbf{x})$ (\cref{fig:stage_4}), which is unaffected by the motion blur. 
In contrast, conventional image binarizations hardly handle the motion of blurry images, resulting in unclear boundaries and degraded context, as shown in \cref{fig:stage_5,fig:stage_6,fig:stage_7,fig:stage_8,fig:stage_9,fig:stage_10,fig:stage_11}.

\begin{algorithm}[t] 
  \caption{Unidirectional Integration (Video)}
  \KwData{Initial latent binary image $B_L(t_\text{s},\mathbf{x})$, events set $\mathcal{E}$ with event $\mathbf{e}_k = (\mathbf{x}_k, t_k, p_k)$, threshold $\theta_{e}$\label{alg:sdi}
  }
  \KwResult{Binary images at time $t_k$, \ie, $B_L(t_k,\mathbf{x})$} 
  $I_\text{pos}(\mathbf{x}) = I_\text{neg}(\mathbf{x}) = 0$ \;
    \For{each event $\mathbf{e}_k$ in the event set $\mathcal{E}$}{
         \If{$B_L(t_{k-1},\mathbf{x}_{k-1}) = 0$ and $p_k = +1$}
         {
              $ I_\text{pos}(\mathbf{x}_k) = I_\text{pos}(\mathbf{x}_k) + c$\;
              \If{$I_\text{pos}(\mathbf{x}_k) > \theta_{e}$ }
              {
                $I_\text{pos}(\mathbf{x}_k) = 0$; $B_L(t_k,\mathbf{x}_k) = 1$;
              }
         }
         \ElseIf{$B_L(t_{k-1},\mathbf{x}_{k-1}) = 1$ and $p_k = -1$}
         {
          $ I_\text{neg}(\mathbf{x}_k) = I_\text{neg}(\mathbf{x}_k) - c$\;
          \If{$I_\text{neg}(\mathbf{x}_k) < -\theta_{e}$ }
          {
            $ I_\text{neg}(\mathbf{x}_k) = 0$; $B_L(t_k,\mathbf{x}_k) = 0$;
          }
         }
         \Else
         {
            $B_L(t_{k},\mathbf{x}_{k}) = B_L(t_{k-1},\mathbf{x}_{k-1})$;
         }
    }
\end{algorithm}
\subsection{High Frame-rate Binary Video Reconstruction}
\label{sec:video}
It is also possible to generate high frame-rate binary video $B_L(t,\mathbf{x})$ using the binary image $B_L(t_\text{s},\mathbf{x})$ and the concurrent events. The binary image $B_L(t_\text{s},\mathbf{x})$ shows the base intensity level of pixel $\mathbf{x}$ at time $t_\text{s}$. $B_L(t_\text{s}, \mathbf{x}) = 0$ indicates the initial intensity level is lower than the threshold, and vice versa. Therefore, for pixels with $B_L(t_\text{s}, \mathbf{x}) = 0$, only when the intensity increases to exceed the threshold, the binary result will switch to $1$, \ie, $B_L(t, \mathbf{x}) = 1$. Therefore, we can integrate events to detect the rising or falling edge and then infer the changing state to update $B_L(t_s,\mathbf{x})$ to generate high frame-rate binary video. However, because the event camera uses two different hardware biases to generate the positive and negative events, the noise of the two event polarities is distributed differently. If one integrates all the events, the unbalanced noise could quickly deviate the integral result so that it cannot be thresholded.

\begin{figure*}[t]
  \centering
  \includegraphics[width=1\textwidth]{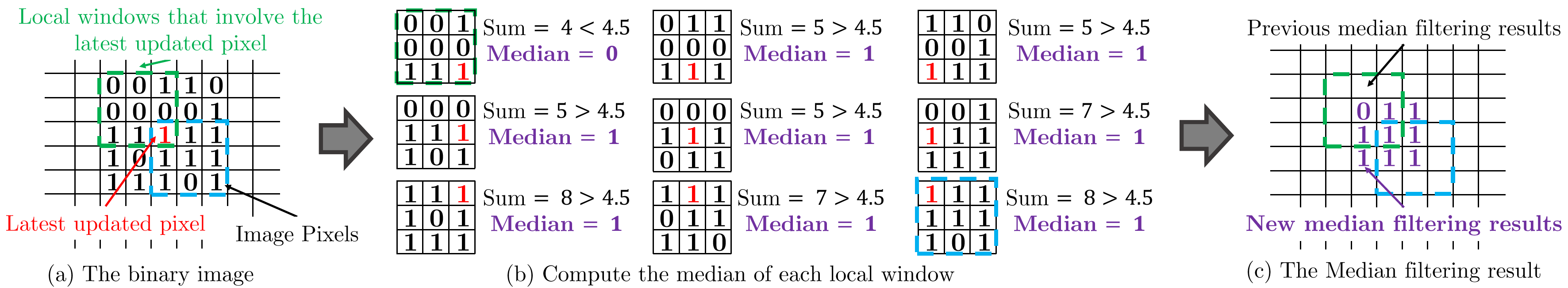}
  \caption{Workflow of the asynchronous median filter. Given the latest updated pixel (red) in (a) the binary image, we can asynchronously compute the median of (b) 9 related local windows by comparing the summation with half of the window size. Then we can produce (c) a new denoised binary image. Best viewed in color.}
  \label{fig:asy}
\end{figure*}
\begin{figure}[t]
  \begin{subfigure}{0.325\linewidth}
    \includegraphics[width=\textwidth]{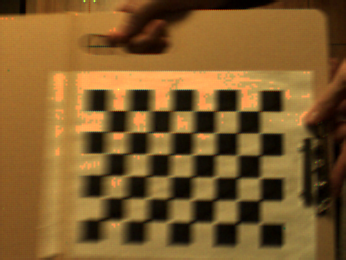}
    \caption{The blurred image}    \label{fig:filtering_6}
  \end{subfigure}
  \hfill
  \begin{subfigure}{0.325\linewidth}
    \includegraphics[width=\textwidth]{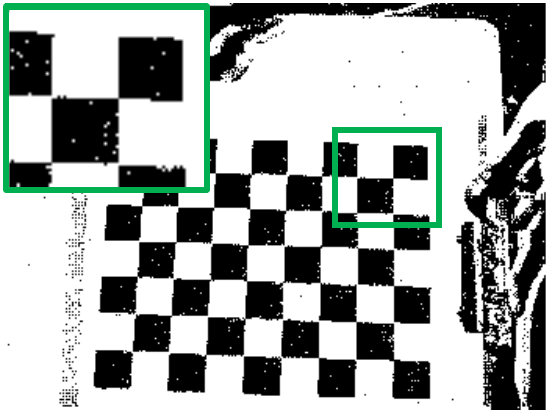}
    \caption{W/o AMF (video)}    \label{fig:filtering_3}
  \end{subfigure}
  \hfill
  \begin{subfigure}{0.325\linewidth}
    \includegraphics[width=\textwidth]{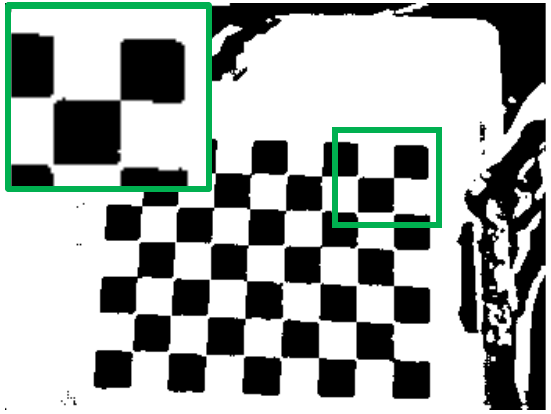}
    \caption{W/ AMF (video)}    \label{fig:filtering_4}
  \end{subfigure}
  \caption{Demonstration of filtering. Our method can generate (b) high frame-rate binary videos of the \textit{Chessboard} sequence, which removes the blurry effects in (a) the blurry inputs. (c) Our asynchronous median filter (AMF) can effectively eliminate the noise.}
  \label{fig:filtering}
\end{figure}
\subsubsection{Unidirectional integration} To this end, we develop unidirectional integration (summarized in \cref{alg:sdi}). We first integrate the single polarity of events based on their current binary state, \eg, integrating positive events when $B_L(t, \mathbf{x})=0$. Then, the binary image is updated once the integration exceeds the threshold $\theta_{e}$. The next integration then uses events with the opposite polarity, \eg, if updates $B_L(t, \mathbf{x})$ to 1, then the next integration only uses negative events. Since only one polarity is considered, the integration result is only a relative measurement of a specific polarity in a short time interval. Therefore, the proposed method suppresses the opposite polarity event noises, allowing the effective generation of high-rate binary video.

\subsubsection{Asynchronous median filtering (AMF)} 

Event noise may produce pepper and salt dots in the binary video (\cref{fig:filtering_3}). The conventional median filter \cite{gonzalez2018digital} requires pre-sorting and operates synchronously, limiting its performance in the generation of high frame-rate videos. To this end, we develop an efficient median filter that operates asynchronously, skipping the time-consuming pre-sorting operation. The median filter uses the median of a local sliding window to replace the centered value and scan the entire image.
\lsj{
As shown in \cref{fig:asy}, when using a $3\times3$ filtering window, each pixel of the image participates in only 9 median calculations (\cref{fig:asy}b). 
Therefore, once the pixel in the binary image is updated asynchronously (\eg, the red 1 in \cref{fig:asy}a), we do not need to apply the median filter to the entire image; instead, we update the local windows around the pixel in different relative positions (\cref{fig:asy}b). Then, we can use the newly updated pixels to create a denoised binary image (\cref{fig:asy}c). Moreover, the sorting in the conventional median filter could be greatly simplified. Since the binary image contains only binary values (0 and 1), sorting to find the median of a local window can be replaced by comparing the positive pixel count over the total number of pixels in the local window.
If this ratio is larger than 0.5, the median is 1 (\eg, in the lower right blue window of \cref{fig:asy}b), otherwise it is 0 (\eg, in the upper left green window of \cref{fig:asy}b). This asynchronous filtering ensures the generation of denoised binary videos at a high frame rate, as shown in \cref{fig:filtering_4}.
}

\begin{figure}[t]
  \begin{subfigure}{0.485\linewidth}
    \includegraphics[width=\textwidth]{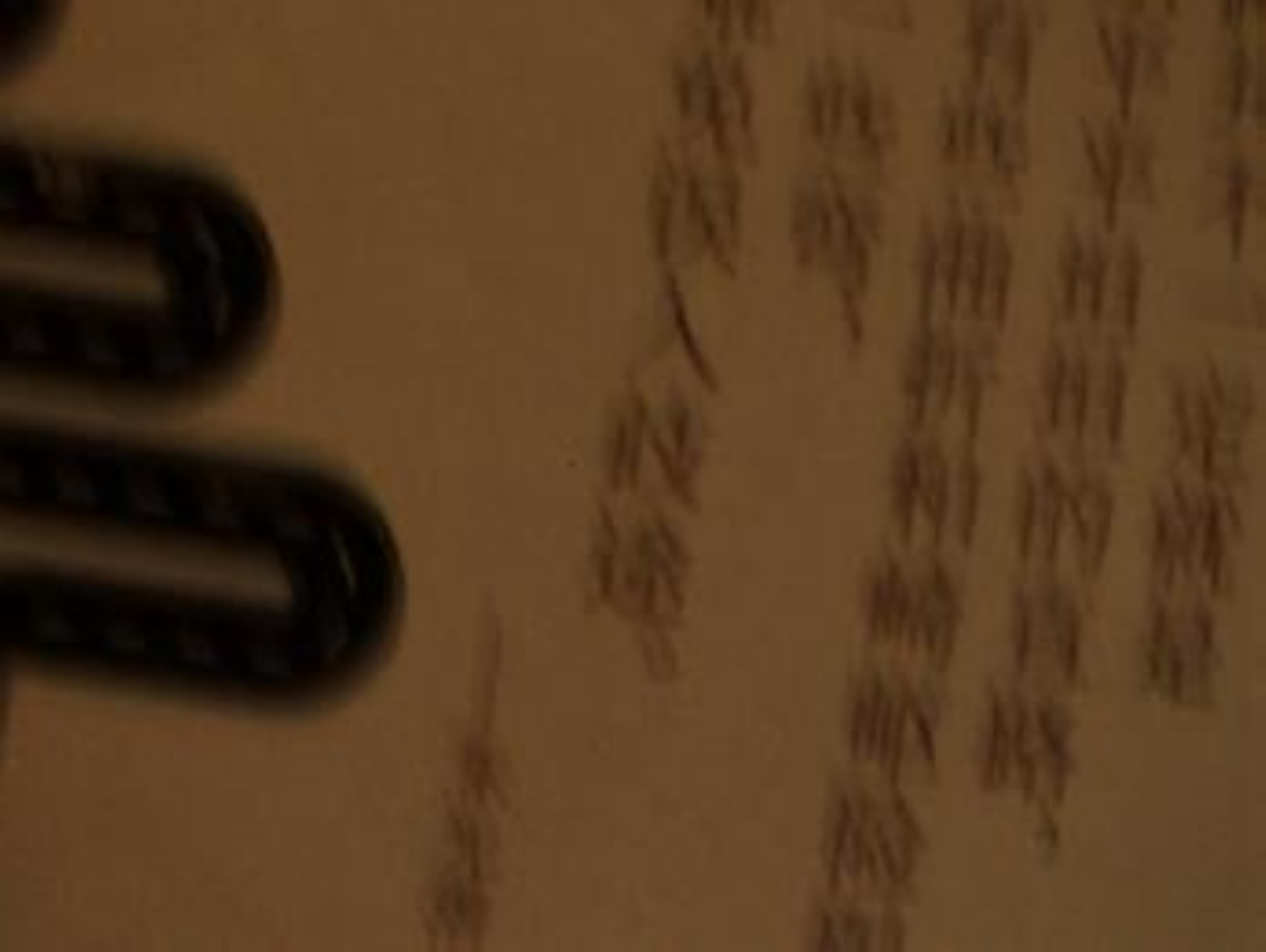}
    \caption{The blurry image}    \label{fig:LDA_1}
  \end{subfigure}
  \hfill
  \begin{subfigure}{0.485\linewidth}
    \includegraphics[width=\textwidth]{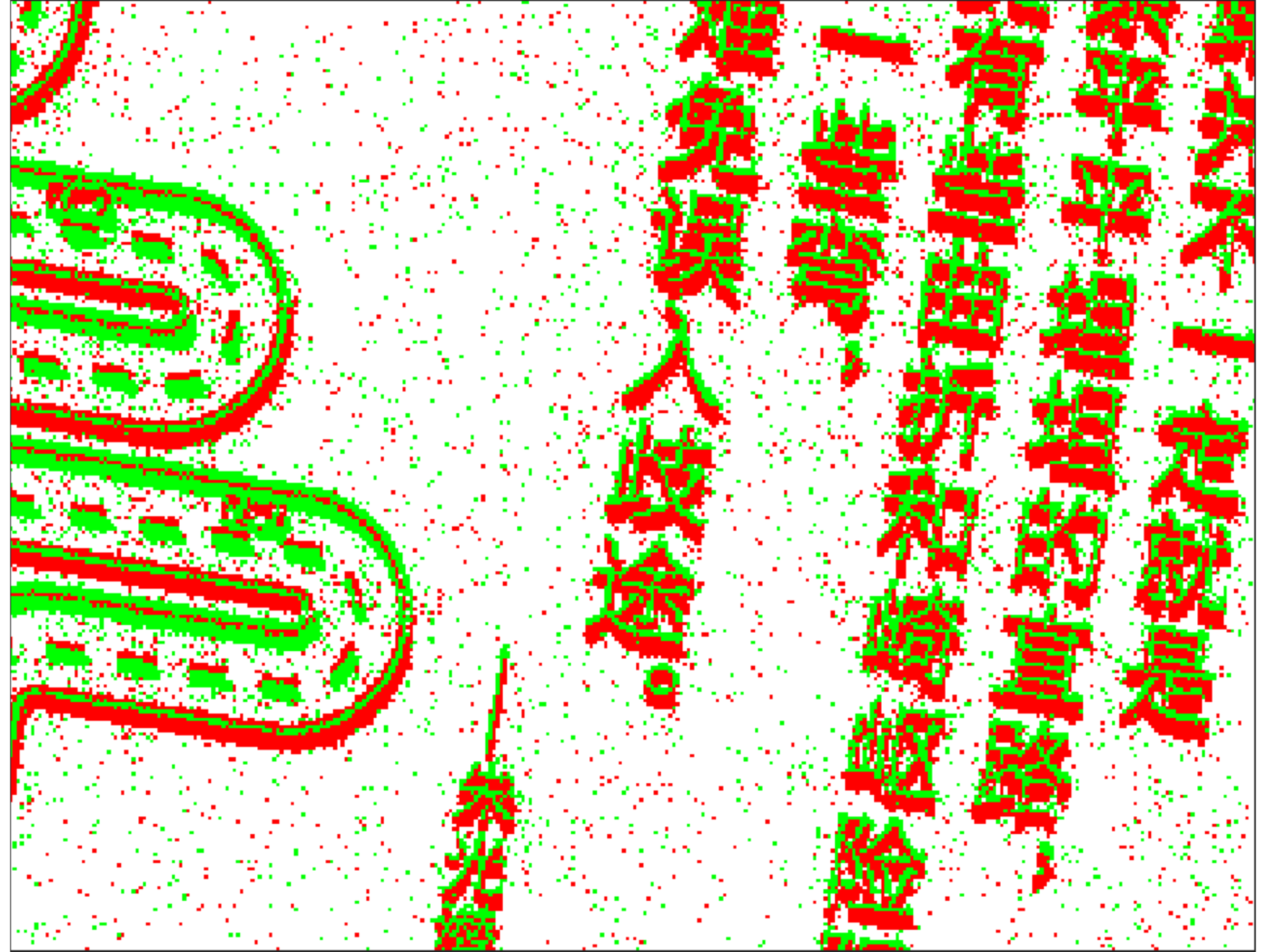}
    \caption{Events}    \label{fig:LDA_2}
  \end{subfigure}
  \\
  \begin{subfigure}{0.485\linewidth}
    \includegraphics[width=\textwidth]{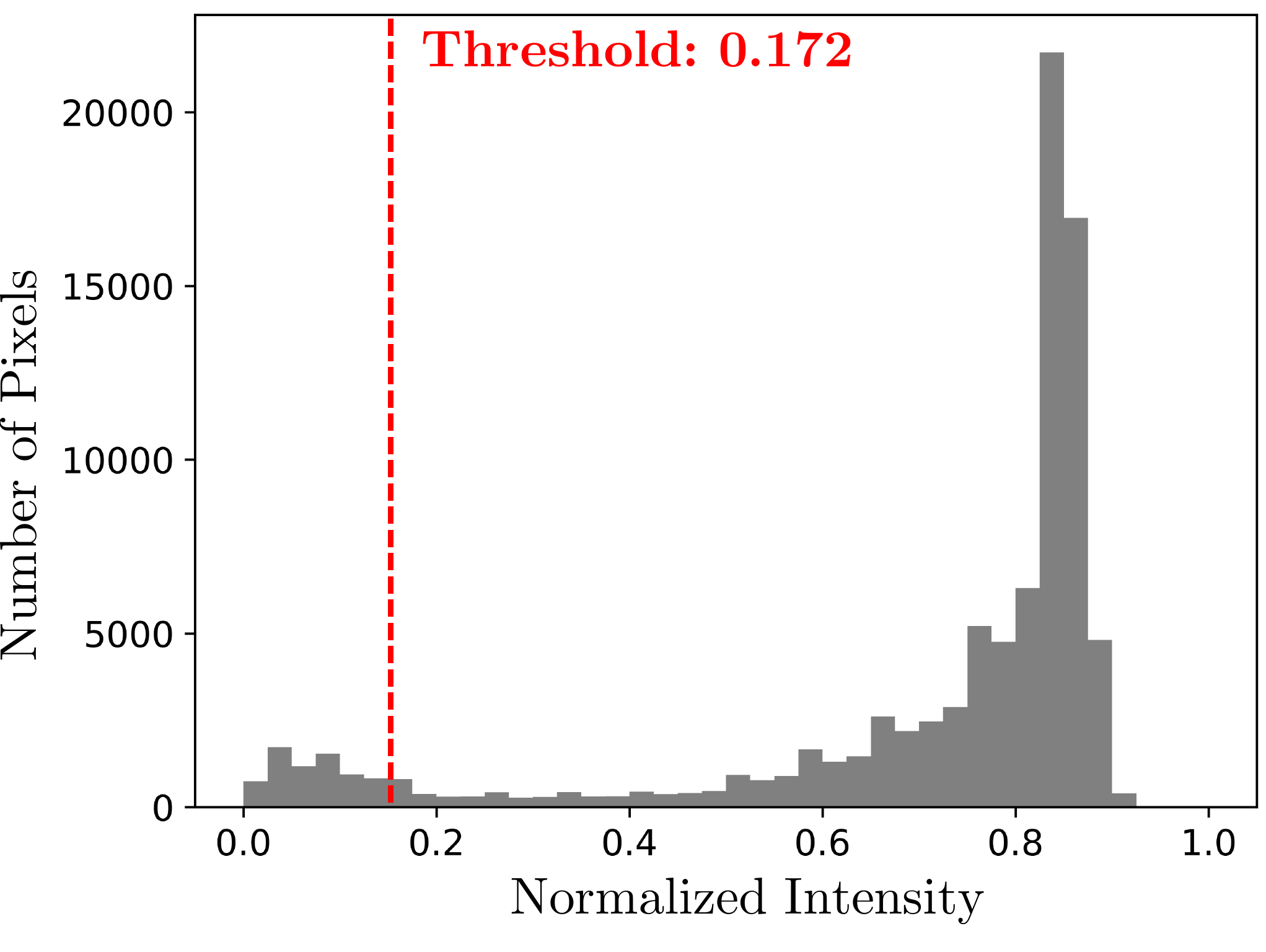}
    \caption{Histogram (without fusion)}    \label{fig:LDA_3}
  \end{subfigure}
  \hfill
  \begin{subfigure}{0.485\linewidth}
    \includegraphics[width=\textwidth]{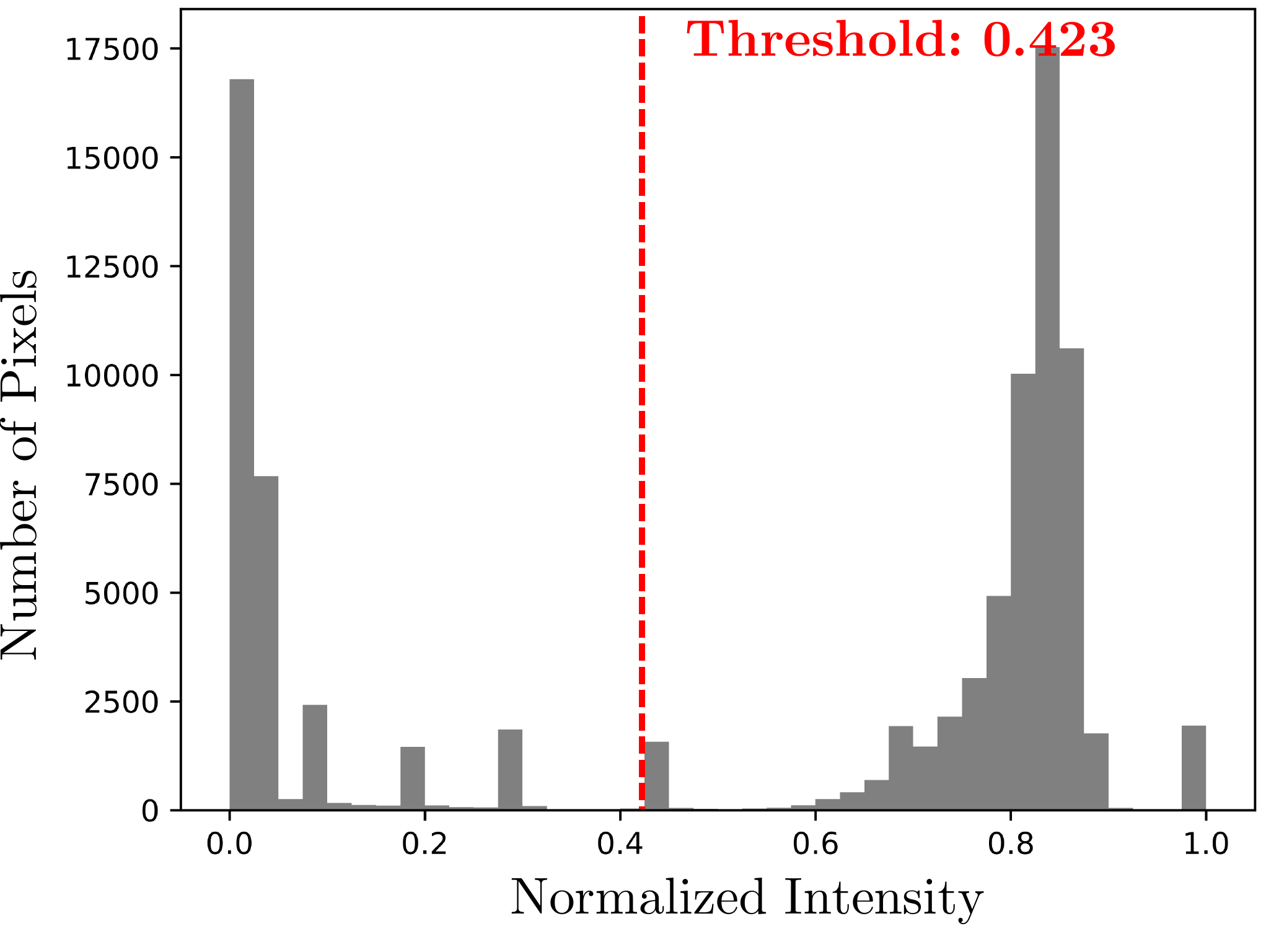}
    \caption{Histogram (with fusion)}    \label{fig:LDA_4}
  \end{subfigure}
  \\
  \begin{subfigure}{0.485\linewidth}
    \includegraphics[width=\textwidth]{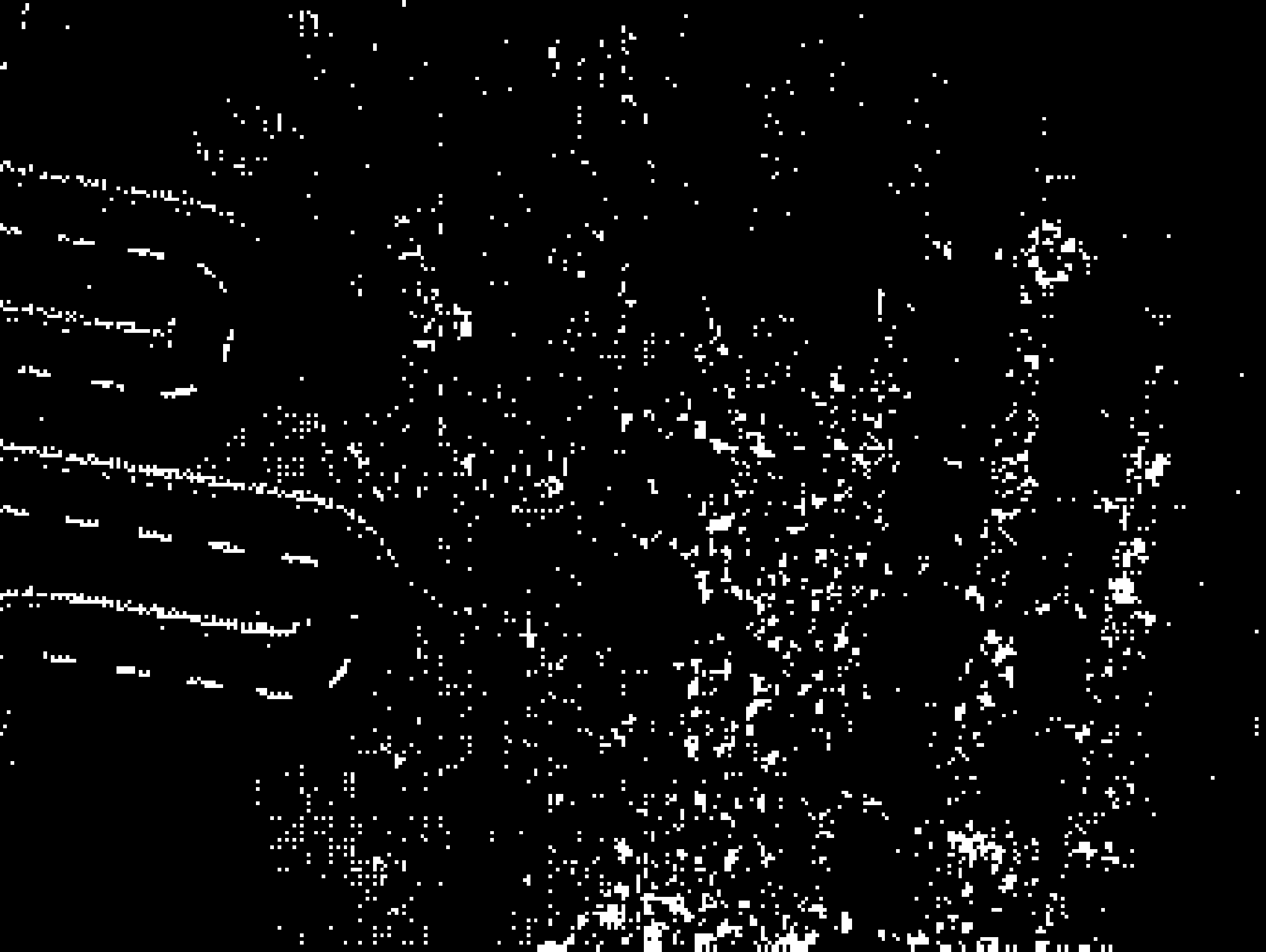}
    \caption{Without fusion}    \label{fig:LDA_5}
  \end{subfigure}
  \hfill
  \begin{subfigure}{0.485\linewidth}
    \includegraphics[width=\textwidth]{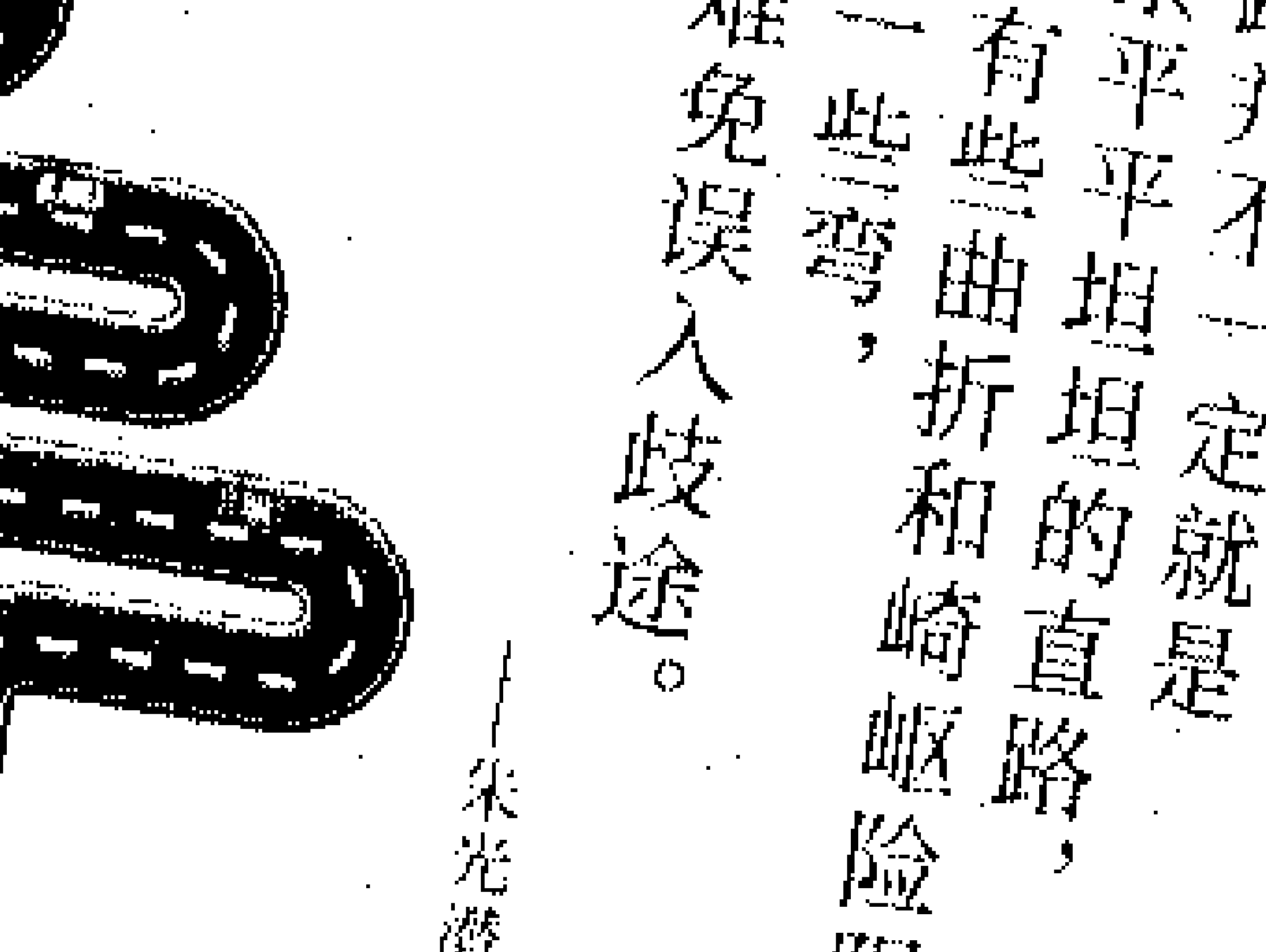}
    \caption{With fusion}    \label{fig:LDA_6}
  \end{subfigure}
  \caption{\lsj{Our fusion module can correct biased statistical information. (a) Motion-blurred images contain biased statistical information, which cannot support accurate thresholding, as shown in (e). We fuse events (b) with the image to construct a motion-invariant histogram (d), allowing for precise threshold optimization (f).}}
  \label{fig:LDA}
\end{figure}

\subsection{Automatic Threshold Estimation}
\label{sec:thresh}

Thus far, we have discussed the recovery of the binary image and its propagation to the high frame-rate video. However, three parameters remain unknown. The first parameter is the contrast of the events camera $c$, which is physically set during the camera's bias configuration and can be obtained using the jAER software \cite{jaer} or data-driven calibration \cite{edipami,wang2020eventcal}. Other parameters are the thresholds $\theta_{e}$ and $\theta_{I}$, which are used to threshold events and images, respectively. Although these parameters can be manually set to achieve effective binarization, variations in natural lighting make manually specified values less robust.  To address this problem, we develop an unsupervised threshold estimation by fusing events and images to generate a motion-invariant histogram and conducting discriminant analysis on it.

\lsj{\subsubsection{Data fusion} Conventionally, the threshold could be determined from the histogram of the intensity image. But when motion blur occurs, the statistical information can be severely biased, as shown in \cref{fig:LDA_1,fig:LDA_3}, resulting in an incorrect threshold for our binarization pipeline, as shown in \cref{fig:LDA_5}. Therefore, to remove the motion blur effect and recover the bimodality from the histogram, we fuse events (\eg, \cref{fig:LDA_2}) into the motion-blurred image (\eg, \cref{fig:LDA_1}) using event integration to approximate the intensity fluctuations caused by motion (as discussed in \cref{eq:logic}).} This allows us to integrate events in each pixel to estimate the first rising or falling integration edge as an approximation of the intensity variation during motion blur and then reconstruct the latent intensity from the variation. 
\lsj{
To this end, we generate the first integration edge image $E(\mathbf{x})$ using events:
\begin{align}\label{eq:fst_edge}
  E(\mathbf{x})\approx c\sum_{k = 1}^{N_e} p_k \sigma(\mathbf{x} - \mathbf{x}_k) H(t_k - t_c),
\end{align}
where $\mathbf{x}$ indicates the pixel position, and $t_c$ is the time that the event's polarity at $\mathbf{x}$ changed compared with previously triggered events. For example, given a pixel $\mathbf{x}$ in the event sensor, if the first event being triggered at this pixel since the exposure start is positive ($p_k = +1$), we will keep integral positive events at this pixel, until one negative event ($p_k = -1$) is triggered, meaning the polarity changed, and this integration is vice versa for the negative events.
}
We compute the mean of the first edge image and set values beyond three standard deviations to zero to remove extreme integration edges induced by hot pixels \cite{dvs128}. Subsequently, we reconstruct the latent intensity $L_e(\mathbf{x})$ using the first integration edge image:
\begin{align}\label{eq:fst_edge_image}
  L_e(\mathbf{x})\approx \left\{\begin{array}{ll}
    \exp(E^{p}_{max} - E(\mathbf{x})), & \text{if }  E(\mathbf{x}) > 0, \\
     \exp(E^{n}_{max}-E(\mathbf{x})), & \text{if }  E(\mathbf{x}) < 0,
     \end{array}\right.
\end{align}
where $E^{p}_{max}$ and $E^{n}_{max}$ denote the maximum integration values for the positive and negative edges, respectively, indicating an approximation of the maximum intensity range in the events space. Therefore, \cref{eq:fst_edge_image} is a reverse deduction from the maximum range, allowing us to infer the latent intensity for recovering bimodality in the histogram \cref{fig:LDA_4}.

Then, we apply the min-max normalization to $L_e(\mathbf{x})$ to generate a normalized latent intensity $\tilde{L}_e(\mathbf{x})$, which relaxes the need for accurate contrast $c$ estimation and alleviates the inaccurate quantification problem of event cameras at high contrast boundaries. We also apply the min-max normalization to the blurry image $I(\mathbf{x})$ to generate the normalized blurry image $\tilde{I}(\mathbf{x})$. The normalization ensures the image and events data are independently scaled to the same range \ie, $[0,1]$ for thresholding. Then we assign pixels $\mathbf{x}$ with $\tilde{L}_e(\mathbf{x}) > 0$ to the normalized blurry image $\tilde{I}(\mathbf{x})$, and produce the fuse image $\tilde{I}_f(\mathbf{x})$. 
Finally, the fused image could reduce the effect of motion blur for robust threshold estimation and binarization (\cref{fig:LDA_4,fig:LDA_6}).

\lsj{\subsubsection{Optimization} From \cref{fig:LDA_3,fig:LDA_4}, we can see that after the fusion process, the histogram reveals clear bimodality. That means the bimodality from the events and the image is well aligned. Therefore, this histogram allows us to estimate a reliable threshold to segment both the events and images into a binary space. Using the histogram of the fused image $\tilde{I}_f(\mathbf{x})$, we aim to find an optimal threshold $\theta^*$ that can classify the fused histogram into two classes, \ie $C_{f}$ (representing intensity levels $[1, \cdots, \theta^*]$) and $C_{b}$ (representing intensity levels $[\theta^*, \cdots, L]$), corresponding to the foreground and background, respectively. That is to say, the threshold could be treated as a decision boundary used in the discriminant analysis \cite{otsu1979threshold} to separate the histogram into two classes. To achieve this, we can use Fisher's discriminant analysis, \ie, maximizing the ratio of the between-class variance $\sigma_B^2$ and within-class variance $\sigma_W^2$:
\begin{equation}\label{eq:ratio}
    \lambda(\theta) = \sigma_B^2(\theta) / \sigma_W^2(\theta),
\end{equation}
where $\sigma_B^2$ is the between-class variance and $\sigma_W^2$ is the within-class variance, defined as:
\begin{align*}
&\sigma_B^2(\theta)=\frac{\left[\mu_H \omega(\theta)-\mu(\theta)\right]^2}{\omega(\theta)[1-\omega(\theta)]}, \\
&\sigma_W^2(\theta)= \sum_{i=1}^\theta (i-\frac{\mu(\theta)}{\omega(\theta)})^2 \mathcal{P}_i + \sum_{i=\theta+1}^L (i-\frac{\mu_H - \mu(\theta)}{1-\omega(\theta)})^2 \mathcal{P}_i,
\end{align*}
}
where, 
\begin{equation}
\omega(\theta)=\sum_{i=1}^\theta \mathcal{P}_i, \quad \mu(\theta)=\sum_{i=1}^\theta i \mathcal{P}_i, 
\end{equation}
are the zeroth-order and first-order cumulative moments of the histogram up to the $\theta$-th level, respectively. And the probability distribution $\mathcal{P}_i$ is given by
\begin{equation}
\mathcal{P}_i = n_i/N,\quad \text{with} \quad  \mathcal{P}_i \geq 0, \quad \sum^L_{i=1} \mathcal{P}_i = 1,
\end{equation}
where, $N$ is the total number of pixels, while $n_i$ denotes the pixel count at each intensity level.  $\mu_H = \mu(L)$ is the mean value of the entire histogram, $L=256$ denotes the fused histogram containing 256 intensity levels.

\lsj{We can maximize the $\lambda$ to identity the optimum threshold. However, one basic relation is that the total variance is equal to the sum of the between-class variance and within-class variance \ie, $\sigma_T^2 = \sigma_B^2(\theta) + \sigma_W^2(\theta)$, where $\sigma_T^2$ is the total variance of levels, given by,
\begin{align}
\sigma_T^2 = \sum_{i=1}^L (i - \mu_H)^2 \mathcal{P}_i,
\end{align}
where we can see that $\sigma_T^2$ is unrelated to the choice of the threshold. Thus, maximizing the \cref{eq:ratio} is equivalent to maximizing the between-class variance $\sigma_B^2$, \ie,
\begin{align}\label{eq:thresEst}
   \lambda(\theta^*) =  \sigma_B^2(\theta^*) / (\sigma_T^2 - \sigma_B^2(\theta^*)) = \max_{1 \leq \theta<L} \sigma_B^2(\theta).
\end{align} 
We solve \cref{eq:thresEst} using the sequential search.} Finally, the thresholds for our event-based binary reconstruction in \cref{sec:dual,sec:video} can be given by:
\begin{equation}
\theta_{I} = \theta^*,\quad  \theta_{e} = \frac{\theta^*}{L} \cdot \text{max}(|E(\mathbf{x})|)
\end{equation}
$\theta^*$ can be directly employed for thresholding the image, and $\theta_{e}$ is scaled to the log space using the maximum value of integration edges for thresholding the events. 

\subsection{Complexity Analysis}
\label{sec:complexity}
\subsubsection{Time Complexity} 
\lsj{
We analyze the computational time regarding three major parts of the proposed algorithm:
1) dual-stage binarization, 2) binary video generation, and 3) threshold estimation. In the dual-stage binarization, the program iterates over the events to perform event integration. Since the integration at each pixel halts upon detecting the first edge, no further events are processed at that pixel position. 
Since such pixels are termed false pixels (\cref{sec:sep}), the time complexity of the dual-stage binarization is \textit{linear} to the number of pixels $\mathcal{O}(N_f)$, where $N_f$ is the number of false pixels.
}

\lsj{In the binary video generation, each event is processed once in an asynchronous manner. This involves executing \cref{alg:sdi} and the asynchronous median filter for each event. Both components entail only a limited number of operations. Consequently, the time complexity of generating the binary video is \textit{linear} with respect to the number of events $N_e$, denoted as $\mathcal{O}(N_e)$. }

\lsj{The threshold estimation process involves scanning all possible intensity levels using a one-dimensional histogram with $L=256$ levels. Therefore, it has a constant time complexity of $\mathcal{O}(1)$. In summary, the time complexity of our algorithms can be expressed as $\mathcal{O}(N_f + N_e + 1)$. Considering that the number of events ($N_e$) typically exceeds the number of false pixels ($N_f$), the overall time complexity is approximately $\mathcal{O}(N_e)$. }

\lsj{In comparison, the event-based double integral (EDI) \cite{edipami}, which serves as the principal method for relating the events and images for intensity reconstruction, requires doubly integral to generate the intensity, resulting in quadratic time complexity $\mathcal{O}(N_e^2)$. Learning-based methods like eSL \cite{yu2023learning} or LEDVDI \cite{lin2020learning} demonstrate linear time complexity when applied to a fixed frame-rate determined by the network (\eg, deblurring a single image). However, to increase the frame-rates, operations such as nested inference \cite{yu2023learning} can lead to quadratic complexity. Furthermore, these methods typically involve converting events into voxel-like representations, which necessitate operations such as event integration, polarity flipping, and normalization. These operations significantly impair their runtime performance. 
}

\subsubsection{Space Complexity}
\lsj{Two types of data are buffered for processing: the first one is an image stored in an array-like container, and the second one is the events stored in a list-like container. Additionally, we need to store the integration image (\ie, \cref{eq:altered}) which is updated during processing. Since the size of these two images remains fixed, their complexity is $\mathcal{O}(1)$. The events change dynamically due to motion, resulting in a space complexity of $\mathcal{O}(N_e)$ for storing them. In summary, the overall space complexity is $\mathcal{O}(N_e)$.}

\begin{figure*}[ht!]
\begin{minipage}{0.325\linewidth}\centering
  \begin{subfigure}{0.4905\linewidth}
    \includegraphics[width=\textwidth]{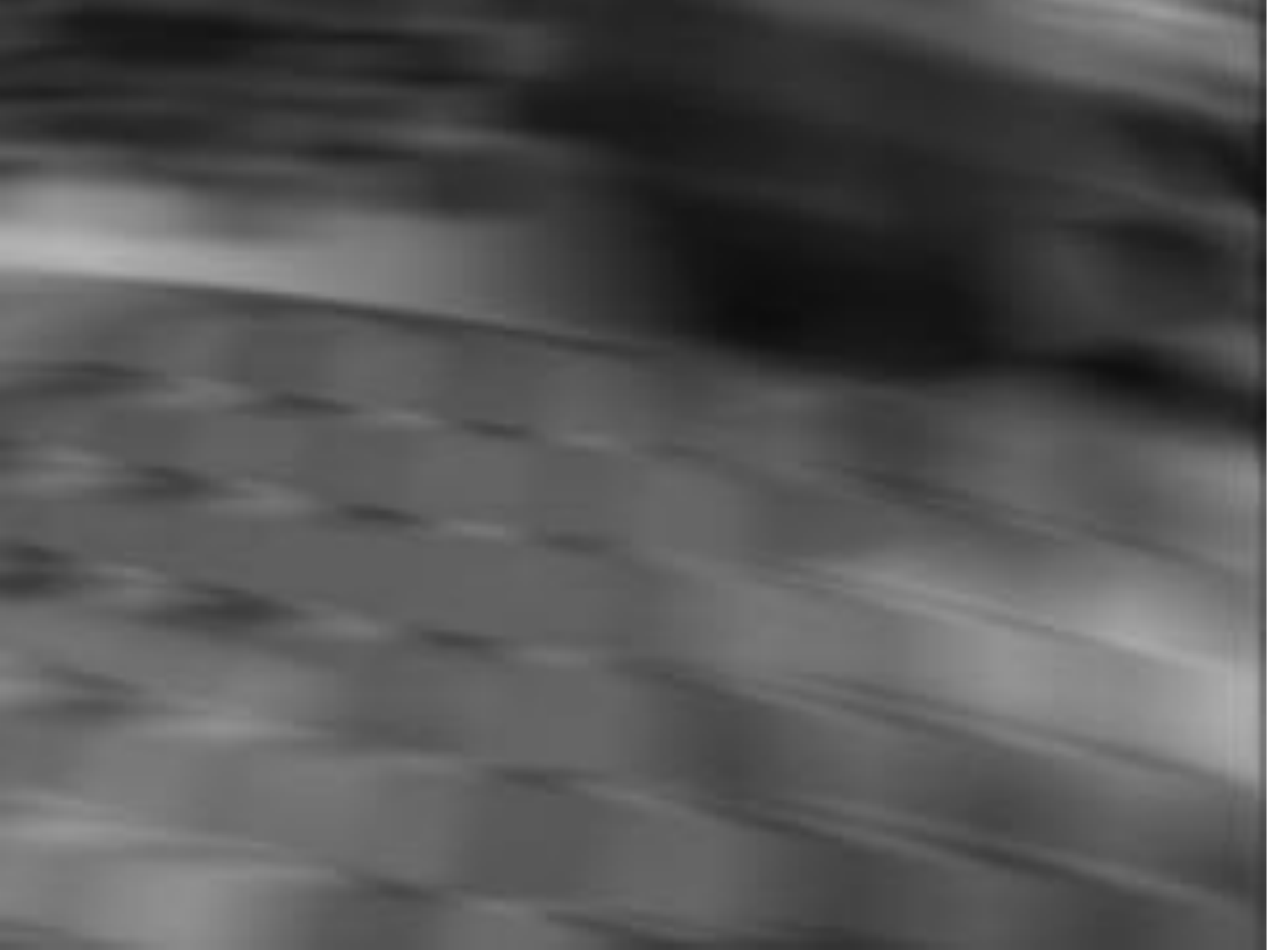}
    \caption{The blurred image}    \label{fig:image_bin_hqf_1}
  \end{subfigure}
  \begin{subfigure}{0.4905\linewidth}
    \includegraphics[width=\textwidth]{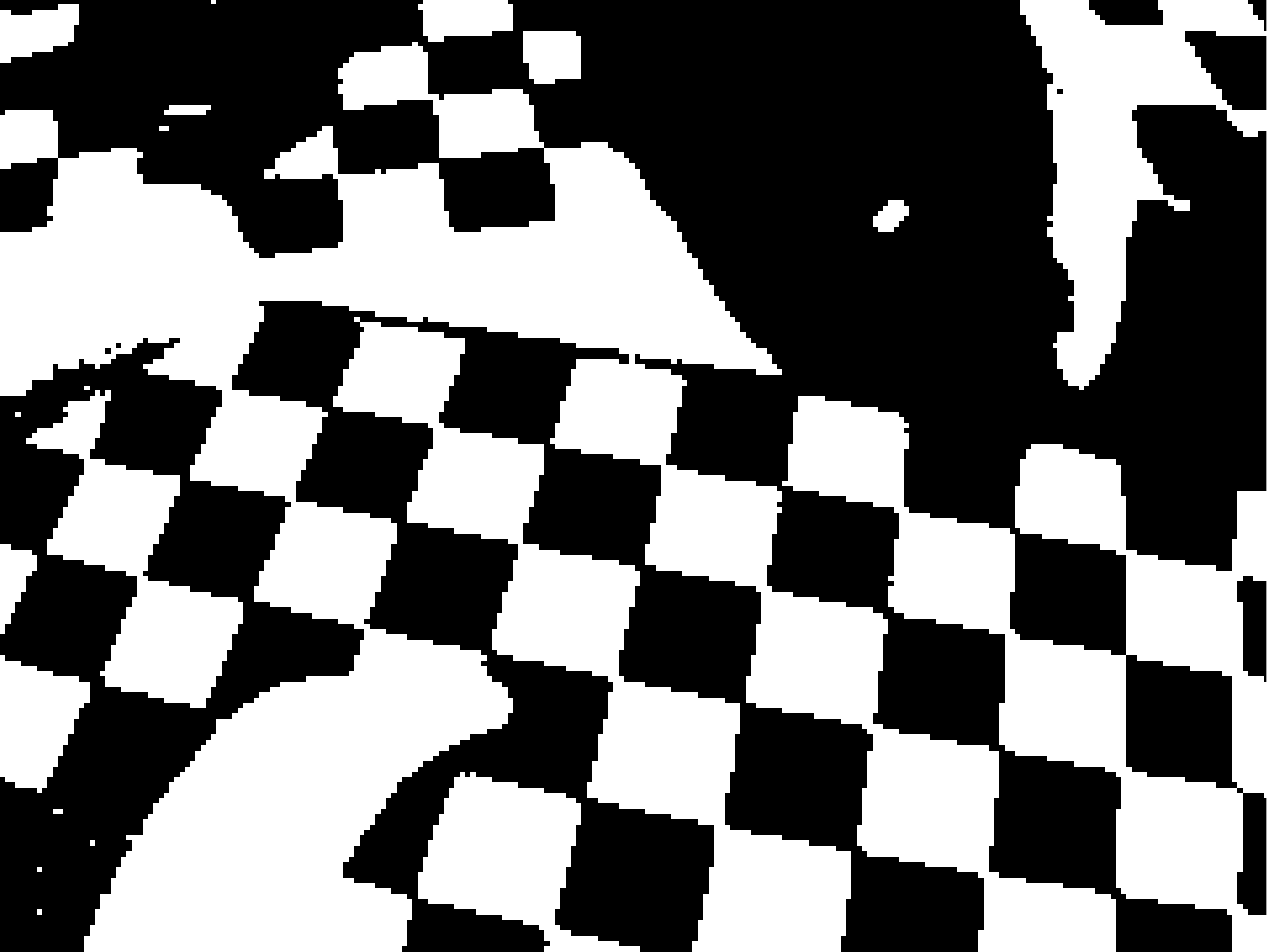}
    \caption{Ground-truth}    \label{fig:image_bin_hqf_2}
  \end{subfigure}
  \begin{subfigure}{0.4905\linewidth}
    \includegraphics[width=\textwidth]{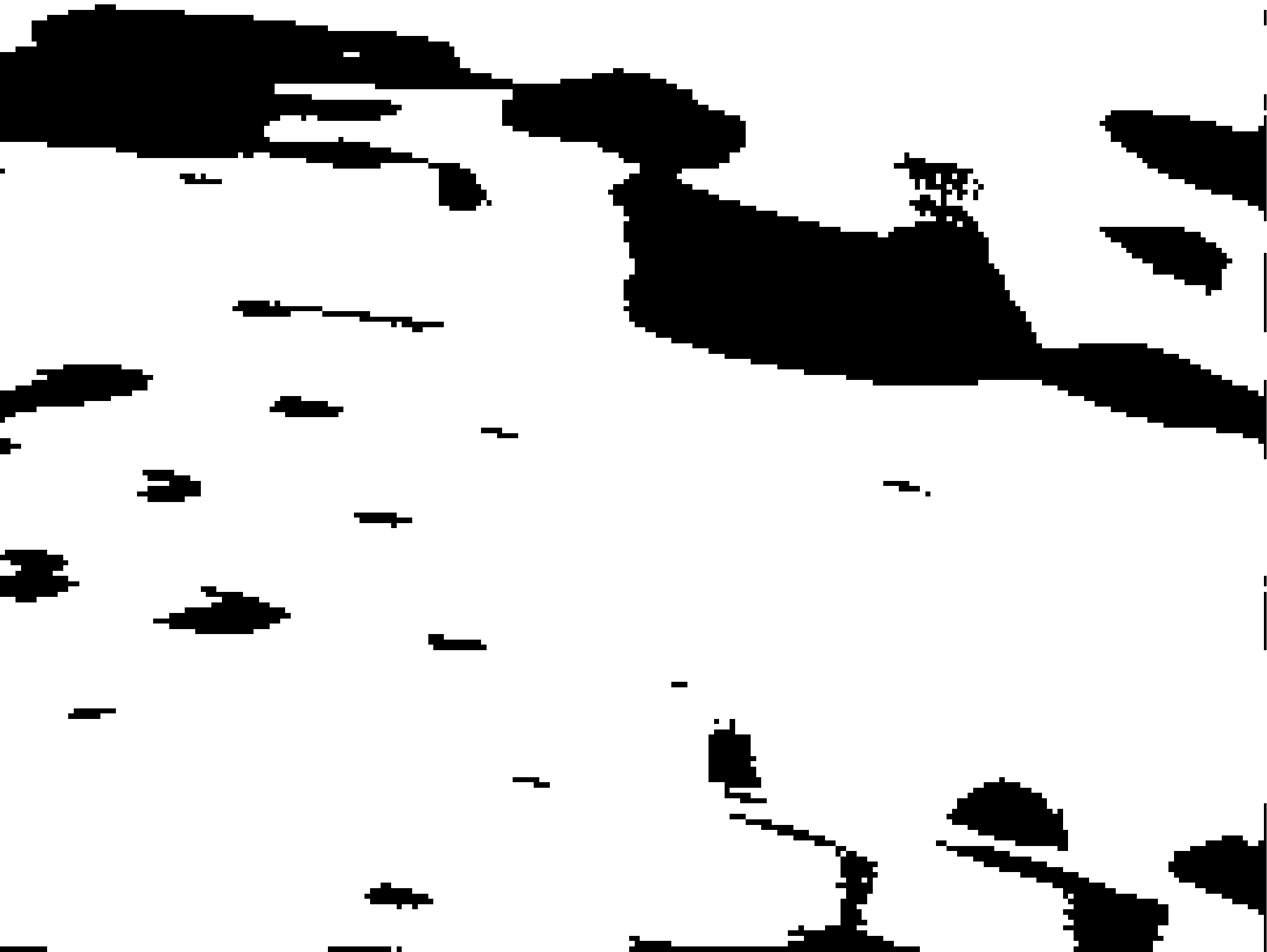}
    \caption{Nick \cite{khurshid2009comparison}}    \label{fig:image_bin_hqf_3}
  \end{subfigure}
  \begin{subfigure}{0.4905\linewidth}
    \includegraphics[width=\textwidth]{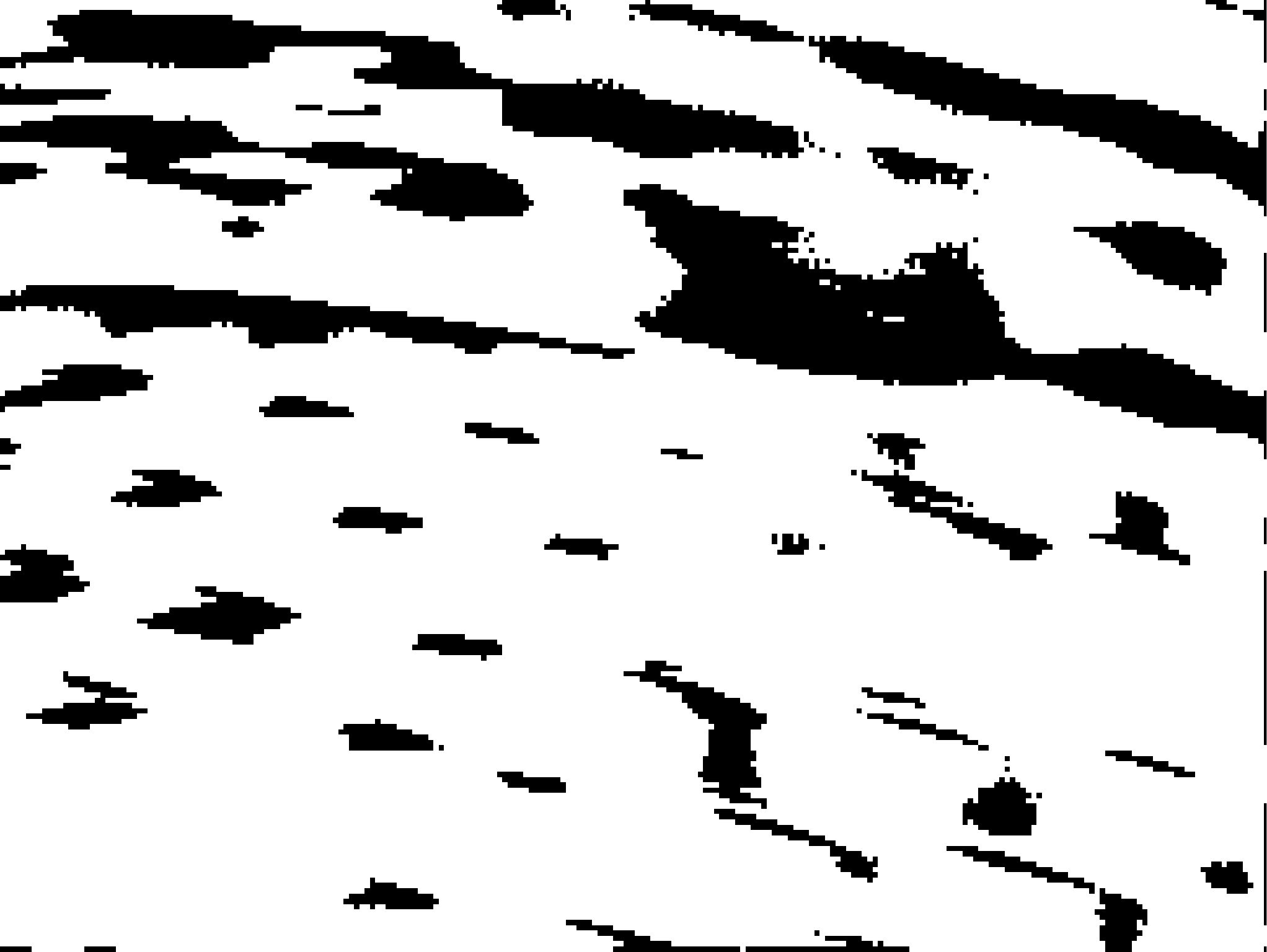}
    \caption{Adaptive \cite{bradley2007adaptive}}    \label{fig:image_bin_hqf_4}
  \end{subfigure}
  \begin{subfigure}{0.4905\linewidth}
    \includegraphics[width=\textwidth]{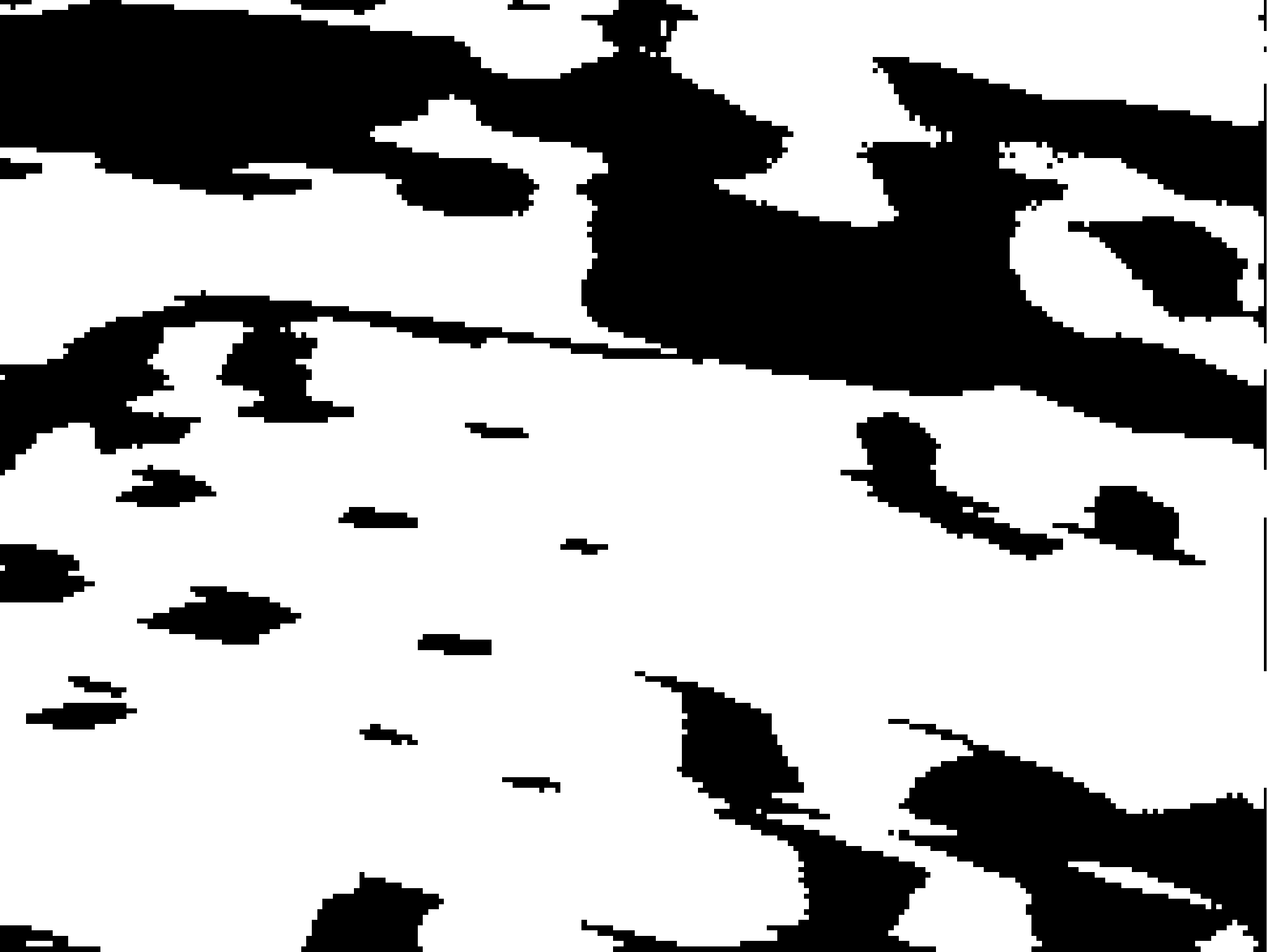}
    \caption{Wolf \cite{wolf2004extraction}}    \label{fig:image_bin_hqf_5}
  \end{subfigure}
  \begin{subfigure}{0.4905\linewidth}
    \includegraphics[width=\textwidth]{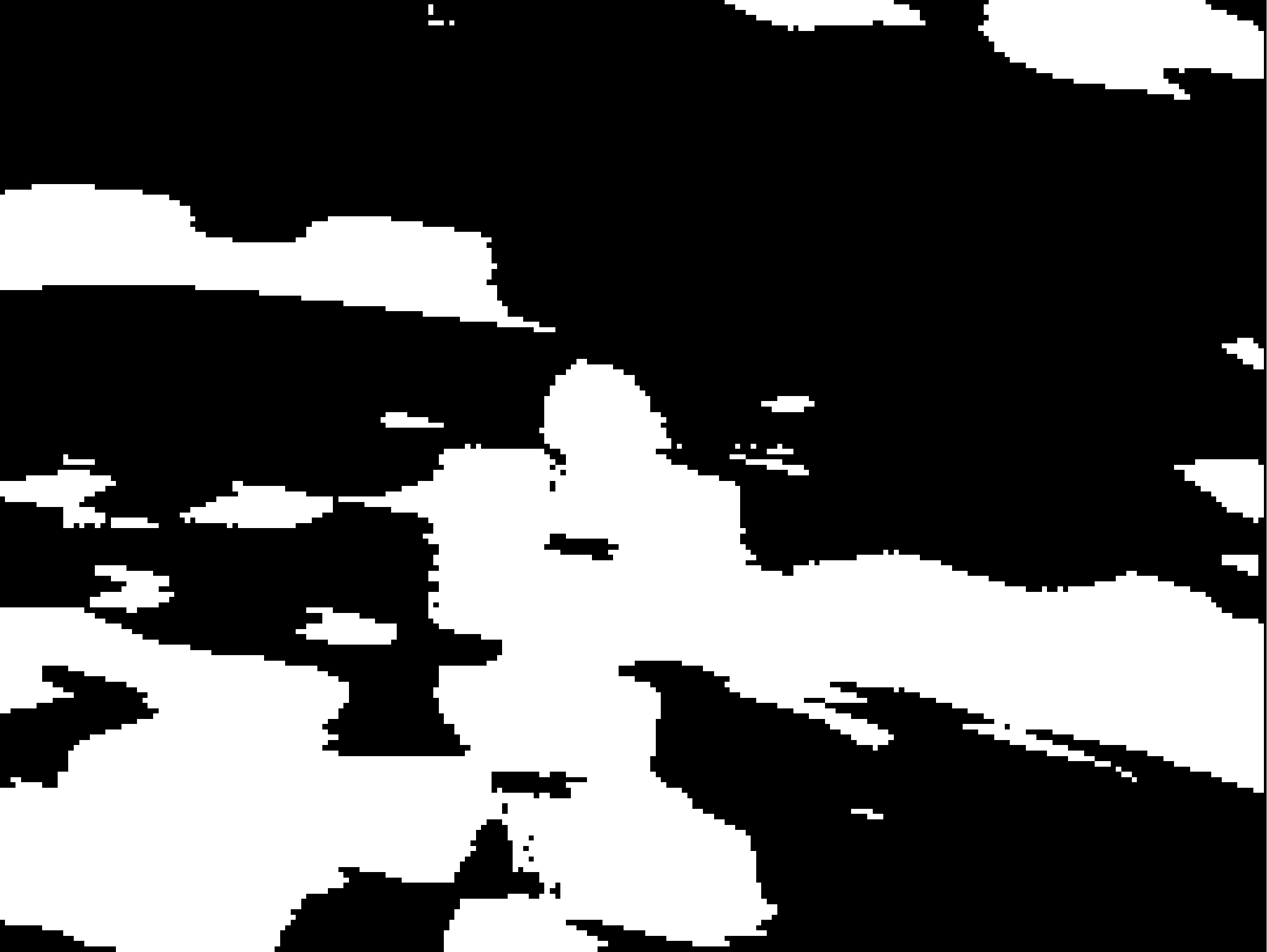}
    \caption{Wan \cite{mustafa2018binarization}}    \label{fig:image_bin_hqf_6}
  \end{subfigure}
    \begin{subfigure}{0.4905\linewidth}
    \includegraphics[width=\textwidth]{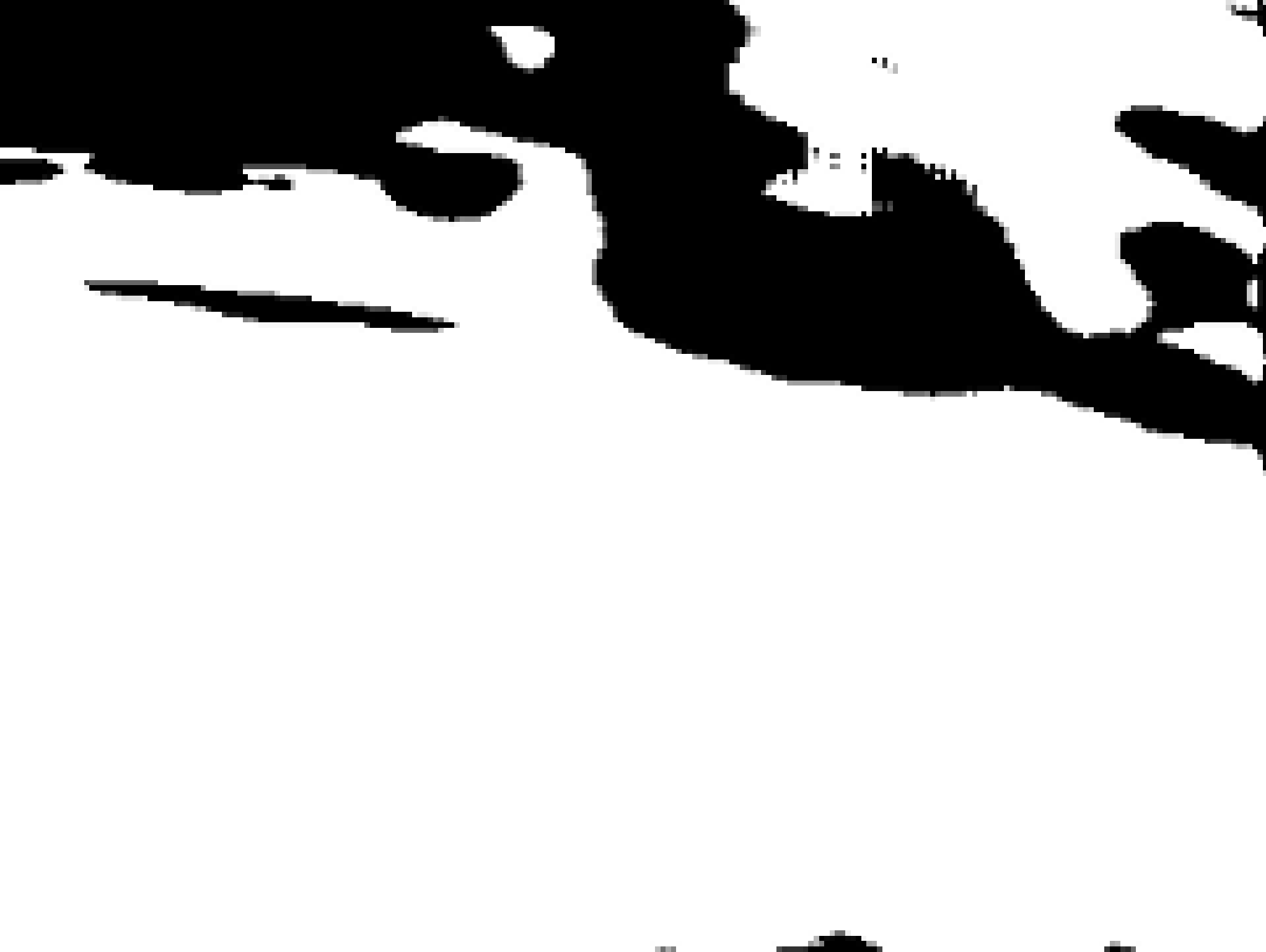}
    \caption{Auto-encoder \cite{calvo2019selectional}}    \label{fig:image_bin_hqf_7}
  \end{subfigure}
  \begin{subfigure}{0.4905\linewidth}
    \includegraphics[width=\textwidth]{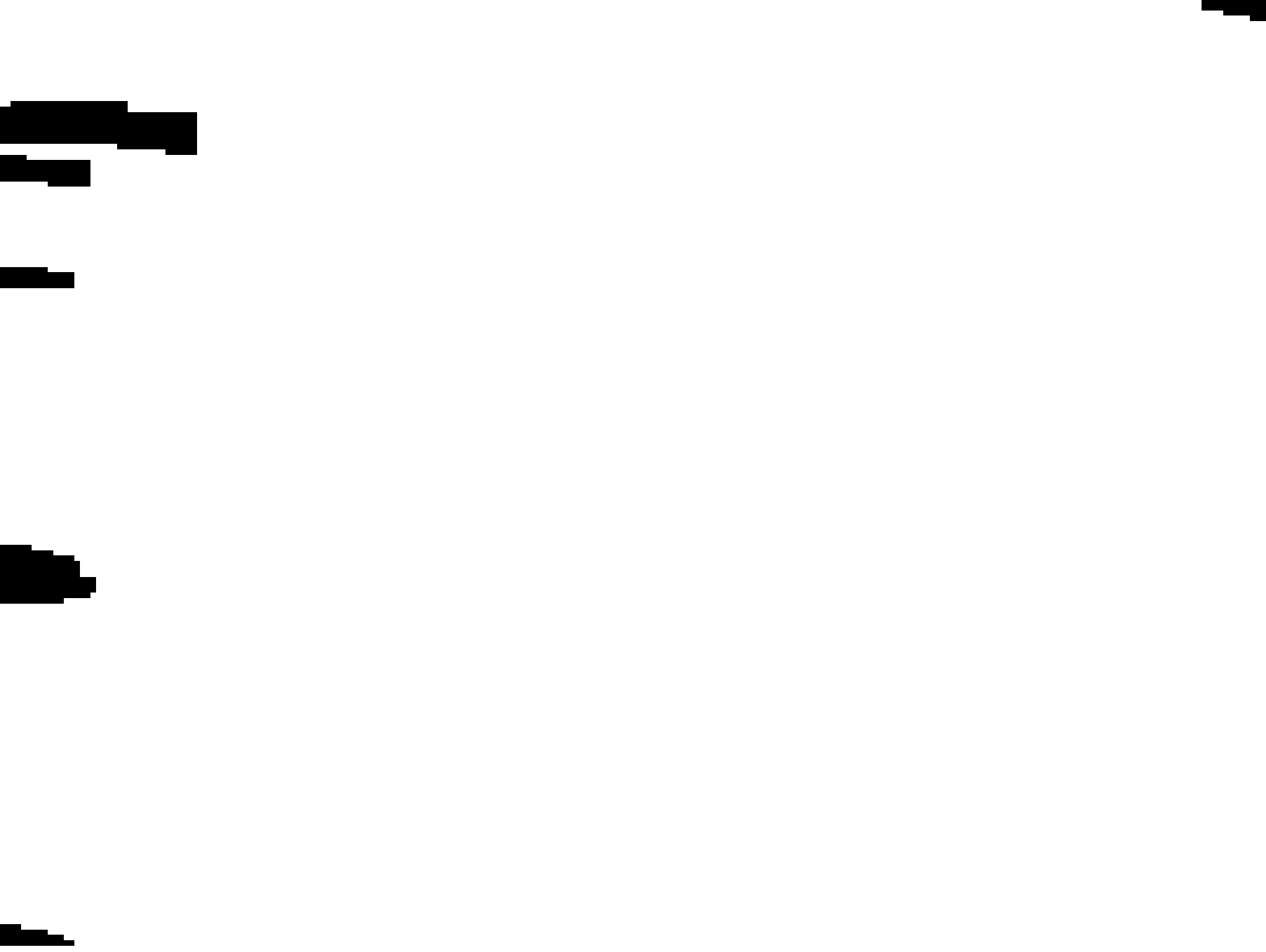}
    \caption{Howe \cite{howe2013document}}    \label{fig:image_bin_hqf_9}
  \end{subfigure}
  \begin{subfigure}{0.4905\linewidth}
    \includegraphics[width=\textwidth]{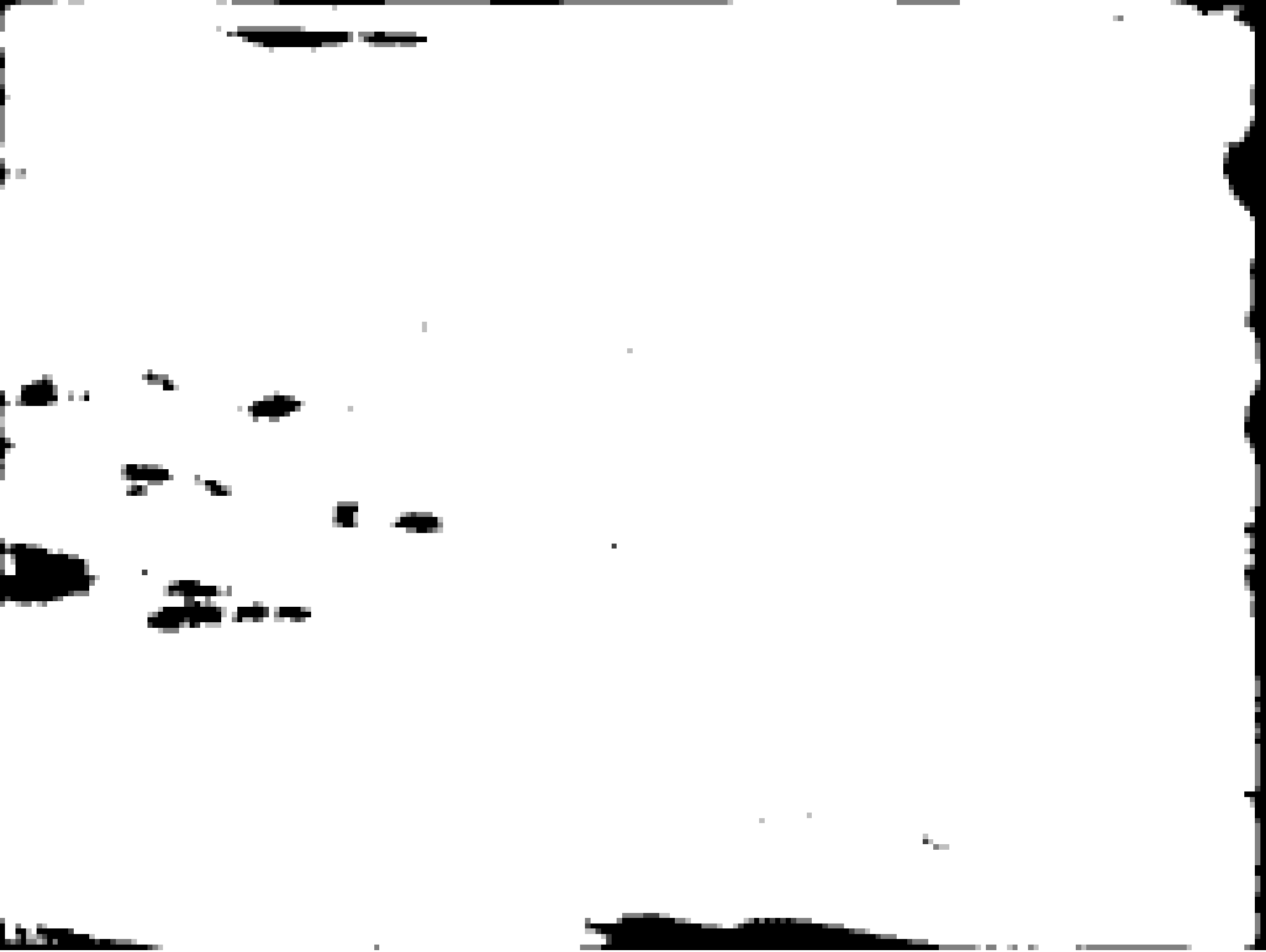}
    \caption{Dplink-Net \cite{xiong2021dp}}    \label{fig:image_bin_hqf_10}
  \end{subfigure}
  \begin{subfigure}{0.4905\linewidth}
    \includegraphics[width=\textwidth]{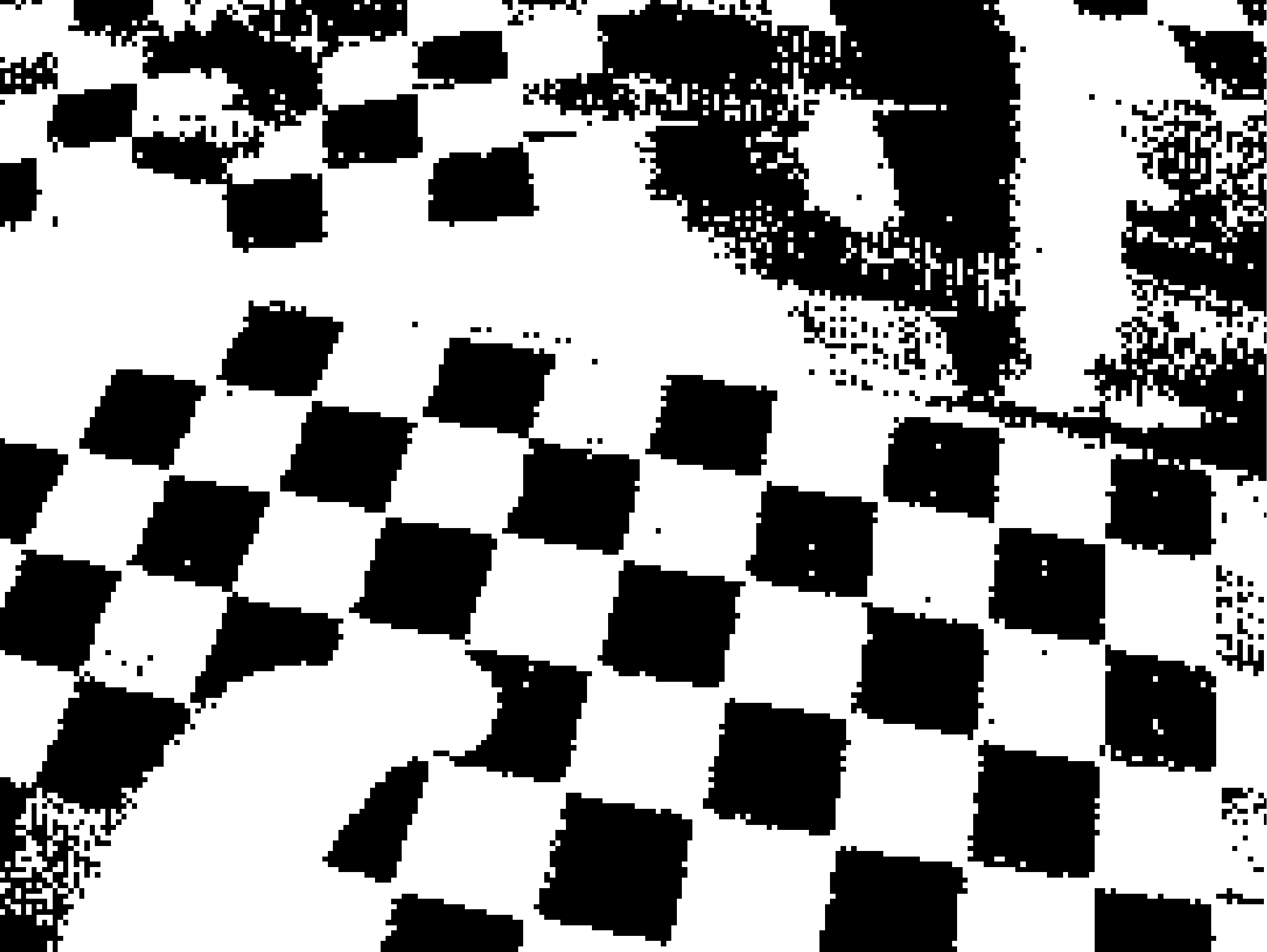}
    \caption{\textbf{Ours}}    \label{fig:image_bin_hqf_8}
  \end{subfigure}
\end{minipage}
\hfill
\begin{minipage}{0.325\linewidth}\centering
  \begin{subfigure}{0.4905\linewidth}
    \includegraphics[width=\textwidth]{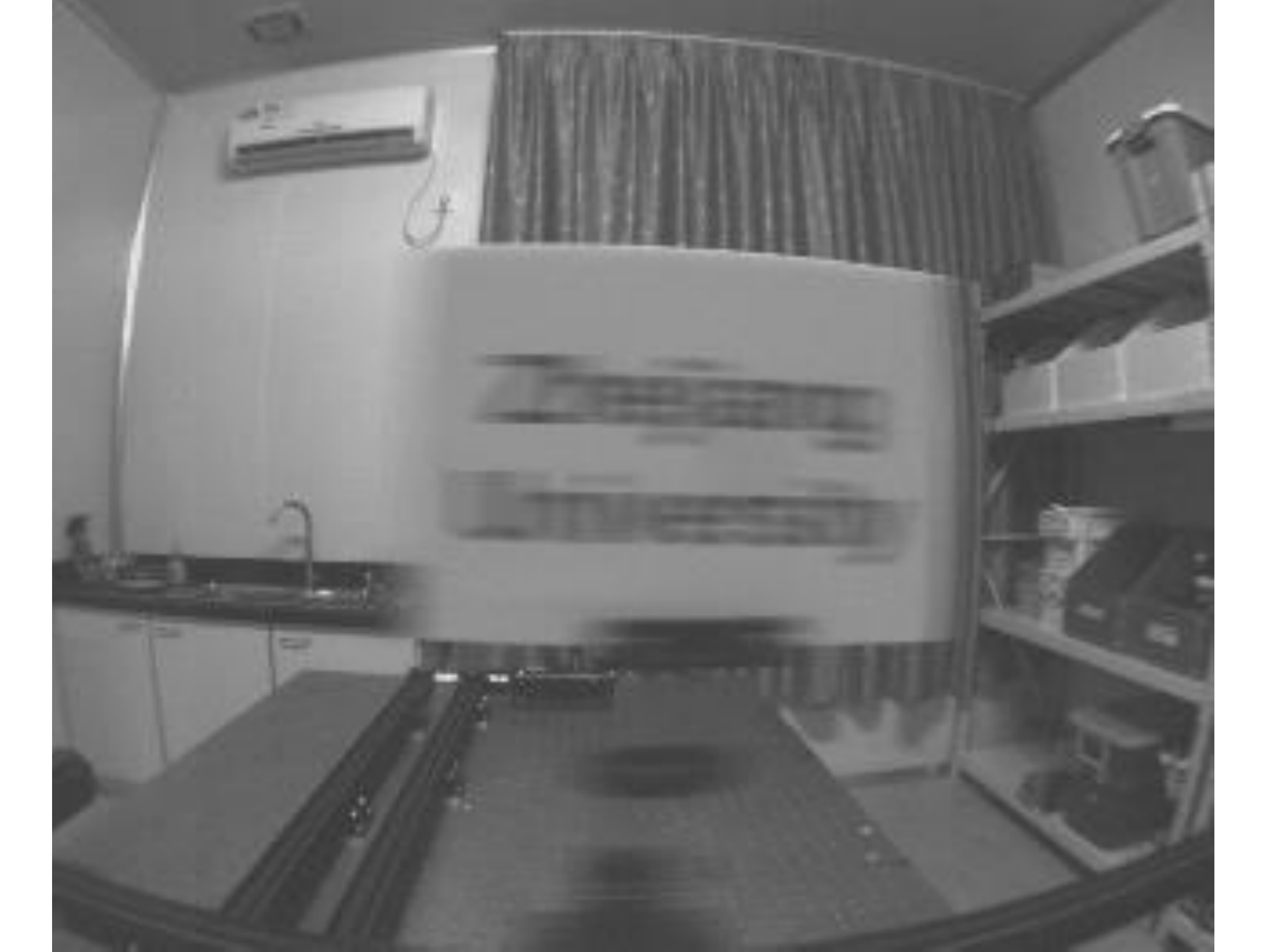}
    \caption{The blurred image}    \label{fig:}
  \end{subfigure}
  \begin{subfigure}{0.4905\linewidth}
    \includegraphics[width=\textwidth]{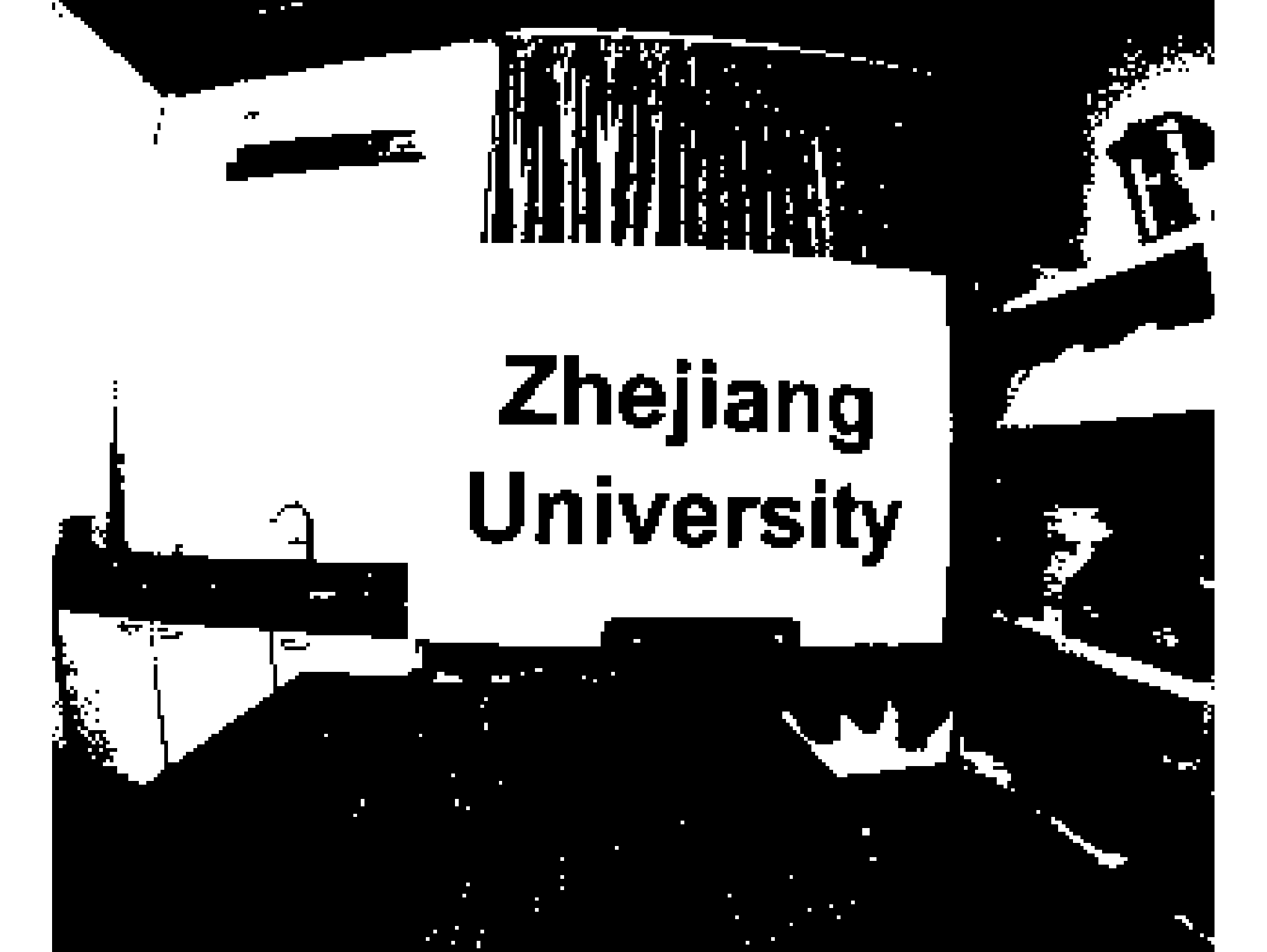}
    \caption{Ground-truth}    \label{fig:}
  \end{subfigure}
  \begin{subfigure}{0.4905\linewidth}
    \includegraphics[width=\textwidth]{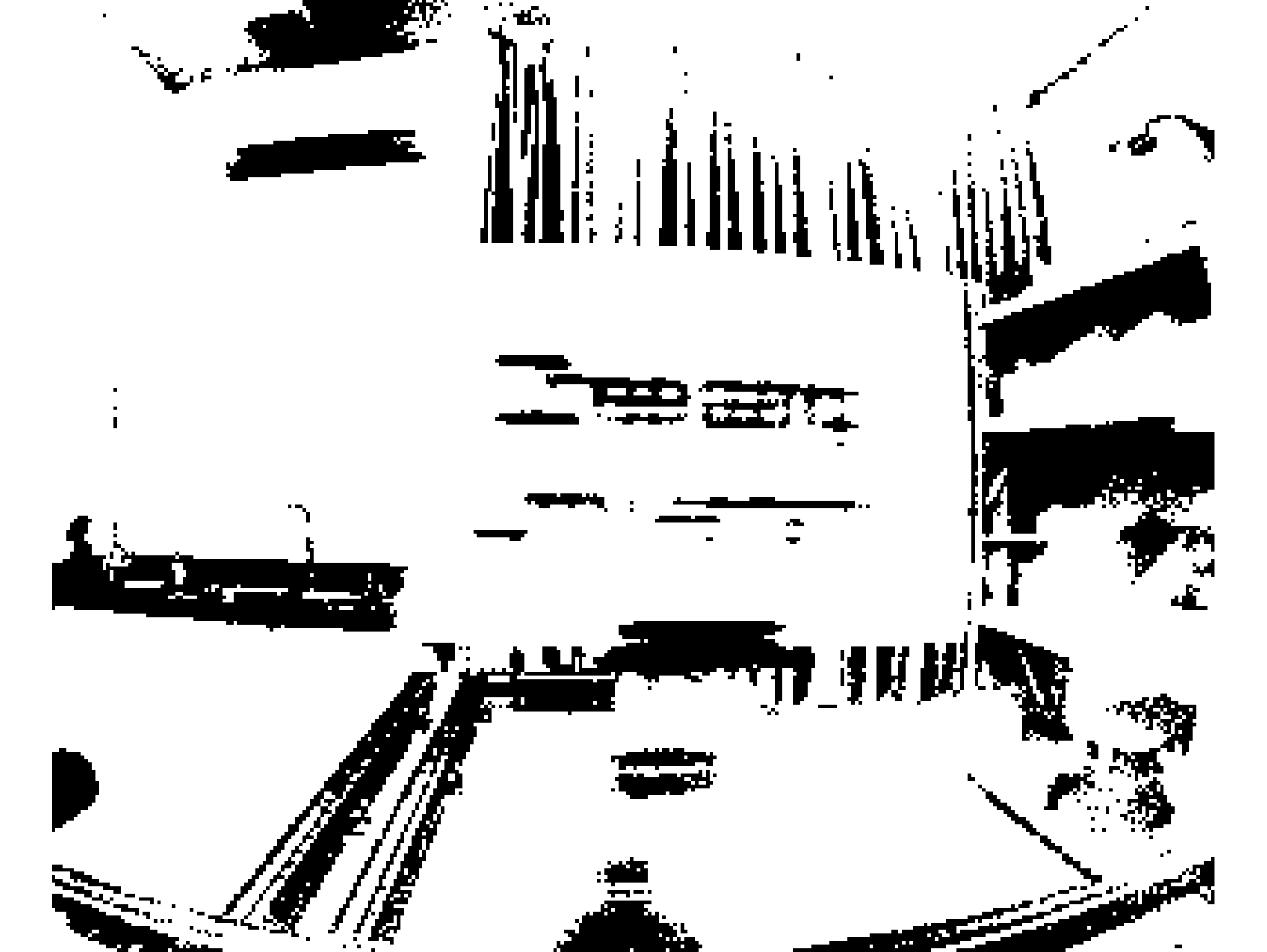}
    \caption{Nick \cite{khurshid2009comparison}}    \label{fig:}
  \end{subfigure}
  \begin{subfigure}{0.4905\linewidth}
    \includegraphics[width=\textwidth]{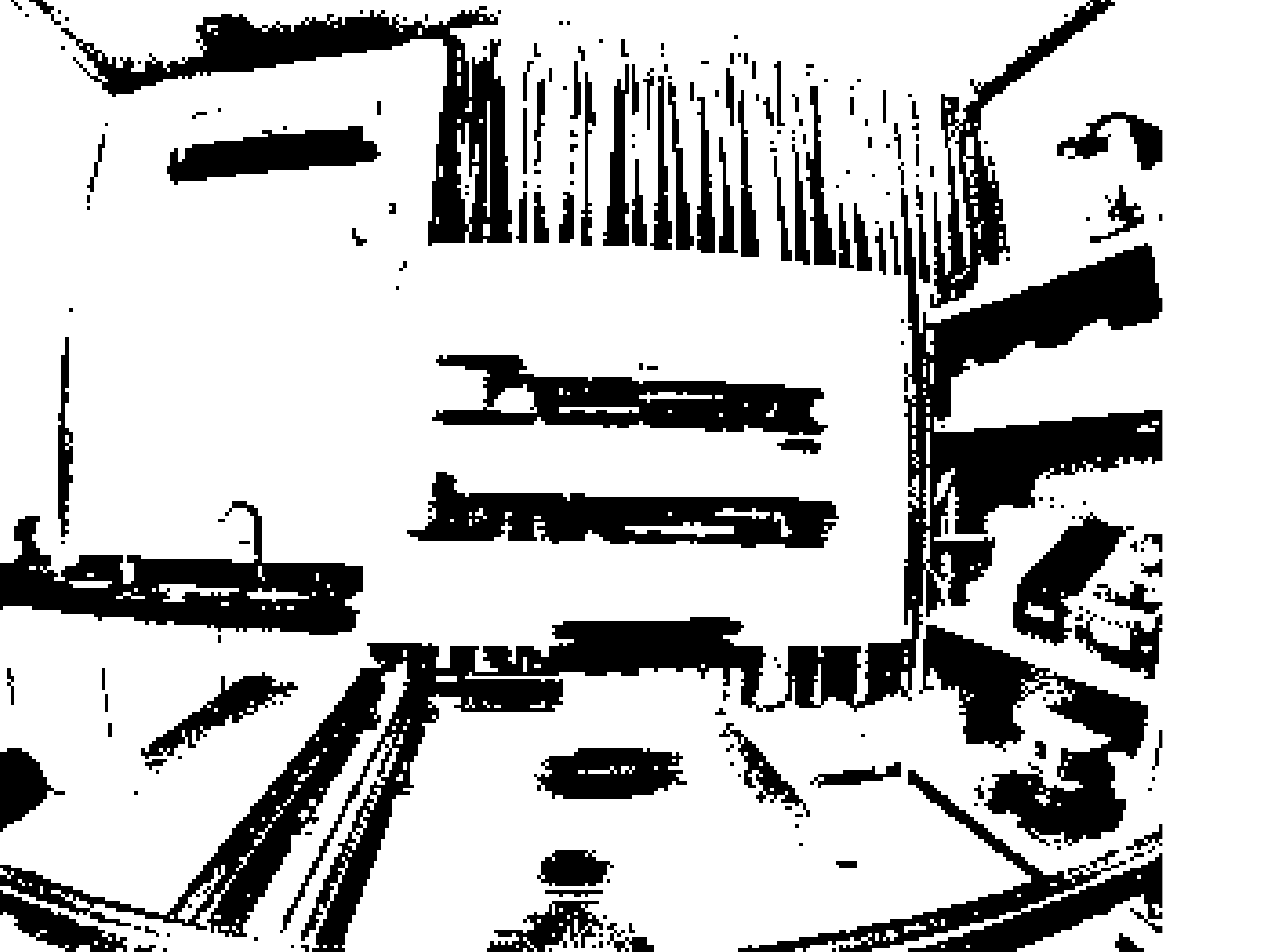}
    \caption{Adaptive \cite{bradley2007adaptive}}    \label{fig:}
  \end{subfigure}
  \begin{subfigure}{0.4905\linewidth}
    \includegraphics[width=\textwidth]{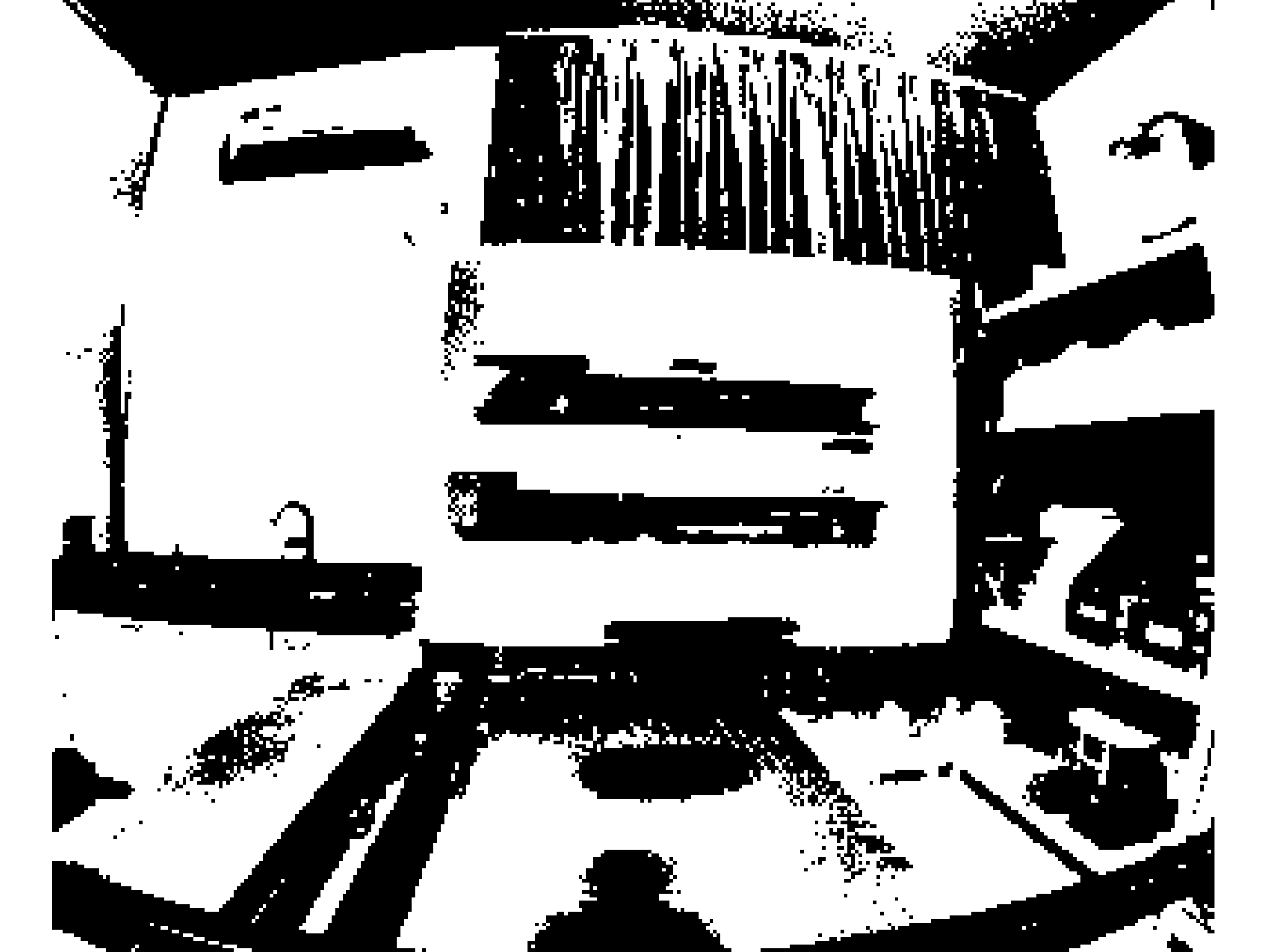}
    \caption{Wolf \cite{wolf2004extraction}}    \label{fig:}
  \end{subfigure}
  \begin{subfigure}{0.4905\linewidth}
    \includegraphics[width=\textwidth]{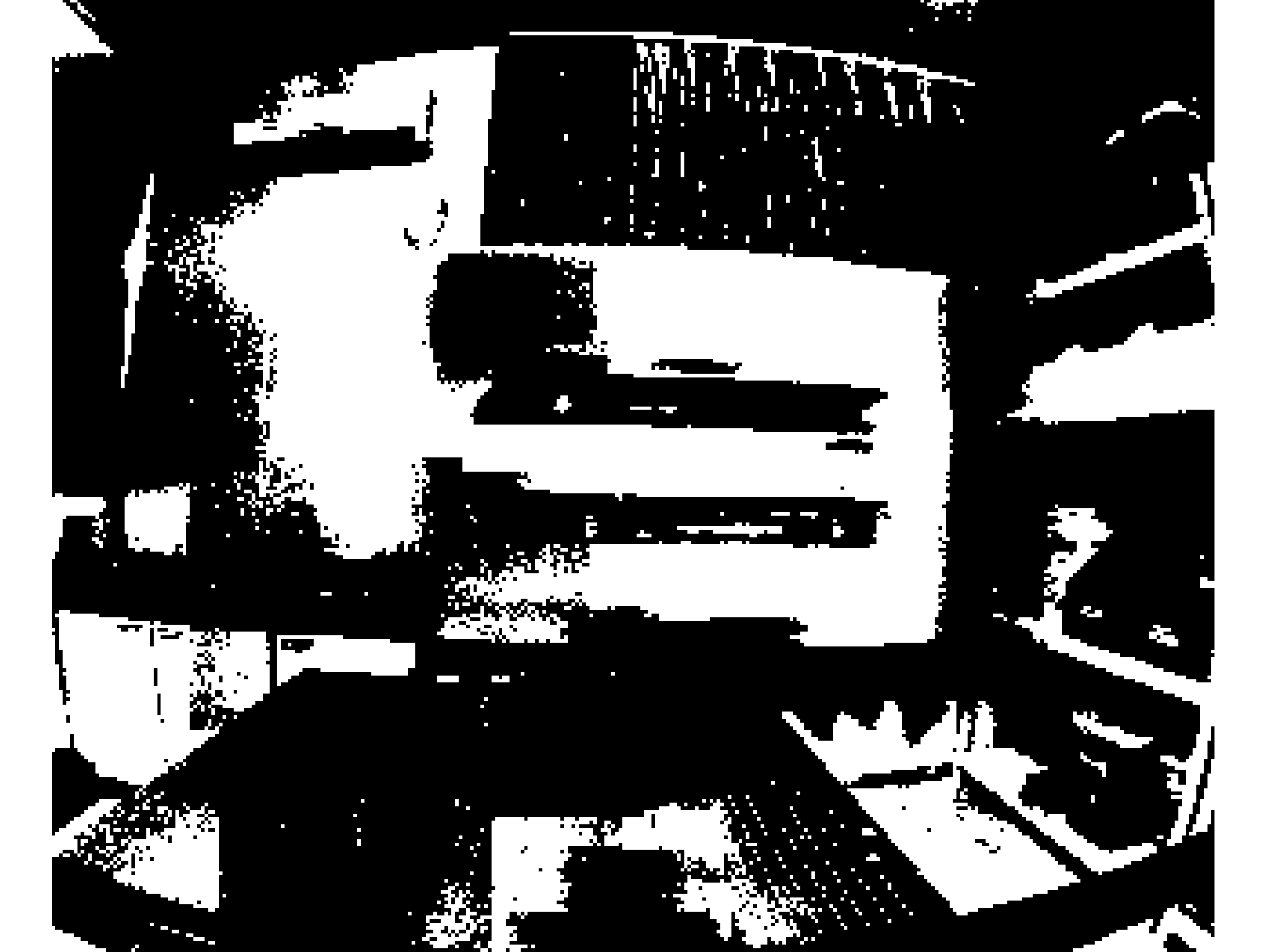}
    \caption{Wan \cite{mustafa2018binarization}}    \label{fig:}
  \end{subfigure}
   \begin{subfigure}{0.4905\linewidth}
    \includegraphics[width=\textwidth]{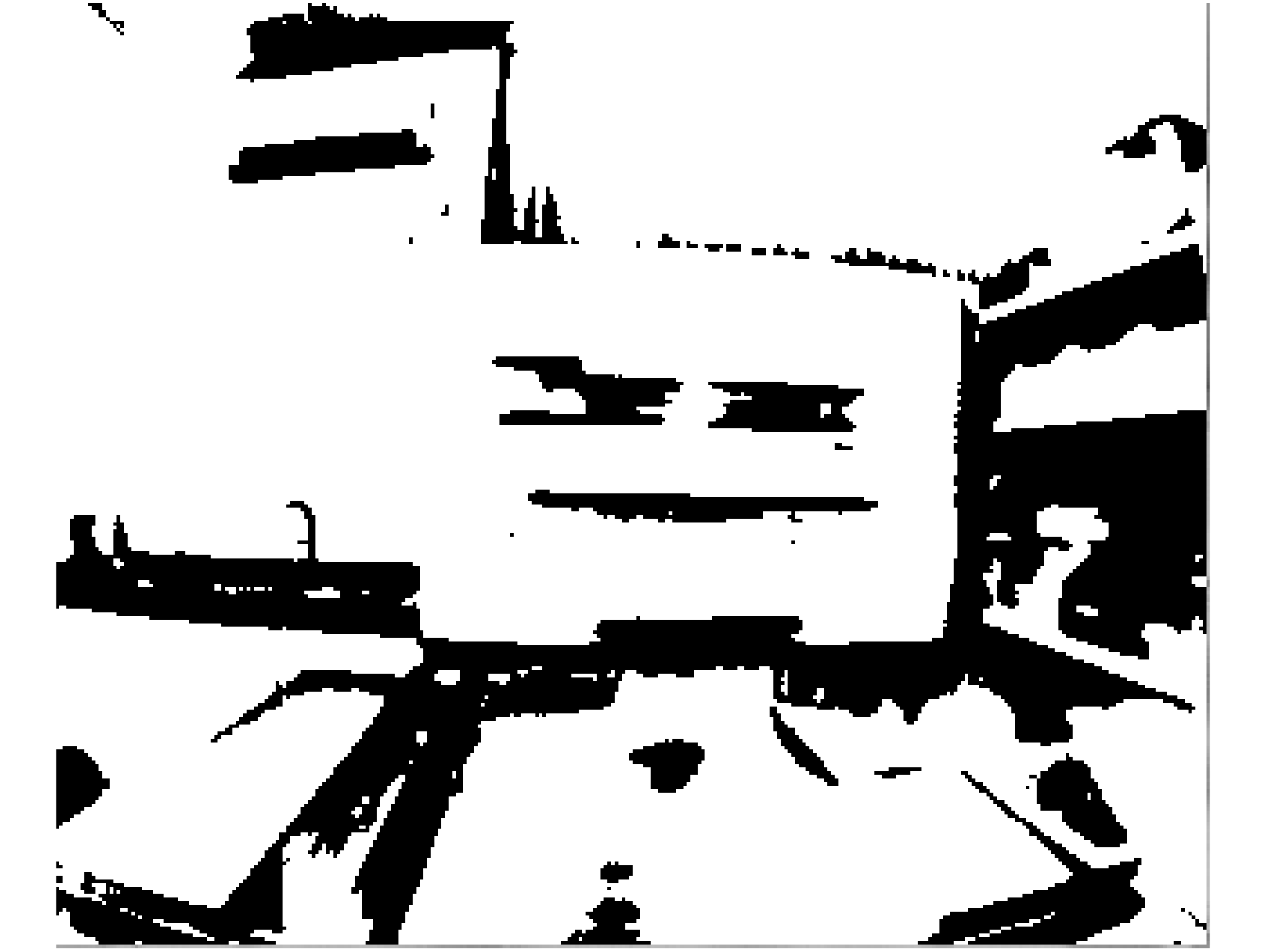}
    \caption{Auto-encoder \cite{calvo2019selectional}}    \label{fig:}
  \end{subfigure}
  \begin{subfigure}{0.4905\linewidth}
    \includegraphics[width=\textwidth]{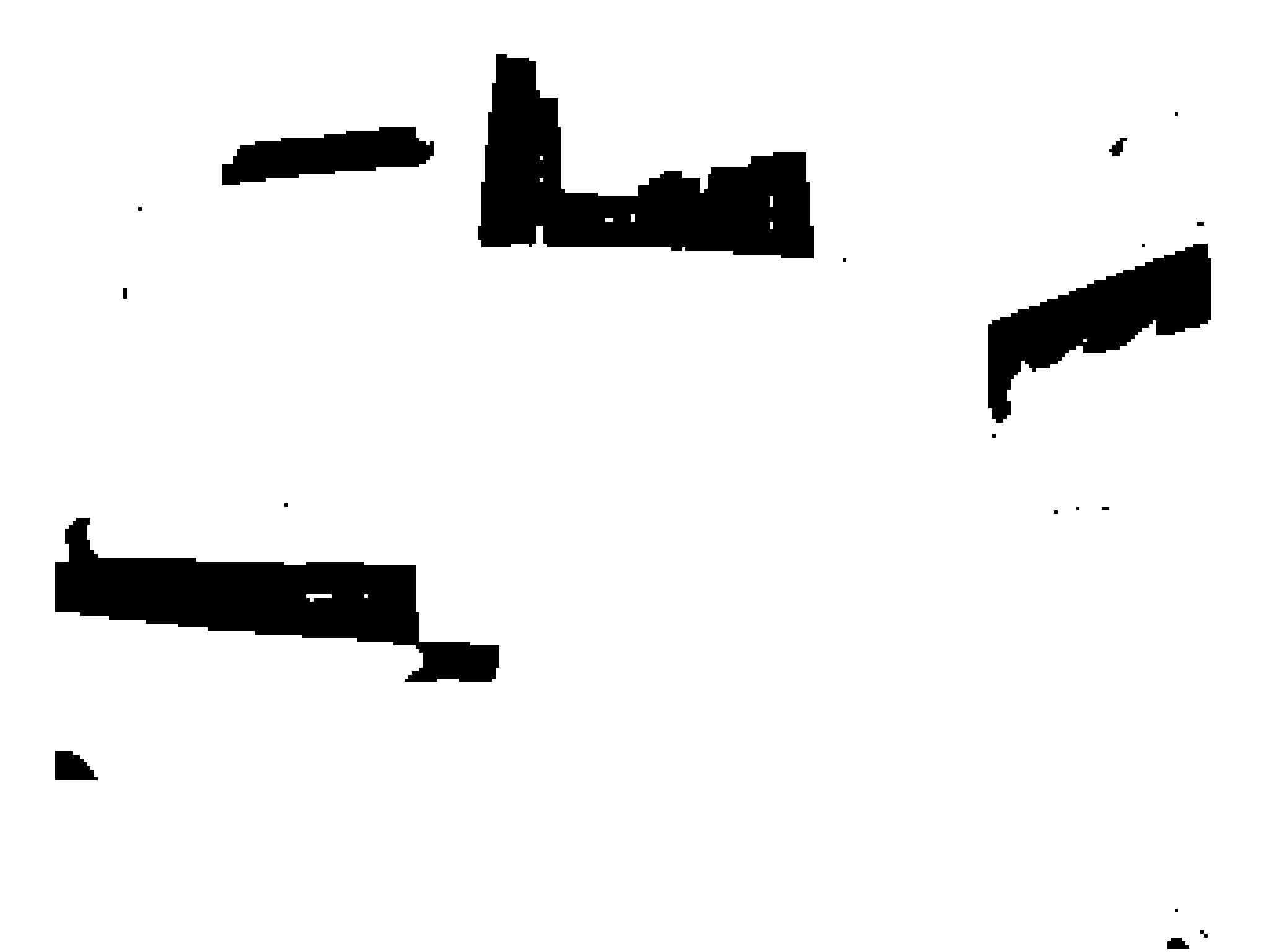}
    \caption{Howe \cite{howe2013document}}    \label{fig:image_bin_reblur_8}
  \end{subfigure}
   \begin{subfigure}{0.4905\linewidth}
    \includegraphics[width=\textwidth]{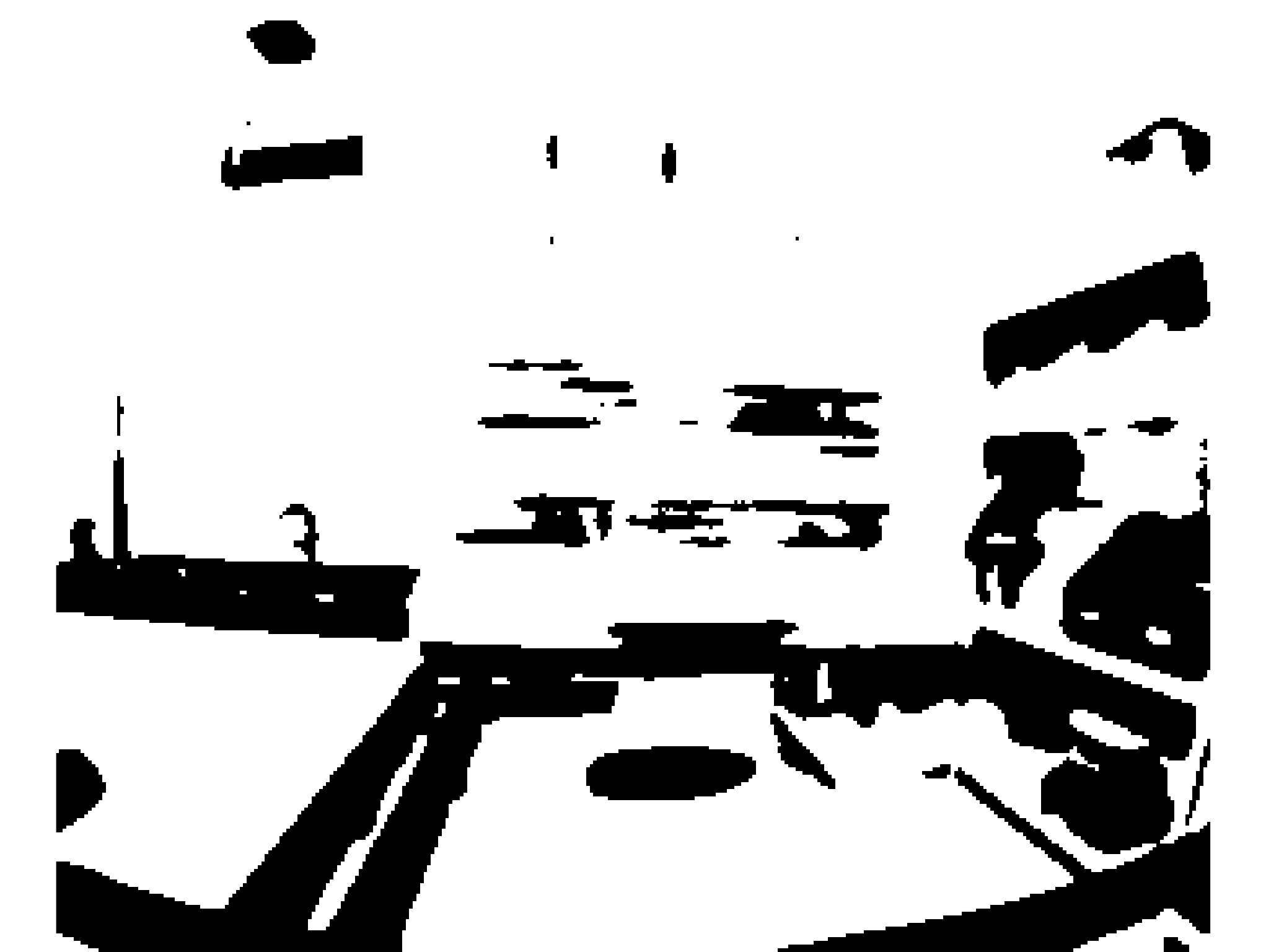}
    \caption{Dplink-Net \cite{xiong2021dp}}    \label{fig:image_bin_reblur_9}
  \end{subfigure}
  \begin{subfigure}{0.4905\linewidth}
    \includegraphics[width=\textwidth]{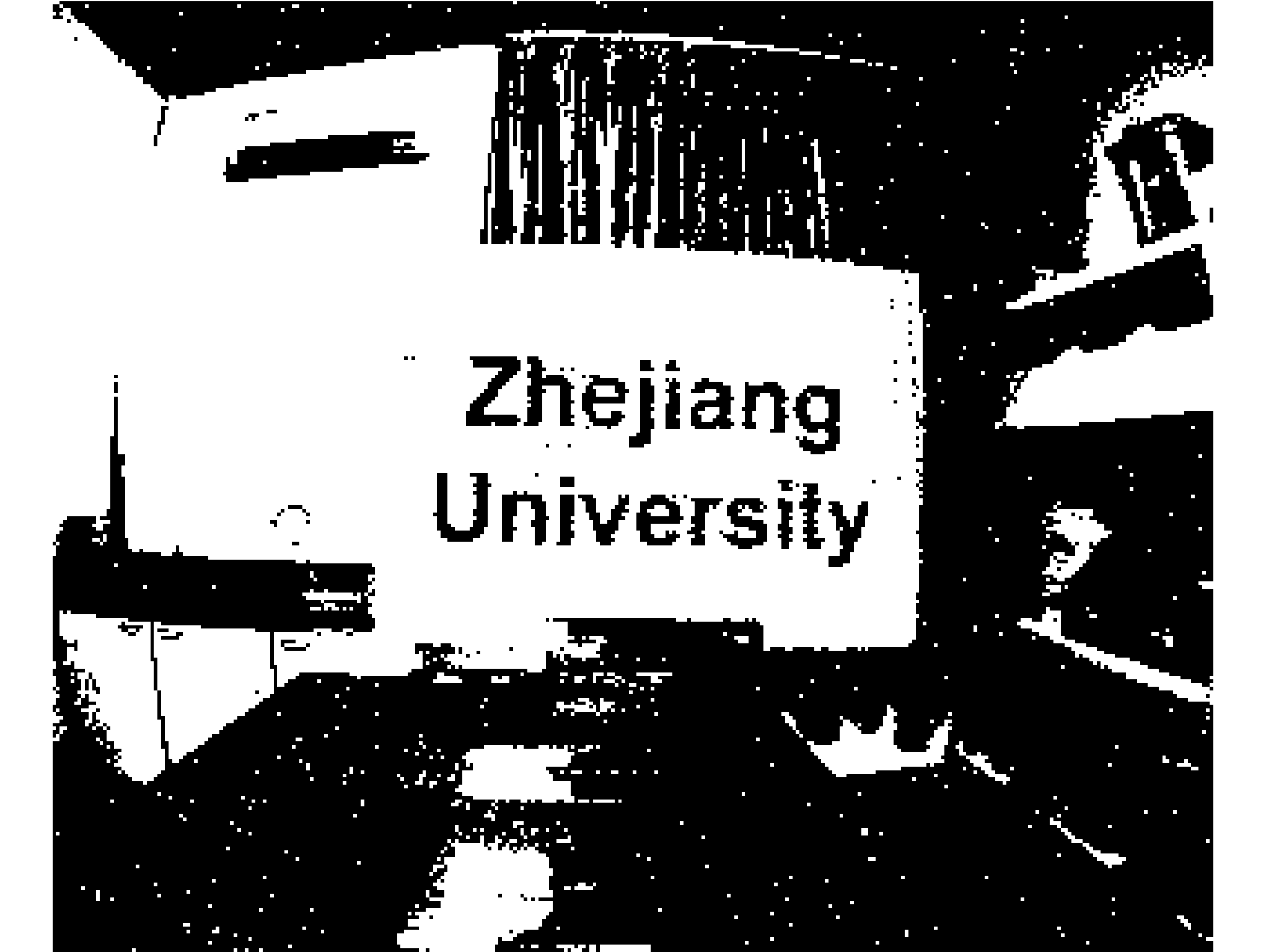}
    \caption{\textbf{Ours}}    \label{fig:}
  \end{subfigure}
\end{minipage}
\hfill
\hfill\begin{minipage}{0.325\linewidth}\centering
  \begin{subfigure}{0.4905\linewidth}
    \includegraphics[width=\textwidth]{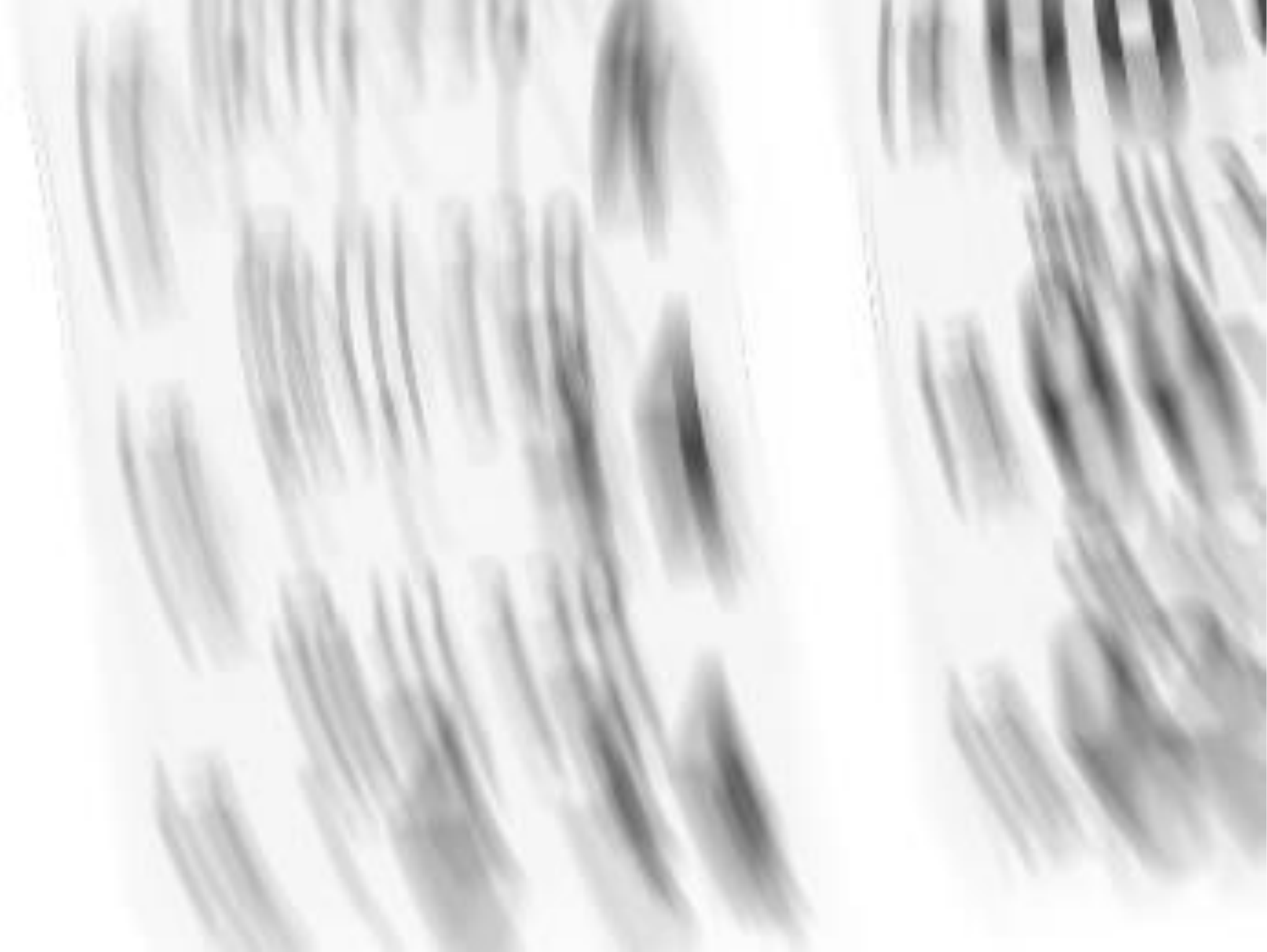}
    \caption{The blurred image}    \label{fig:image_bin_ebt_1}
  \end{subfigure}
  \begin{subfigure}{0.4905\linewidth}
    \includegraphics[width=\textwidth]{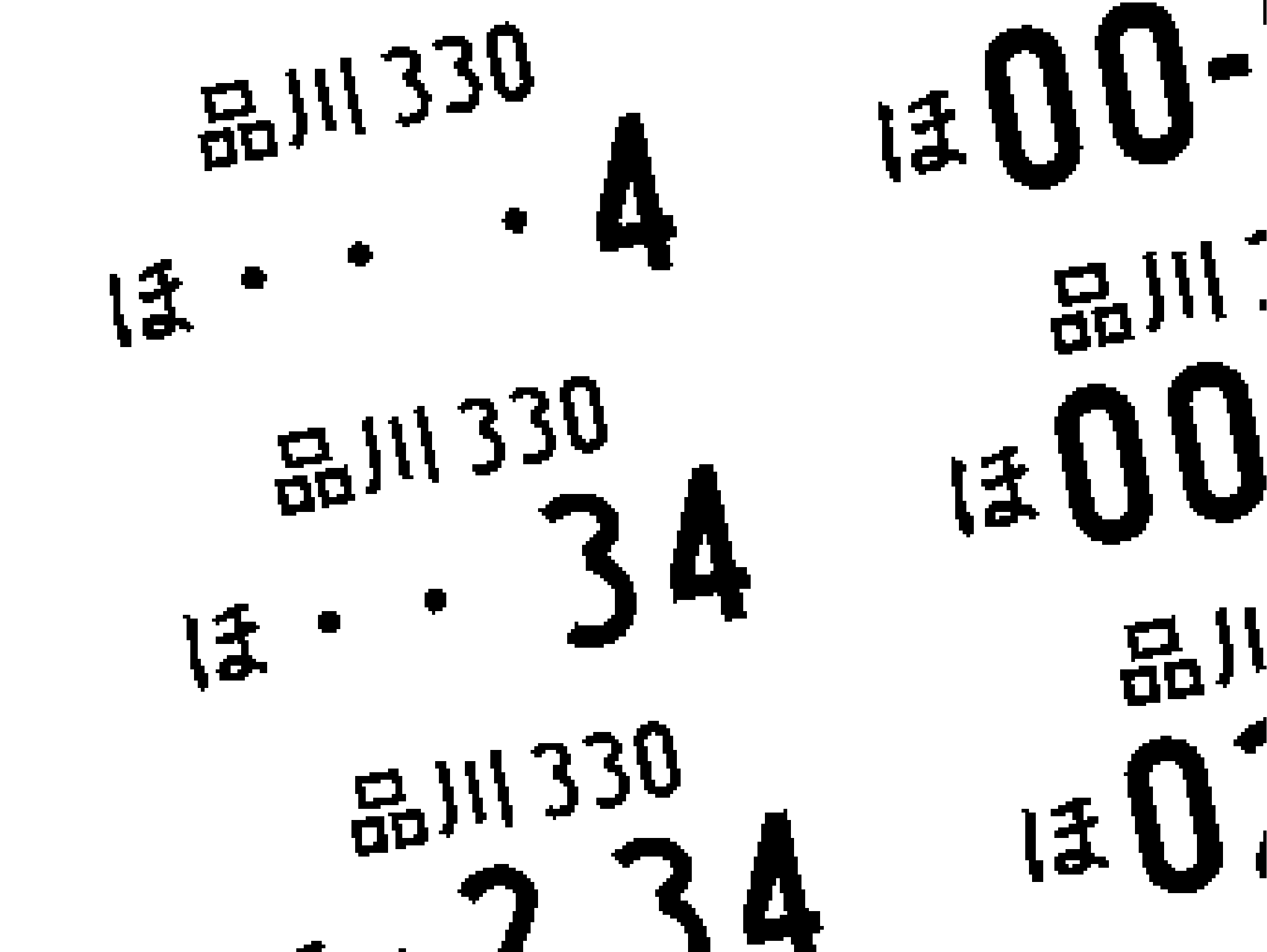}
    \caption{Ground-truth}    \label{fig:image_bin_ebt_2}
  \end{subfigure}
  \begin{subfigure}{0.4905\linewidth}
    \includegraphics[width=\textwidth]{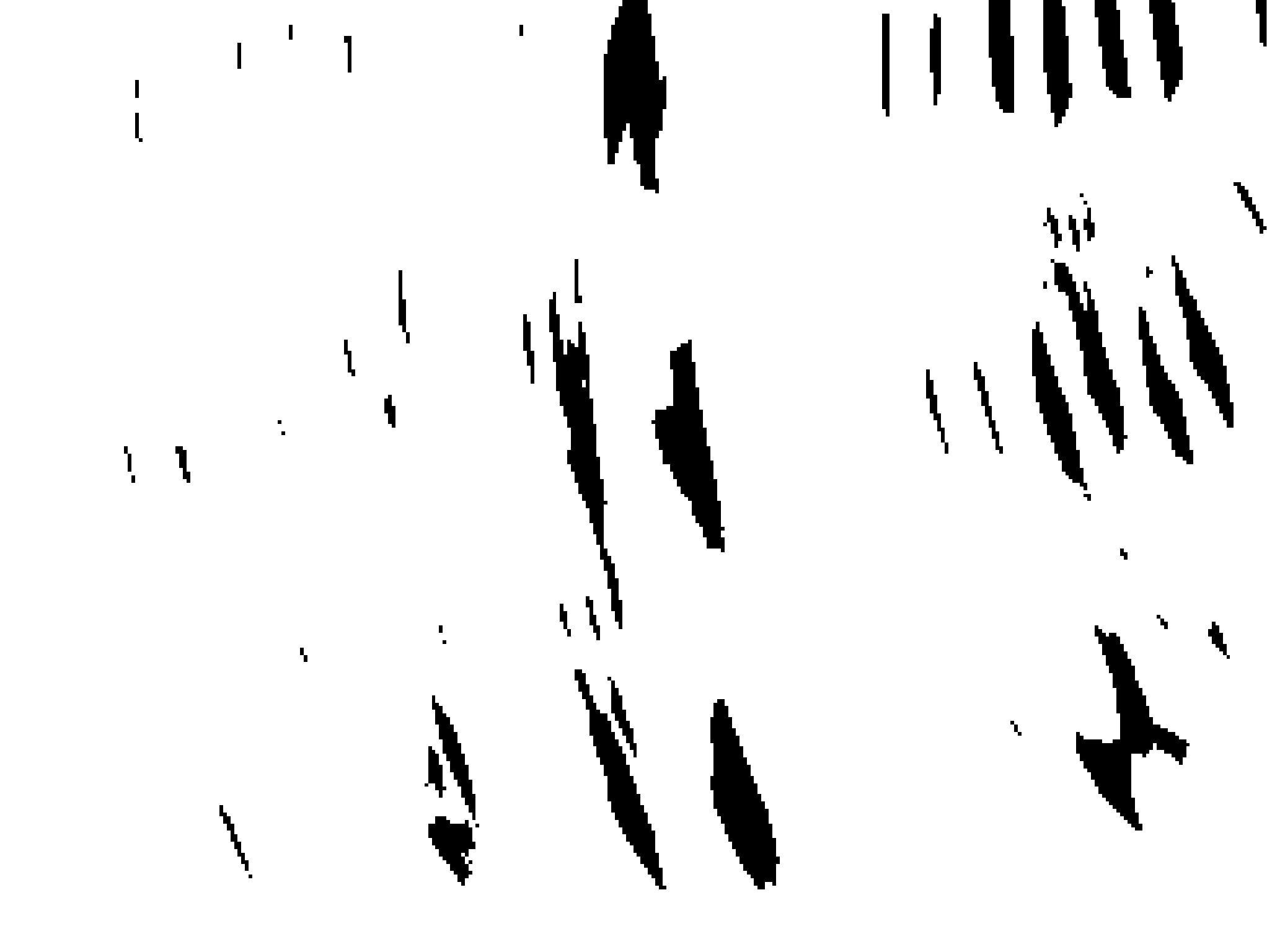}
    \caption{Nick \cite{khurshid2009comparison}}    \label{fig:image_bin_ebt_3}
  \end{subfigure}
  \begin{subfigure}{0.4905\linewidth}
    \includegraphics[width=\textwidth]{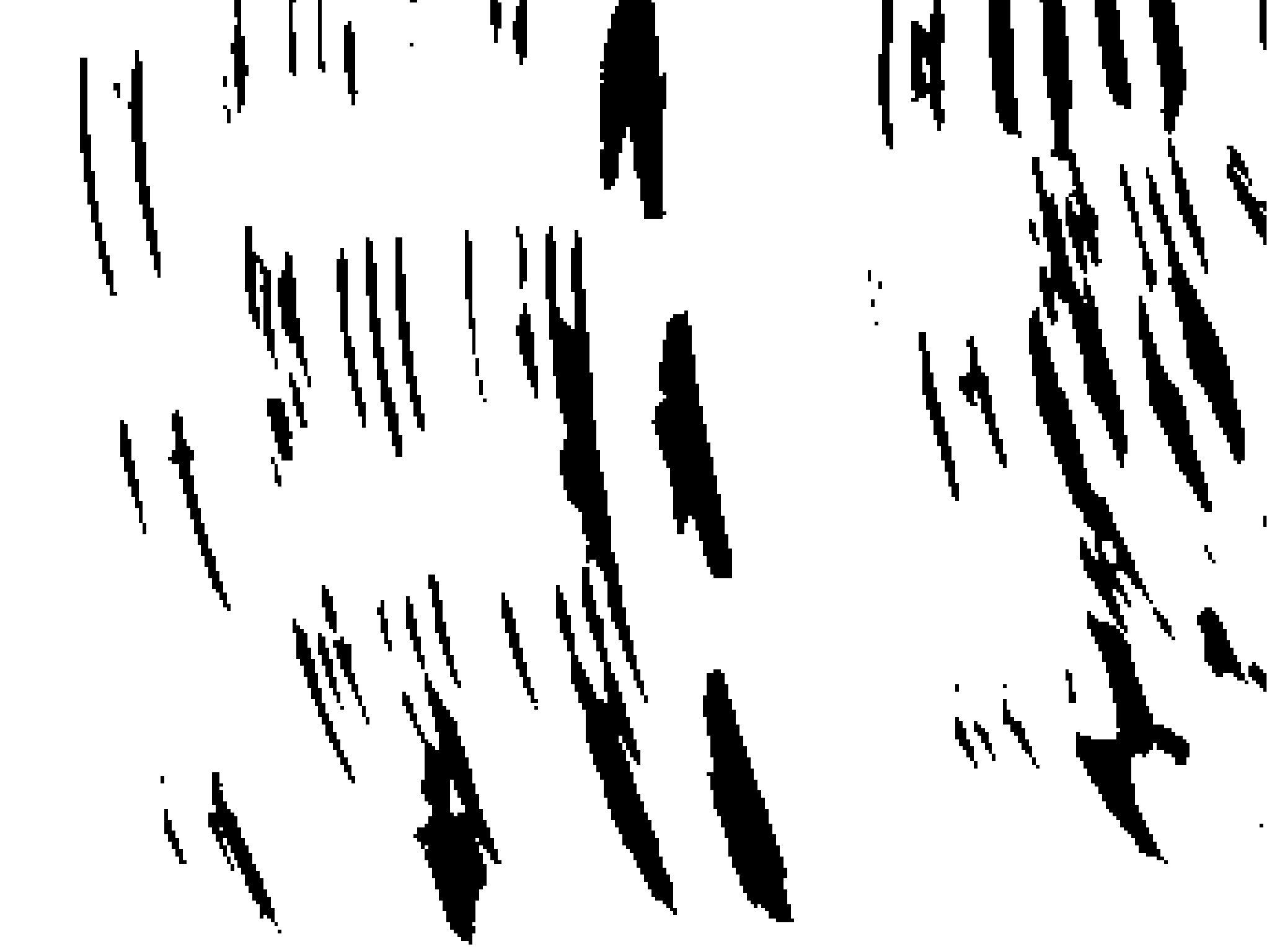}
    \caption{Adaptive \cite{bradley2007adaptive}}    \label{fig:image_bin_ebt_4}
  \end{subfigure}
  \begin{subfigure}{0.4905\linewidth}
    \includegraphics[width=\textwidth]{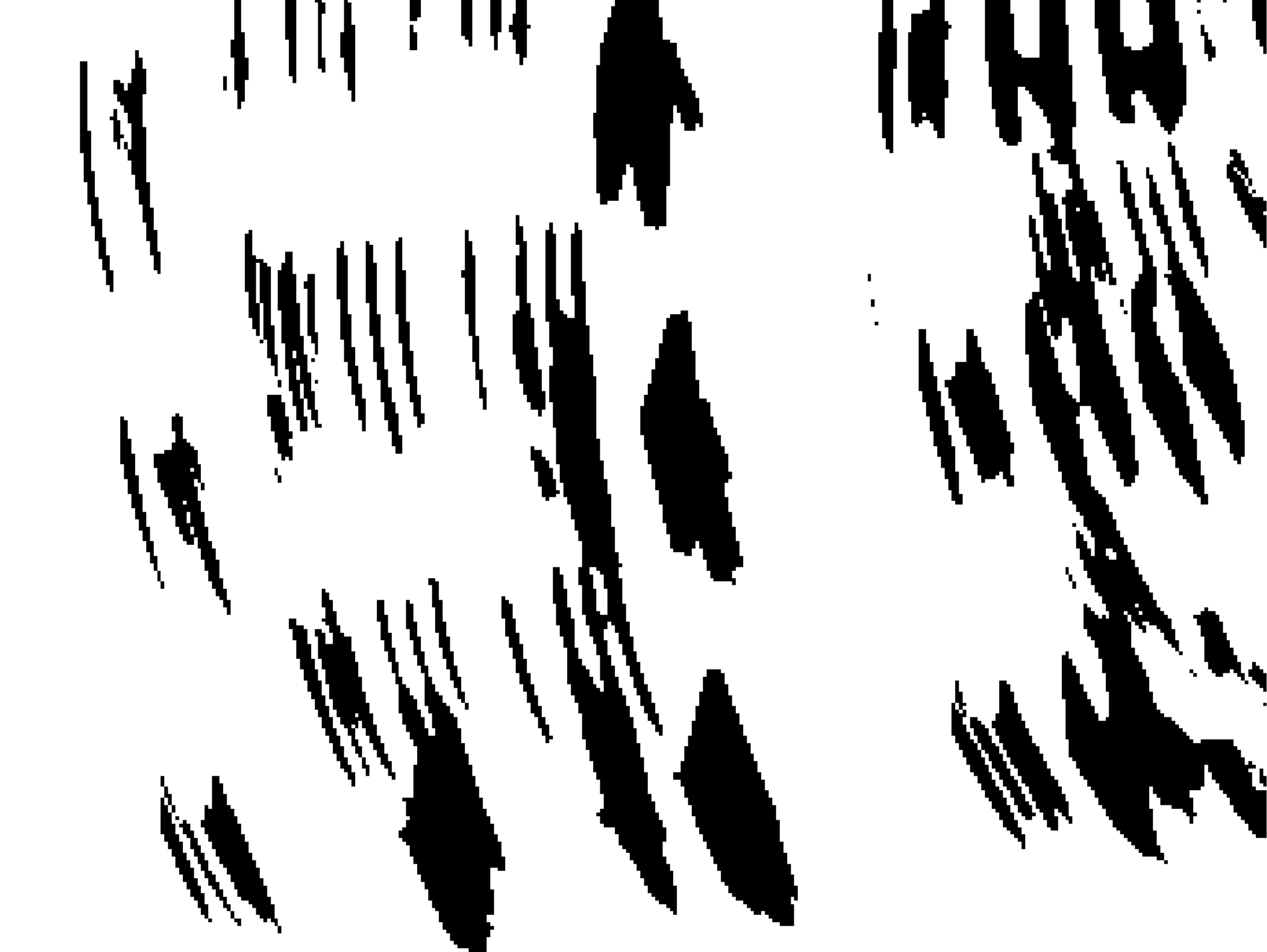}
    \caption{Wolf \cite{wolf2004extraction}}    \label{fig:image_bin_ebt_5}
  \end{subfigure}
  \begin{subfigure}{0.4905\linewidth}
    \includegraphics[width=\textwidth]{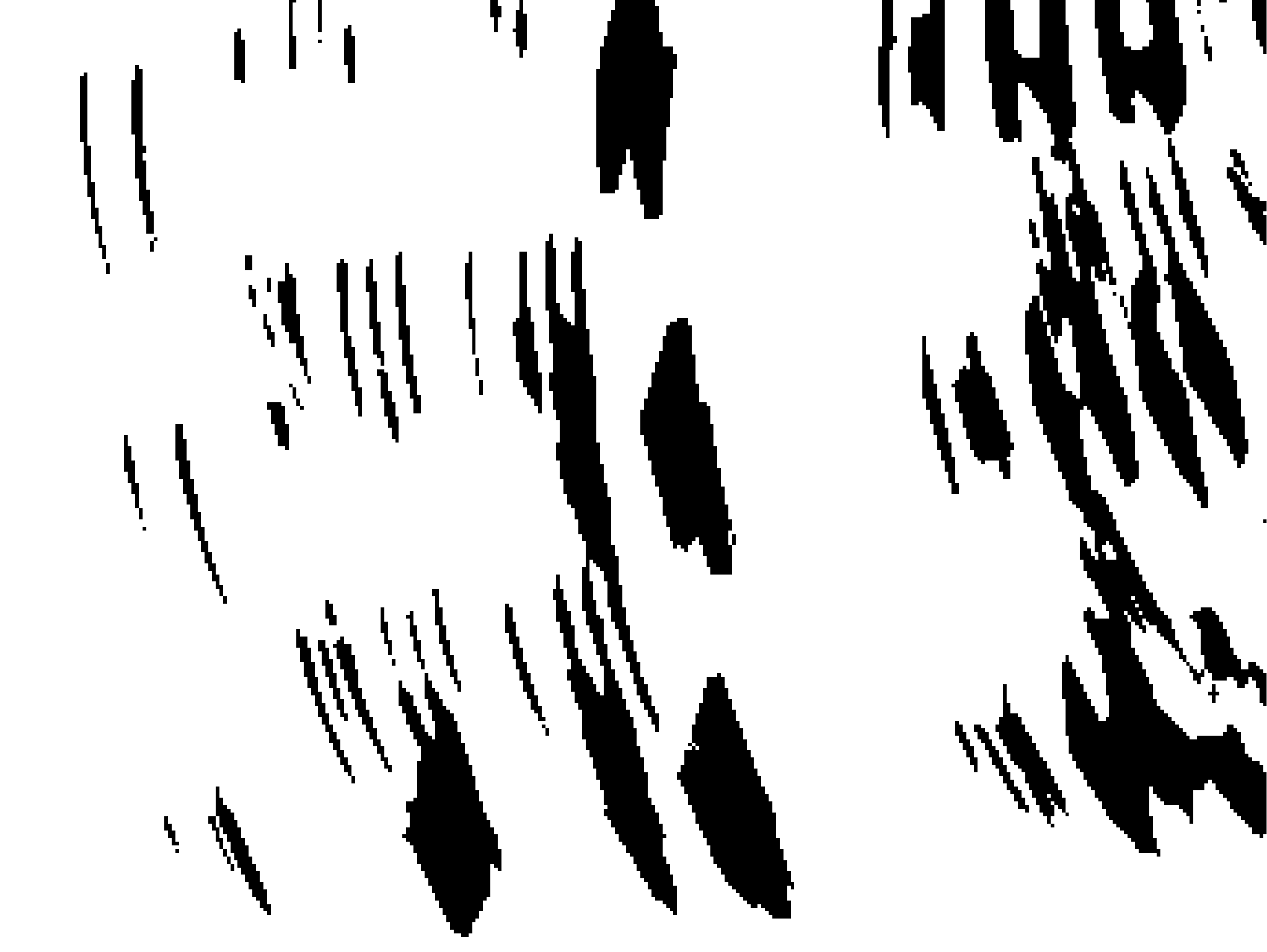}
    \caption{Wan \cite{mustafa2018binarization}}    \label{fig:image_bin_ebt_6}
  \end{subfigure}
    \begin{subfigure}{0.4905\linewidth}
    \includegraphics[width=\textwidth]{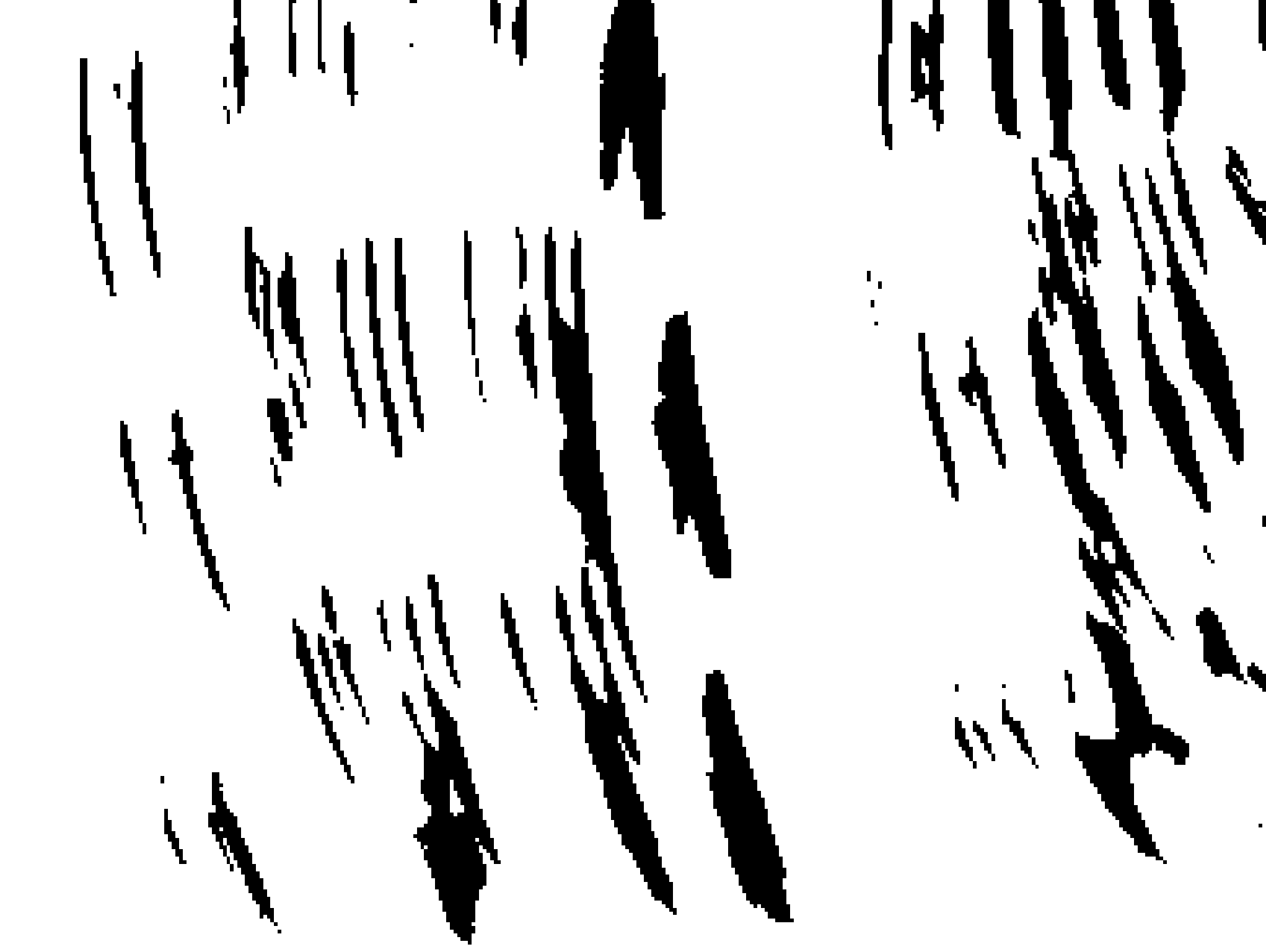}
    \caption{Auto-encoder \cite{calvo2019selectional}}    \label{fig:image_bin_ebt_7}
  \end{subfigure}
  \begin{subfigure}{0.4905\linewidth}
    \includegraphics[width=\textwidth]{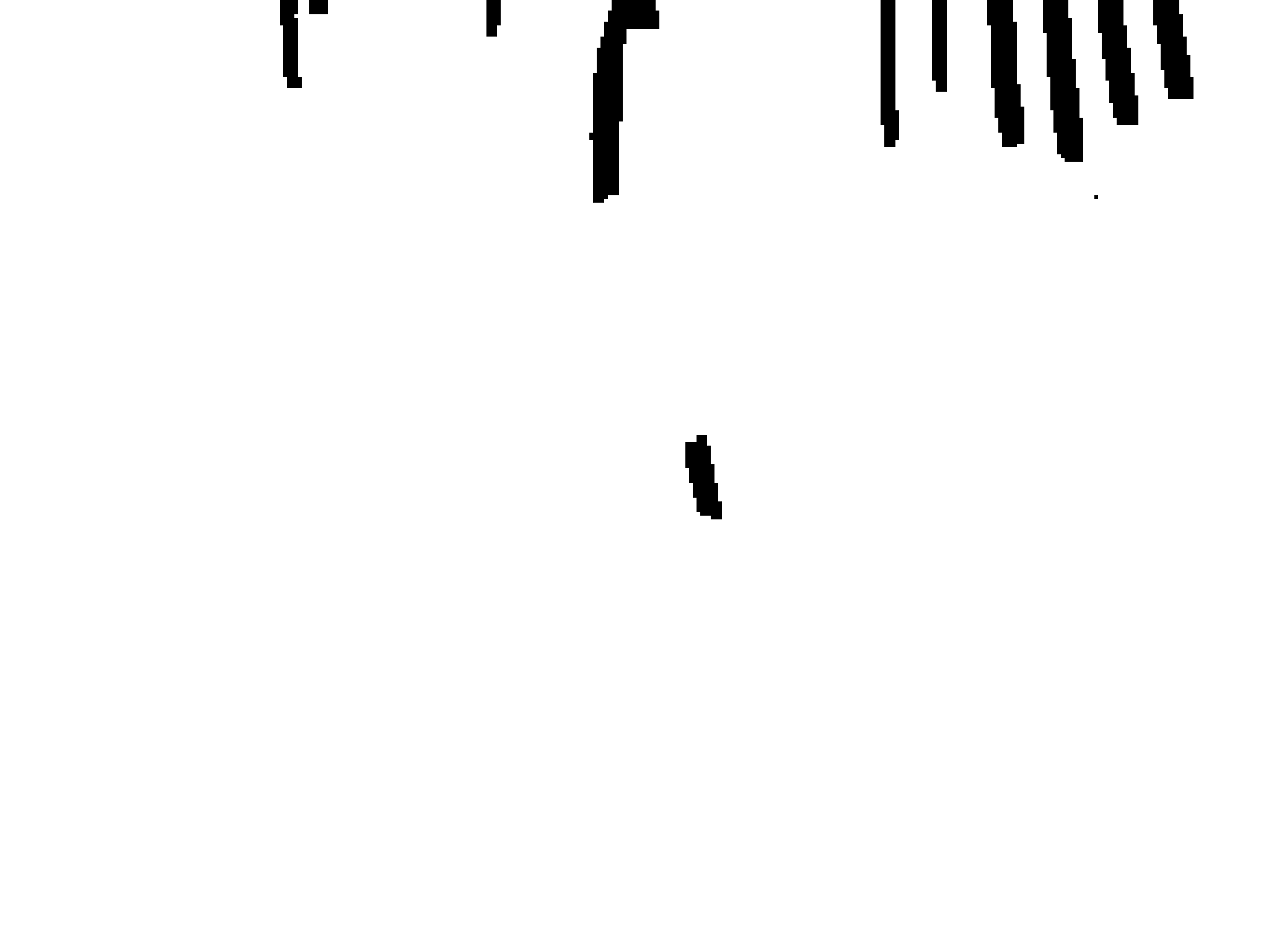}
    \caption{Howe \cite{howe2013document}}    \label{fig:image_bin_ebt_8}
  \end{subfigure}
   \begin{subfigure}{0.4905\linewidth}
    \includegraphics[width=\textwidth]{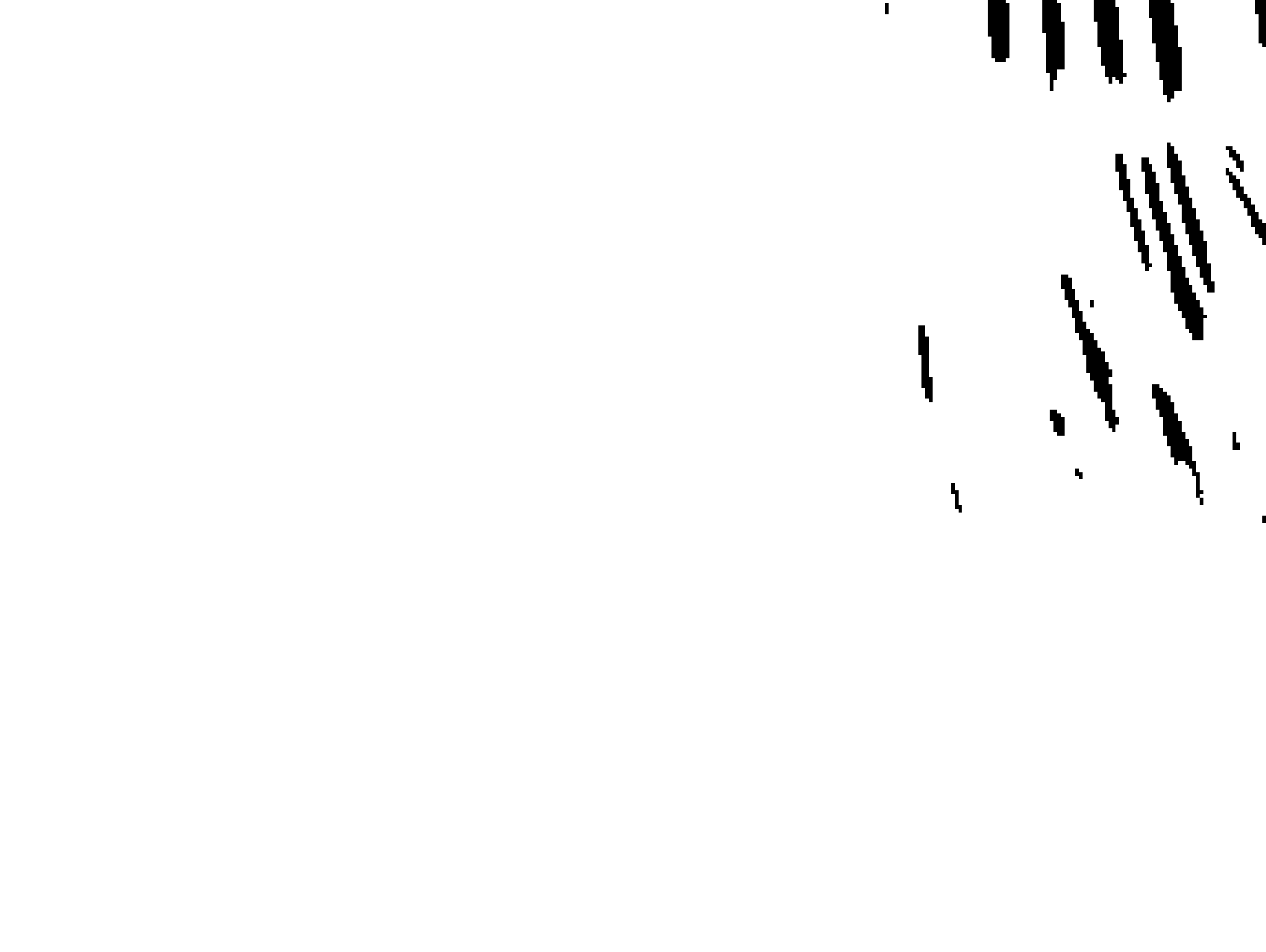}
    \caption{Dplink-Net \cite{xiong2021dp}}    \label{fig:image_bin_ebt_9}
  \end{subfigure}
  \begin{subfigure}{0.4905\linewidth}
    \includegraphics[width=\textwidth]{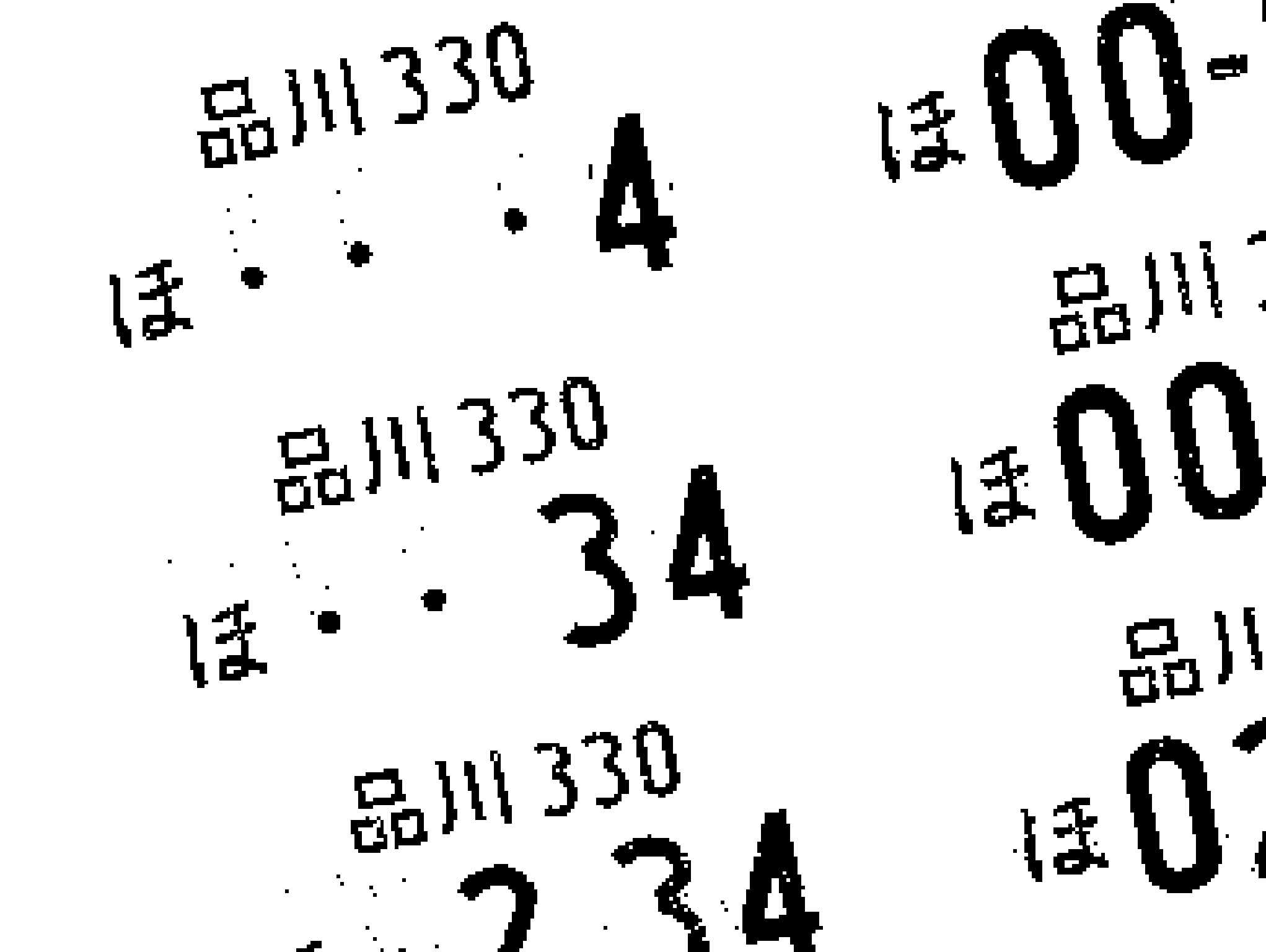}
    \caption{\textbf{Ours}}    \label{fig:image_bin_ebt_10}
  \end{subfigure}
\end{minipage}
\caption{\lsj{Comparison with state-of-the-art image binarization methods on (a)-(j) HQF \cite{stoffregen2020reducing} dataset, (k)-(t) Reblur \cite{sun2022event} dataset, and (u)-(ad) EBT dataset. Conventional image binarization methods (c)-(i), (m)-(s), and (w)-(ac) fail in motion blurry inputs and lose most details. (j) (t) (ad) Our method can produce sharp binary output and retain the find textual and geometry details.}}
\label{fig:ib} 
\end{figure*}

\begin{figure*}[t]
\begin{minipage}{0.325\linewidth}\centering
  \begin{subfigure}{\linewidth}
    \includegraphics[width=\textwidth]{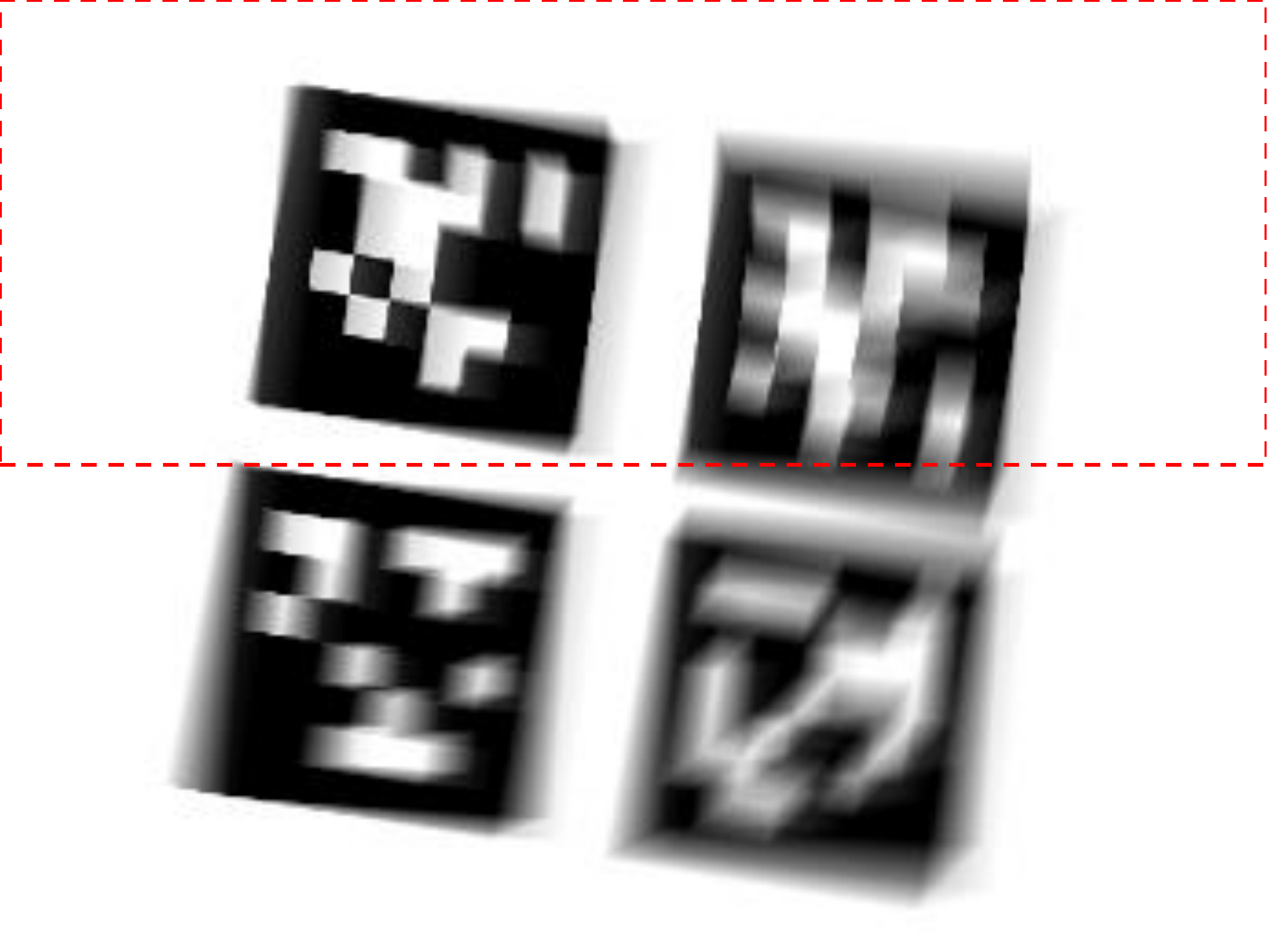}
    \caption{The blurred image}    \label{fig:deblur_bin_ebt}
  \end{subfigure}
  \begin{subfigure}{0.485\linewidth}
    \includegraphics[width=\textwidth]{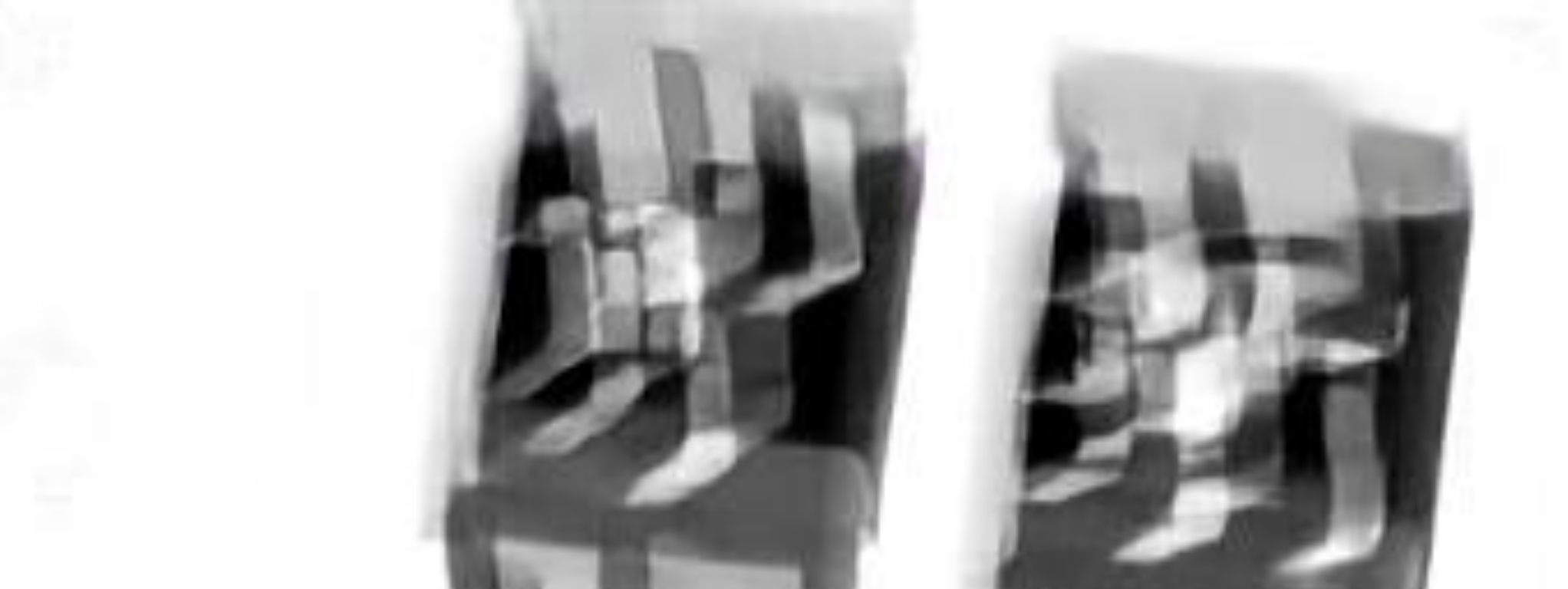}
    \caption{Jin \cite{jin2018learning}}    \label{fig:deblur_bin_ebt_halfscale_1}
  \end{subfigure}
  \begin{subfigure}{0.485\linewidth}
    \includegraphics[width=\textwidth]{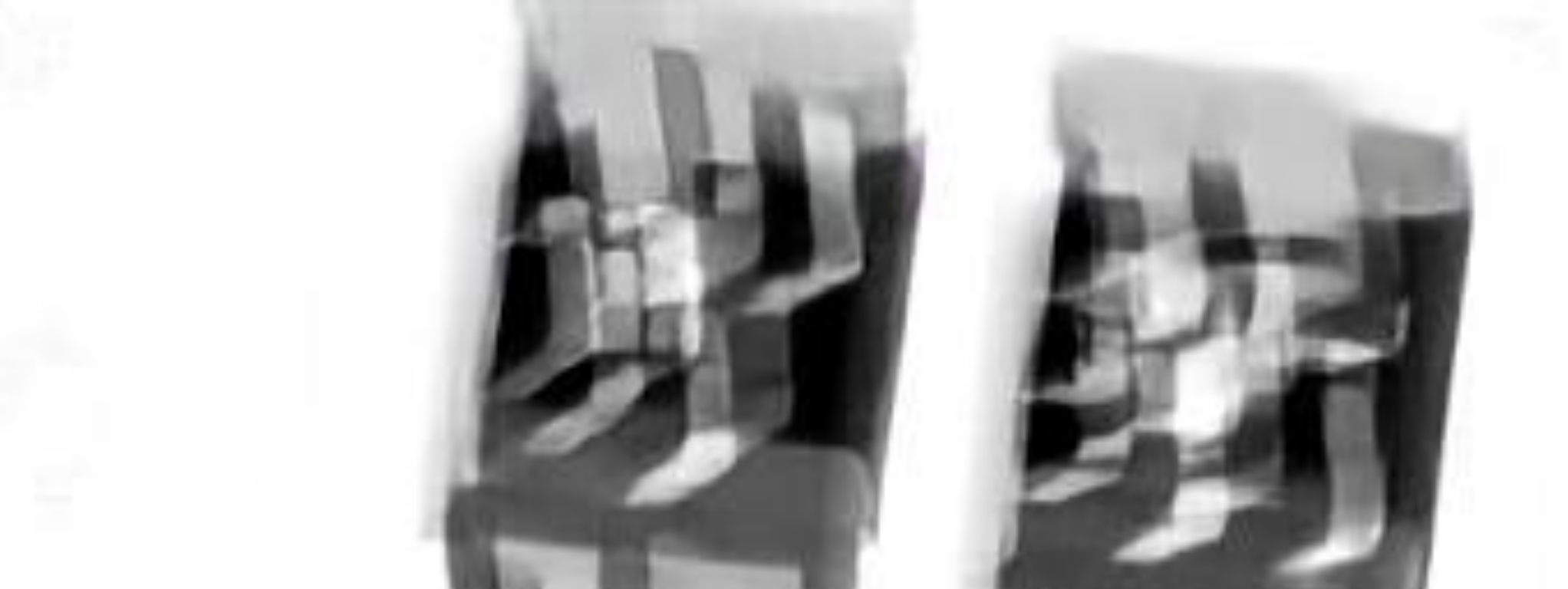}
    \caption*{Jin + WAN \cite{mustafa2018binarization}}   
  \end{subfigure}
  \begin{subfigure}{0.485\linewidth}
    \includegraphics[width=\textwidth]{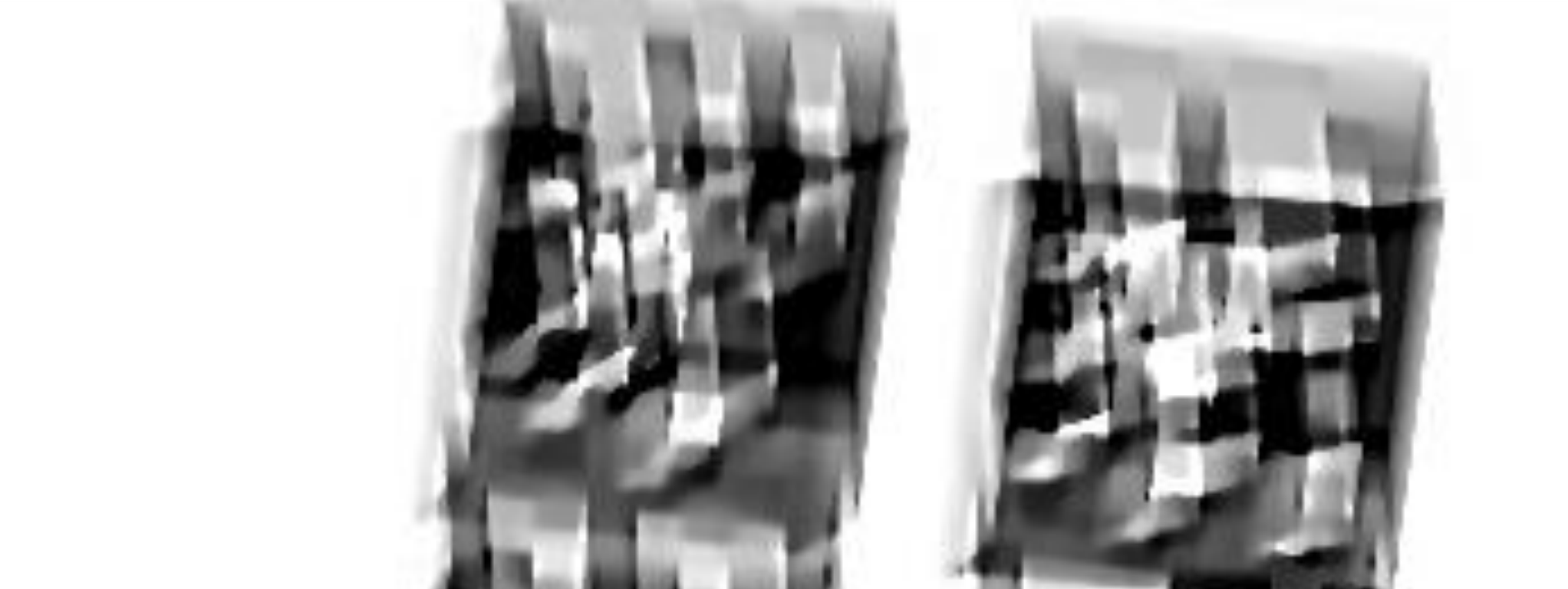}
    \caption{L0-reg \cite{textdeblur2014}}    \label{fig:deblur_bin_ebt_halfscale_3}
  \end{subfigure}
  \begin{subfigure}{0.485\linewidth}
    \includegraphics[width=\textwidth]{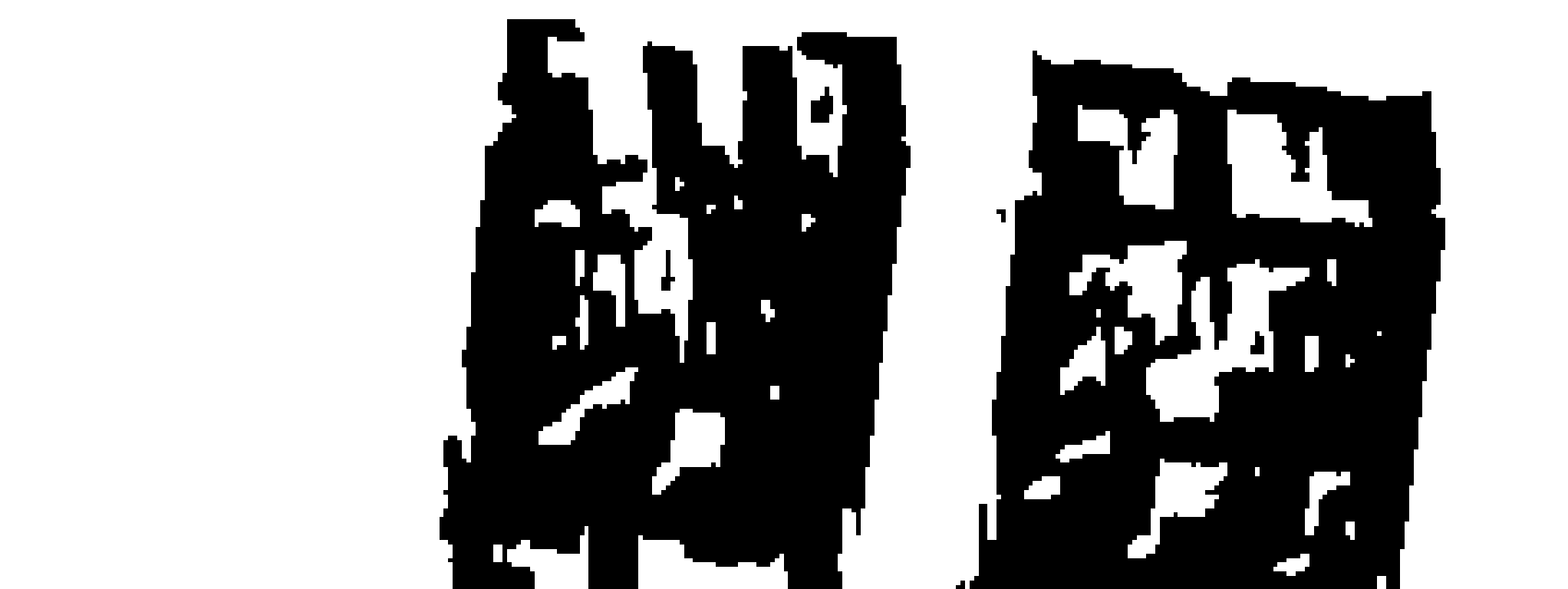}
    \caption*{L0-reg + WAN \cite{mustafa2018binarization}}    
  \end{subfigure}
  \begin{subfigure}{0.485\linewidth}
    \includegraphics[width=\textwidth]{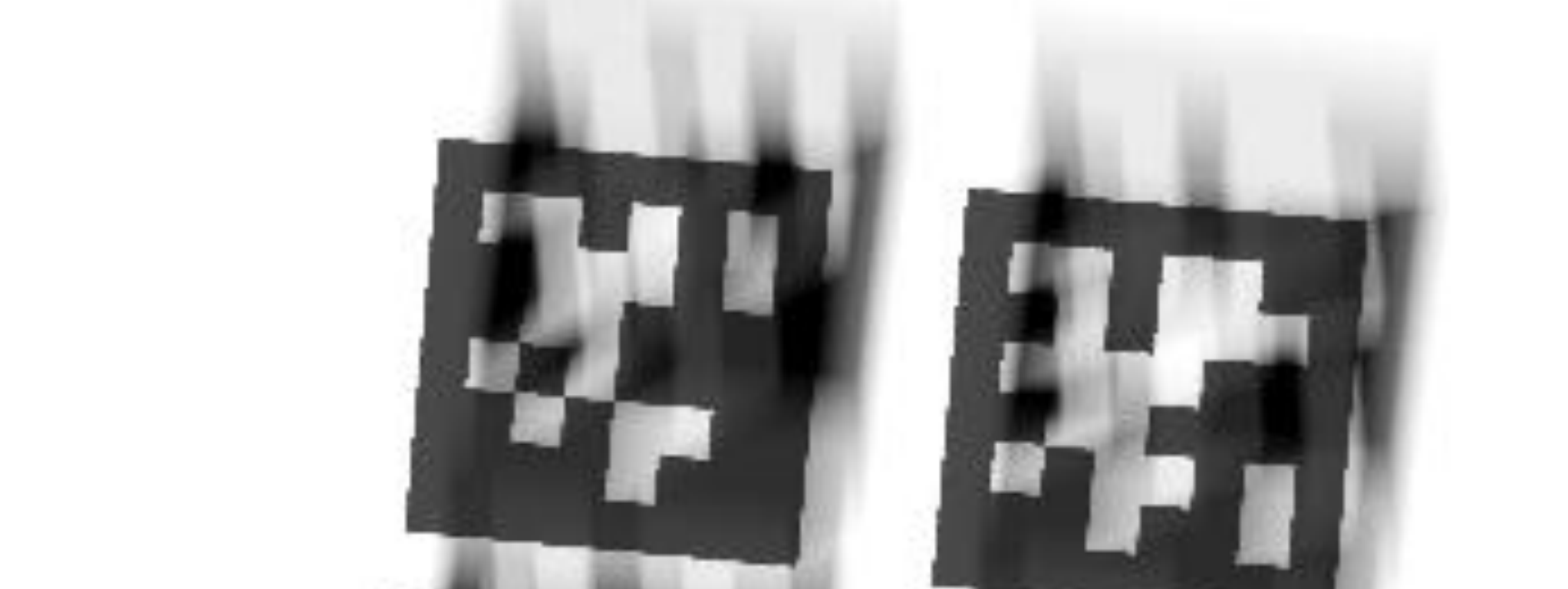}
    \caption{EDI \cite{edipami}}    \label{fig:deblur_bin_ebt_halfscale_5}
  \end{subfigure}
  \begin{subfigure}{0.485\linewidth}
    \includegraphics[width=\textwidth]{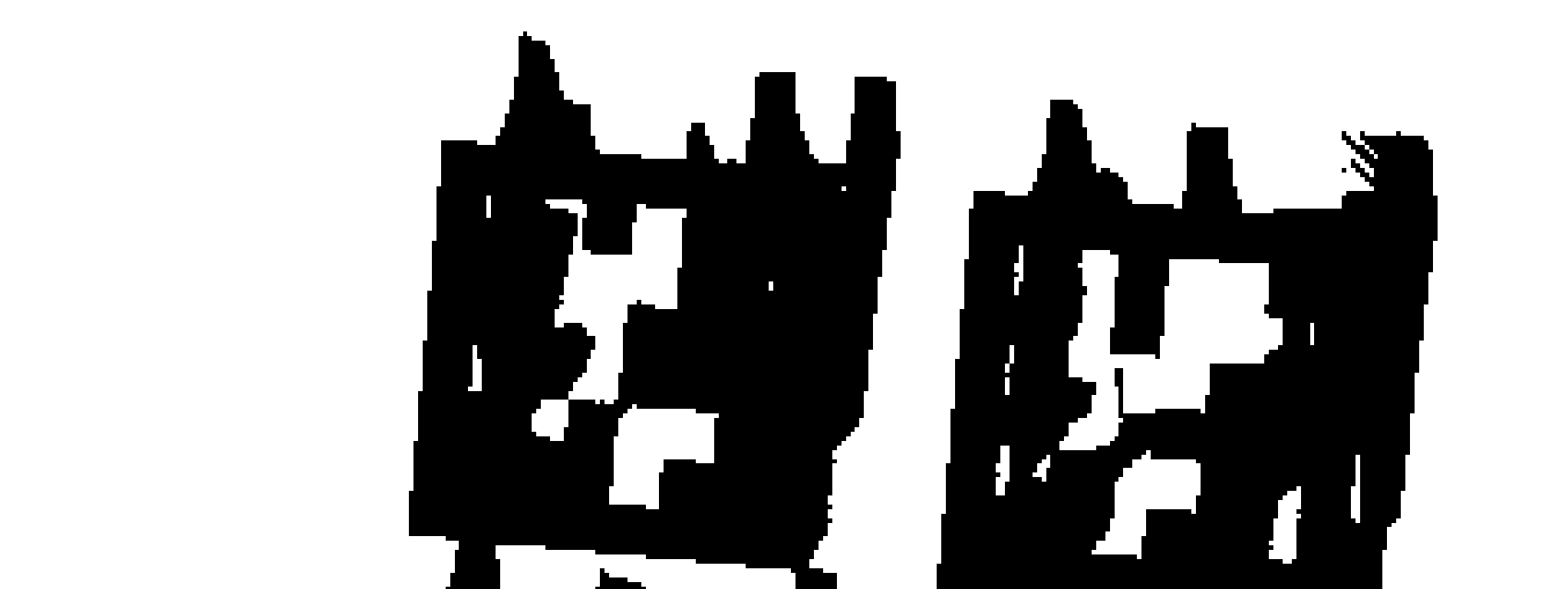}
    \caption*{EDI + WAN \cite{mustafa2018binarization}}  
  \end{subfigure}
  \begin{subfigure}{0.485\linewidth}
    \includegraphics[width=\textwidth]{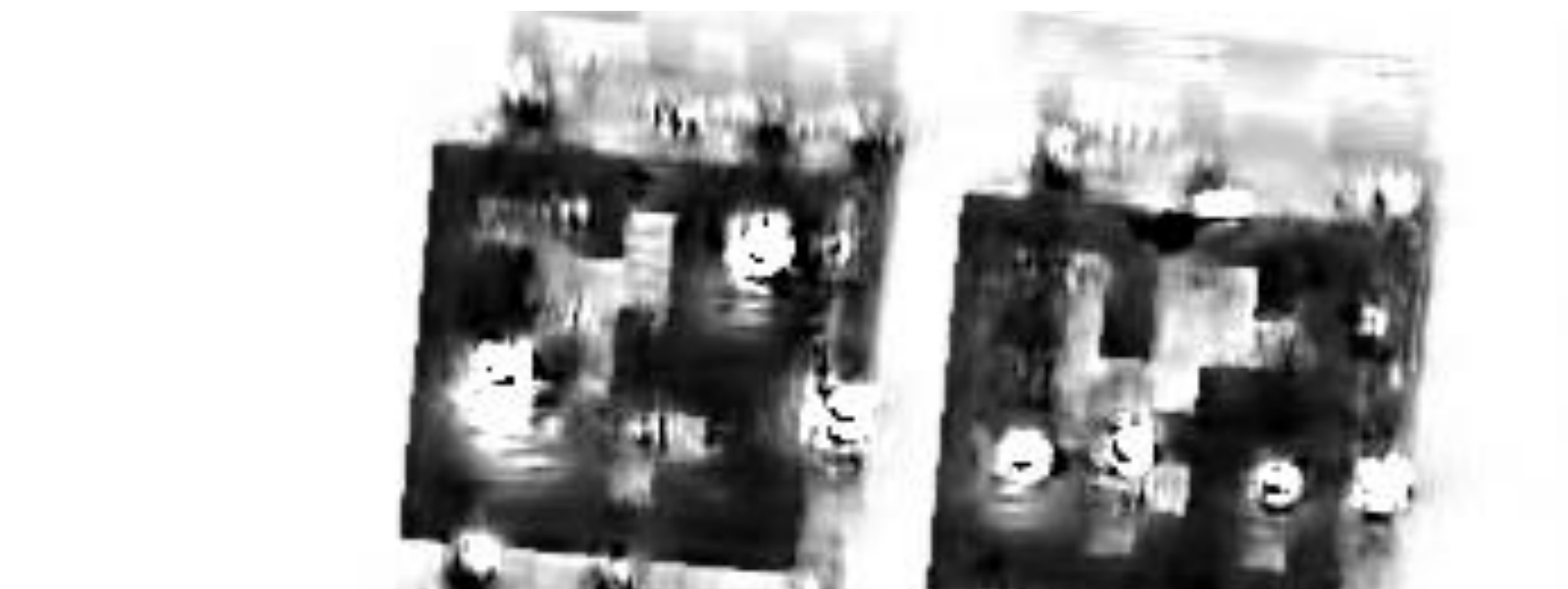}
    \caption{eSL \cite{yu2023learning}}    \label{fig:deblur_bin_ebt_halfscale_7}
  \end{subfigure}
    \begin{subfigure}{0.485\linewidth}
    \includegraphics[width=\textwidth]{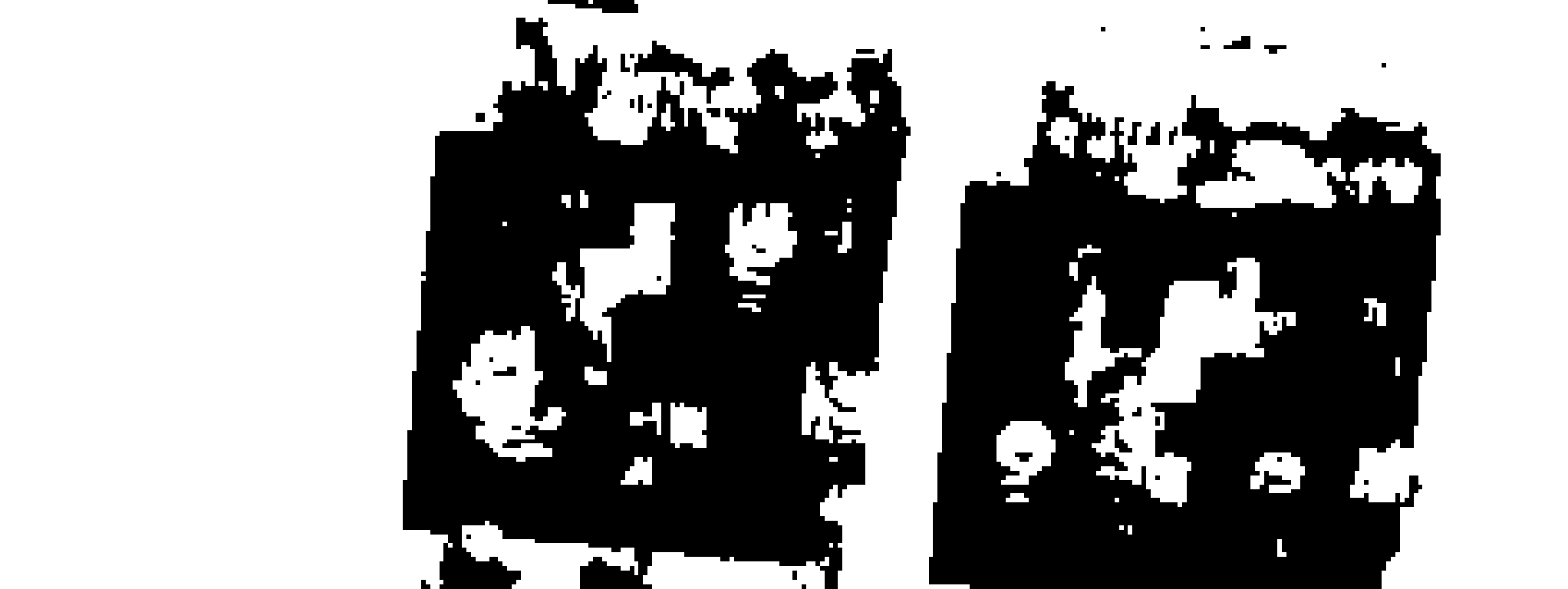}
    \caption*{eSL + WAN \cite{mustafa2018binarization}}   
  \end{subfigure}
  
  \begin{subfigure}{0.485\linewidth}
    \includegraphics[width=\textwidth]{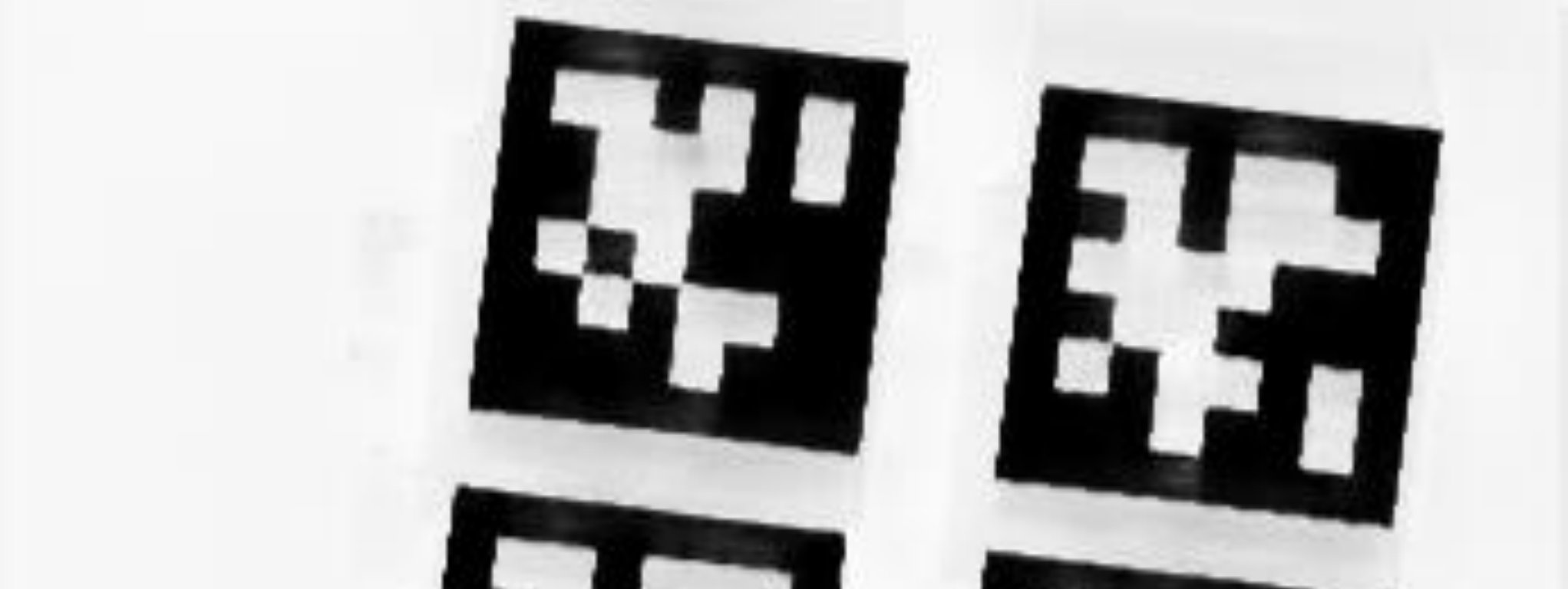}
    \caption{LEDVDI \cite{lin2020learning}}    \label{fig:deblur_bin_ebt_halfscale_9}
  \end{subfigure}
    \begin{subfigure}{0.485\linewidth}
    \includegraphics[width=\textwidth]{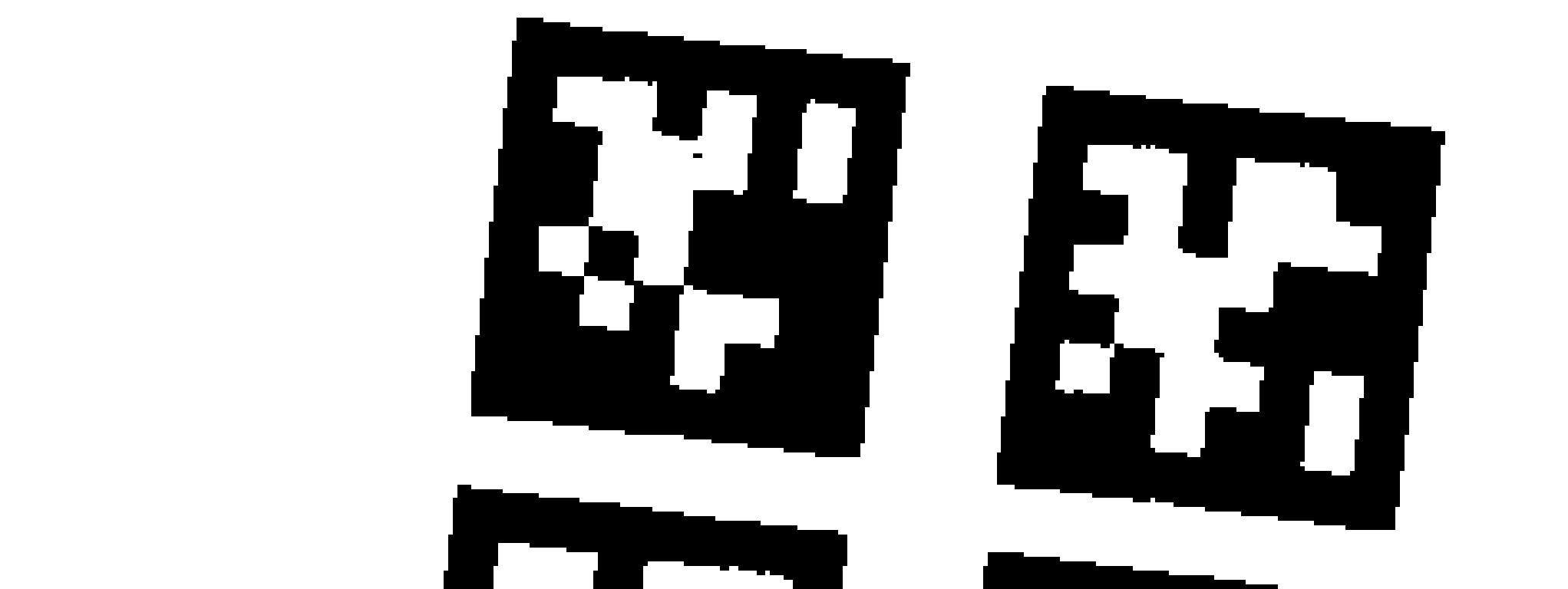}
    \caption*{LEDVDI + WAN \cite{mustafa2018binarization}}    \label{fig:deblur_bin_ebt_halfscale_10}
  \end{subfigure}
    \begin{subfigure}{0.485\linewidth}
    \includegraphics[width=\textwidth]{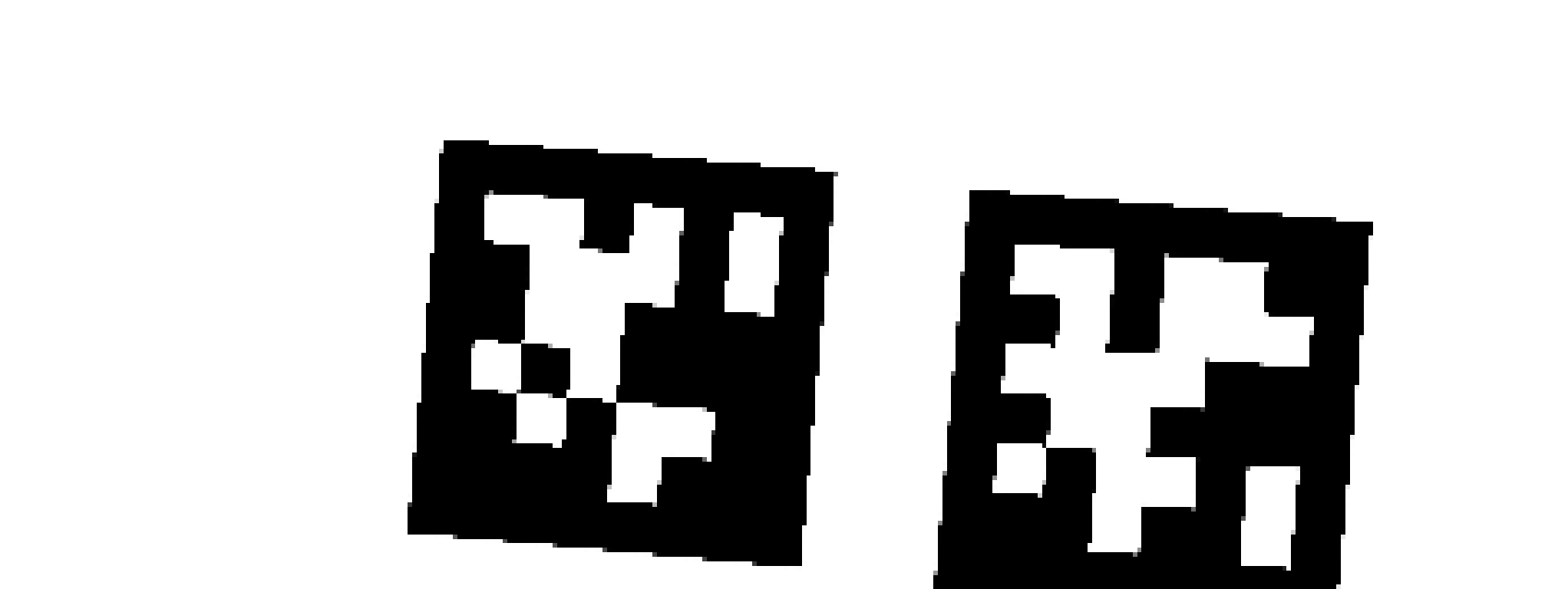}
    \caption{Ground truth}    \label{fig:deblur_bin_ebt_halfscale_11}
  \end{subfigure}
    \begin{subfigure}{0.485\linewidth}
    \includegraphics[width=\textwidth]{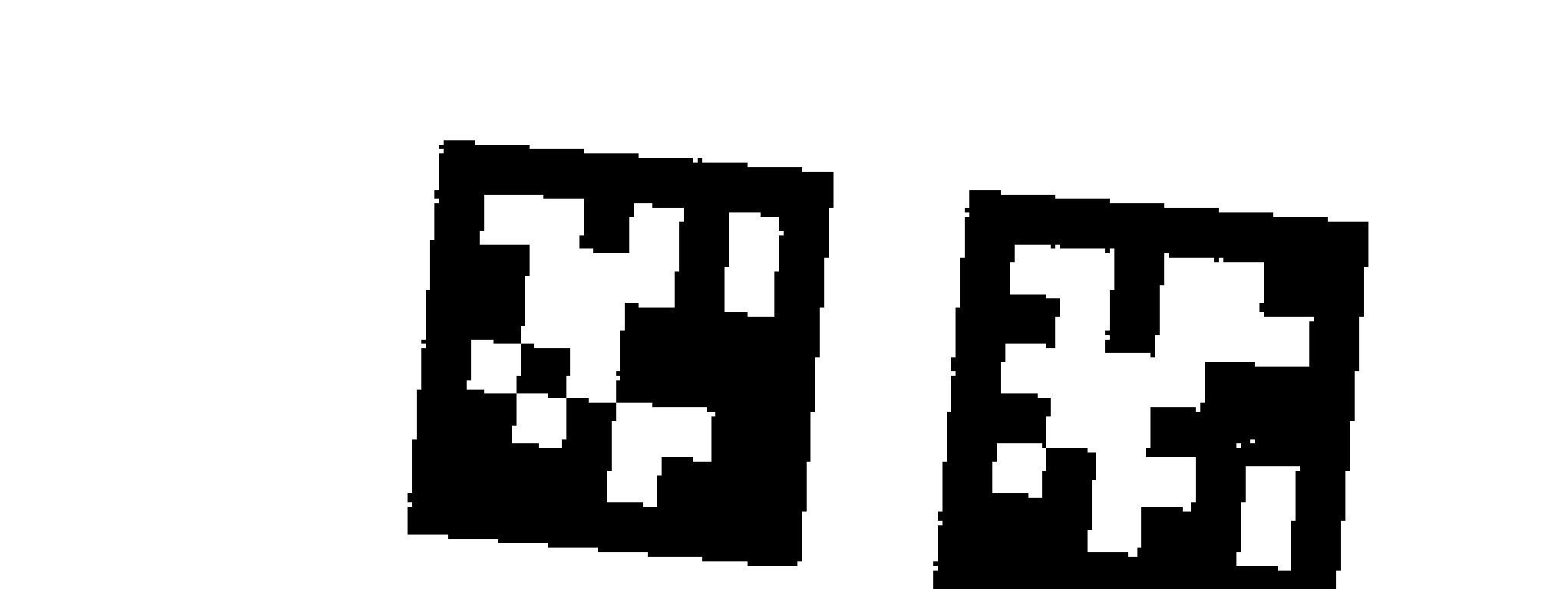}
    \caption{\textbf{Ours}}    \label{fig:deblur_bin_ebt_halfscale_12}
  \end{subfigure}
\end{minipage}
\hfill
\begin{minipage}{0.325\linewidth}\centering
  \begin{subfigure}{\linewidth}
    \includegraphics[width=\textwidth]{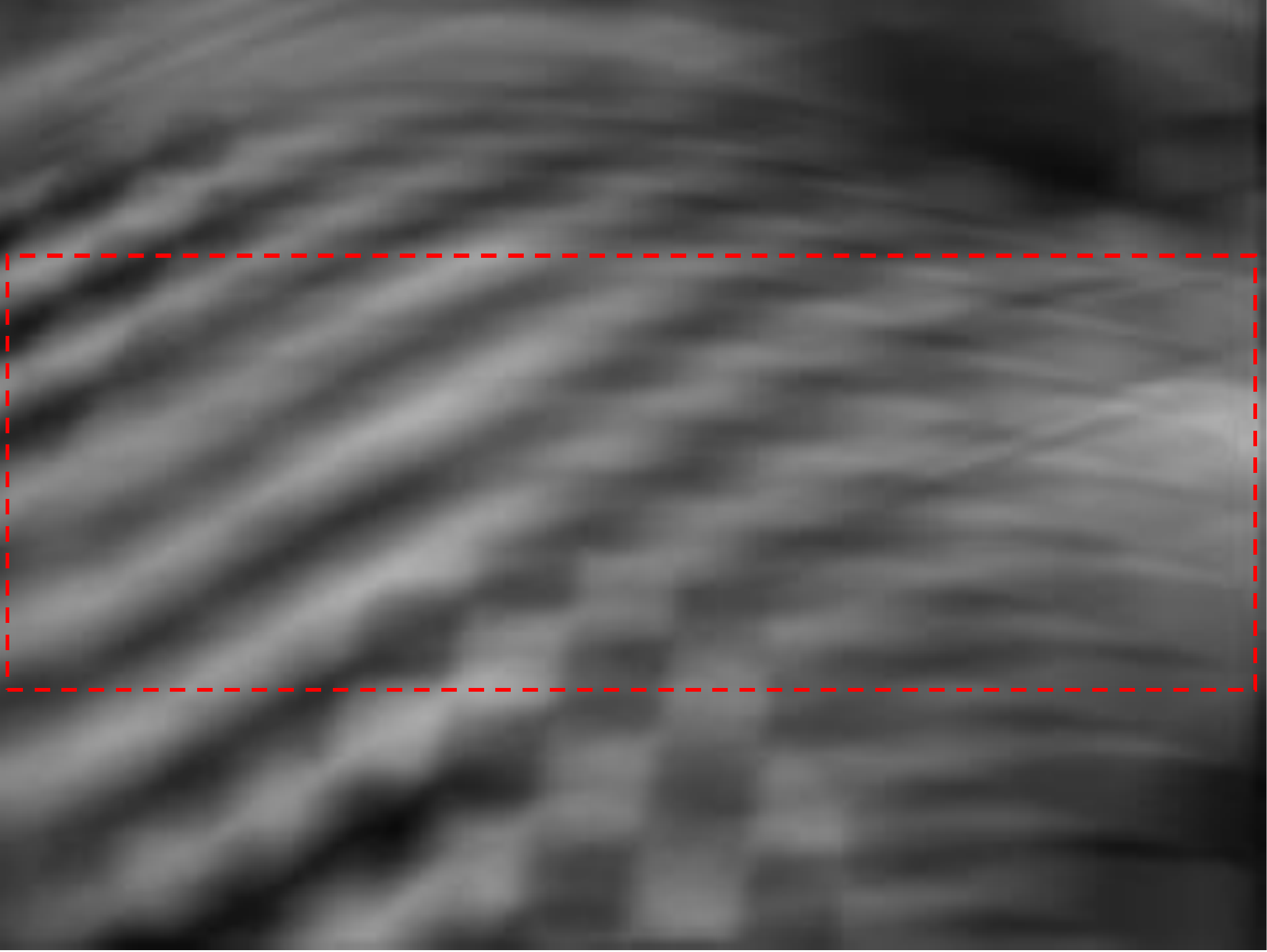}
    \caption{The blurred image}    \label{fig:deblur_bin_hqf}
  \end{subfigure}
  \begin{subfigure}{0.485\linewidth}
    \includegraphics[width=\textwidth]{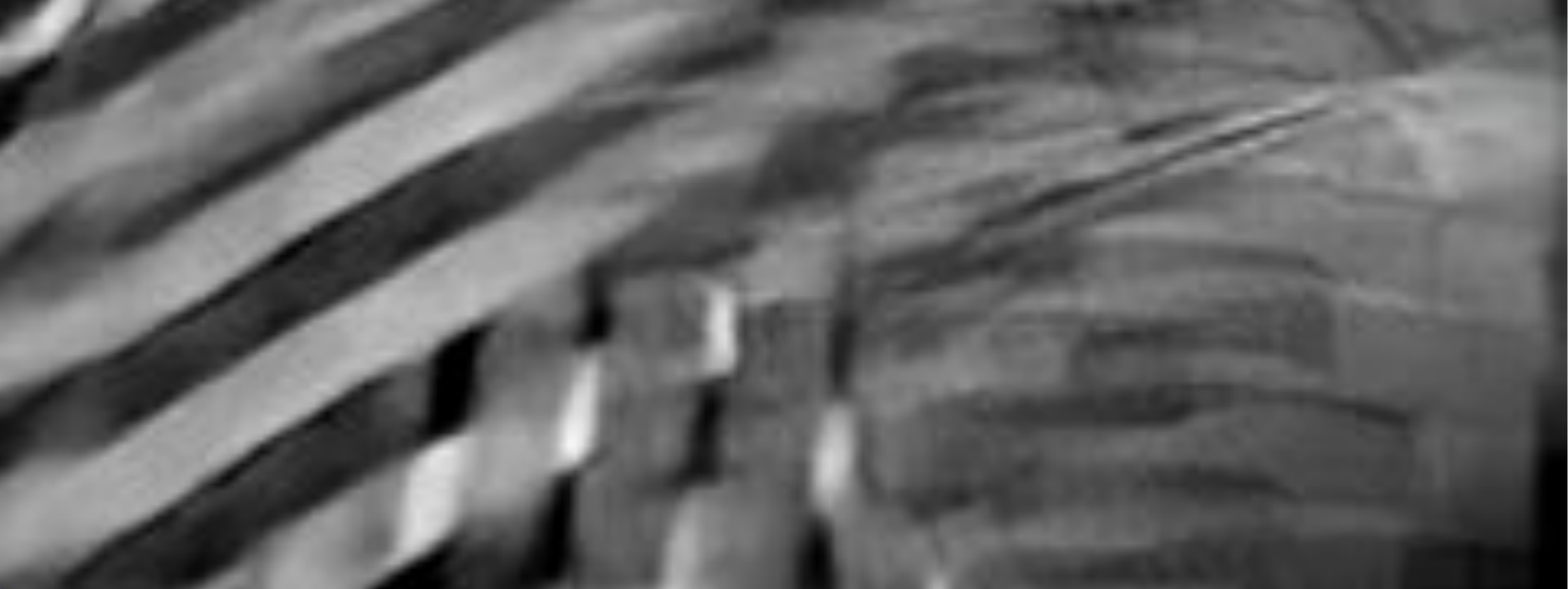}
    \caption{Jin \cite{jin2018learning}}    \label{fig:deblur_bin_hqf_halfscale_1}
  \end{subfigure}
  \begin{subfigure}{0.485\linewidth}
    \includegraphics[width=\textwidth]{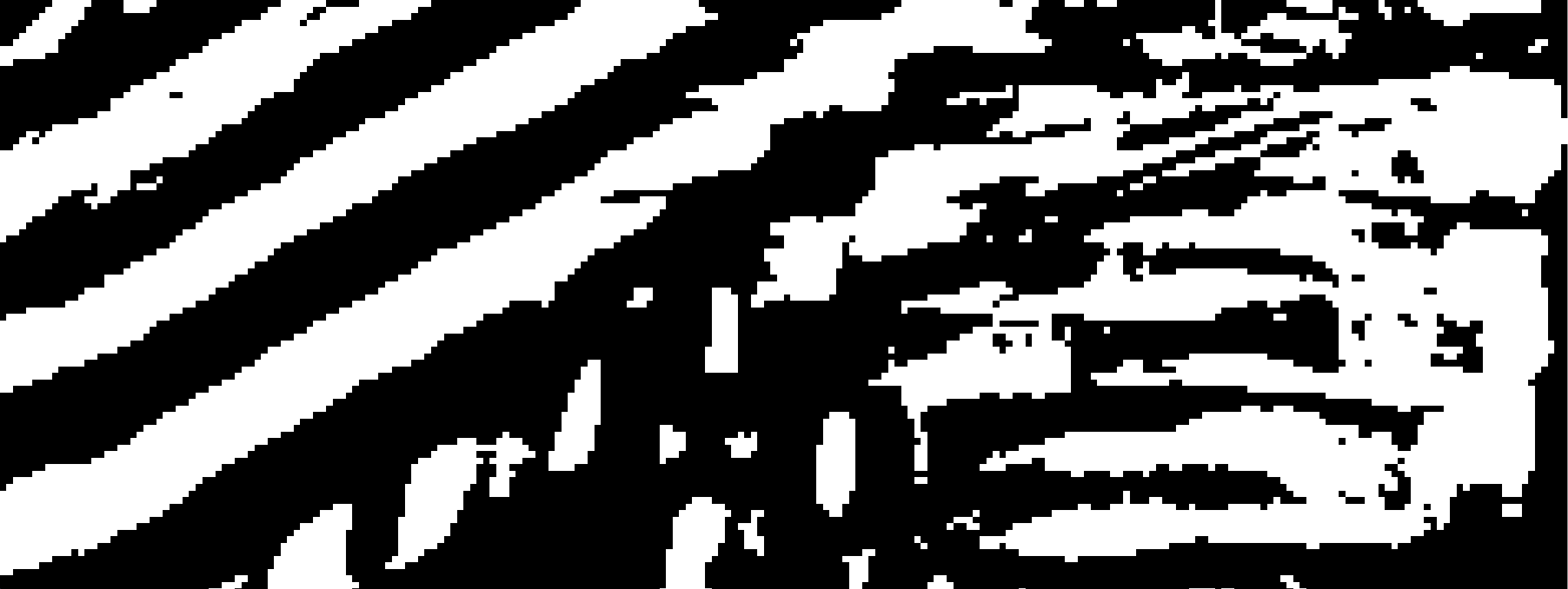}
    \caption*{Jin + WAN \cite{mustafa2018binarization}}    \label{fig:deblur_bin_hqf_halfscale_2}
  \end{subfigure}
  \begin{subfigure}{0.485\linewidth}
    \includegraphics[width=\textwidth]{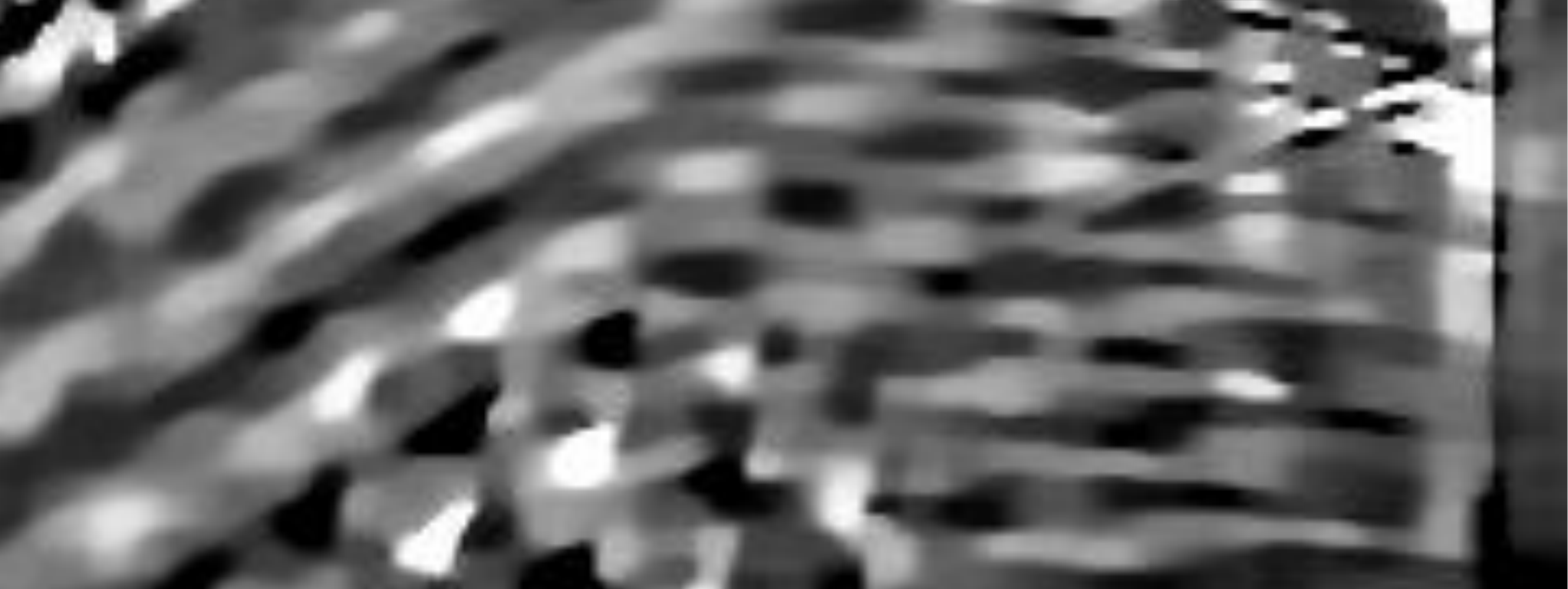}
    \caption{L0-reg \cite{textdeblur2014}}    \label{fig:deblur_bin_hqf_halfscale_3}
  \end{subfigure}
  \begin{subfigure}{0.485\linewidth}
    \includegraphics[width=\textwidth]{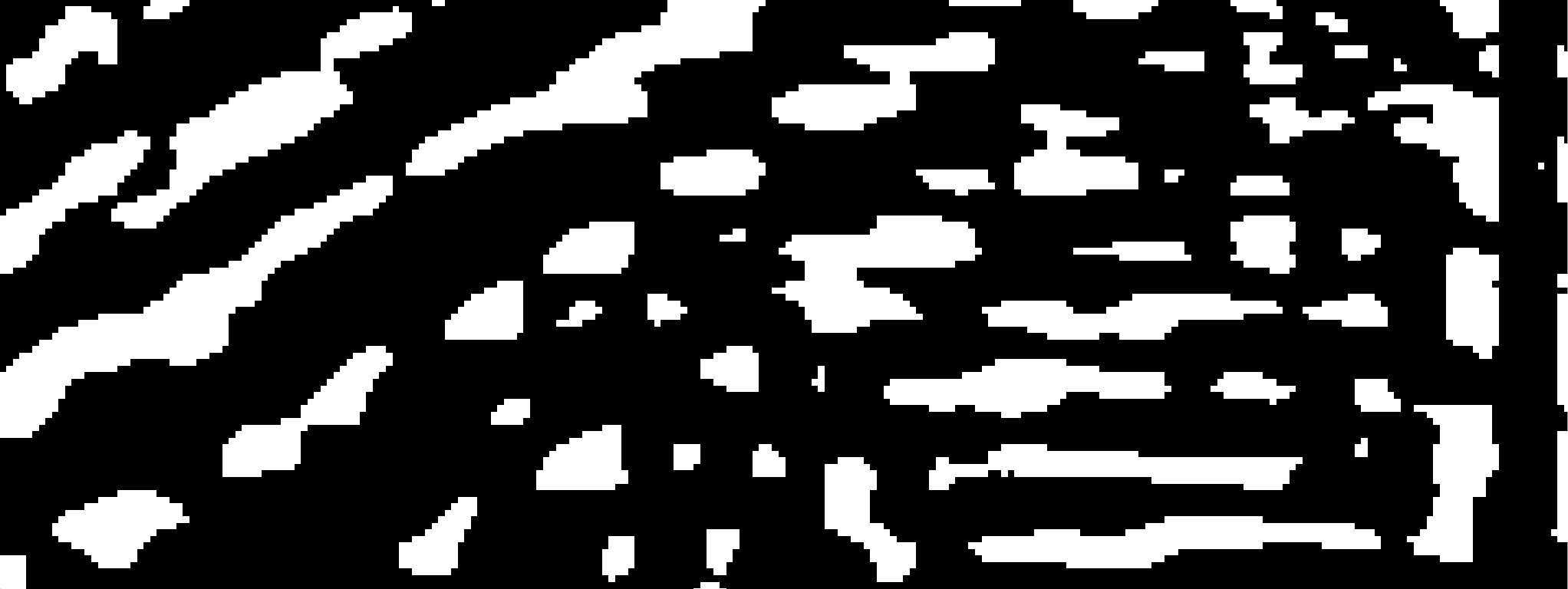}
    \caption{L0-reg + WAN \cite{mustafa2018binarization}}    \label{fig:deblur_bin_hqf_halfscale_4}
  \end{subfigure}
  \begin{subfigure}{0.485\linewidth}
    \includegraphics[width=\textwidth]{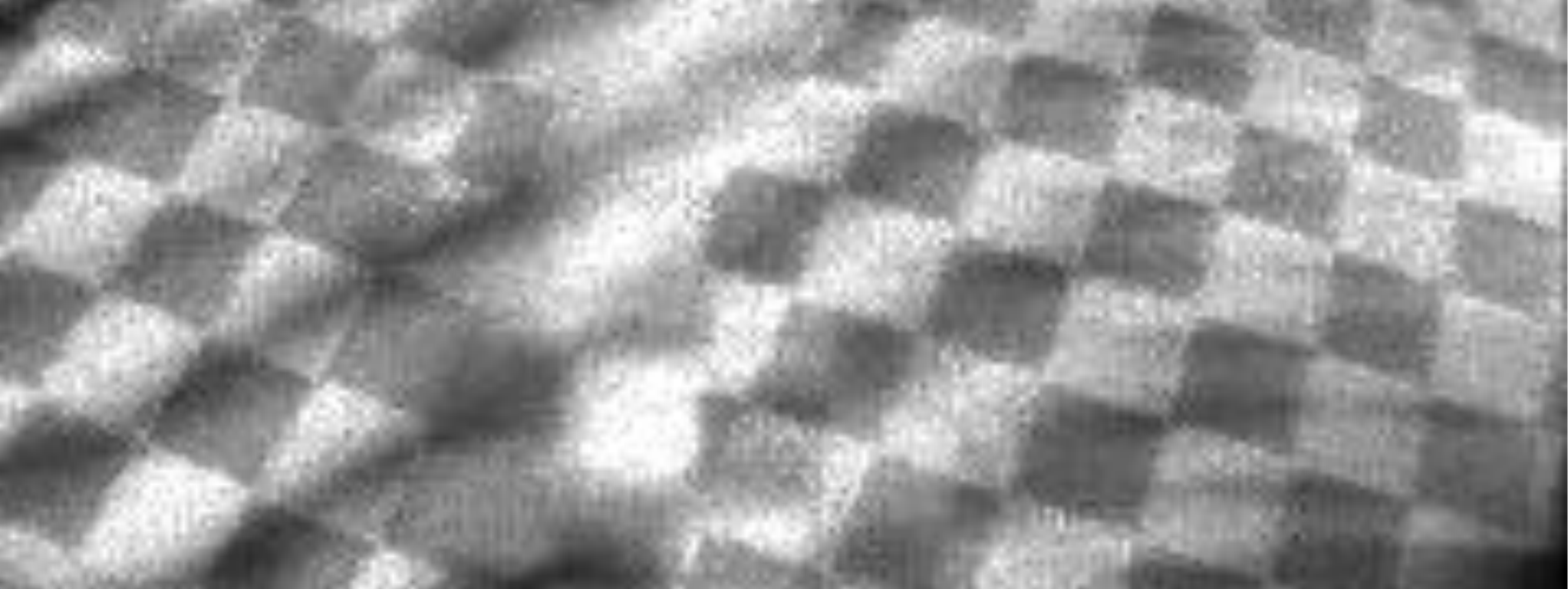}
    \caption{EDI \cite{edipami}}    \label{fig:deblur_bin_hqf_halfscale_5}
  \end{subfigure}
  \begin{subfigure}{0.485\linewidth}
    \includegraphics[width=\textwidth]{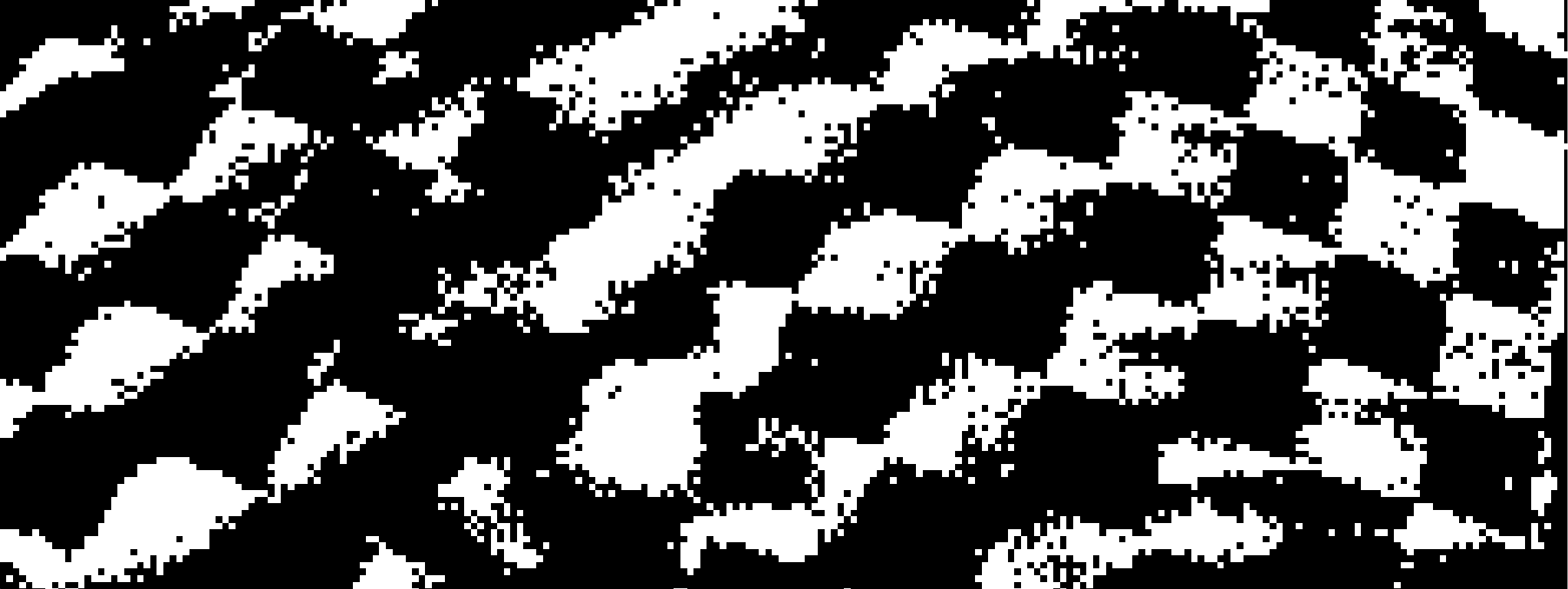}
    \caption*{EDI + WAN \cite{mustafa2018binarization}}    \label{fig:deblur_bin_hqf_halfscale_6}
  \end{subfigure}
  \begin{subfigure}{0.485\linewidth}
    \includegraphics[width=\textwidth]{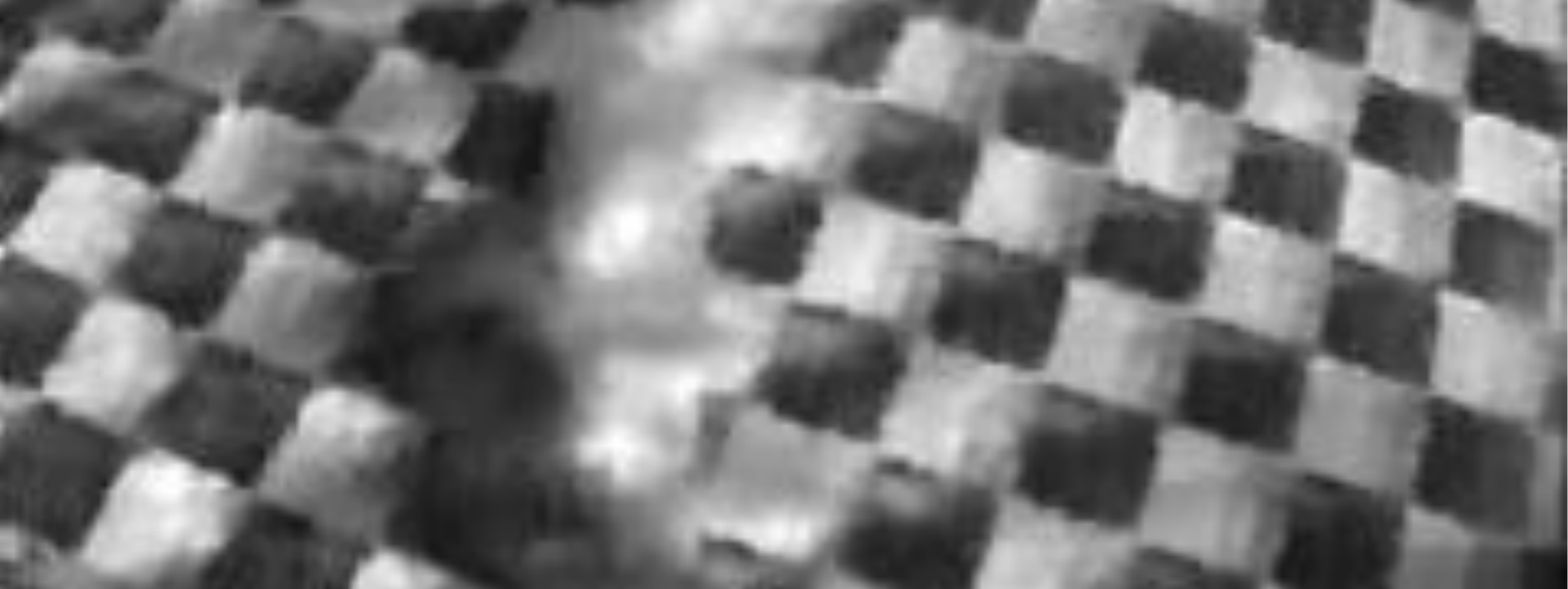}
    \caption{eSL \cite{yu2023learning}}    \label{fig:deblur_bin_hqf_halfscale_7}
  \end{subfigure}
  \begin{subfigure}{0.485\linewidth}
    \includegraphics[width=\textwidth]{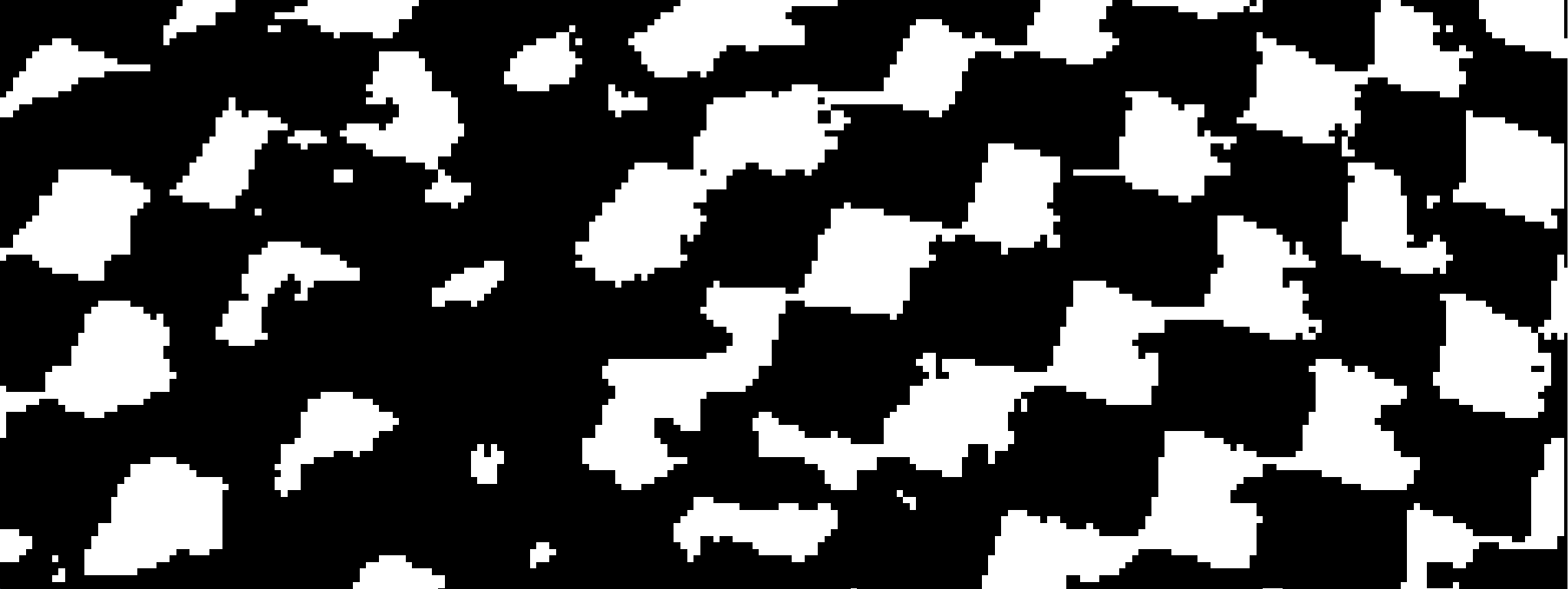}
    \caption*{eSL + WAN \cite{mustafa2018binarization}}    \label{fig:deblur_bin_hqf_halfscale_8}
  \end{subfigure}
  \begin{subfigure}{0.485\linewidth}
    \includegraphics[width=\textwidth]{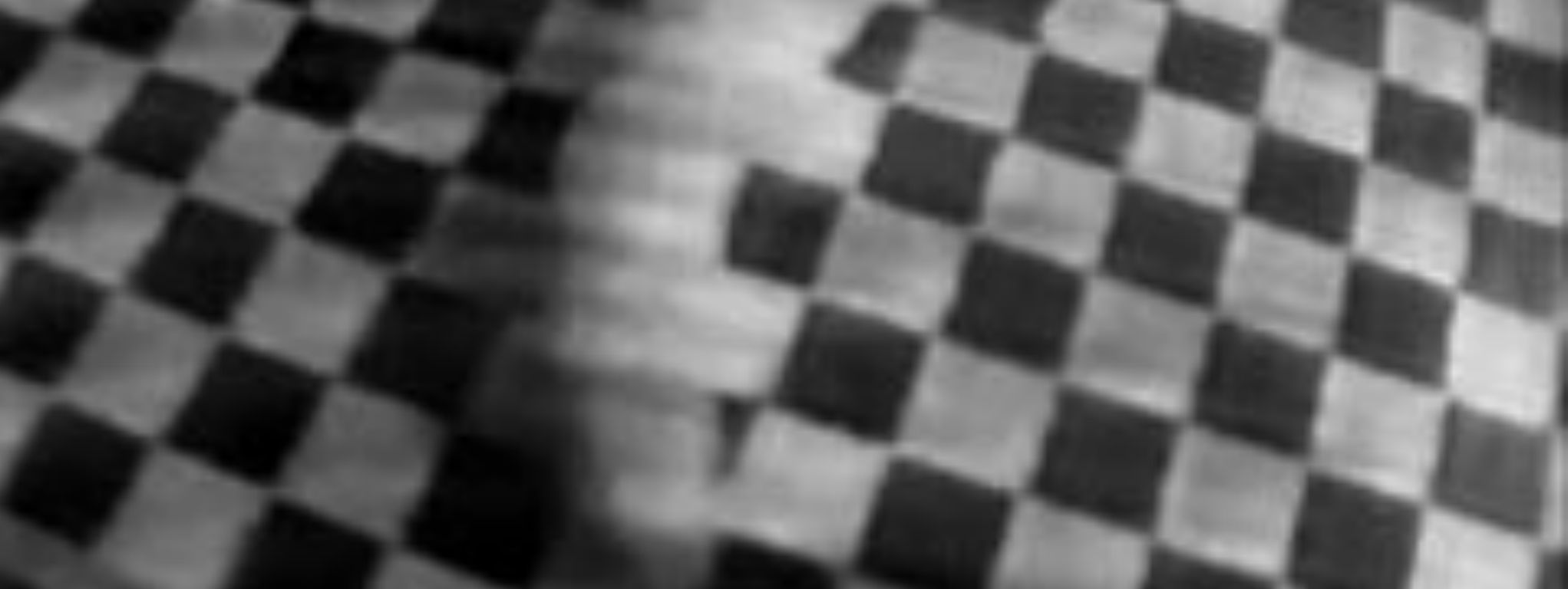}
    \caption{LEDVDI \cite{lin2020learning}}    \label{fig:deblur_bin_hqf_halfscale_9}
  \end{subfigure}
    \begin{subfigure}{0.485\linewidth}
    \includegraphics[width=\textwidth]{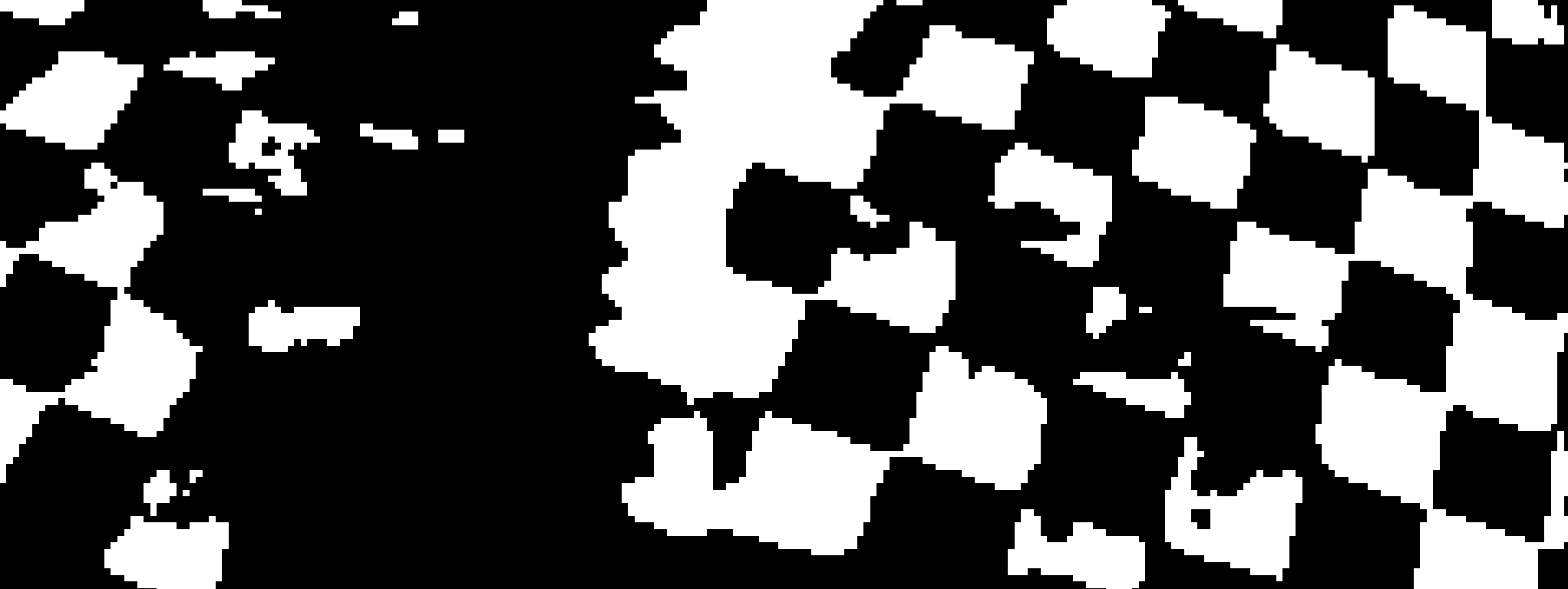}
    \caption*{LEDVDI + WAN \cite{mustafa2018binarization}}    \label{fig:deblur_bin_hqf_halfscale_10}
  \end{subfigure}
  \begin{subfigure}{0.485\linewidth}
    \includegraphics[width=\textwidth]{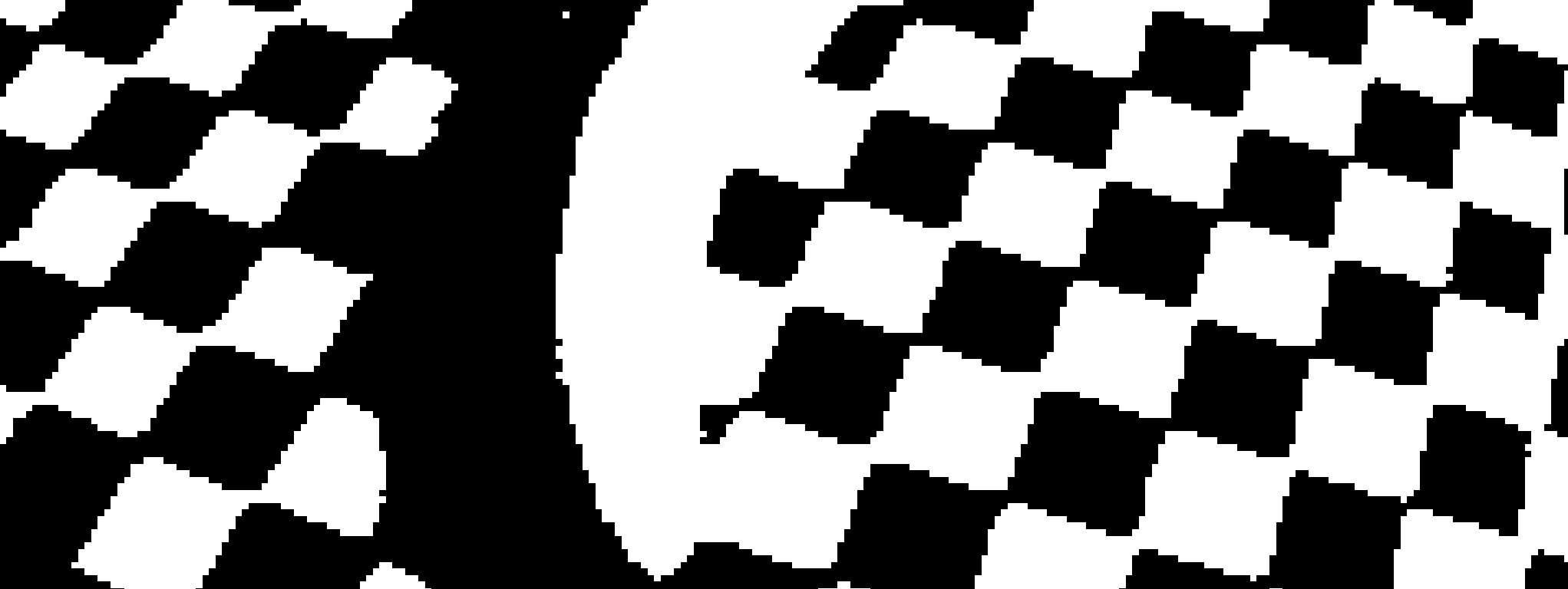}
    \caption{Ground truth}    \label{fig:deblur_bin_hqf_halfscale_11}
  \end{subfigure}
  \begin{subfigure}{0.485\linewidth}
    \includegraphics[width=\textwidth]{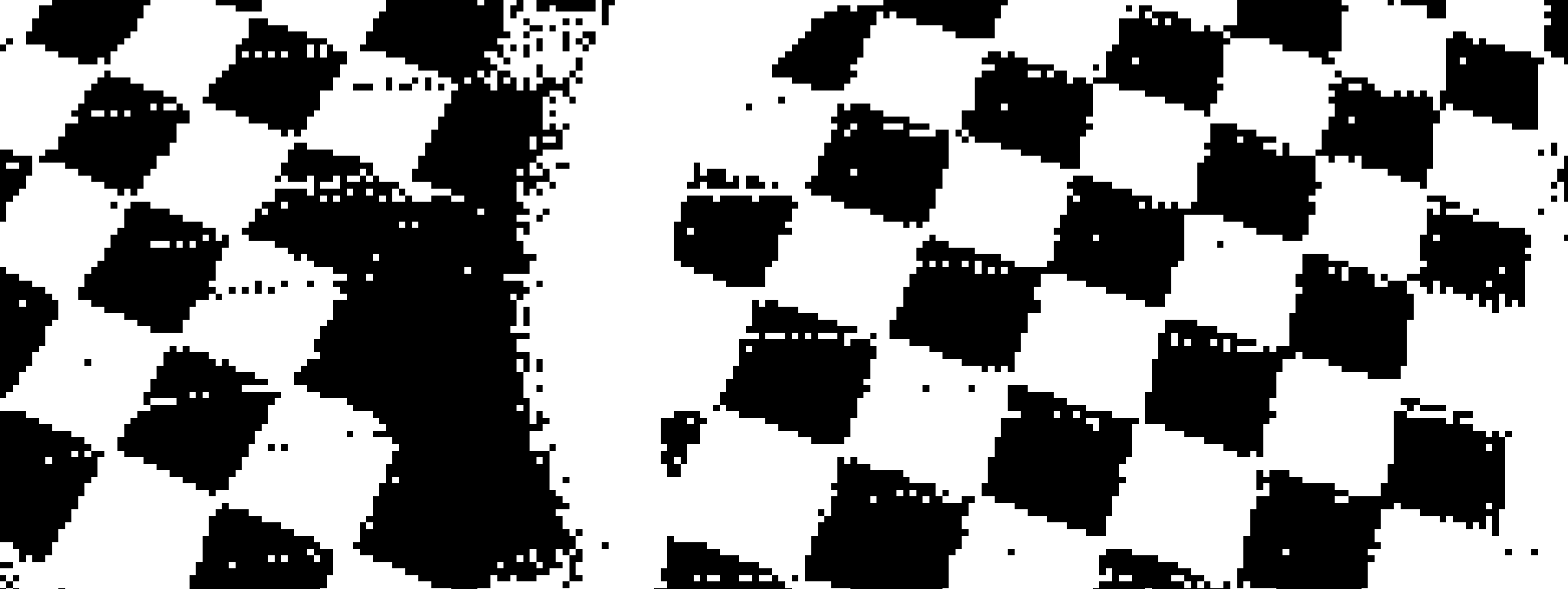}
    \caption{\textbf{Ours}}    \label{fig:deblur_bin_hqf_halfscale_12}
  \end{subfigure}
\end{minipage}
\hfill
\begin{minipage}{0.325\linewidth}\centering
  \begin{subfigure}{\linewidth}
    \includegraphics[width=\textwidth]{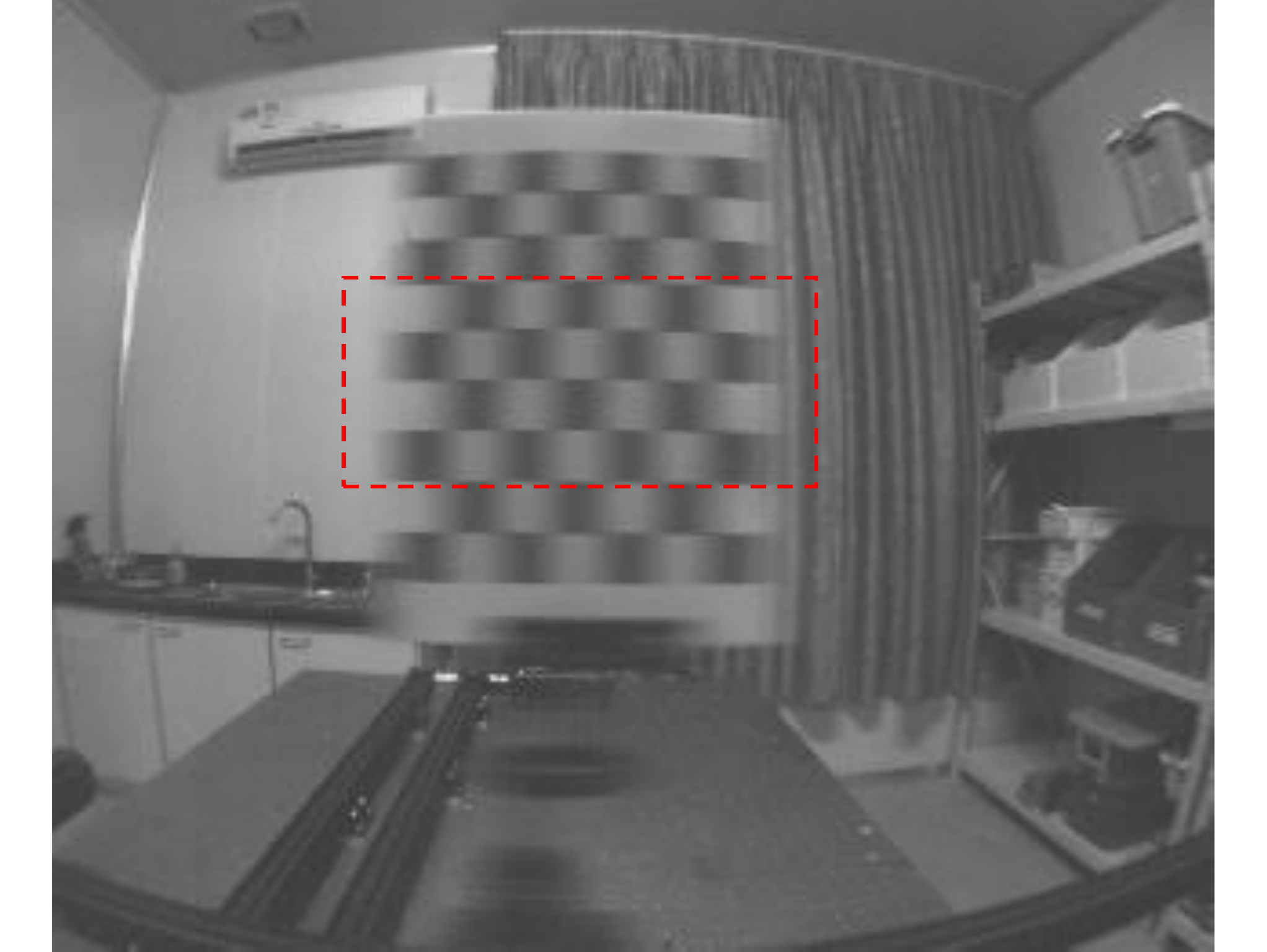}
    \caption{The blurred image}    \label{fig:deblur_bin_reblur}
  \end{subfigure}
  \begin{subfigure}{0.485\linewidth}
    \includegraphics[width=\textwidth]{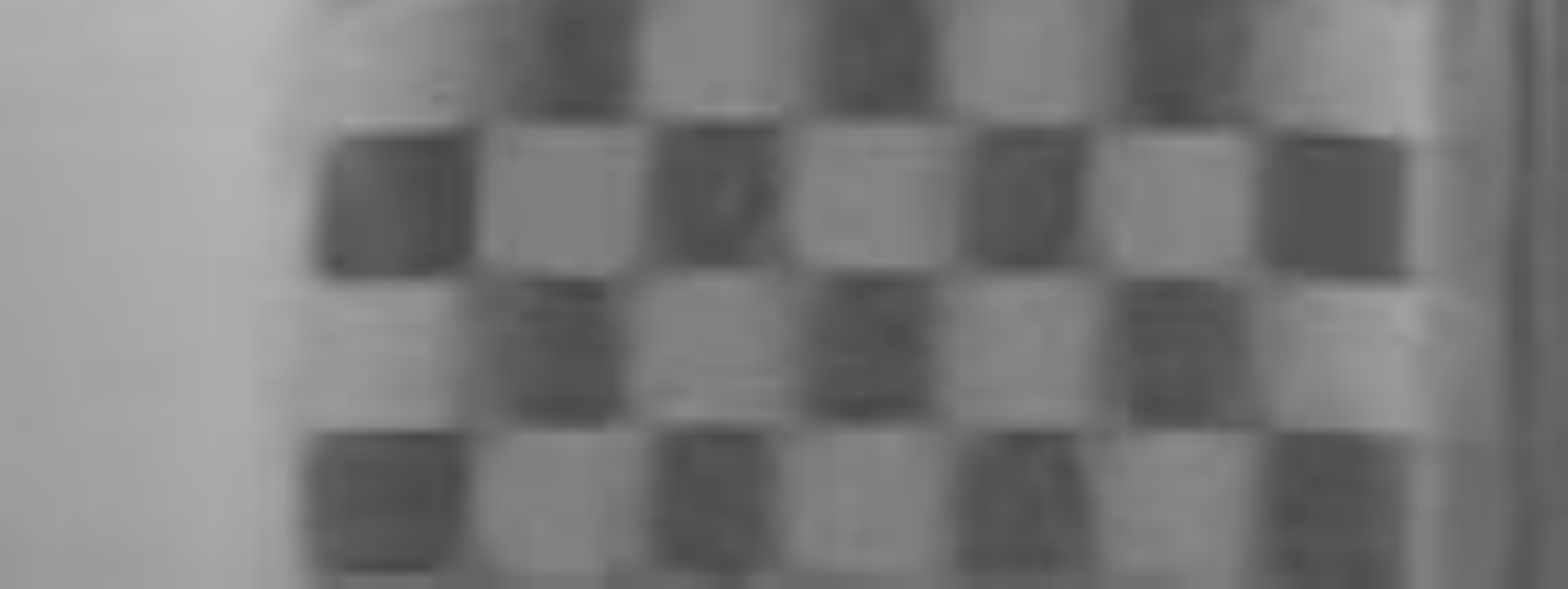}
    \caption{Jin \cite{jin2018learning}}    \label{fig:deblur_bin_reblur_halfscale_1}
  \end{subfigure}
  \begin{subfigure}{0.485\linewidth}
    \includegraphics[width=\textwidth]{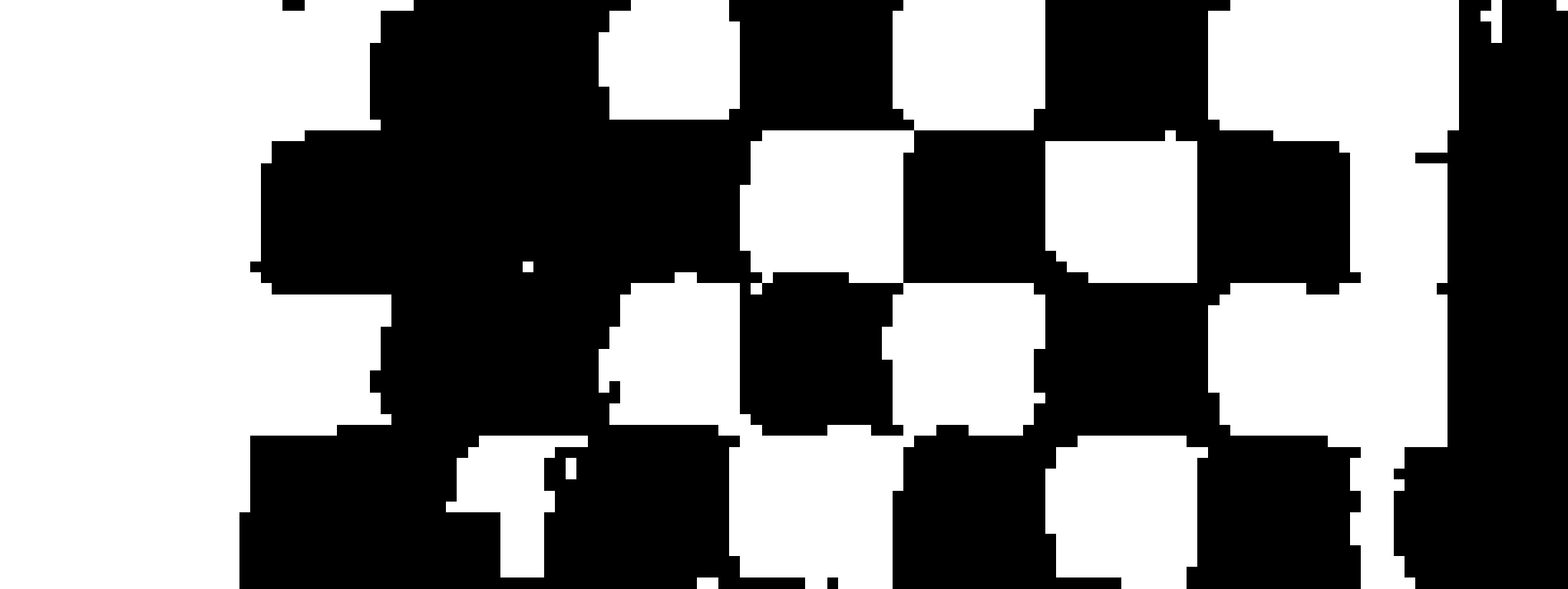}
    \caption*{Jin + WAN \cite{mustafa2018binarization}}    \label{fig:deblur_bin_reblur_halfscale_2}
  \end{subfigure}
  \begin{subfigure}{0.485\linewidth}
    \includegraphics[width=\textwidth]{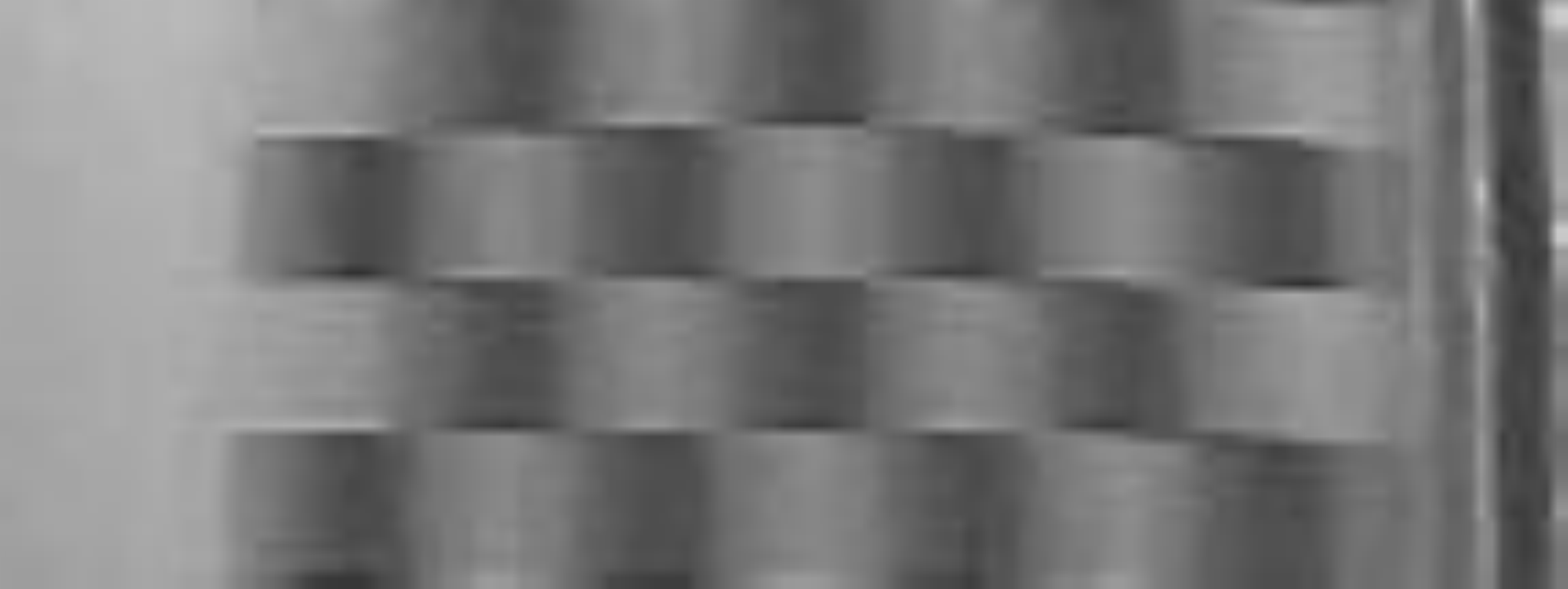}
    \caption{L0-reg \cite{textdeblur2014}}    \label{fig:deblur_bin_reblur_halfscale_3}
  \end{subfigure}
  \begin{subfigure}{0.485\linewidth}
    \includegraphics[width=\textwidth]{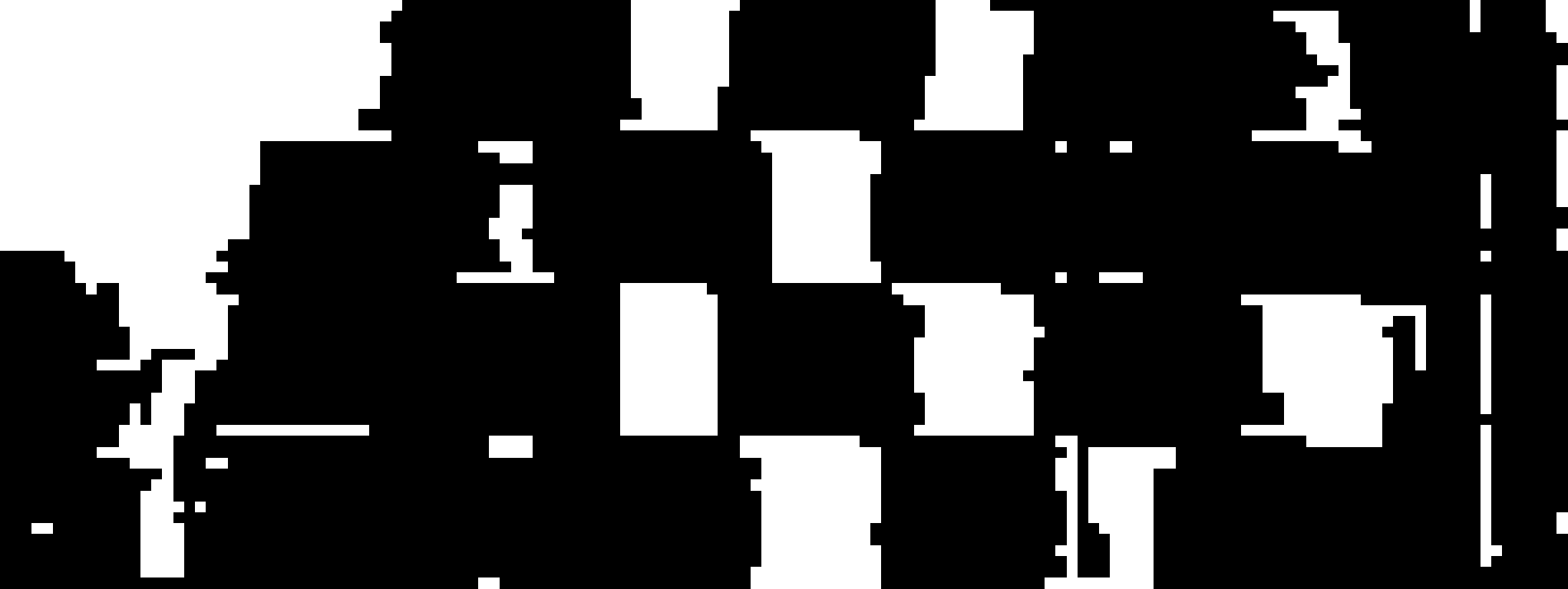}
    \caption*{L0-reg + WAN \cite{mustafa2018binarization}}    \label{fig:deblur_bin_reblur_halfscale_4}
  \end{subfigure}
  \begin{subfigure}{0.485\linewidth}
    \includegraphics[width=\textwidth]{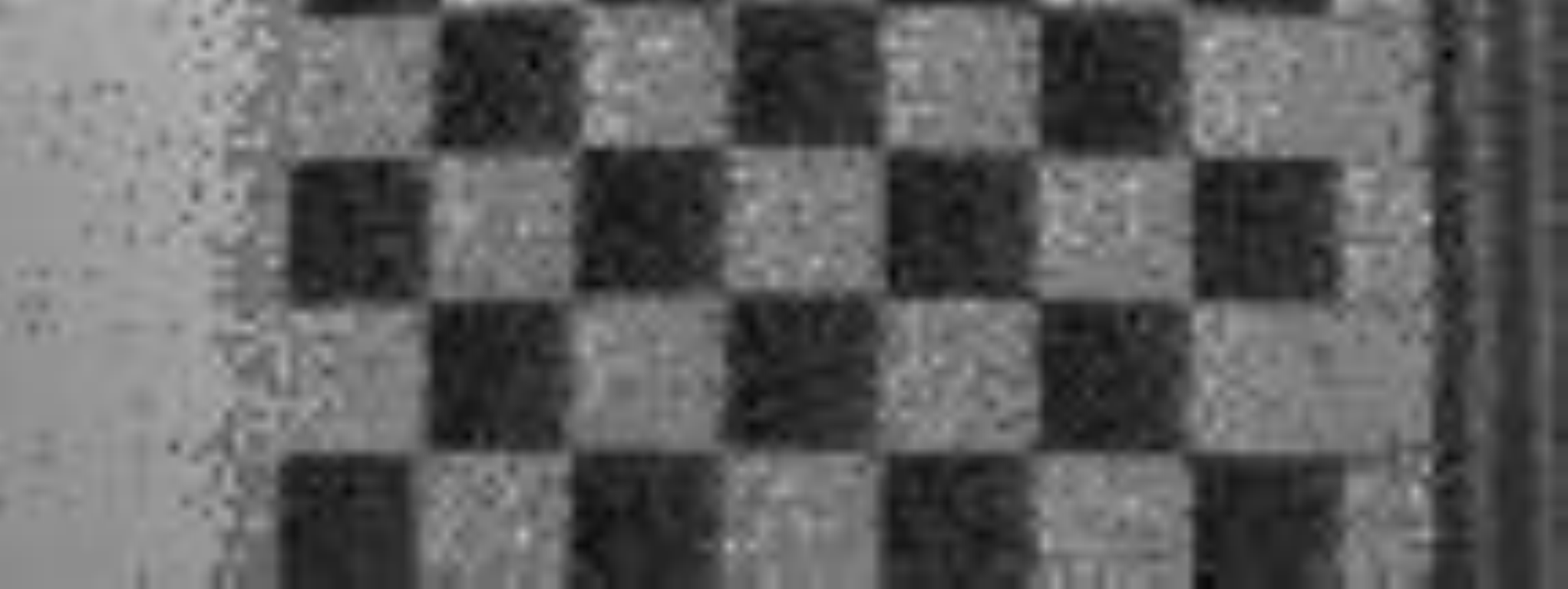}
    \caption{EDI \cite{edipami}}    \label{fig:deblur_bin_reblur_halfscale_5}
  \end{subfigure}
  \begin{subfigure}{0.485\linewidth}
    \includegraphics[width=\textwidth]{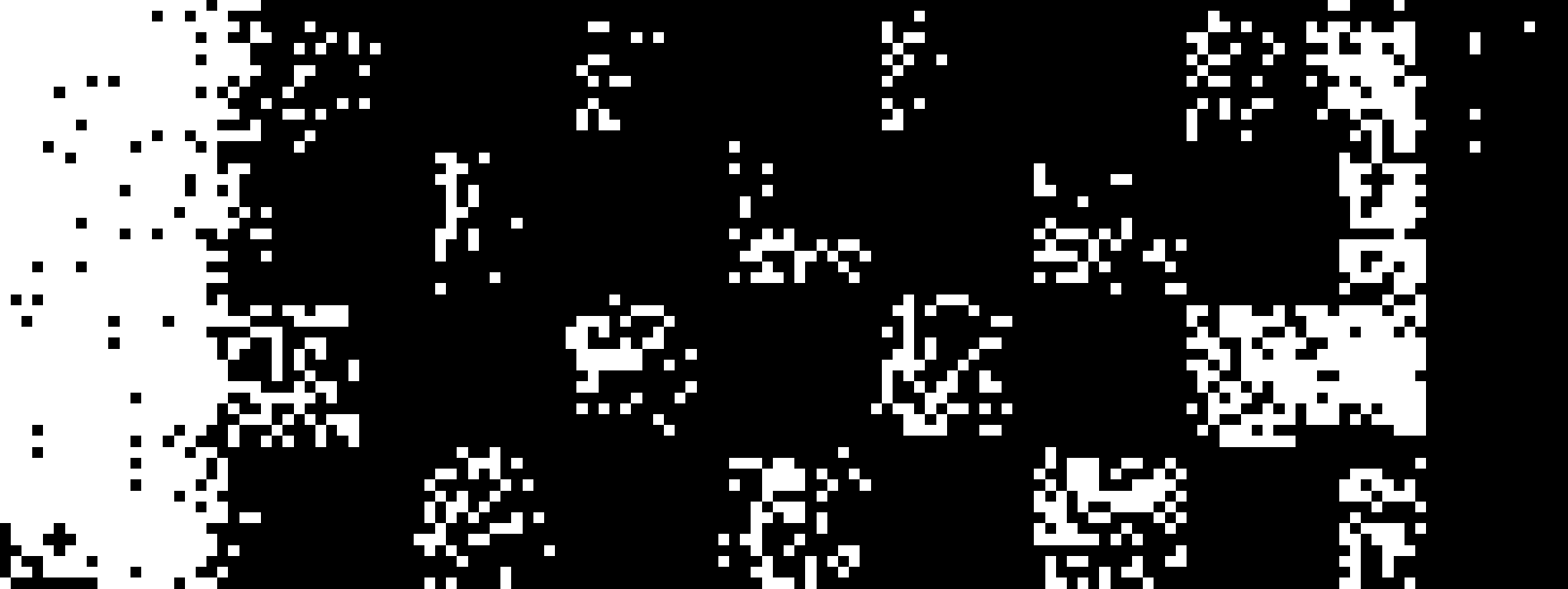}
    \caption*{EDI + WAN \cite{mustafa2018binarization}}    \label{fig:deblur_bin_reblur_halfscale_6}
  \end{subfigure}
    \begin{subfigure}{0.485\linewidth}
    \includegraphics[width=\textwidth]{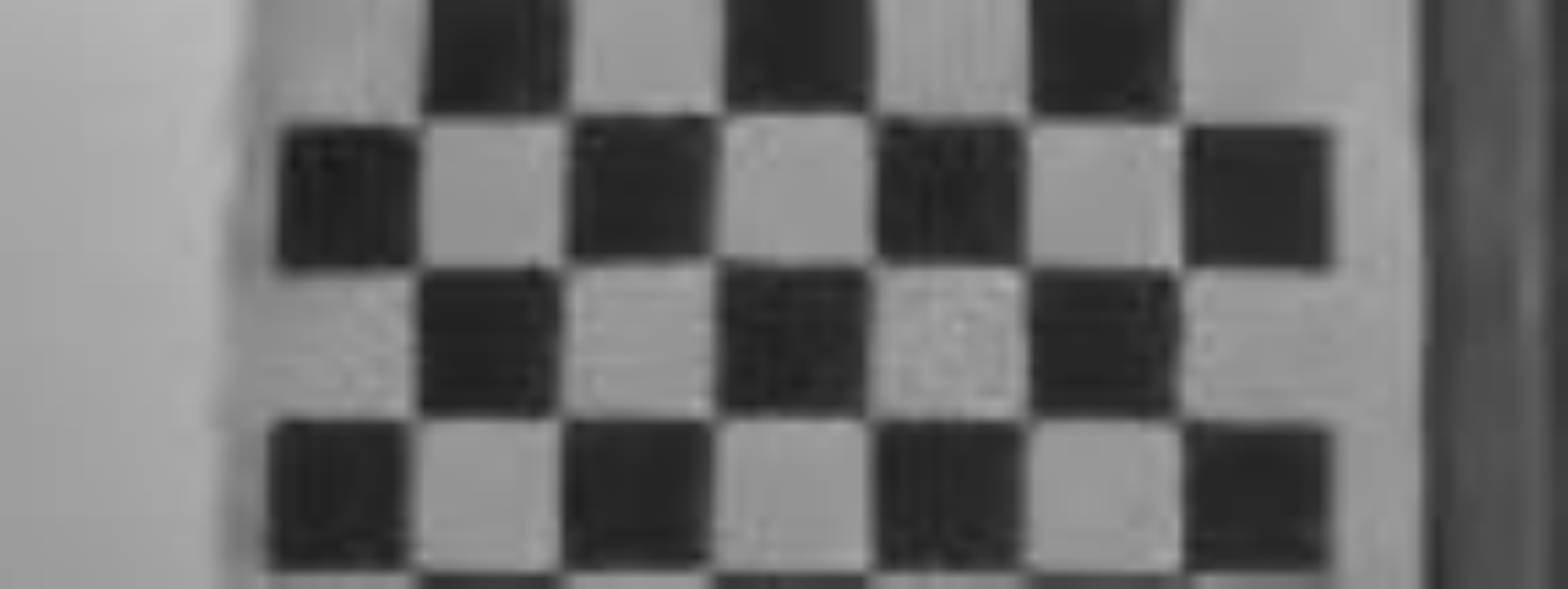}
    \caption{eSL \cite{yu2023learning} }    \label{fig:deblur_bin_reblur_halfscale_7}
  \end{subfigure}
    \begin{subfigure}{0.485\linewidth}
    \includegraphics[width=\textwidth]{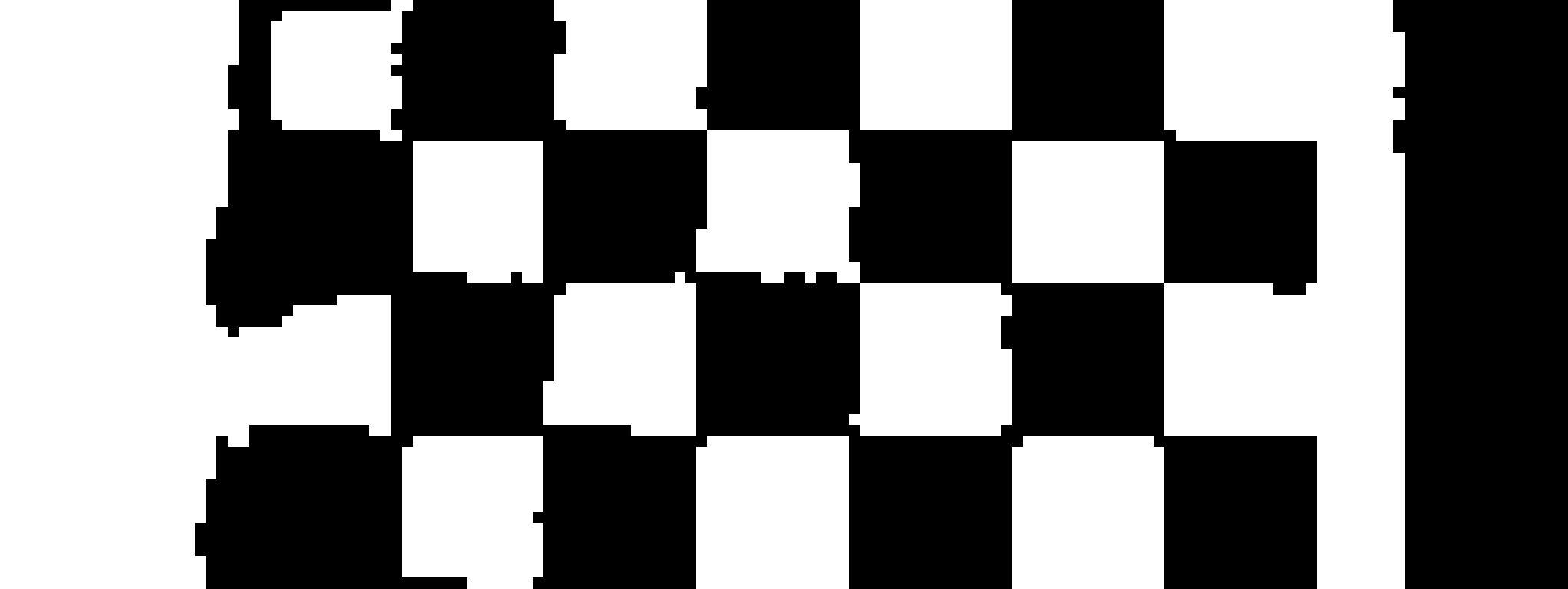}
    \caption*{eSL + WAN \cite{mustafa2018binarization}}    \label{fig:deblur_bin_reblur_halfscale_8}
  \end{subfigure}
  \begin{subfigure}{0.485\linewidth}
    \includegraphics[width=\textwidth]{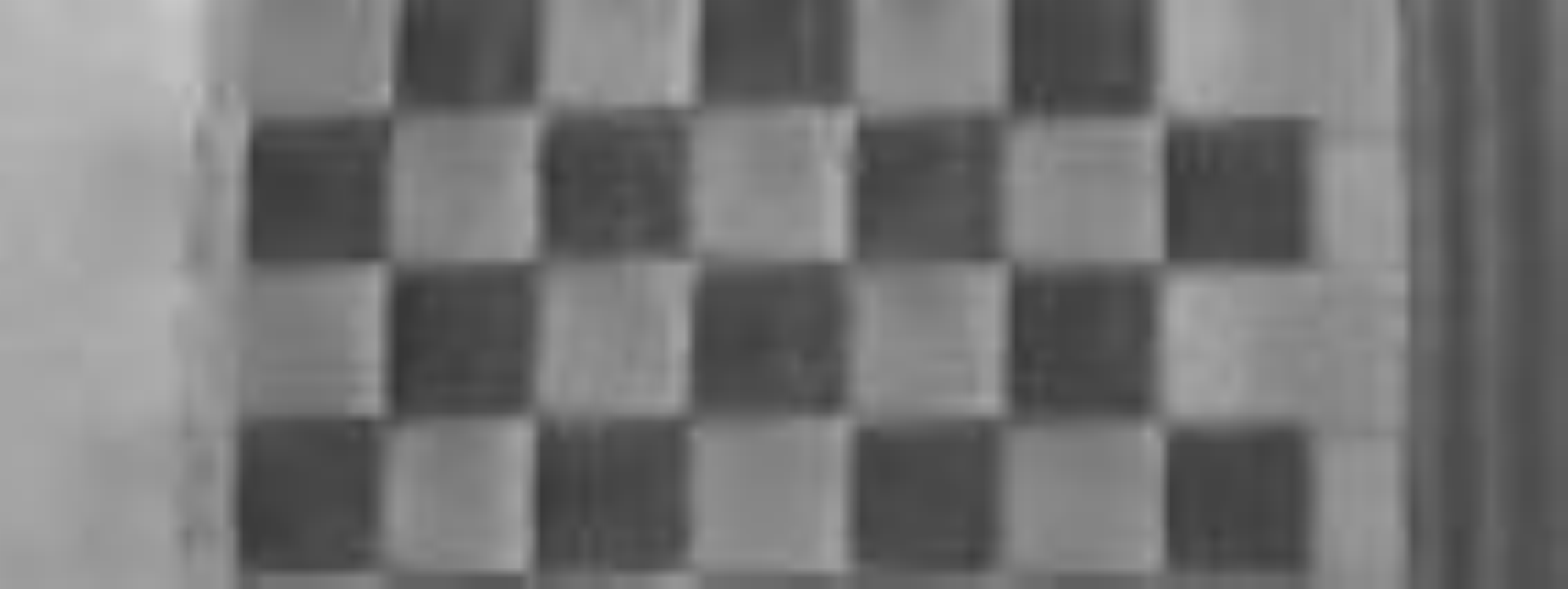}
    \caption{LEDVDI \cite{lin2020learning}}    \label{fig:deblur_bin_reblur_halfscale_9}
  \end{subfigure}
    \begin{subfigure}{0.485\linewidth}
    \includegraphics[width=\textwidth]{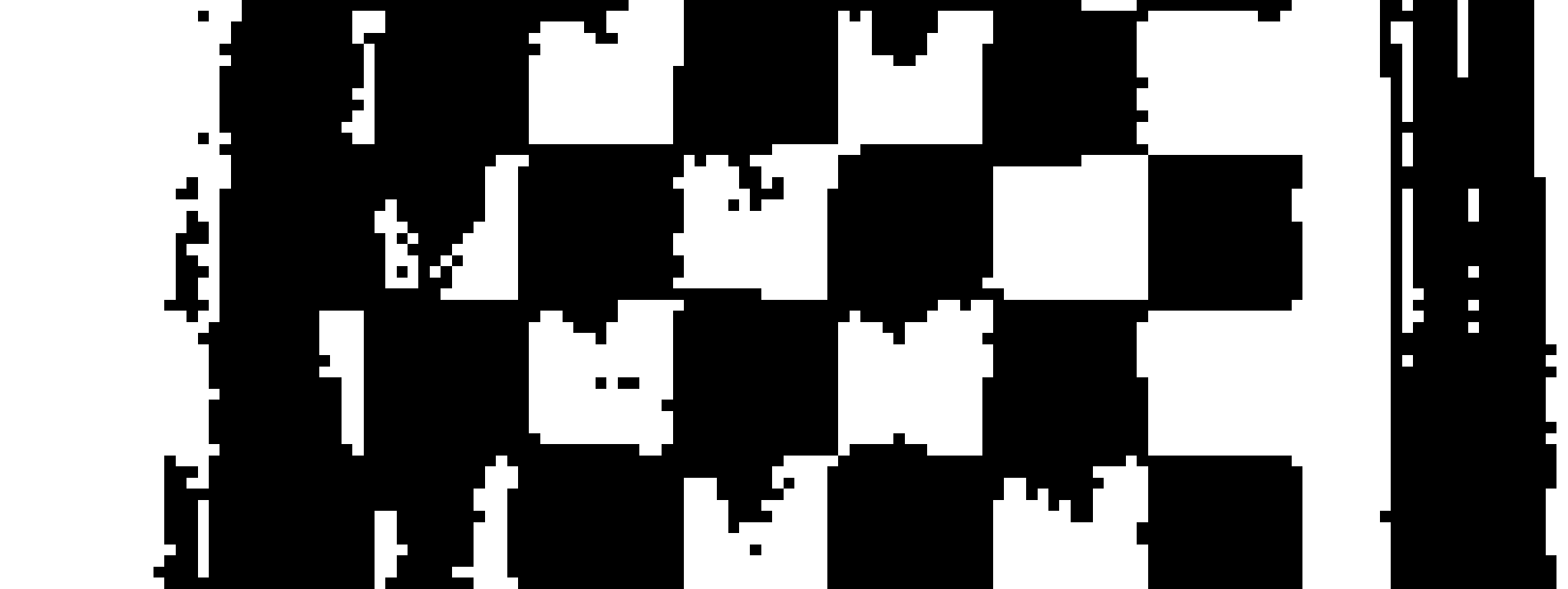}
    \caption*{LEDVDI + WAN \cite{mustafa2018binarization}}    \label{fig:deblur_bin_reblur_halfscale_10}
  \end{subfigure}
  \begin{subfigure}{0.485\linewidth}
    \includegraphics[width=\textwidth]{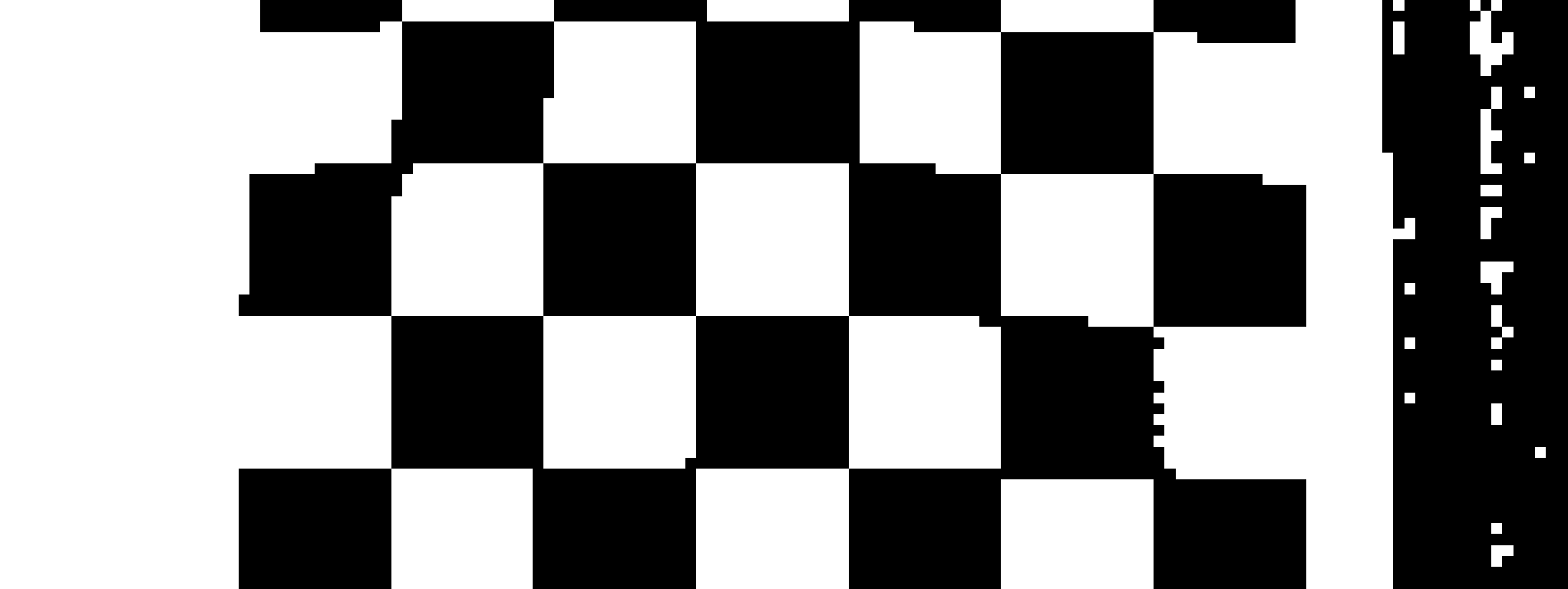}
    \caption{Ground truth}    \label{fig:deblur_bin_reblur_halfscale_9}
  \end{subfigure}
    \begin{subfigure}{0.485\linewidth}
    \includegraphics[width=\textwidth]{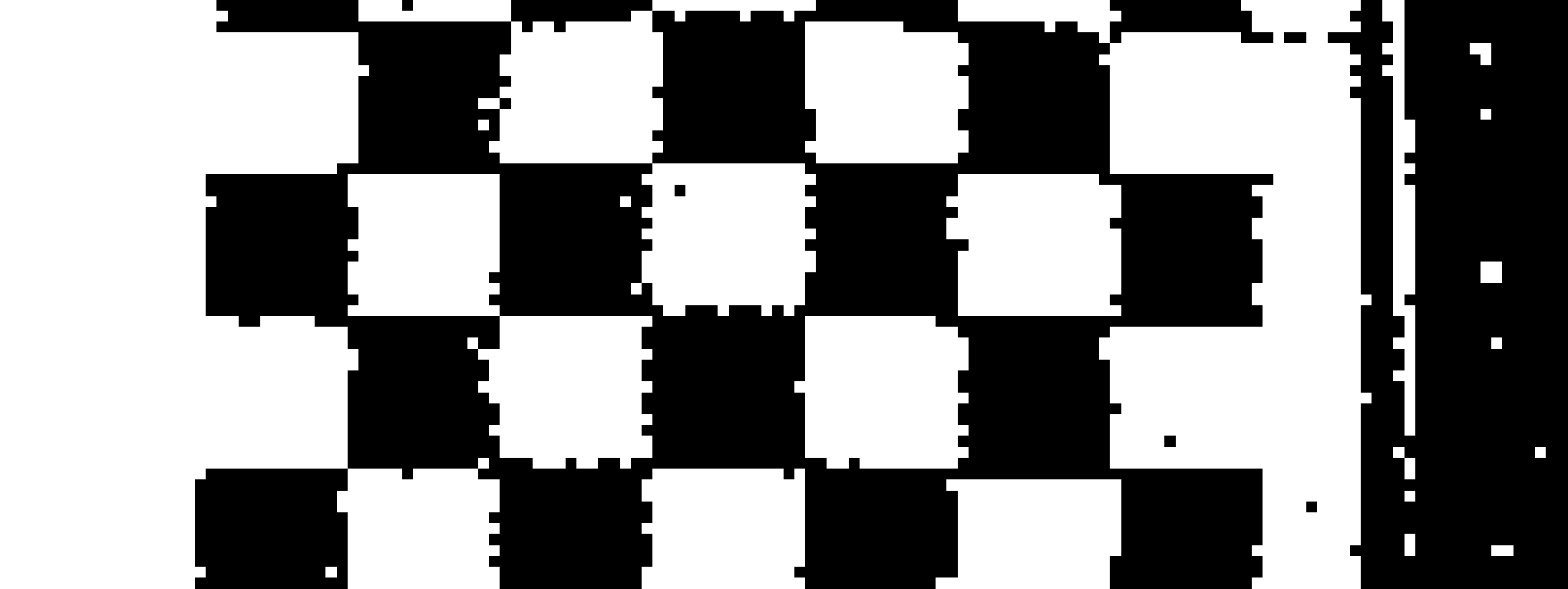}
    \caption{\textbf{Ours}}    \label{fig:deblur_bin_reblur_halfscale_10}
  \end{subfigure}
\end{minipage}
\caption{\lsj{Comparison with state-of-the-art motion deblurring methods for binary images on different datasets: our EBT (left column),  HQF \cite{stoffregen2020reducing} dataset (middle column), and Reblur \cite{sun2022event} dataset (right column). 
}}
\label{fig:deblur_ib}
\end{figure*}

\begin{figure*}[ht!]
\begin{subfigure}{0.335\linewidth}
\includegraphics[width=\textwidth]{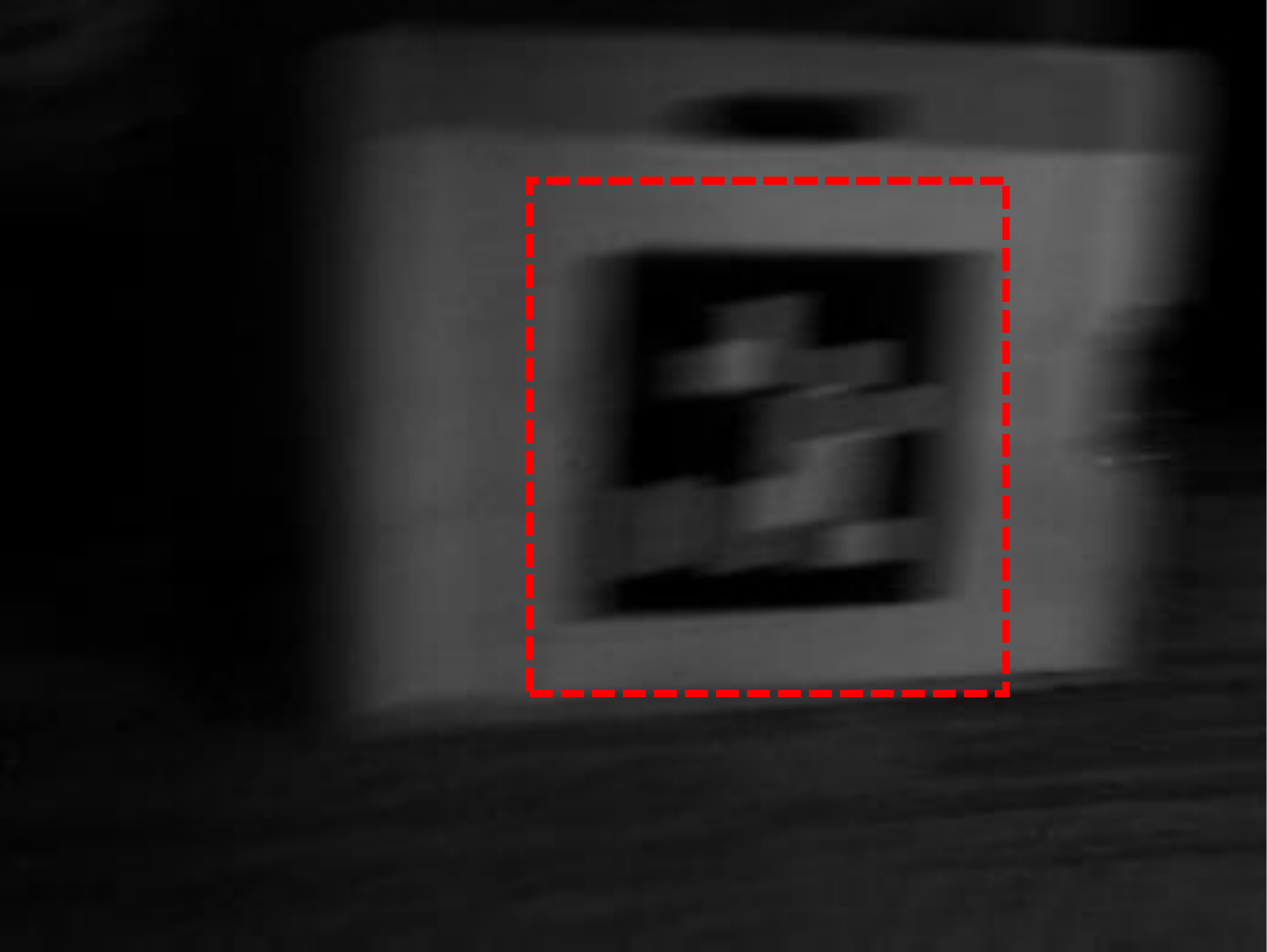}
\caption{The blurred image}    \label{fig:video_ib_ebt_real_1}
\end{subfigure}
\hfill
\begin{subfigure}{0.3\linewidth}
\includegraphics[width=\textwidth]{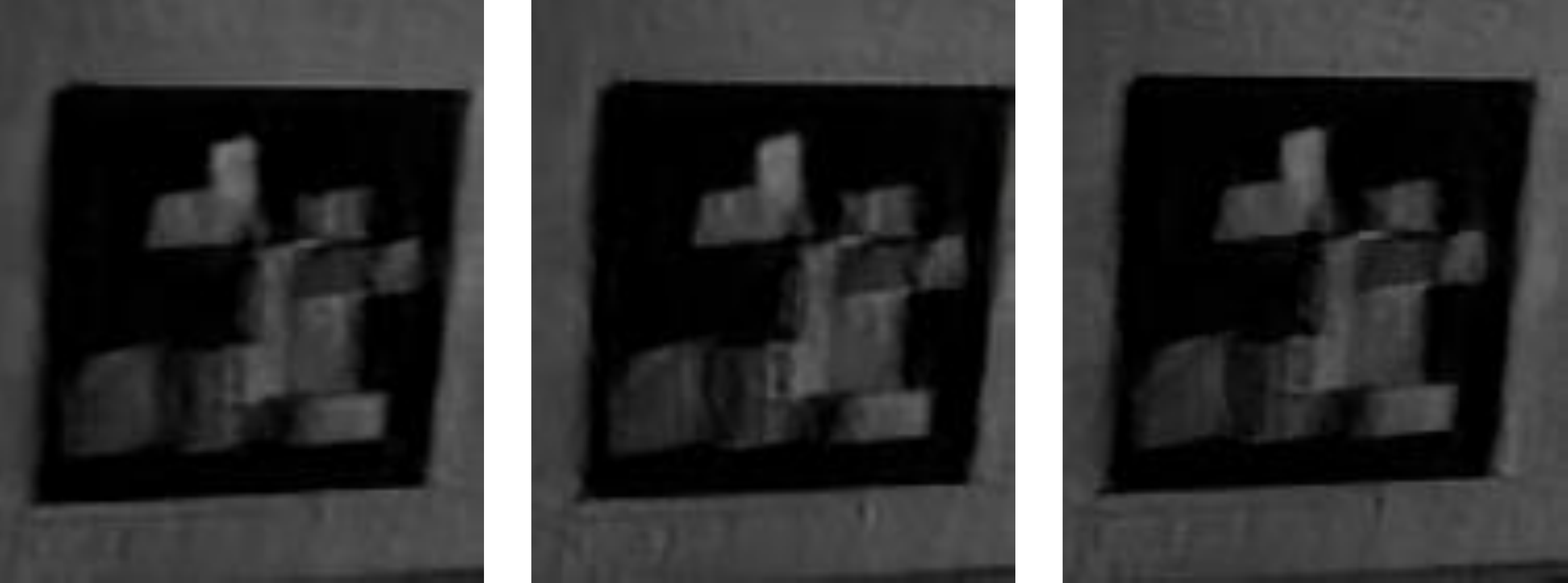}
\caption*{Jin \cite{jin2018learning}}
\includegraphics[width=\textwidth]{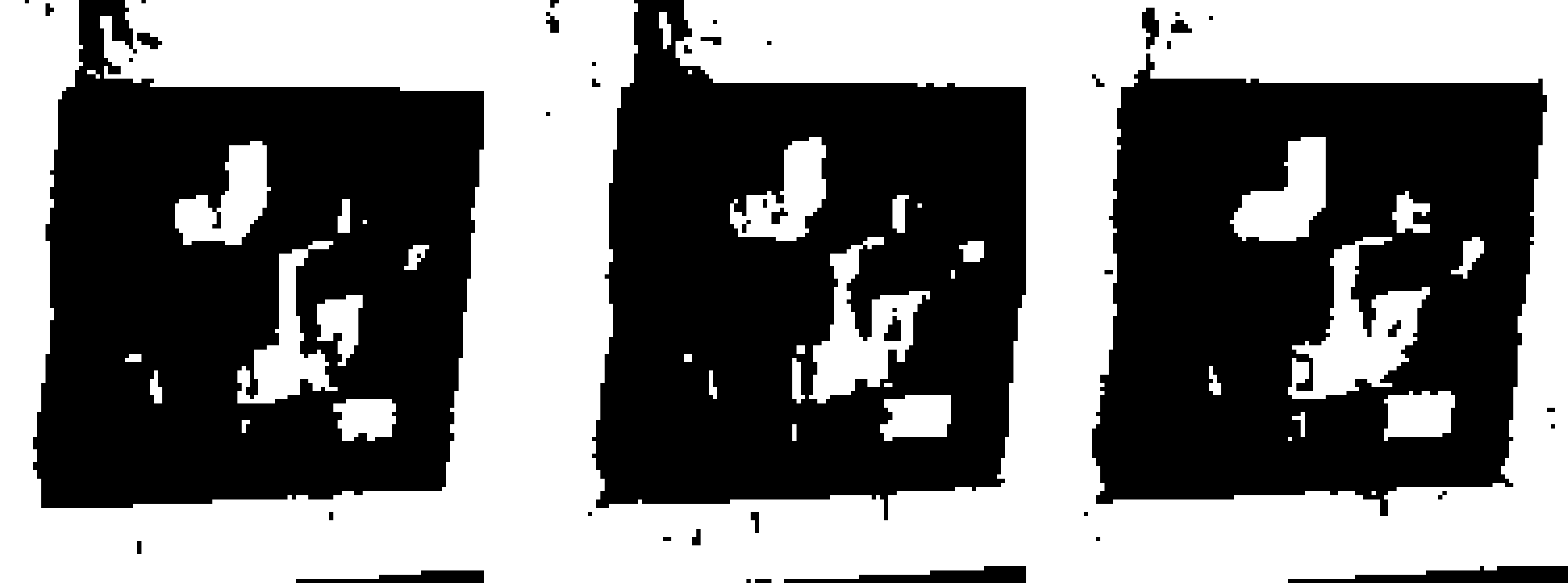}
\caption{Jin + Wan \cite{mustafa2018binarization}}    \label{fig:video_ib_ebt_real_recale_2}
\end{subfigure}
\hfill
\begin{subfigure}{0.3\linewidth}
\includegraphics[width=\textwidth]{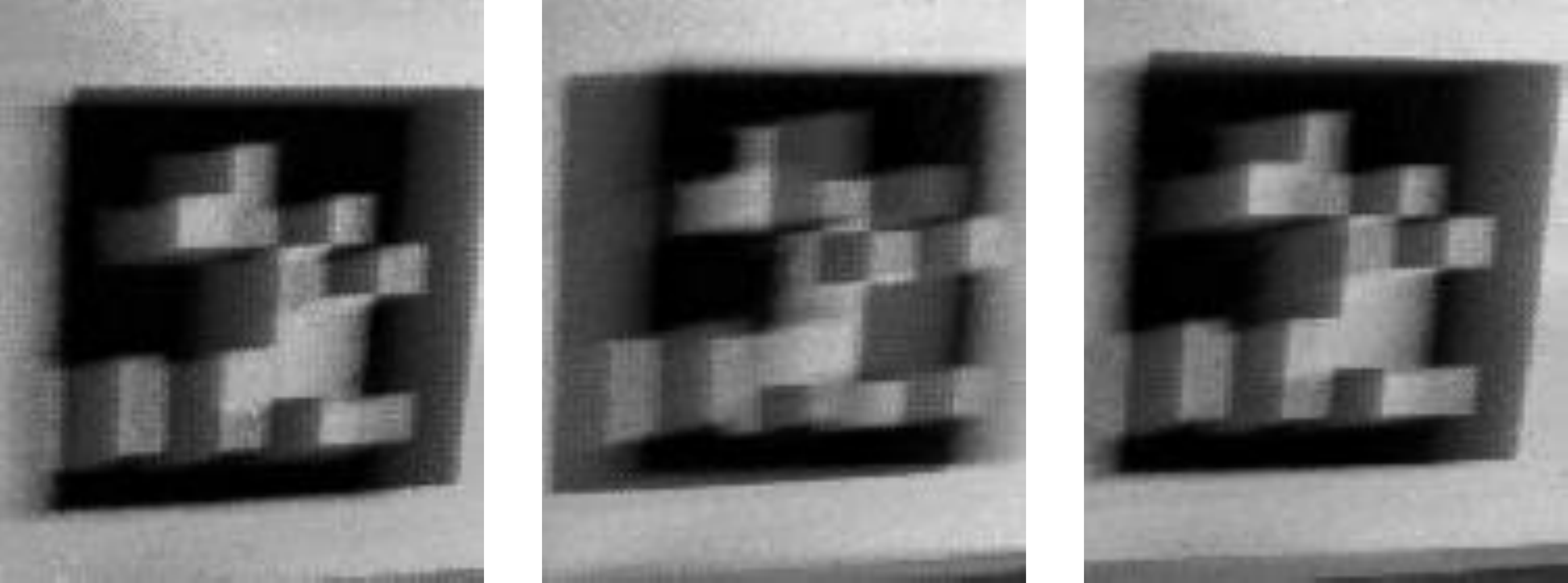}
\caption*{EDI \cite{edipami}}
\includegraphics[width=\textwidth]{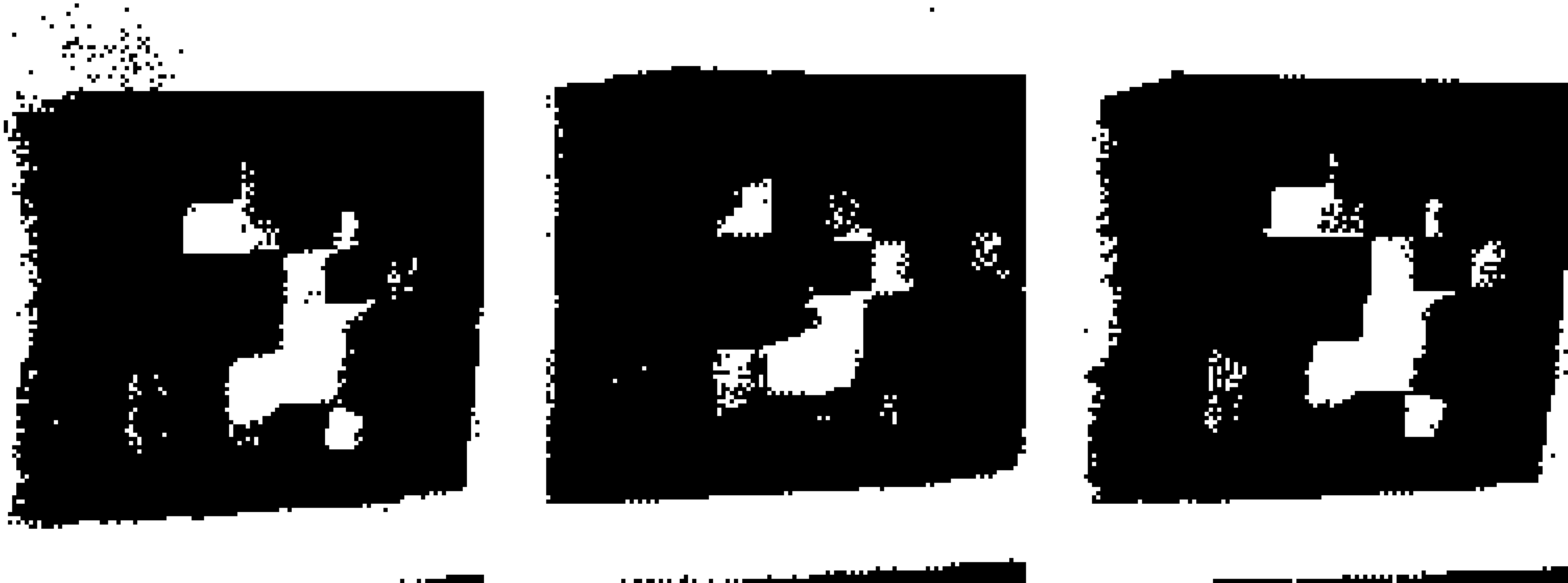}
\caption{EDI + Wan \cite{mustafa2018binarization}}    \label{fig:video_ib_ebt_real_recale_4}
\end{subfigure}
\vfill
\begin{subfigure}{0.3\linewidth}
\includegraphics[width=\textwidth]{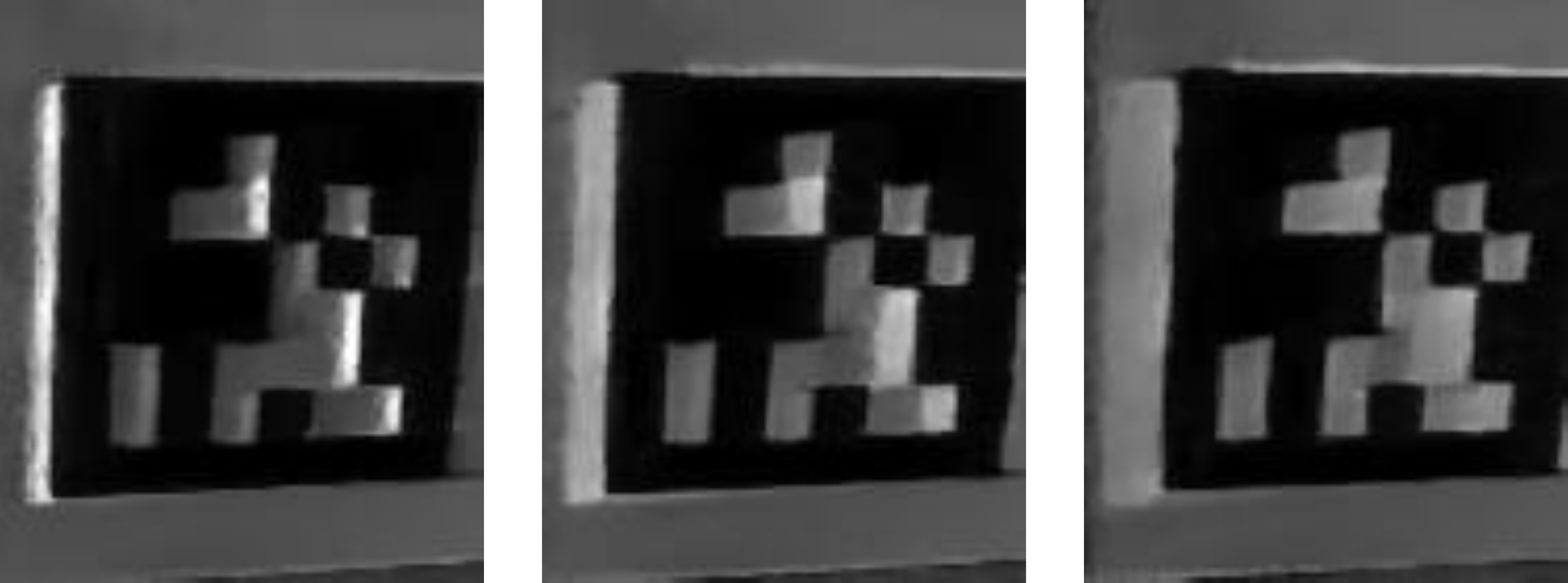}
\caption*{eSL \cite{yu2023learning}} 
\includegraphics[width=\textwidth]{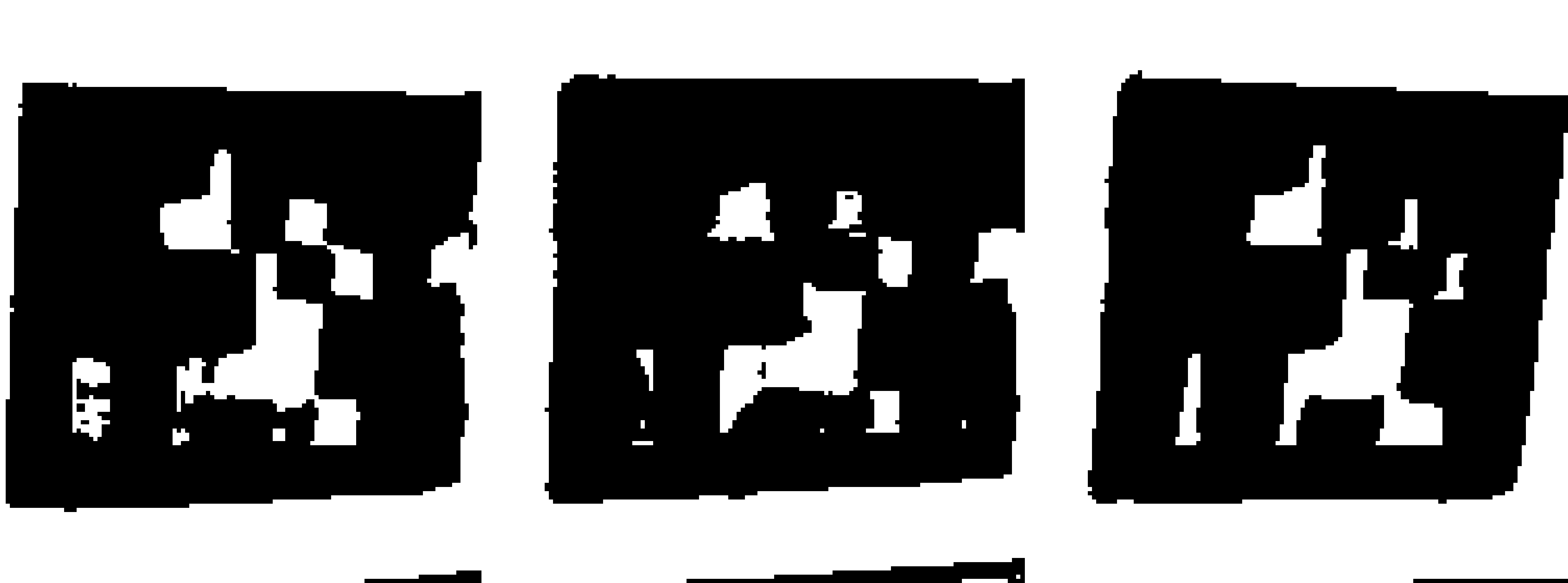}
\caption{eSL + Wan \cite{mustafa2018binarization}}    \label{fig:video_ib_ebt_real_recale_6}
\end{subfigure}
\hfill
\begin{subfigure}{0.3\linewidth}
\includegraphics[width=\textwidth]{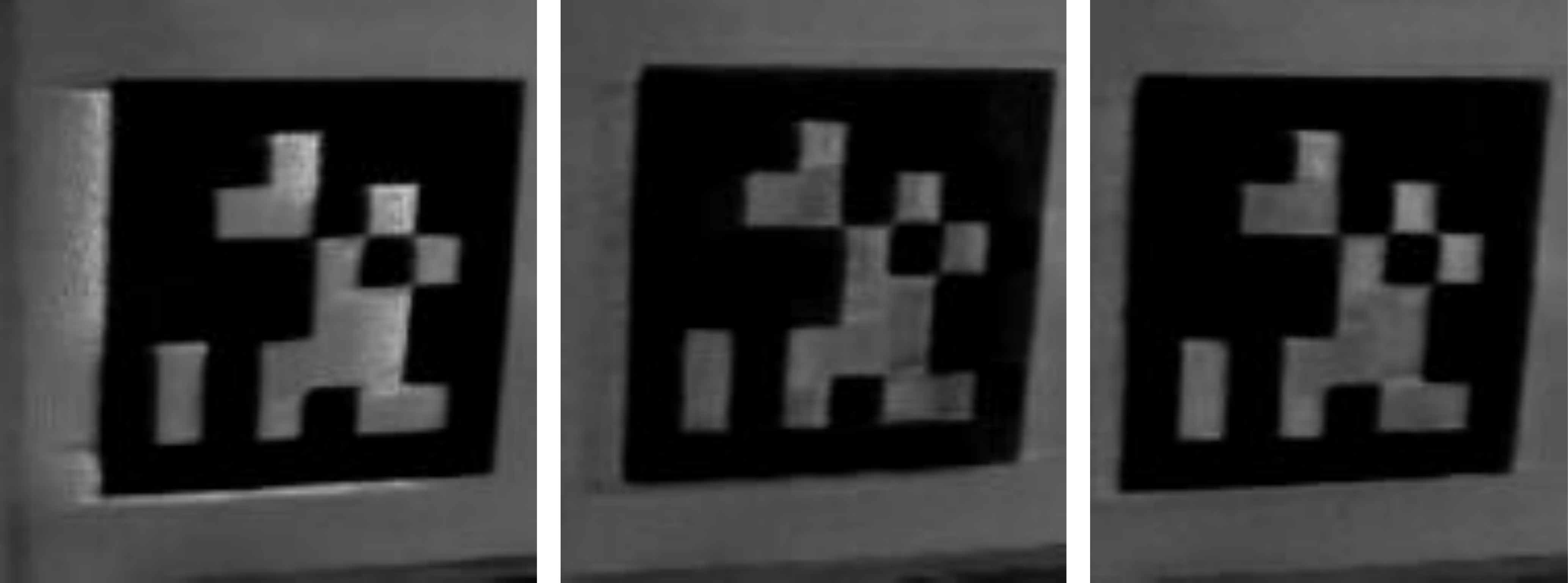}
\caption*{LEDVDI \cite{lin2020learning}} 
\includegraphics[width=\textwidth]{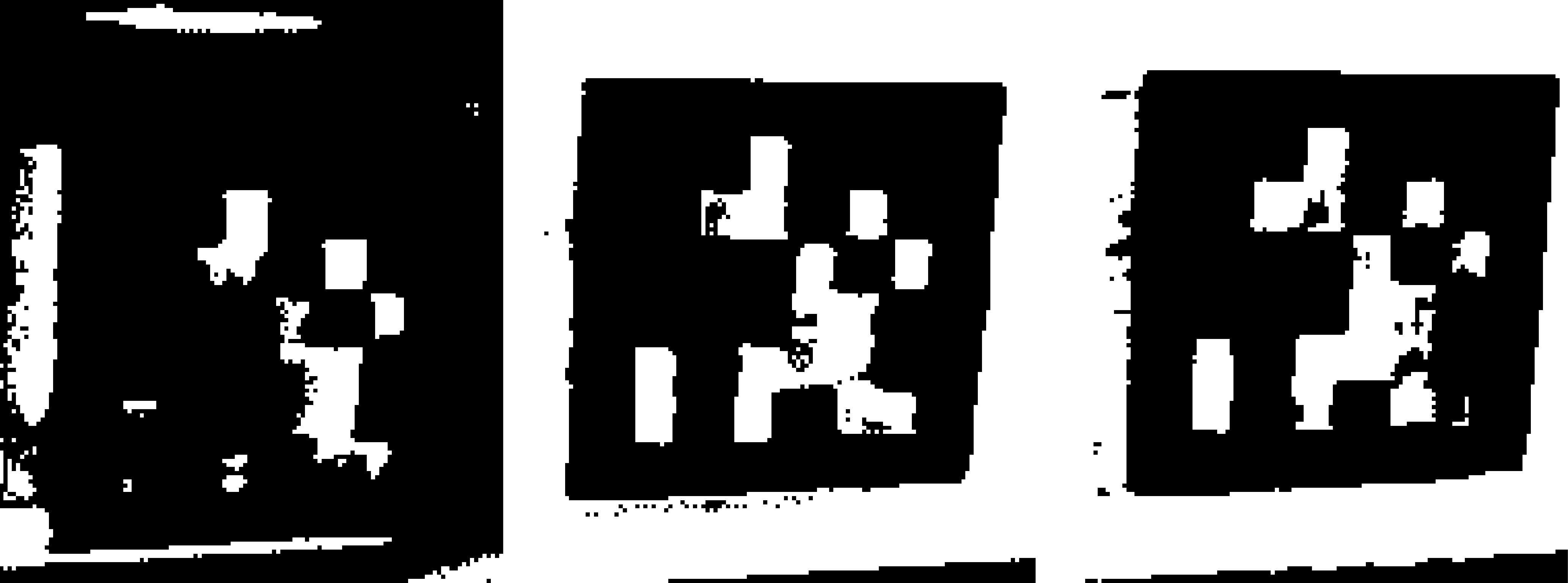}
\caption{LEDVDI + Wan \cite{mustafa2018binarization}}    \label{fig:video_ib_ebt_real_recale_8}
\end{subfigure}
\hfill
\begin{subfigure}{0.3\linewidth}
\includegraphics[width=\textwidth]{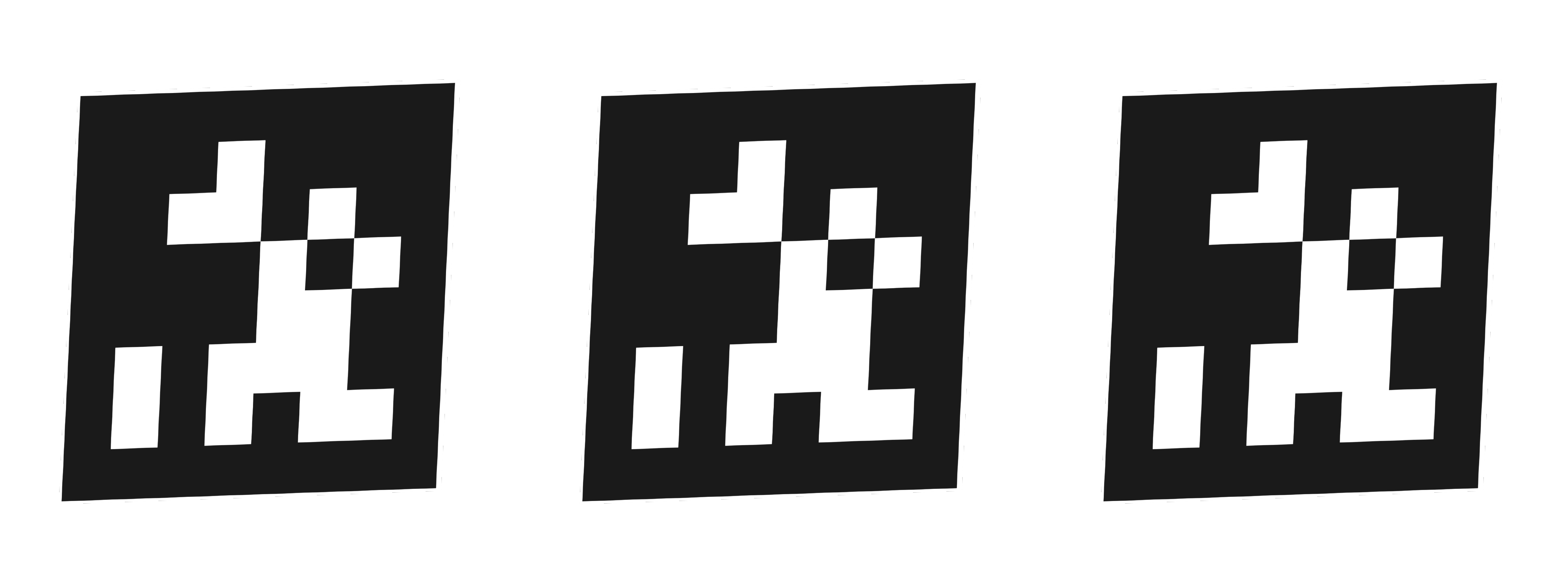}
\caption{Apriltag ID:0}    \label{fig:video_ib_ebt_real_rescale_9}
\includegraphics[width=\textwidth]{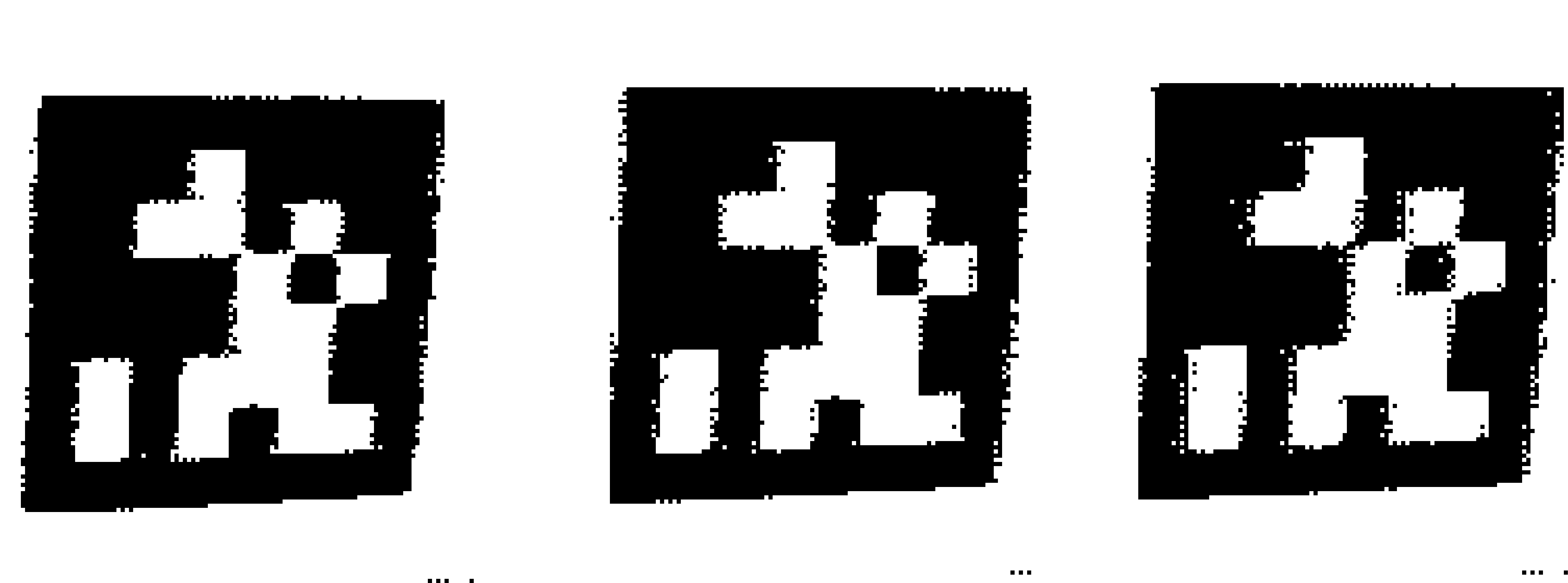}
\caption{\textbf{Ours}}    \label{fig:video_ib_ebt_real_recale_10}
\end{subfigure}
\caption{\lsj{A video of a real-world sequence from the EBT dataset. (b), (c), (d), and (e) represent the cropped video frames of Jin \cite{jin2018learning}, EDI \cite{edipami}, eSL \cite{yu2023learning}, and LEDVDI \cite{yu2023learning}, resp., along with their combination with Wan \cite{mustafa2018binarization}. (f) and (g) are the GT Apirltag and our result, resp.}}
  \label{fig:video_bin_ebt_real}
\end{figure*}

\begin{figure*}[ht!]
\begin{subfigure}{0.335\linewidth}
\includegraphics[width=\textwidth]{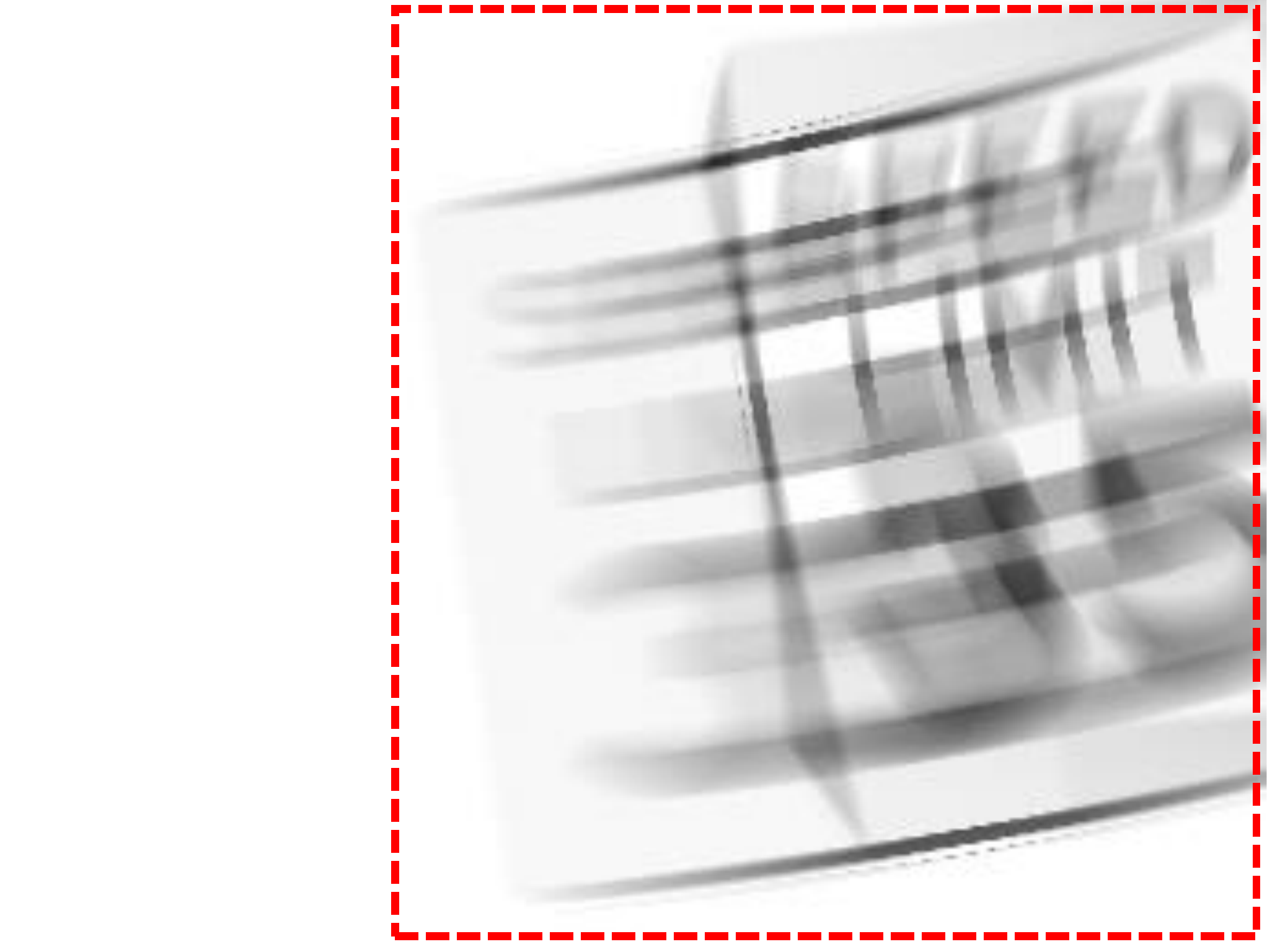}
\caption{The blurred image}    \label{fig:video_ib_ebt_sim}
\end{subfigure}
\hfill
\begin{subfigure}{0.3\linewidth}
\includegraphics[width=\textwidth]{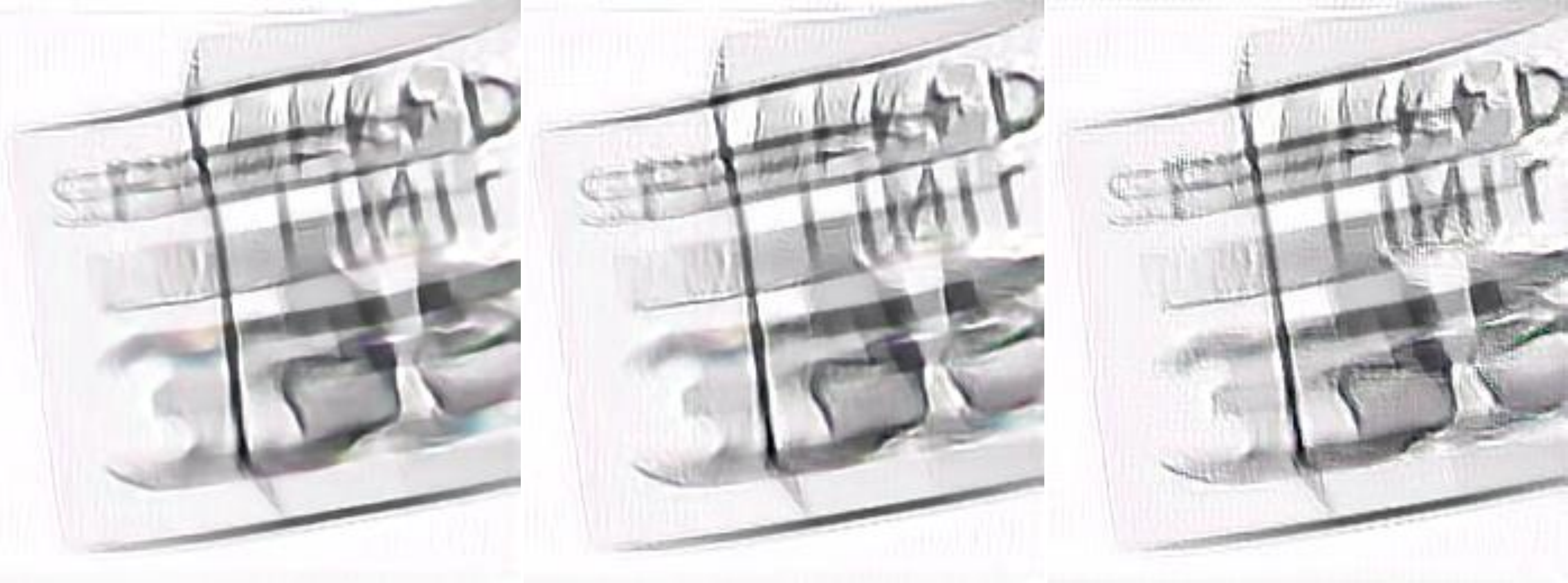}
\caption*{Jin \cite{jin2018learning}}
\includegraphics[width=\textwidth]{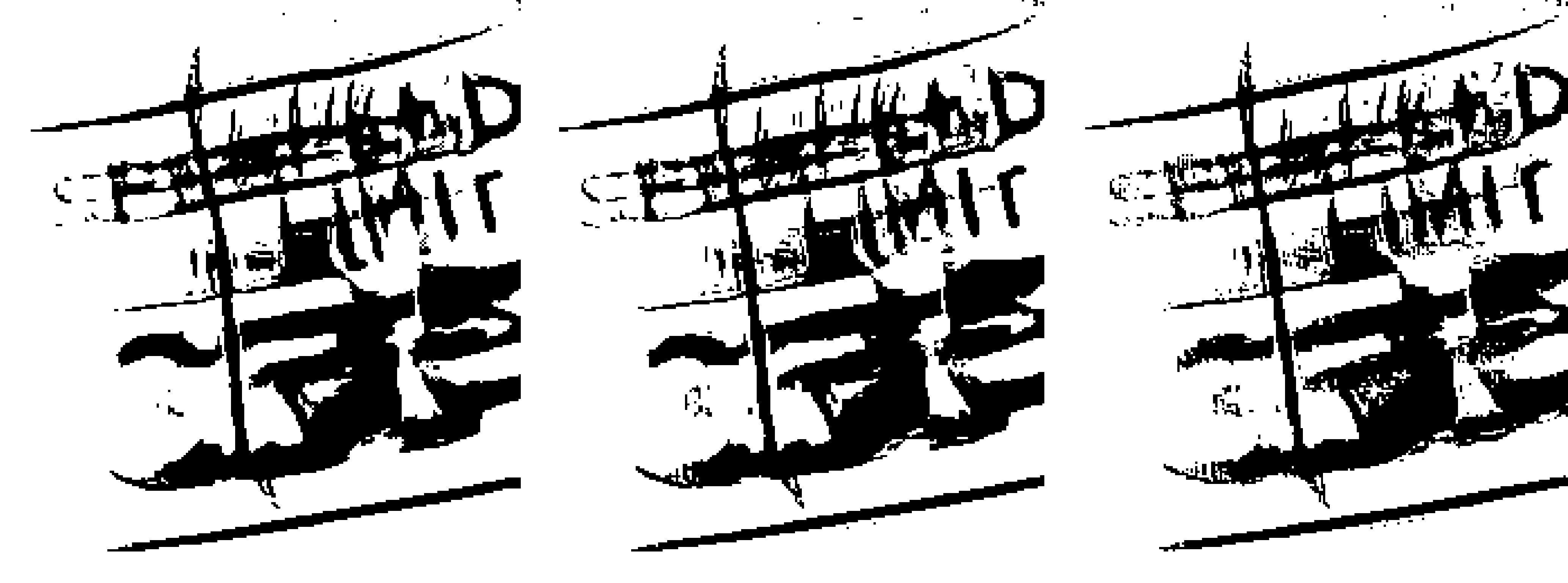}
\caption{Jin + Wan \cite{mustafa2018binarization}}    \label{fig:video_ib_ebt_sim_rescale_2}
\end{subfigure}
\hfill
\begin{subfigure}{0.3\linewidth}
\includegraphics[width=\textwidth]{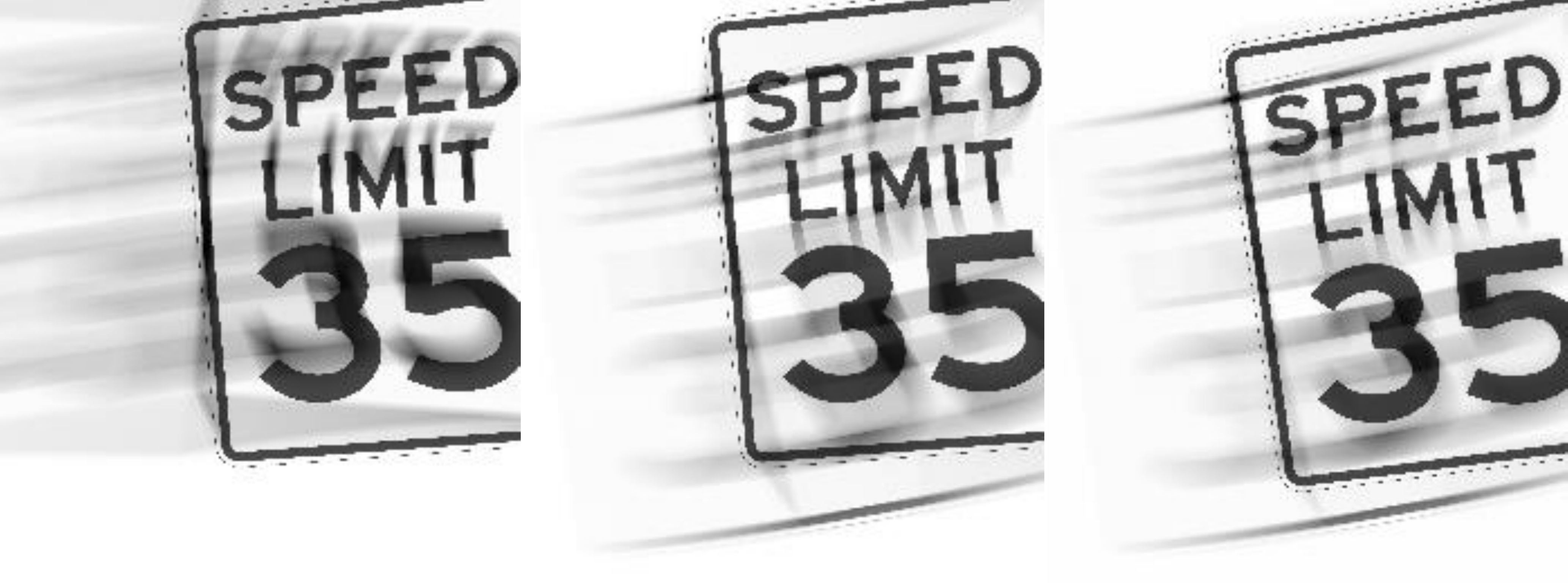}
\caption*{EDI \cite{edipami}}
\includegraphics[width=\textwidth]{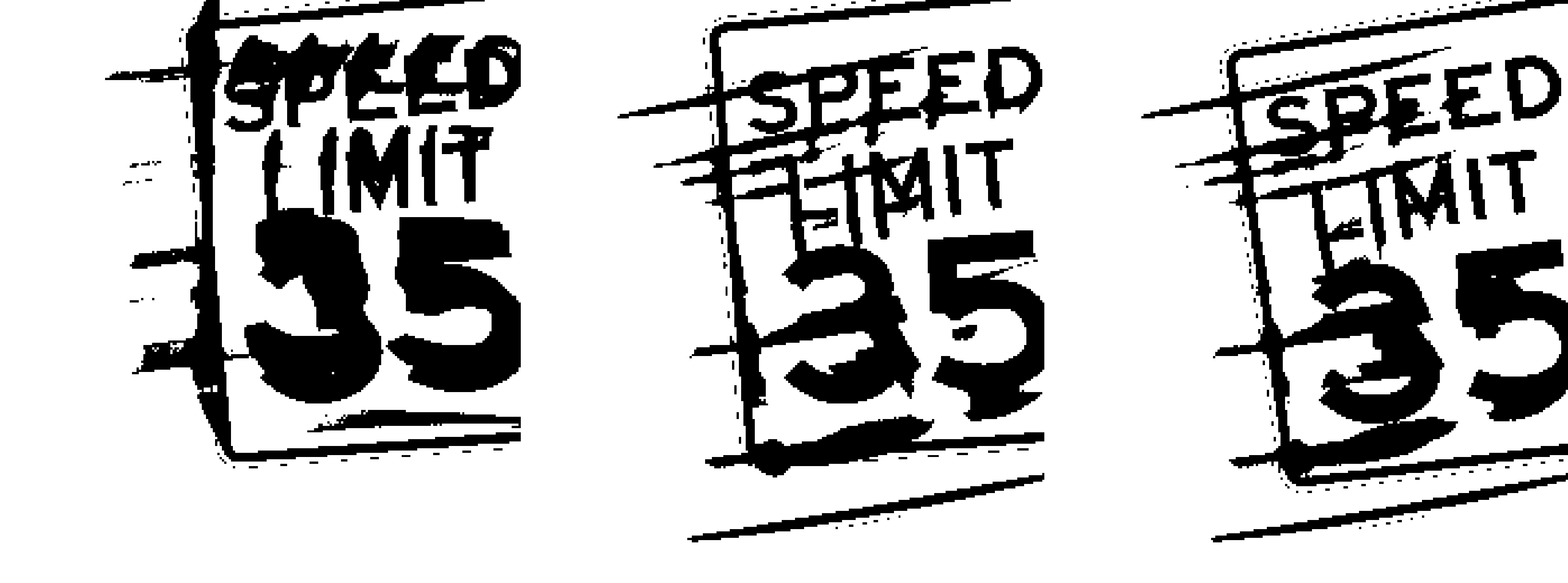}
\caption{EDI + Wan \cite{mustafa2018binarization}}    \label{fig:video_ib_ebt_sim_rescale_4}
\end{subfigure}
\hfill
\begin{subfigure}{0.3\linewidth}
\includegraphics[width=\textwidth]{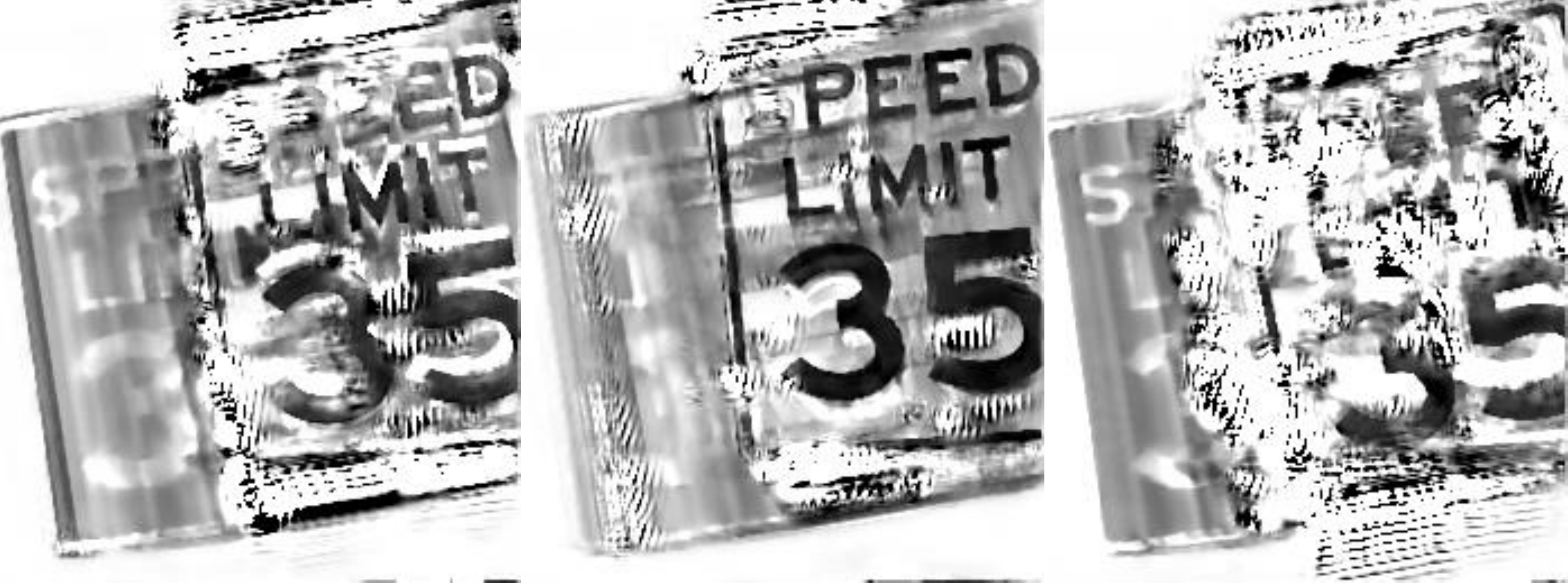}
\caption*{eSL \cite{yu2023learning}} 
\includegraphics[width=\textwidth]{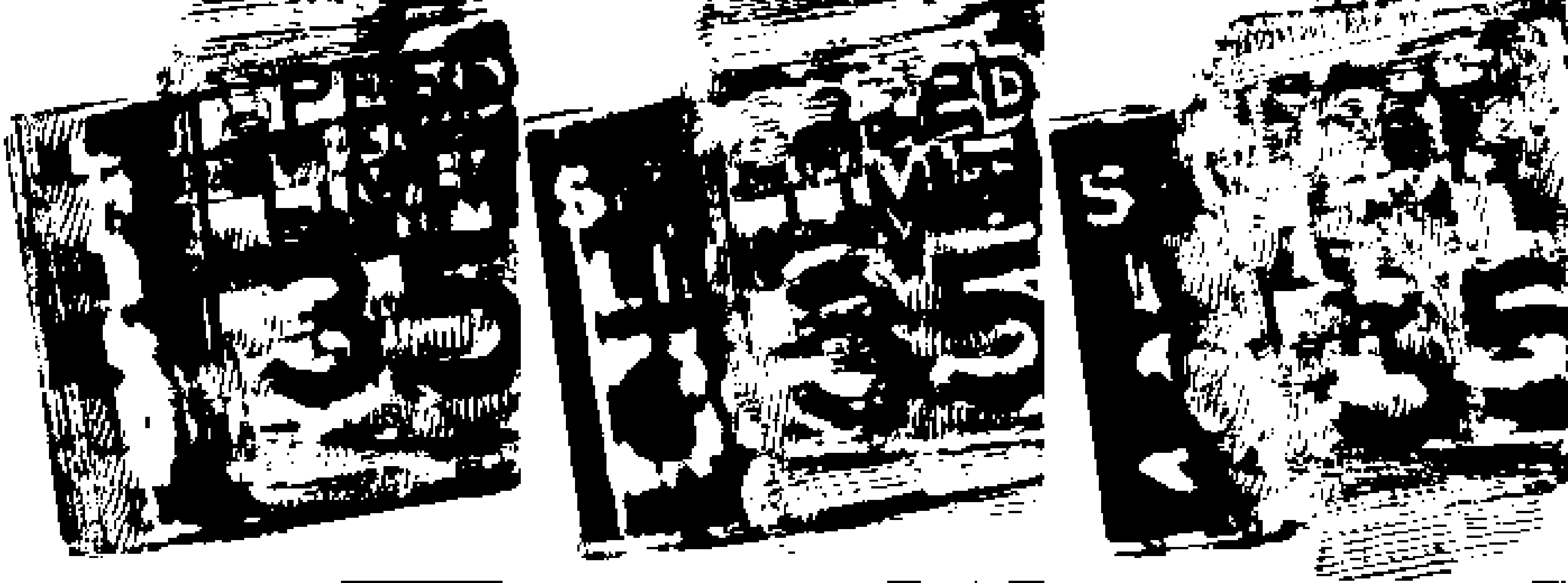}
\caption{eSL + Wan \cite{mustafa2018binarization}}    \label{fig:video_ib_ebt_sim_rescale_6}
\end{subfigure}
\hfill
\begin{subfigure}{0.3\linewidth}
\includegraphics[width=\textwidth]{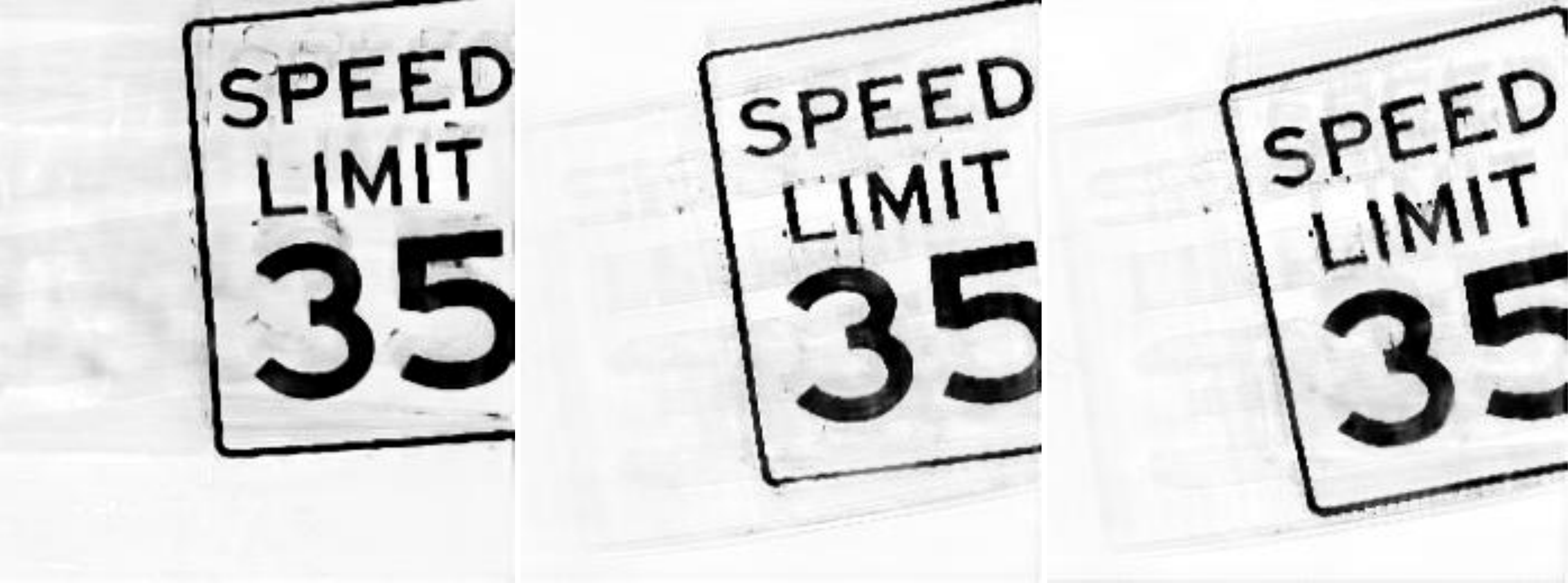}
\caption*{LEDVDI \cite{lin2020learning}}
\includegraphics[width=\textwidth]{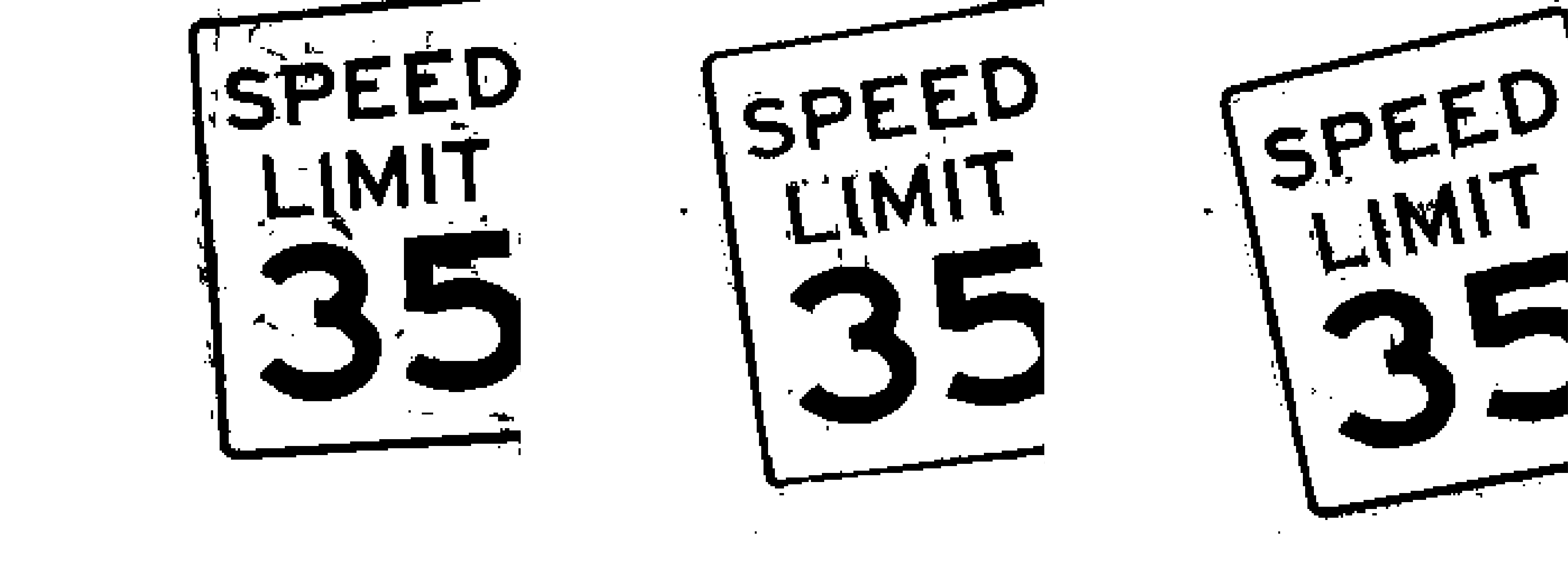}
\caption{LEDVDI + Wan \cite{mustafa2018binarization}}    \label{fig:video_ib_ebt_sim_rescale_8}
\end{subfigure}
\hfill
\begin{subfigure}{0.3\linewidth}
\includegraphics[width=\textwidth]{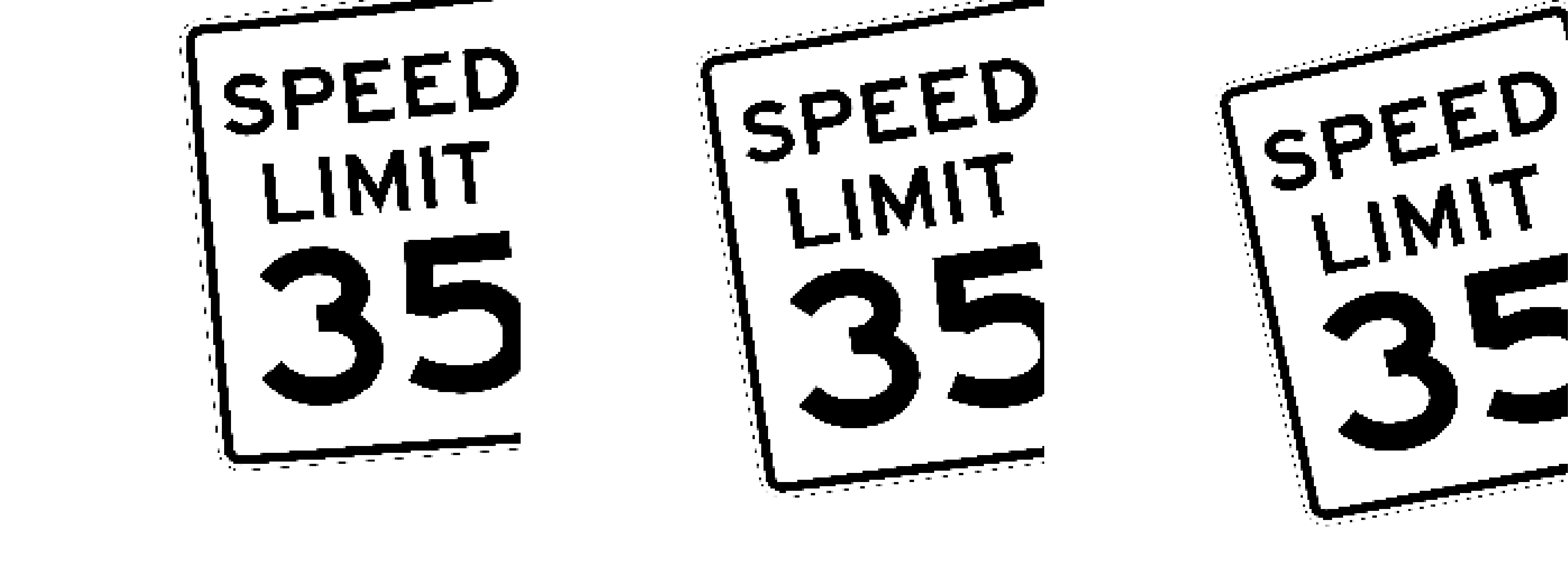}
\caption{Ground truth}
\includegraphics[width=\textwidth]{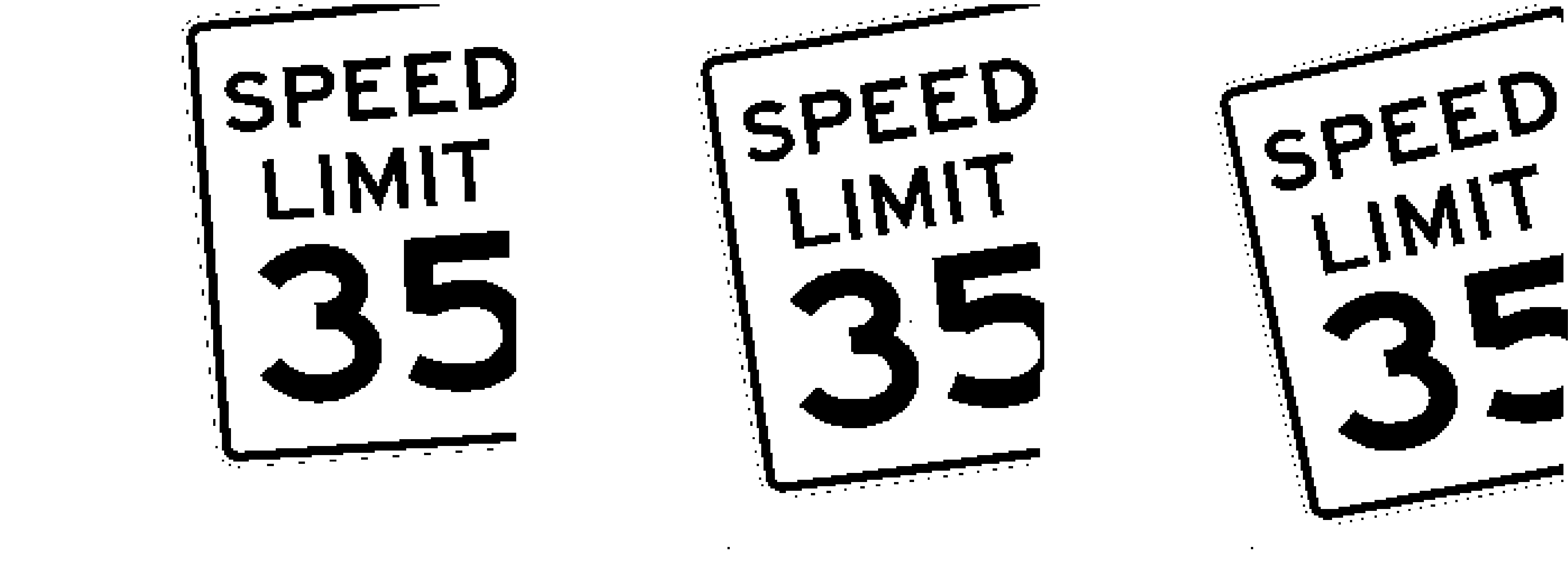}
\caption{\textbf{Ours}}    \label{fig:video_ib_ebt_sim_rescale_8}
\end{subfigure}
\caption{\lsj{A video of a synthetic sequence from the EBT dataset. (b), (c), (d), and (e) represent the cropped video frames of Jin \cite{jin2018learning}, EDI \cite{edipami}, eSL \cite{yu2023learning}, and LEDVDI \cite{yu2023learning}, resp., along with their combination with Wan \cite{mustafa2018binarization}. (f) and (g) are the GT and our result, resp.}}
\label{fig:video_bin_ebt_syn}
\end{figure*}

\begin{figure*}[ht!]
\begin{subfigure}{0.335\linewidth}
\includegraphics[width=\textwidth]{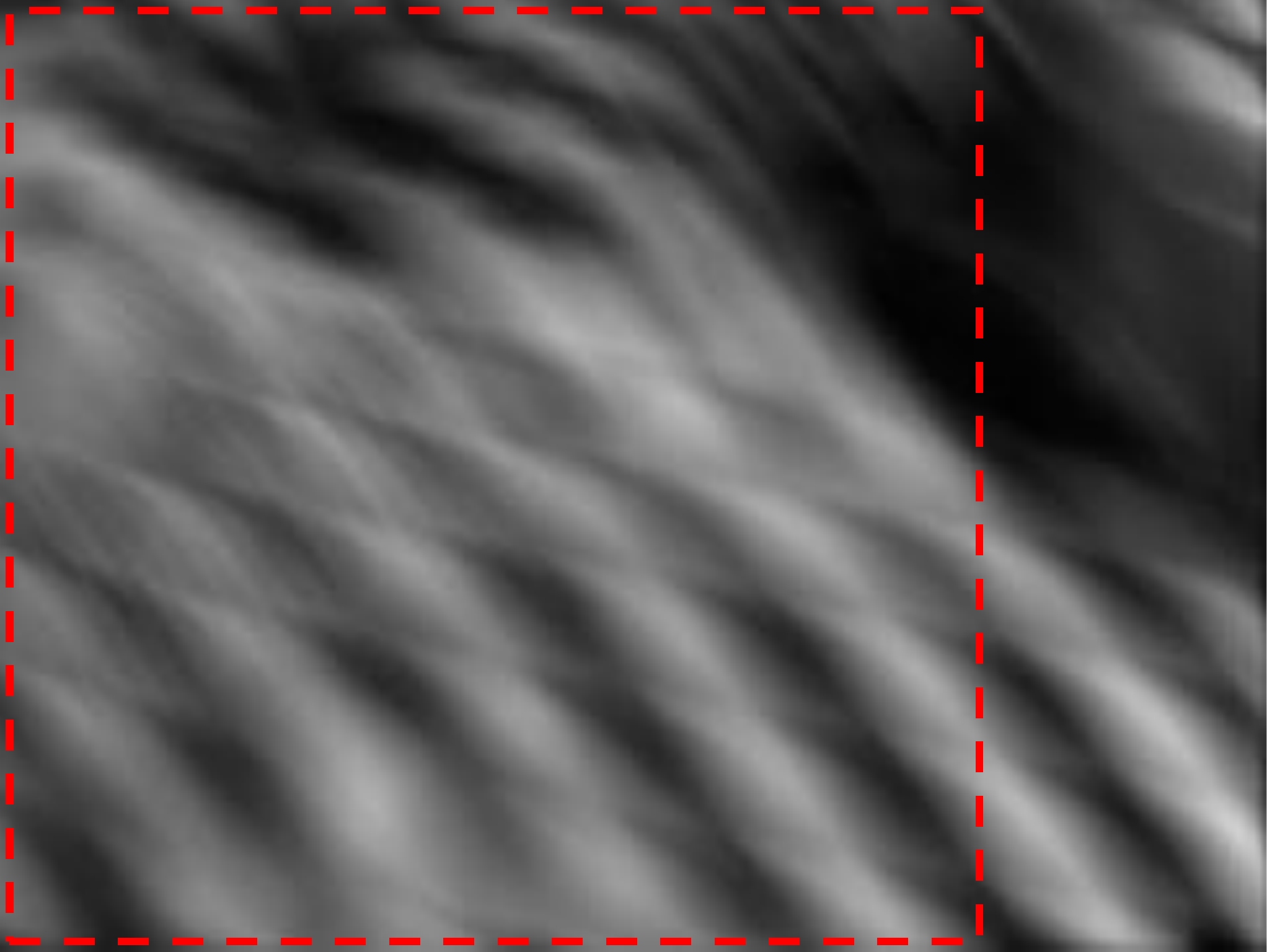}
\caption{The blurred image}    \label{fig:video_ib_hqf}
\end{subfigure}
\hfill
\begin{subfigure}{0.3\linewidth}
\includegraphics[width=\textwidth]{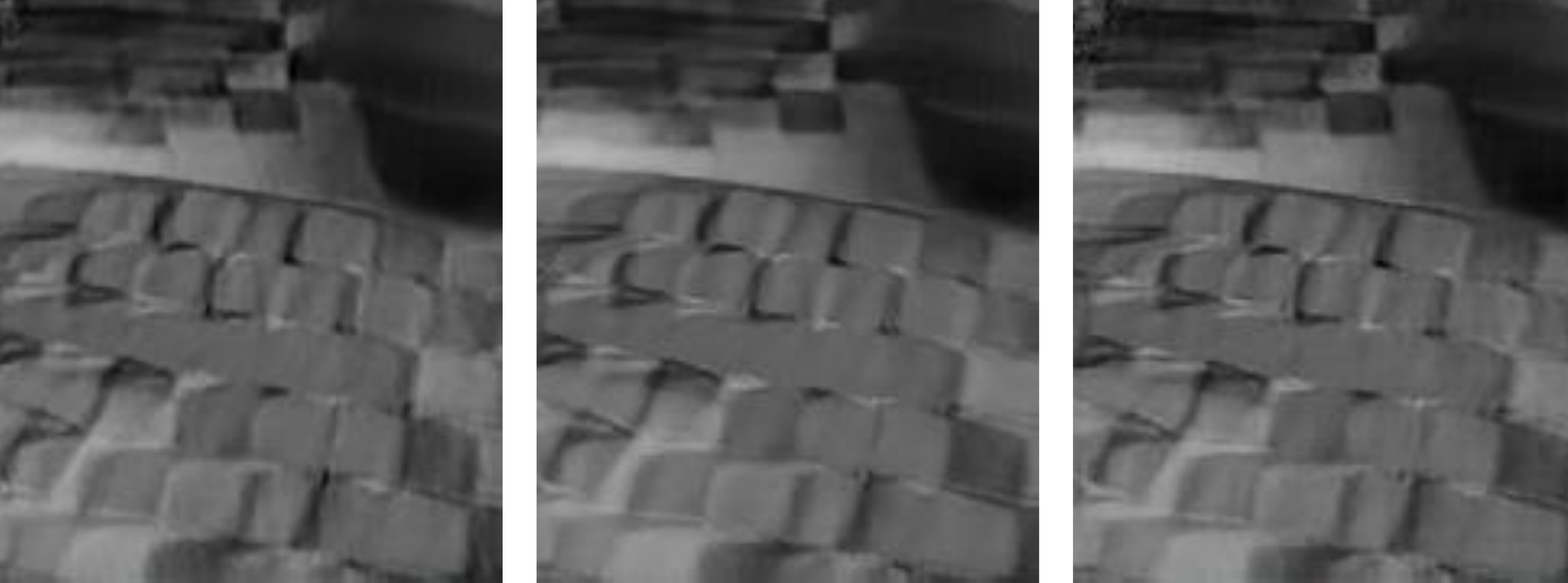}
\caption{Jin \cite{jin2018learning}}
\includegraphics[width=\textwidth]{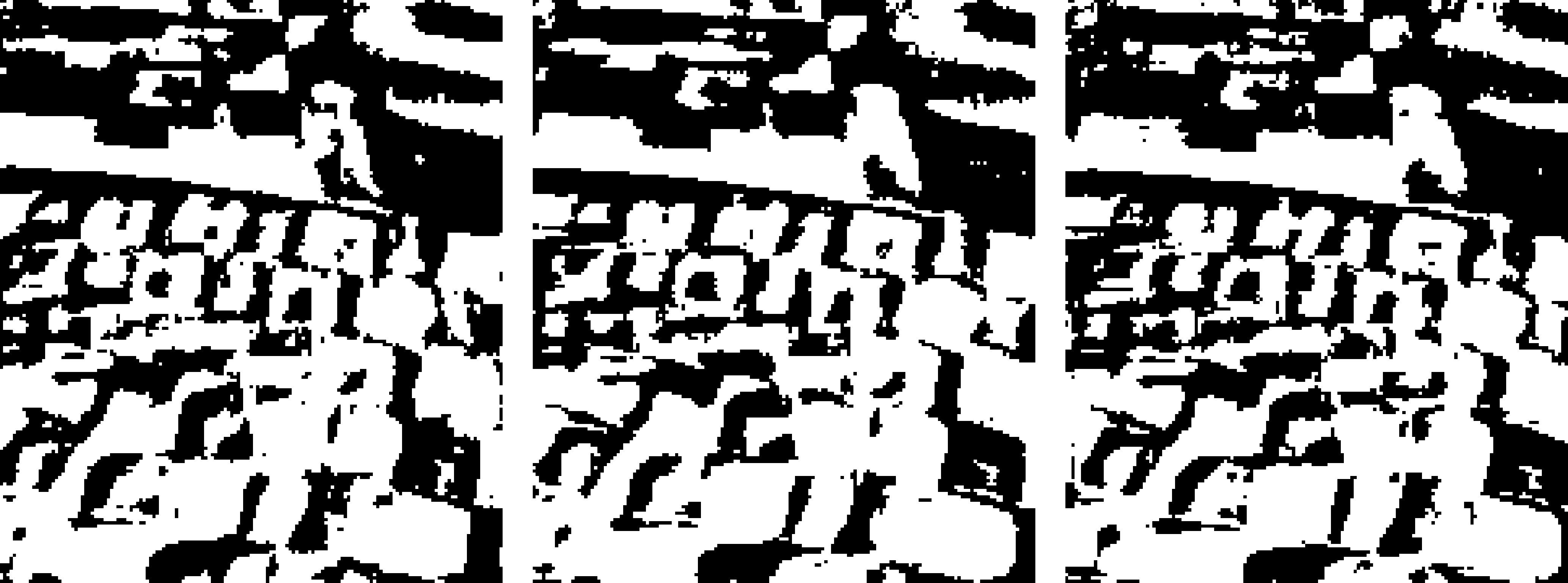}
\caption*{Jin + Wan \cite{mustafa2018binarization}}    \label{fig:video_ib_hqf_rescale_2}
\end{subfigure}
\hfill
\begin{subfigure}{0.3\linewidth}
\includegraphics[width=\textwidth]{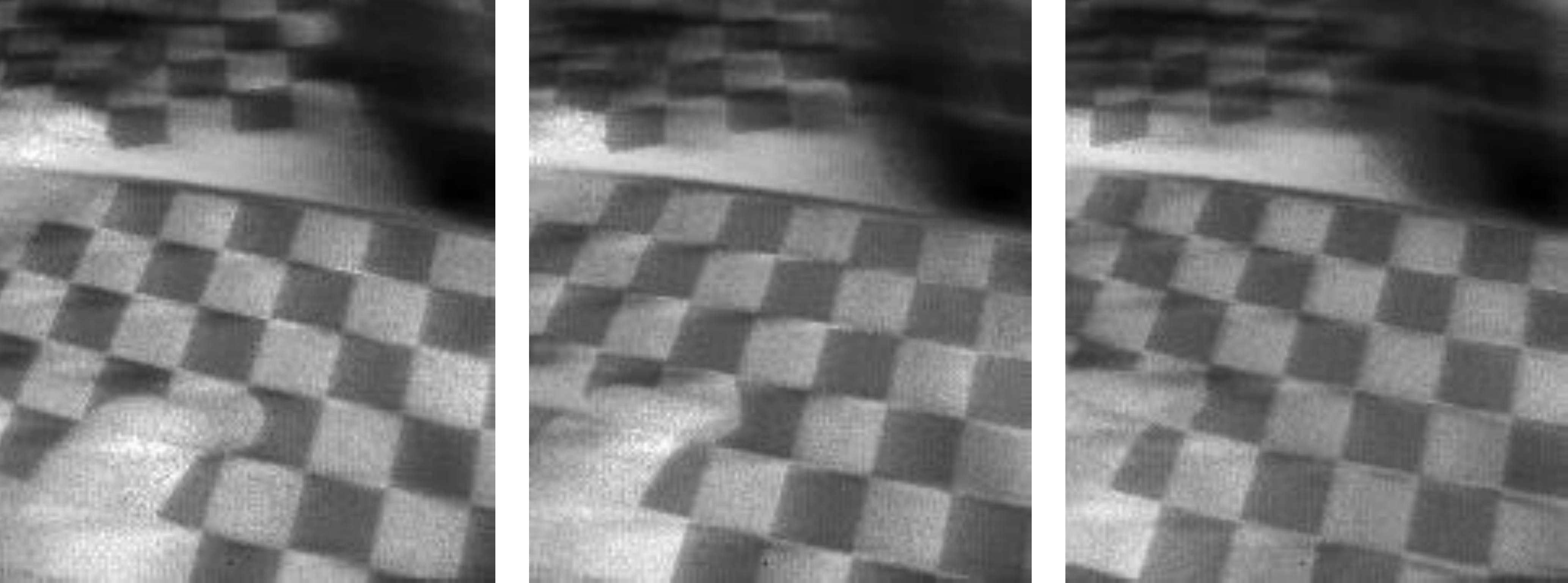}
\caption{EDI \cite{edipami}}
\includegraphics[width=\textwidth]{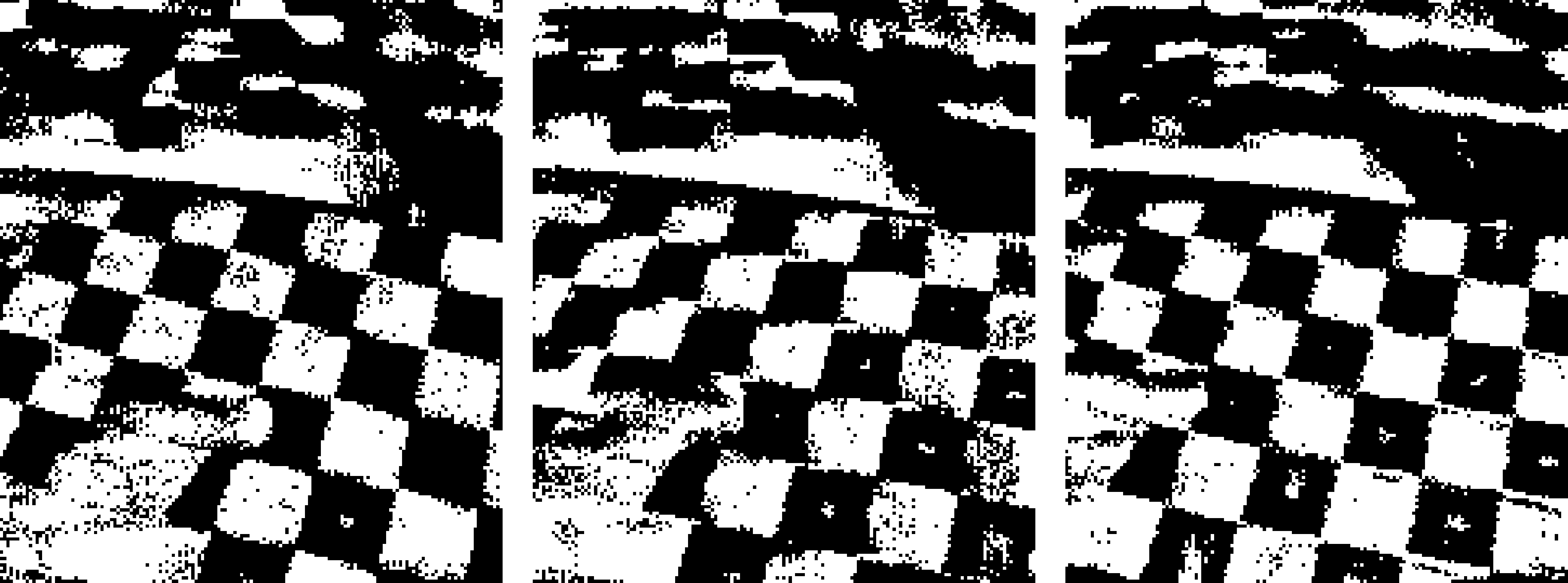}
\caption*{EDI + Wan \cite{mustafa2018binarization}}    \label{fig:video_ib_hqf_rescale_4}
\end{subfigure}
\hfill
\begin{subfigure}{0.3\linewidth}
\includegraphics[width=\textwidth]{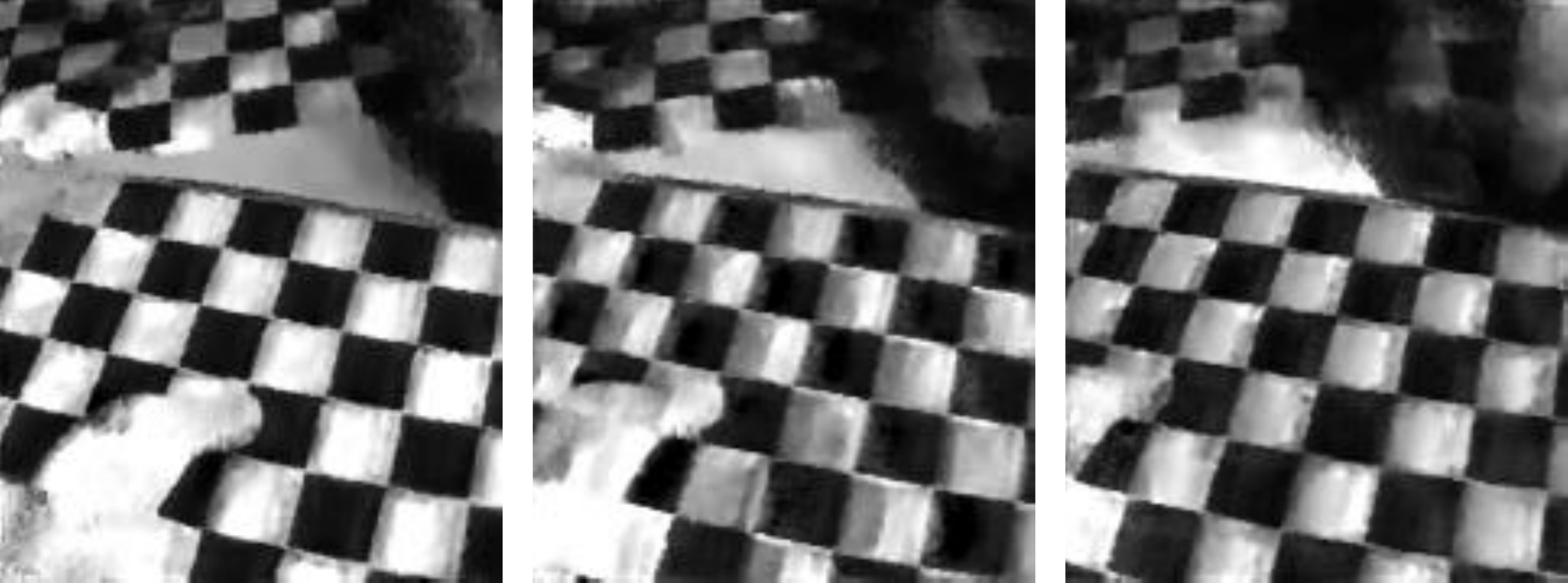}
\caption{eSL \cite{yu2023learning}} 
\includegraphics[width=\textwidth]{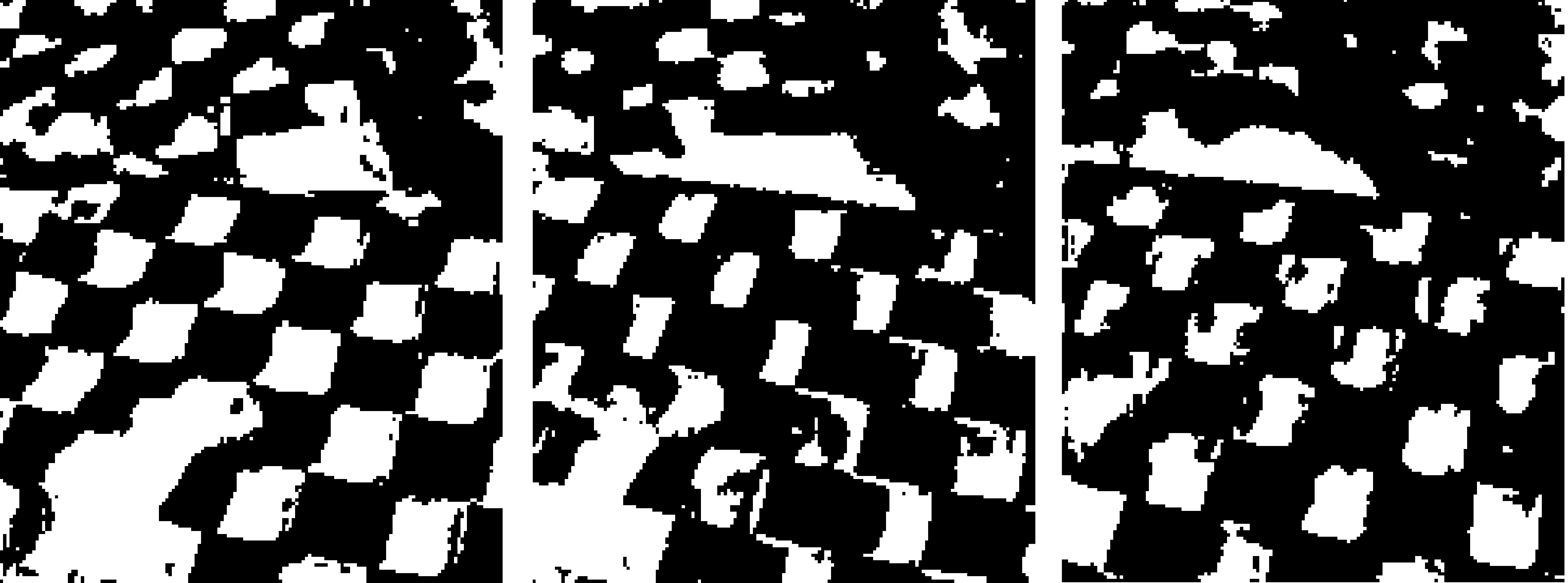}
\caption*{eSL + Wan \cite{mustafa2018binarization}}    \label{fig:video_ib_hqf_rescale_6}
\end{subfigure}
\hfill
\begin{subfigure}{0.3\linewidth}
\includegraphics[width=\textwidth]{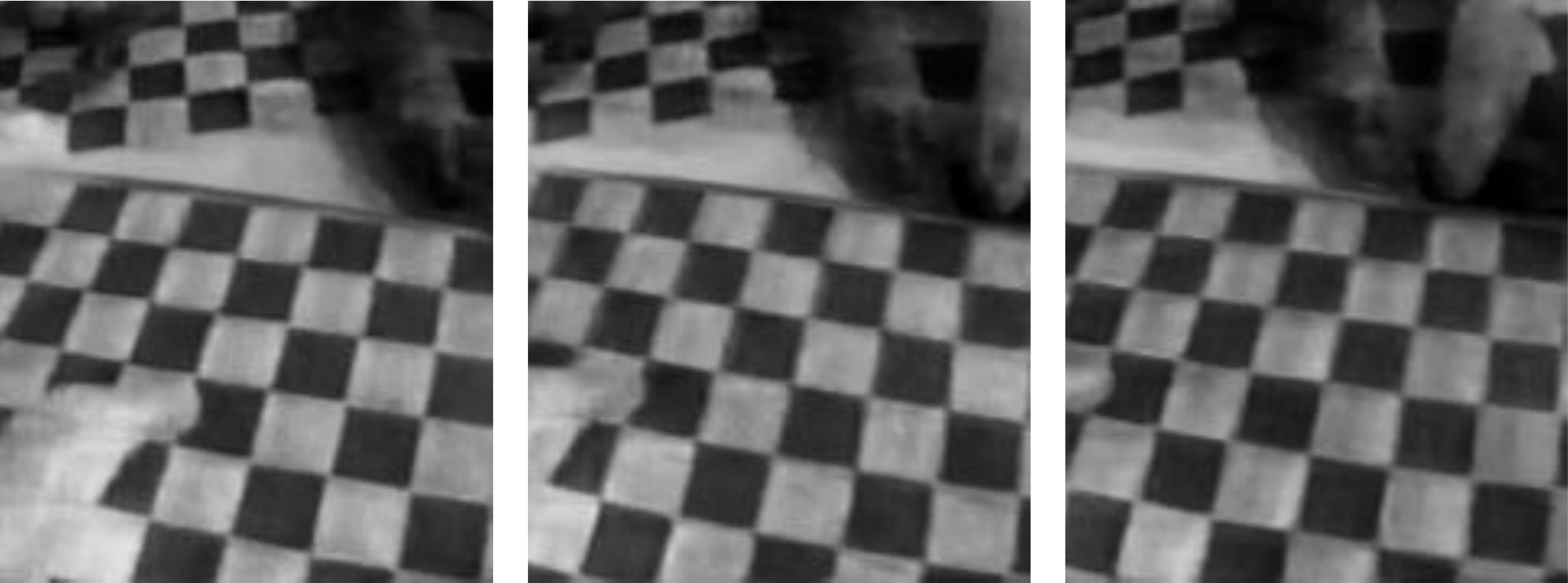}
\caption{LEDVDI \cite{lin2020learning}}
\includegraphics[width=\textwidth]{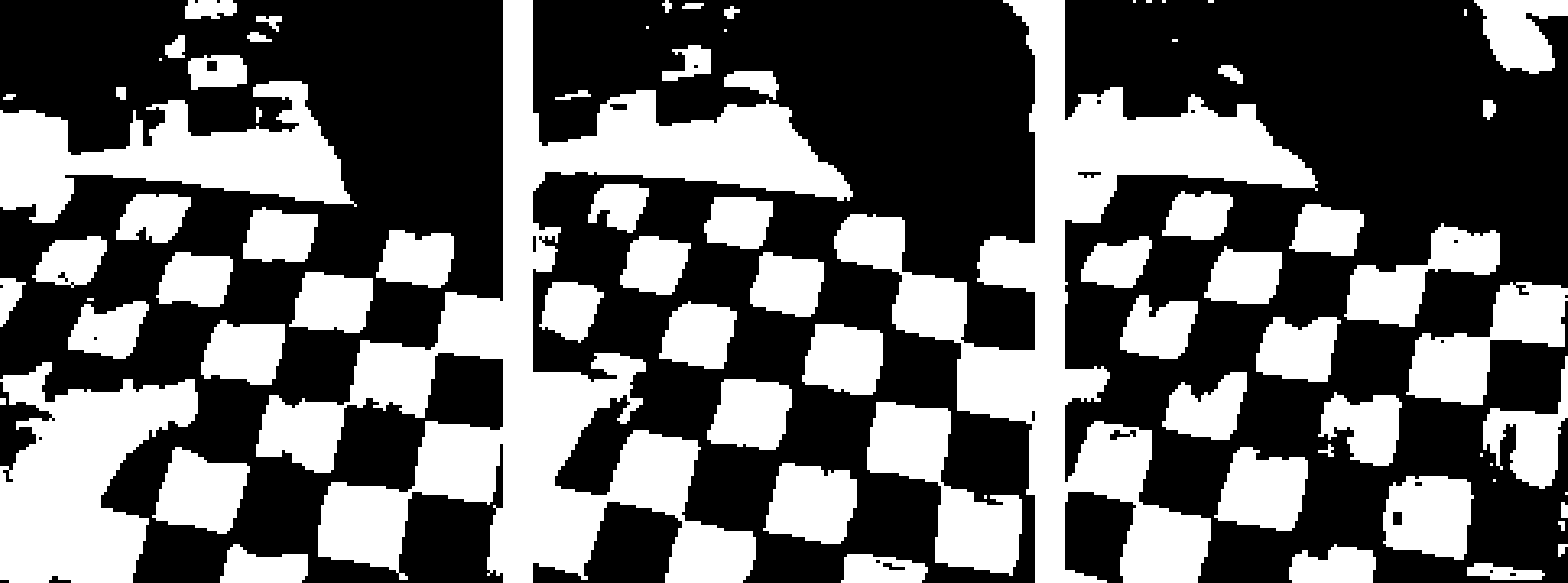}
\caption{LEDVDI + Wan \cite{mustafa2018binarization}}    \label{fig:video_ib_hqf_rescale_8}
\end{subfigure}
\hfill
\begin{subfigure}{0.3\linewidth}
\includegraphics[width=\textwidth]{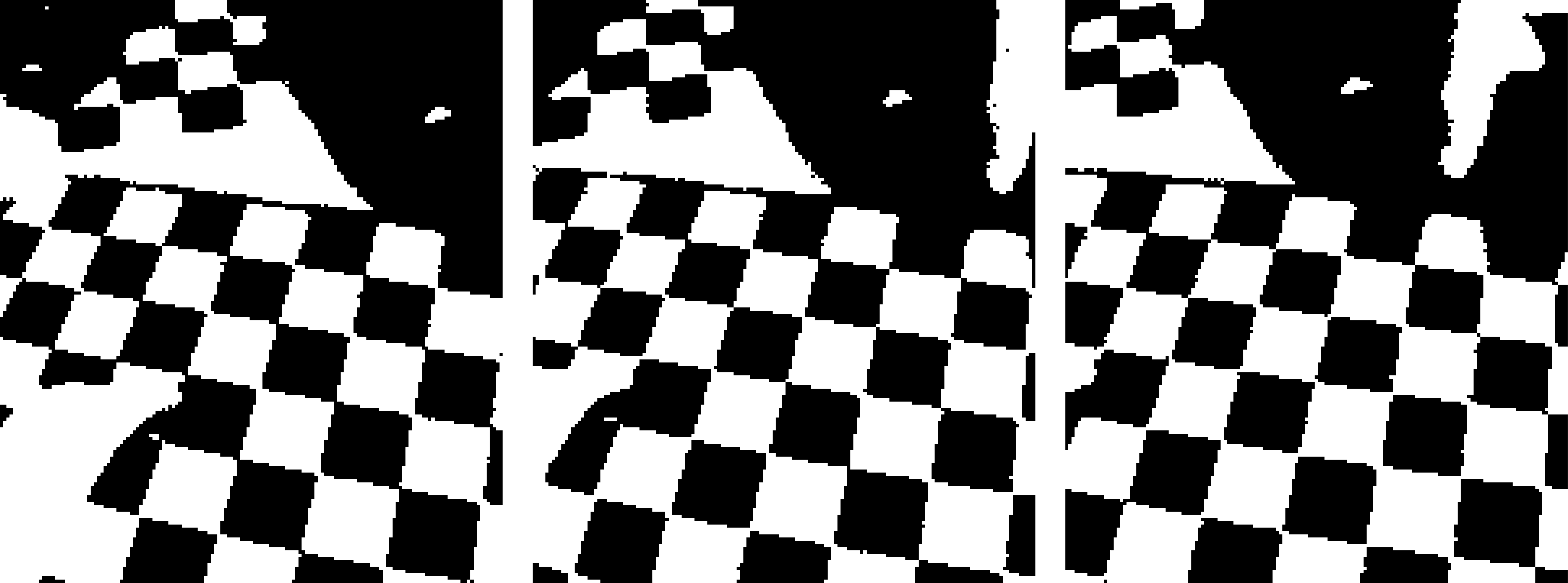}
\caption{Ground truth}
\includegraphics[width=\textwidth]{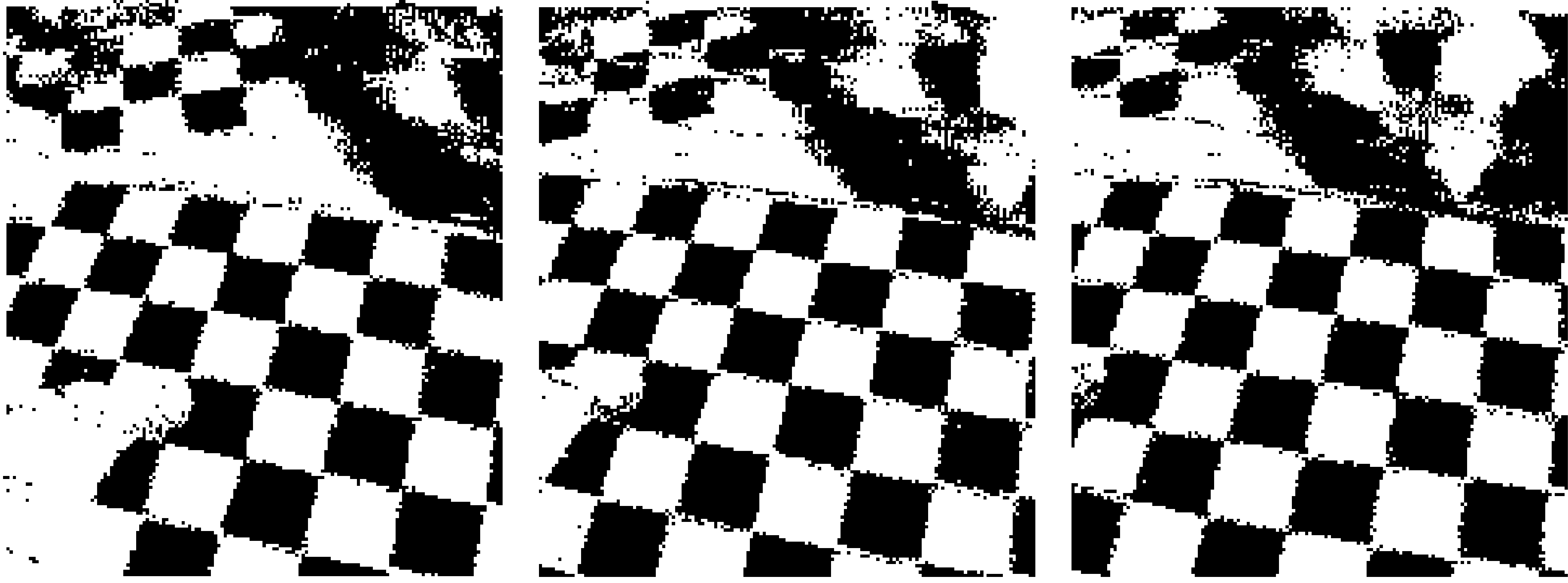}
\caption{\textbf{Ours}}    \label{fig:video_ib_hqf_rescale_10}
\end{subfigure}
\caption{\lsj{Binary videos generated on the HQF dataset. (b), (c), (d), and (e) represent the cropped video frames of Jin \cite{jin2018learning}, EDI \cite{edipami}, eSL \cite{yu2023learning}, and LEDVDI \cite{yu2023learning}, respectively, along with their combination with Wan \cite{mustafa2018binarization}. (f) Ground truth. (g) Ours.}
}
\label{fig:video_bin_hqf}
\end{figure*}

\begin{figure}[ht!]
  \centering
  \begin{subfigure}{\linewidth}
    \includegraphics[width=0.315\textwidth]{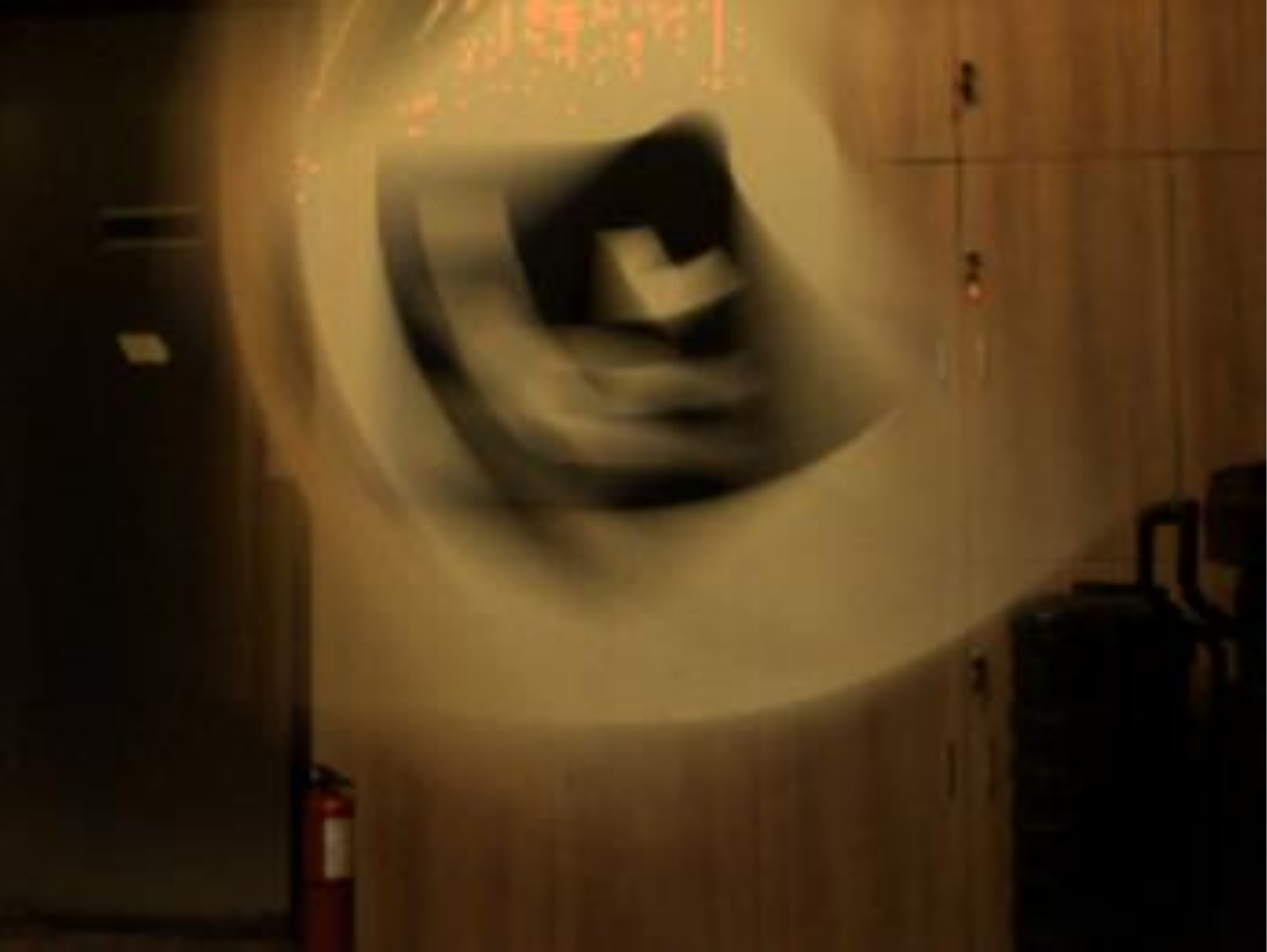}
    \hfill
    \includegraphics[width=0.315\textwidth]{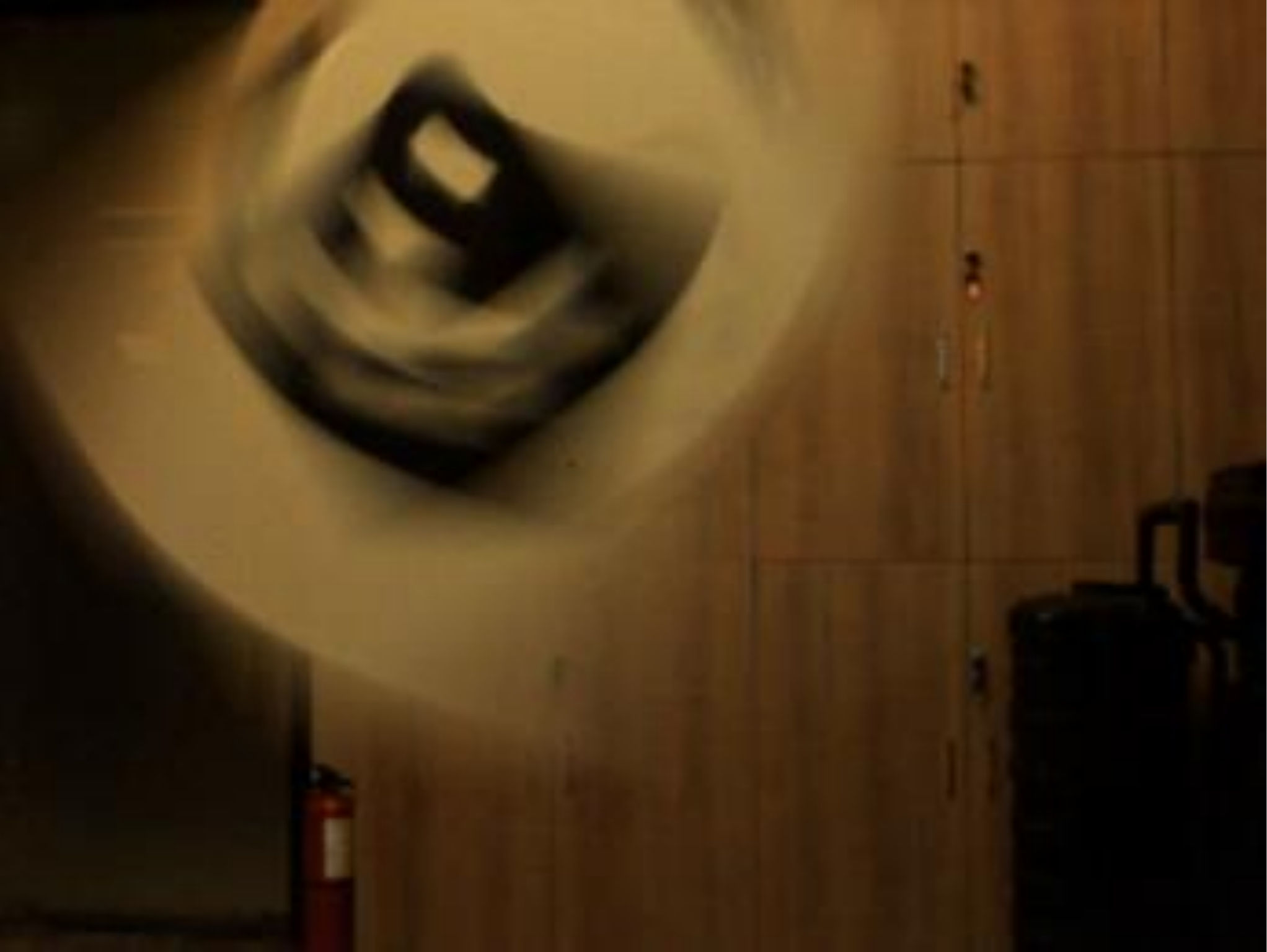}
    \hfill
    \includegraphics[width=0.315\textwidth]{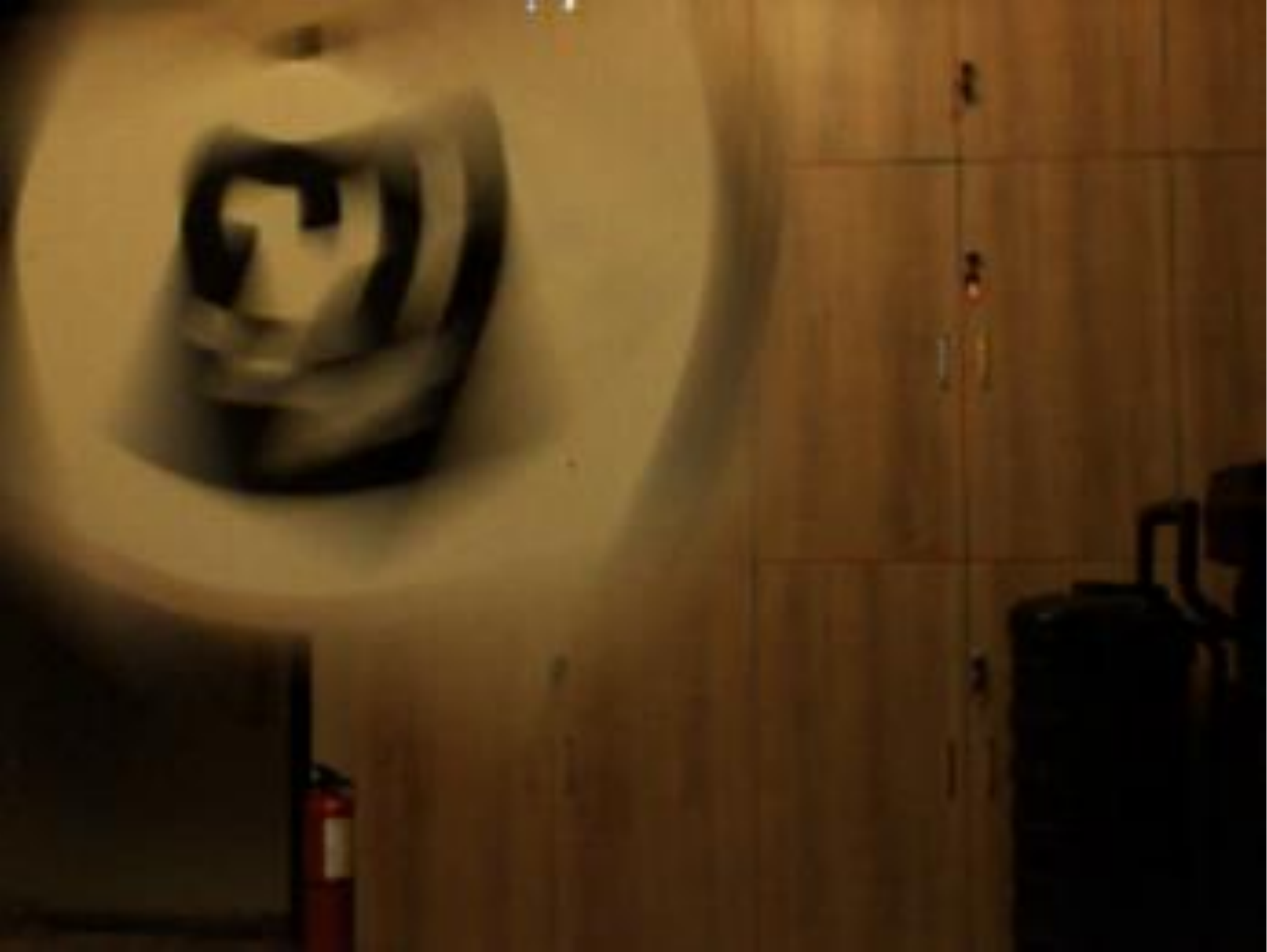}
    \hfill
    \caption{The blurry snapshots from a video}    \label{fig:apriltag_3}
  \end{subfigure}
  \hfill
  \begin{subfigure}{\linewidth}
    \includegraphics[width=0.315\textwidth]{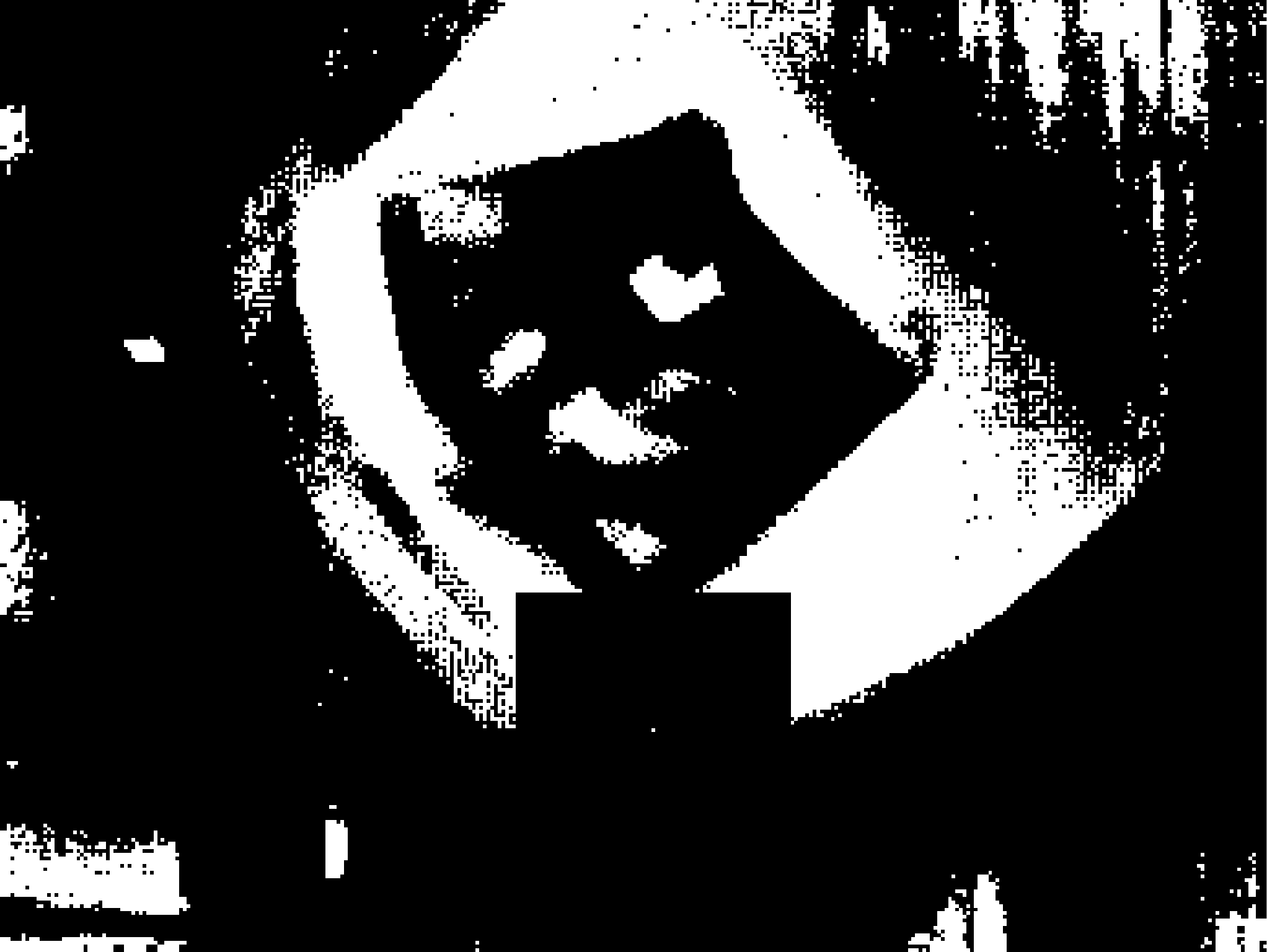}
    \hfill
    \includegraphics[width=0.315\textwidth]{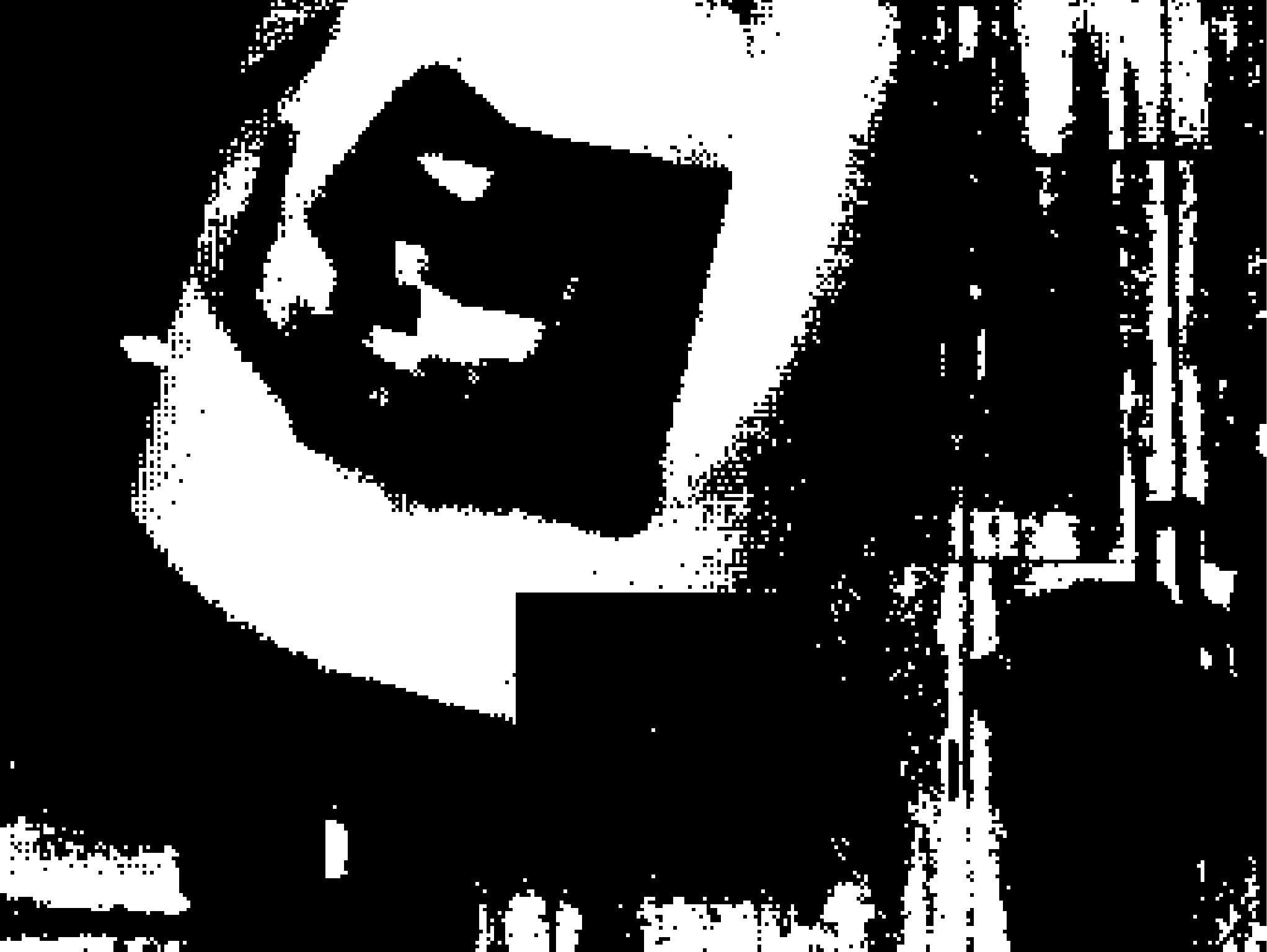}
    \hfill
    \includegraphics[width=0.315\textwidth]{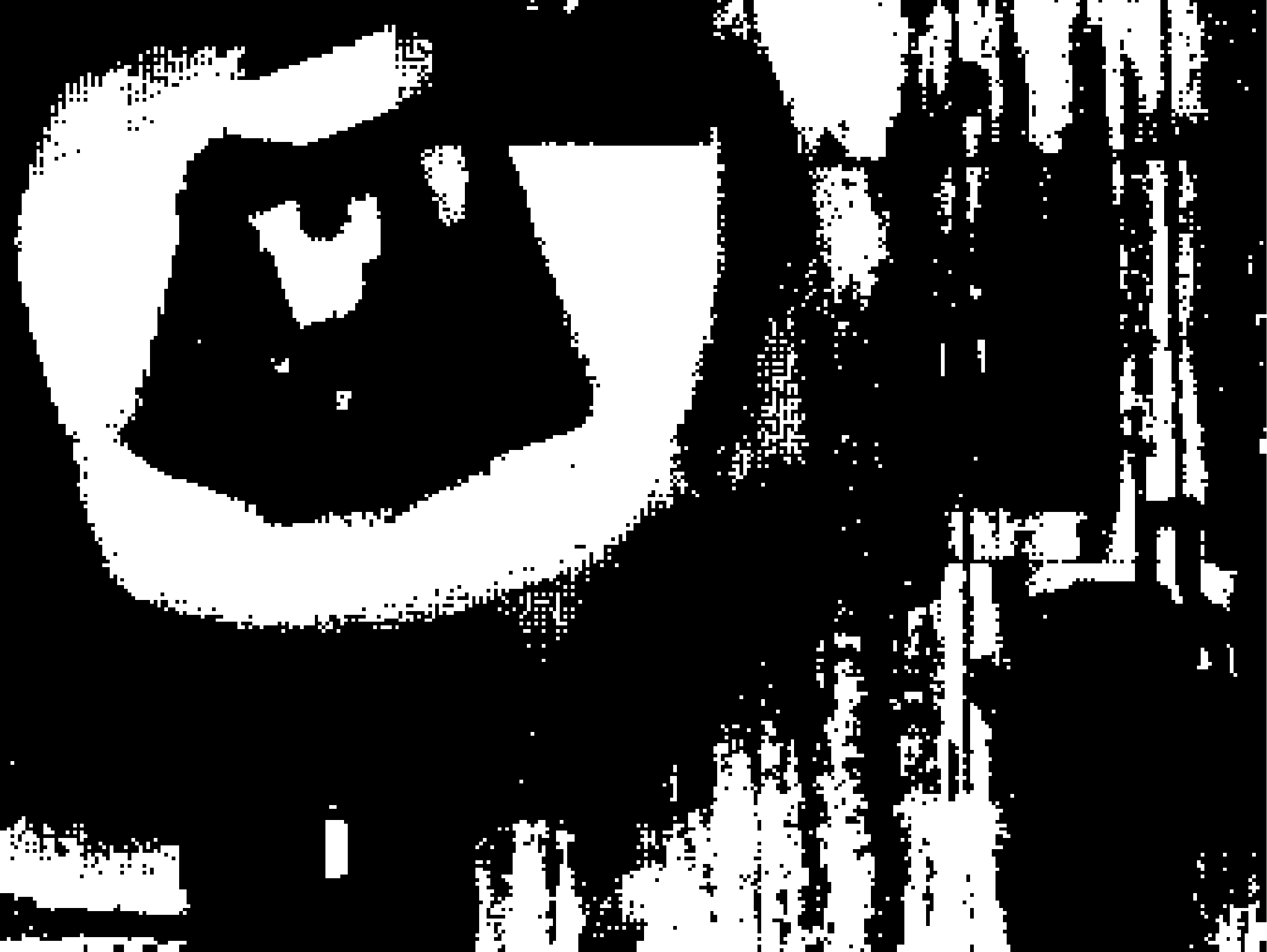}
    \hfill
    \caption{EDI \cite{edipami} + Wan \cite{mustafa2018binarization}}    \label{fig:apriltag_6}
  \end{subfigure}
  \hfill
  \begin{subfigure}{\linewidth}
    \includegraphics[width=0.315\textwidth]{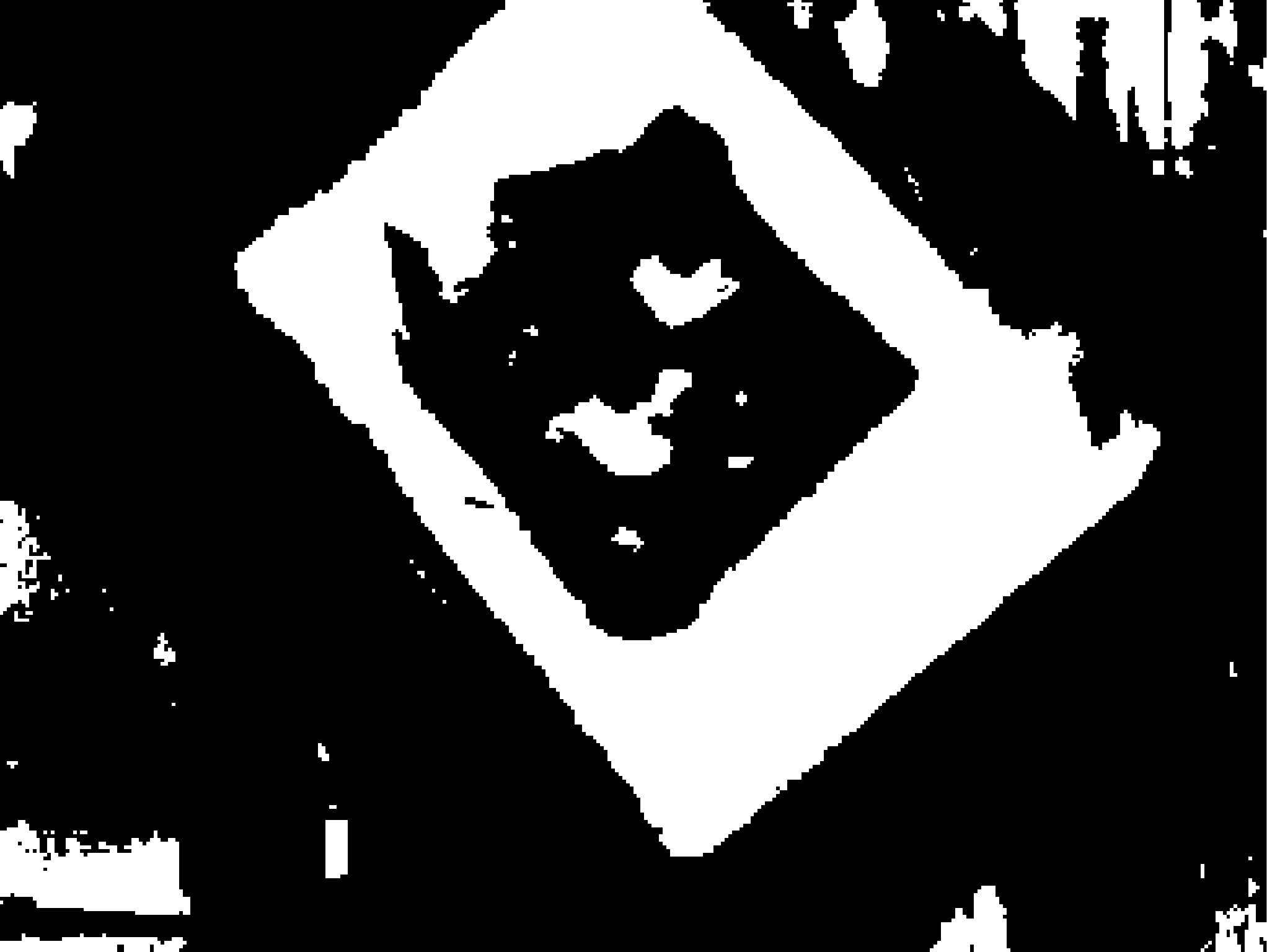}
    \hfill
    \includegraphics[width=0.315\textwidth]{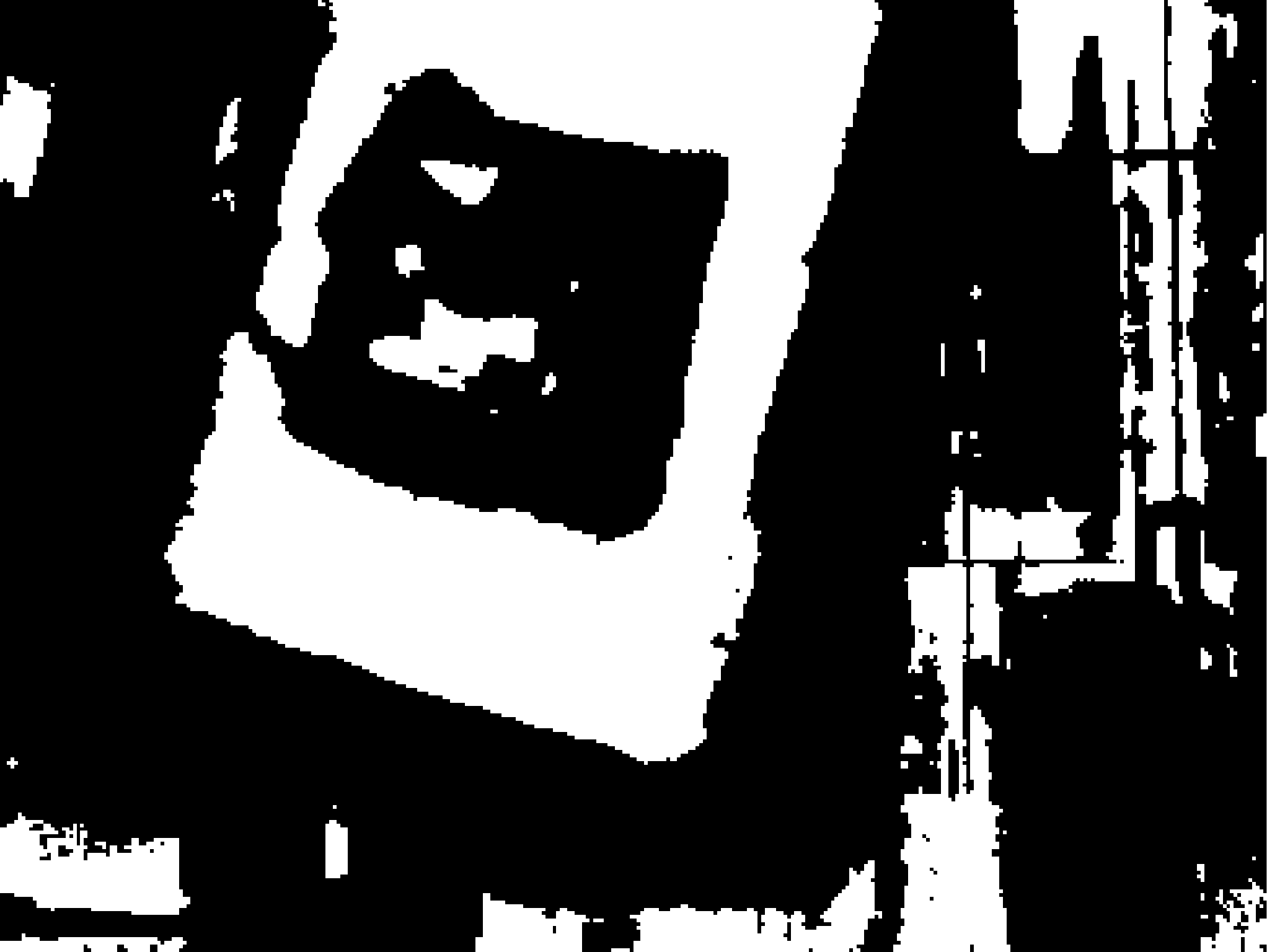}
    \hfill
    \includegraphics[width=0.315\textwidth]{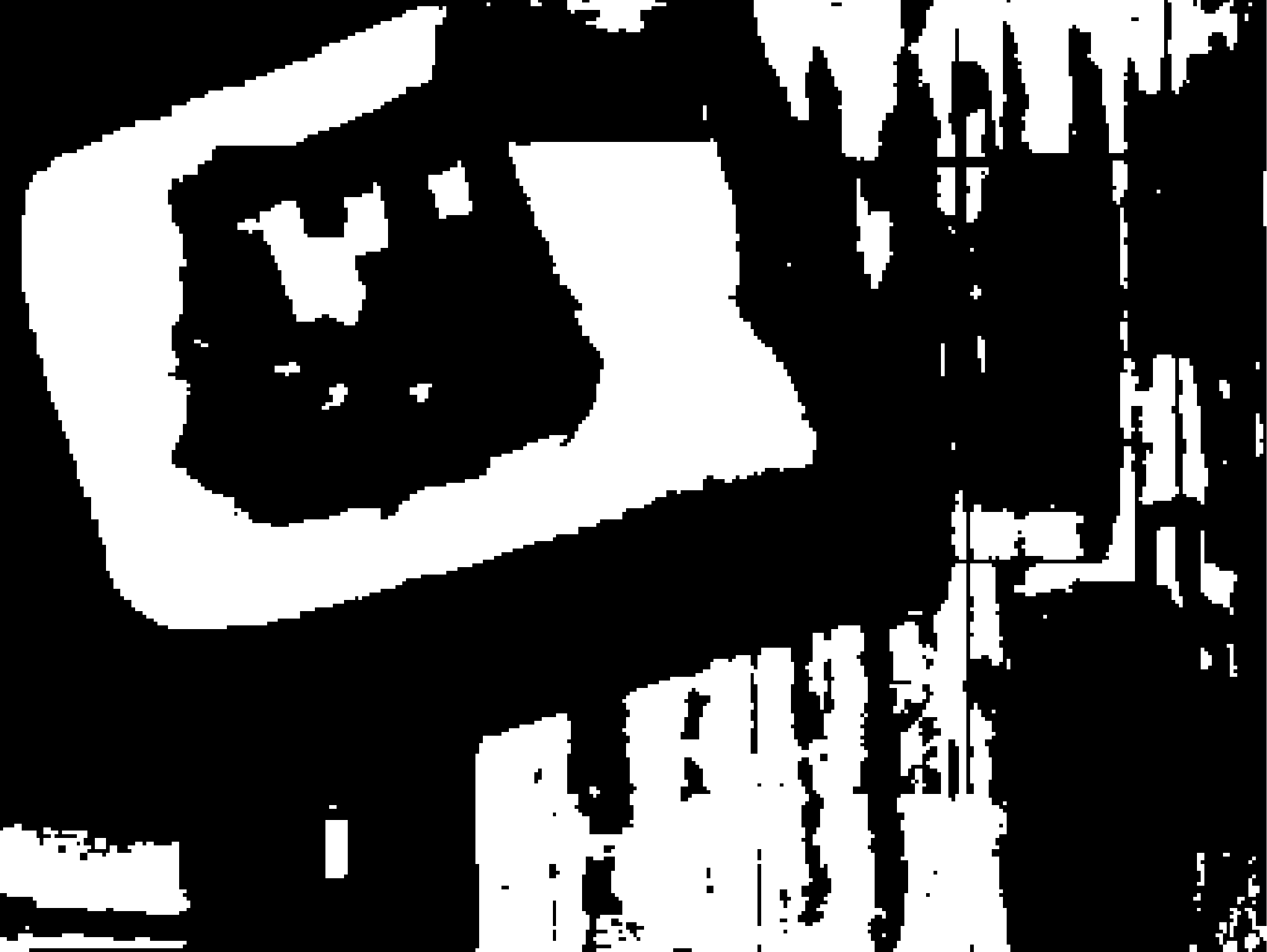}
    \hfill
    \caption{eSL\cite{yu2023learning} + Wan \cite{mustafa2018binarization}}    \label{fig:apriltag_9}
  \end{subfigure}
  \vfill
  \begin{subfigure}{\linewidth}
    \includegraphics[width=0.315\textwidth]{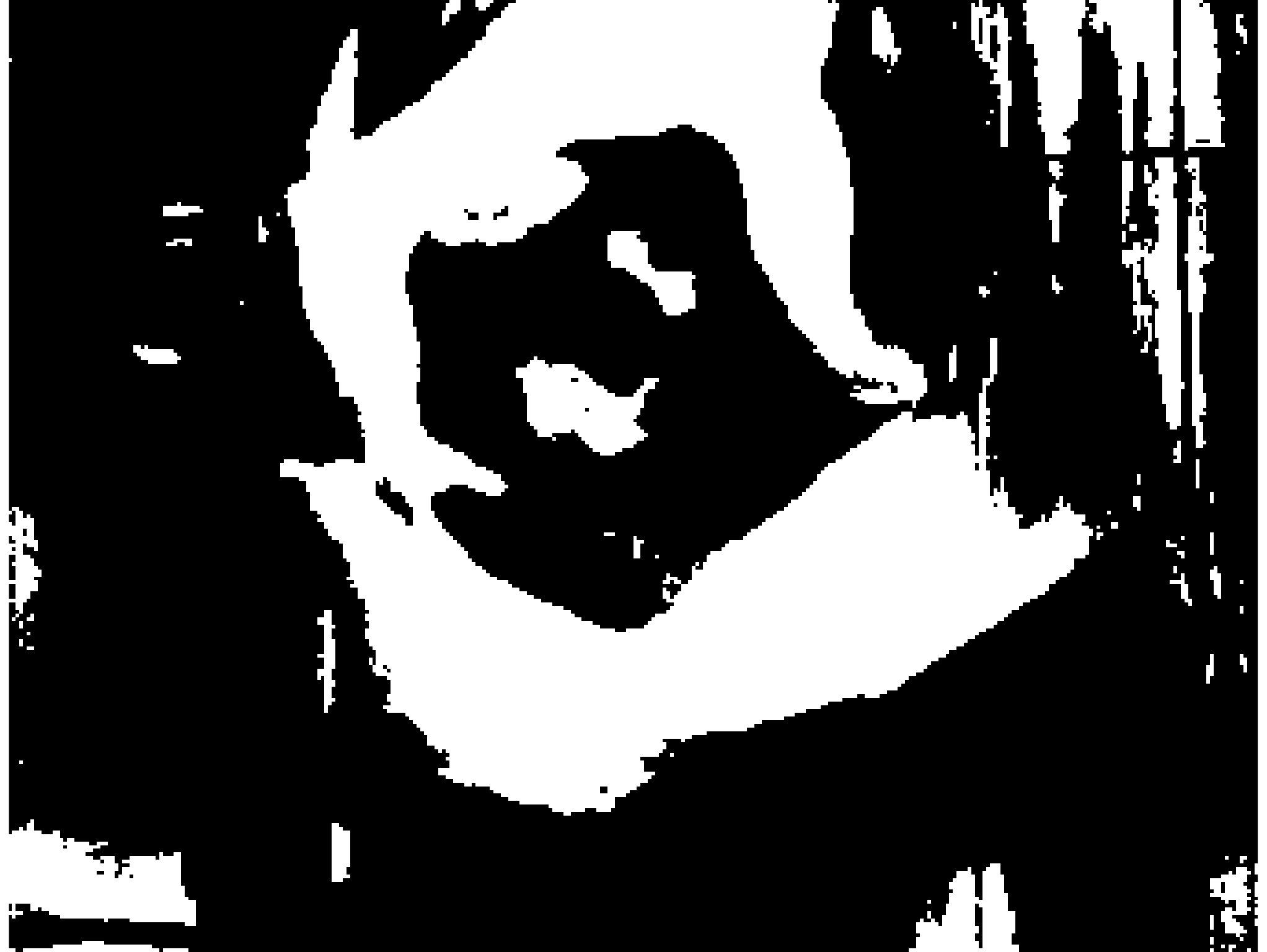}
    \hfill
    \includegraphics[width=0.315\textwidth]{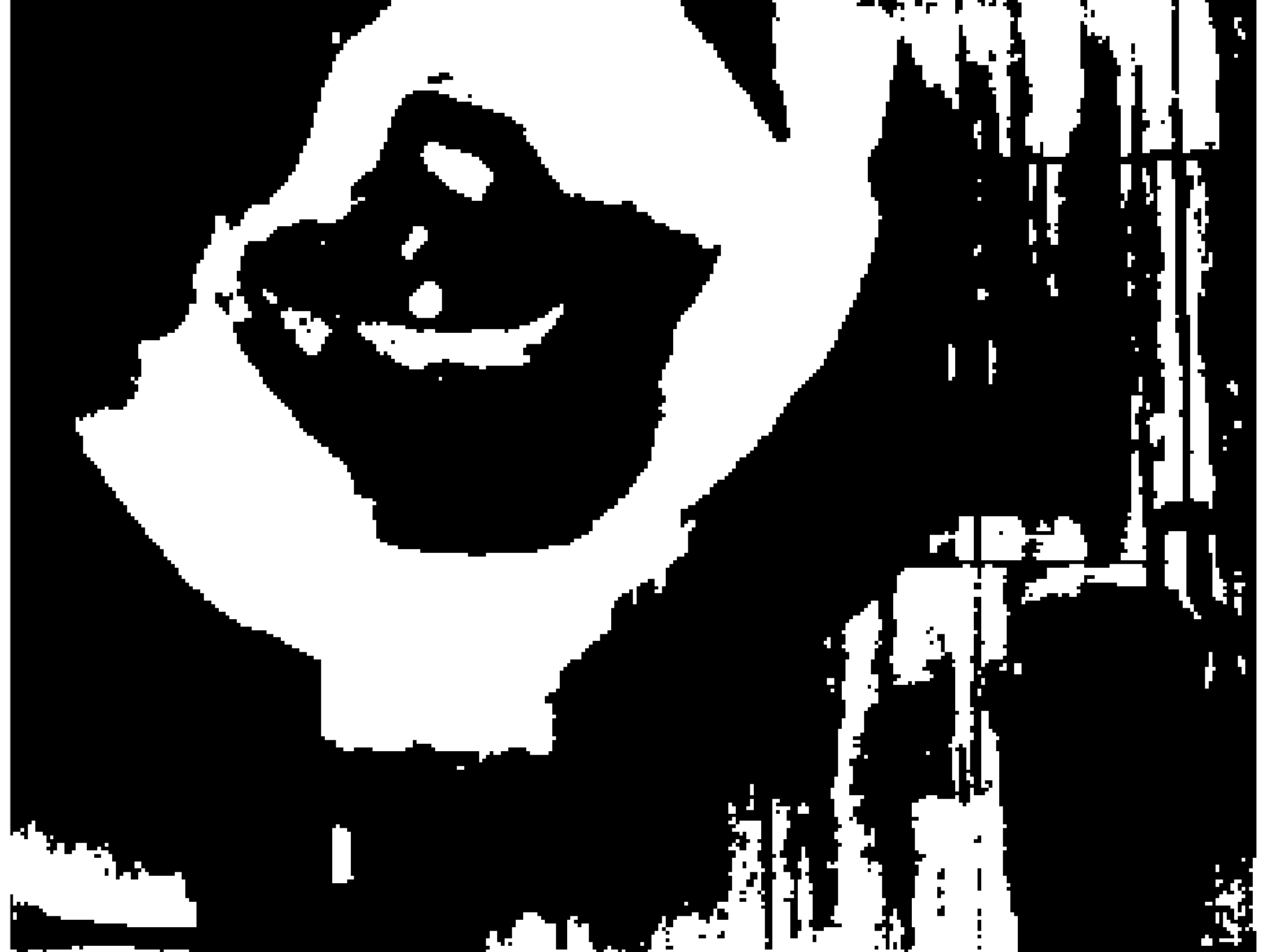}
    \hfill
    \includegraphics[width=0.315\textwidth]{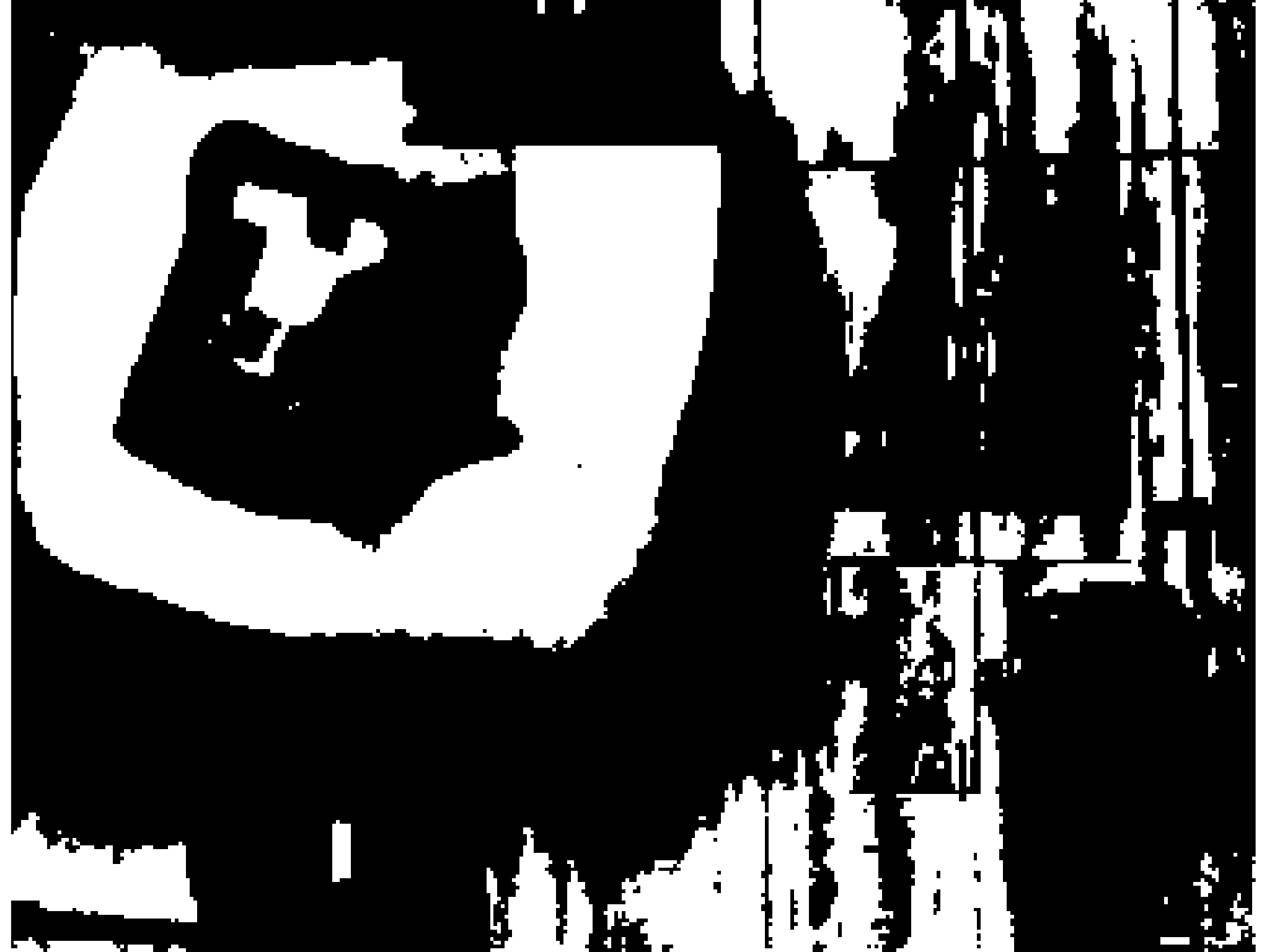}
    \hfill
    \caption{Jin \cite{jin2018learning} + Wan \cite{mustafa2018binarization}}    \label{apriltag_12}
  \end{subfigure}
  \hfill
  \begin{subfigure}{\linewidth}
    \includegraphics[width=0.315\textwidth]{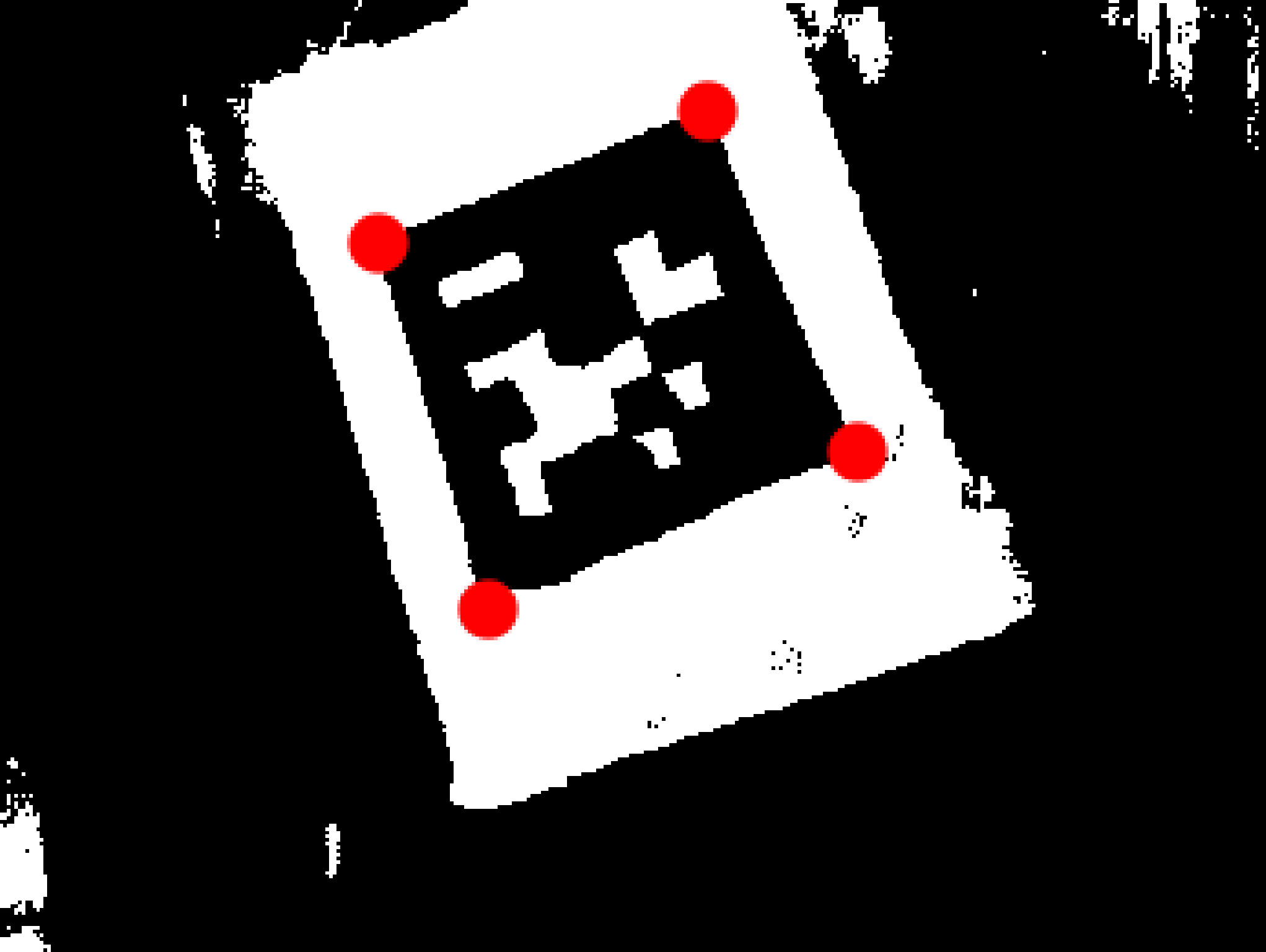}
    \hfill
    \includegraphics[width=0.315\textwidth]{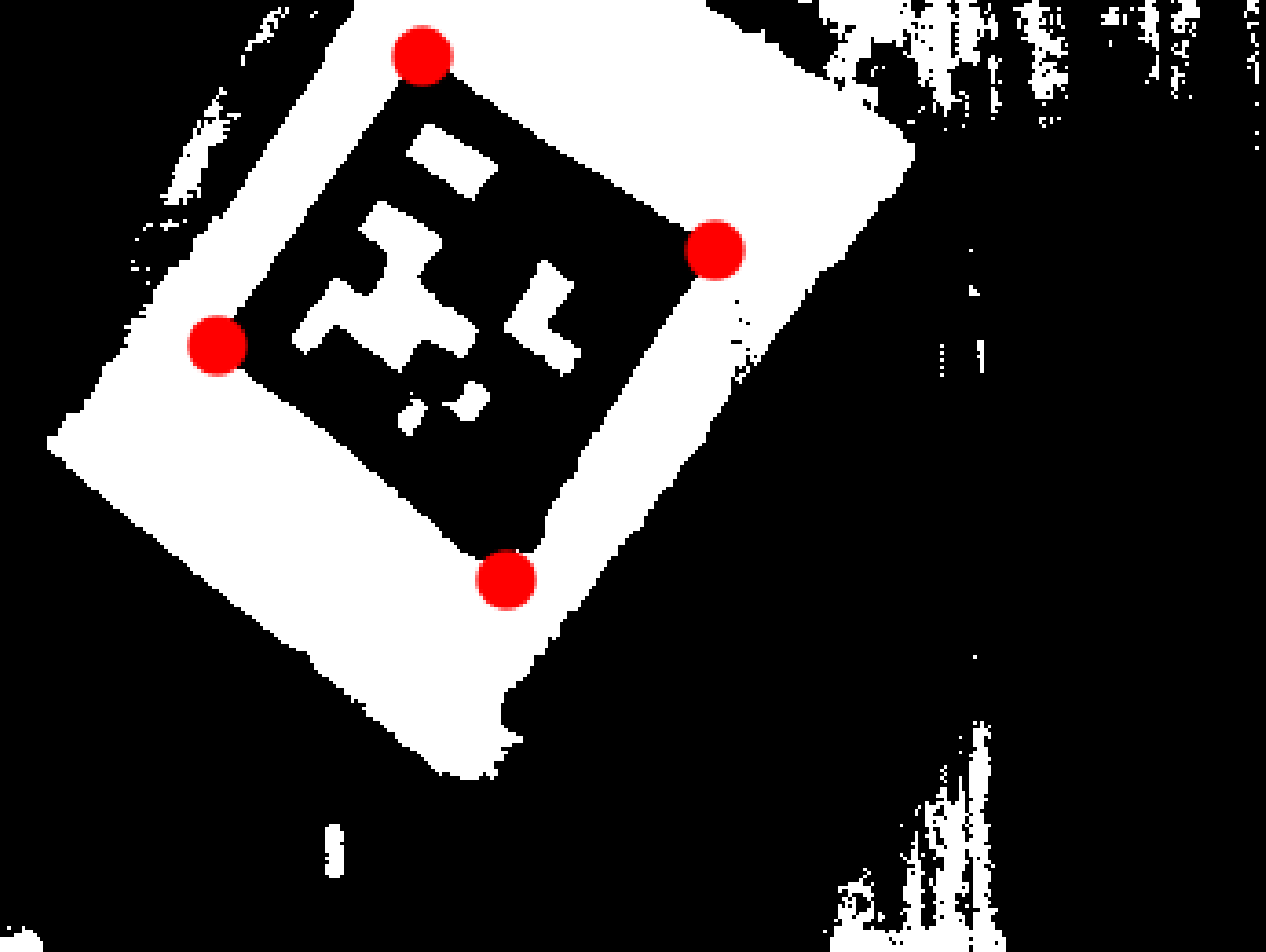}
    \hfill
    \includegraphics[width=0.315\textwidth]{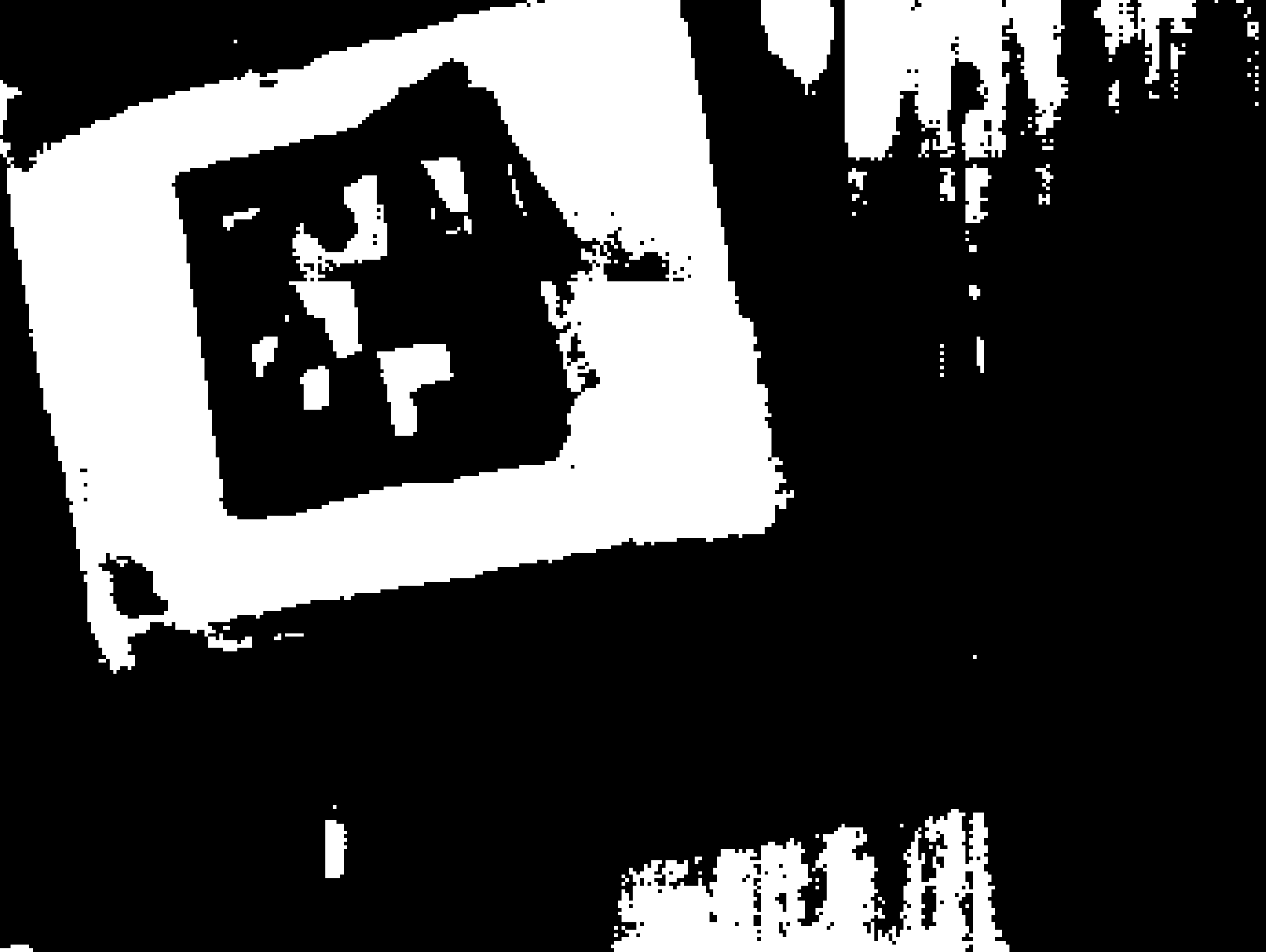}
    \hfill
    \caption{LEDVDI \cite{lin2020learning} + Wan \cite{mustafa2018binarization}}    \label{apriltag_ledvdi}
  \end{subfigure}
  \hfill
  \begin{subfigure}{\linewidth}
    \includegraphics[width=0.315\textwidth]{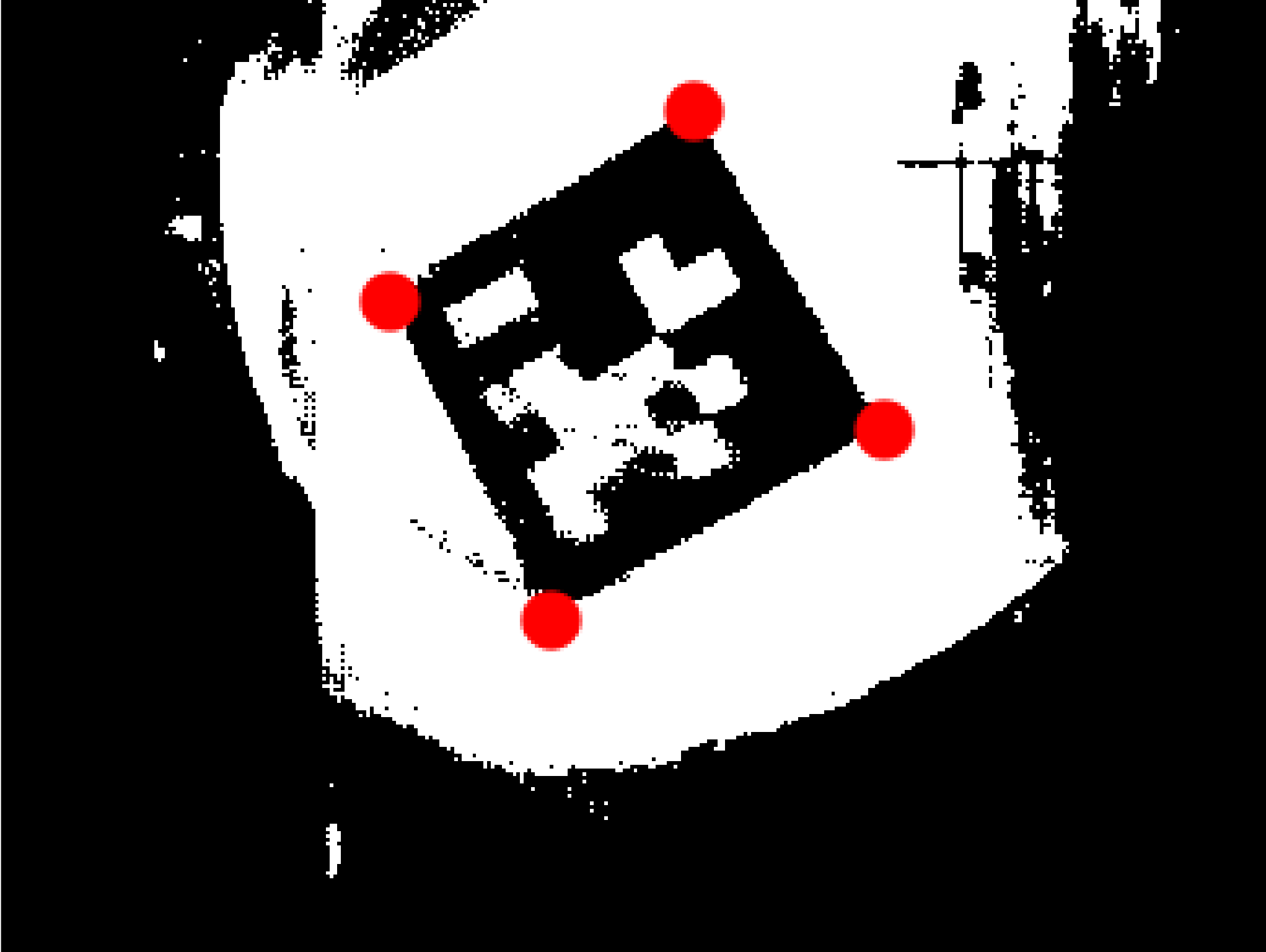}
    \hfill
    \includegraphics[width=0.315\textwidth]{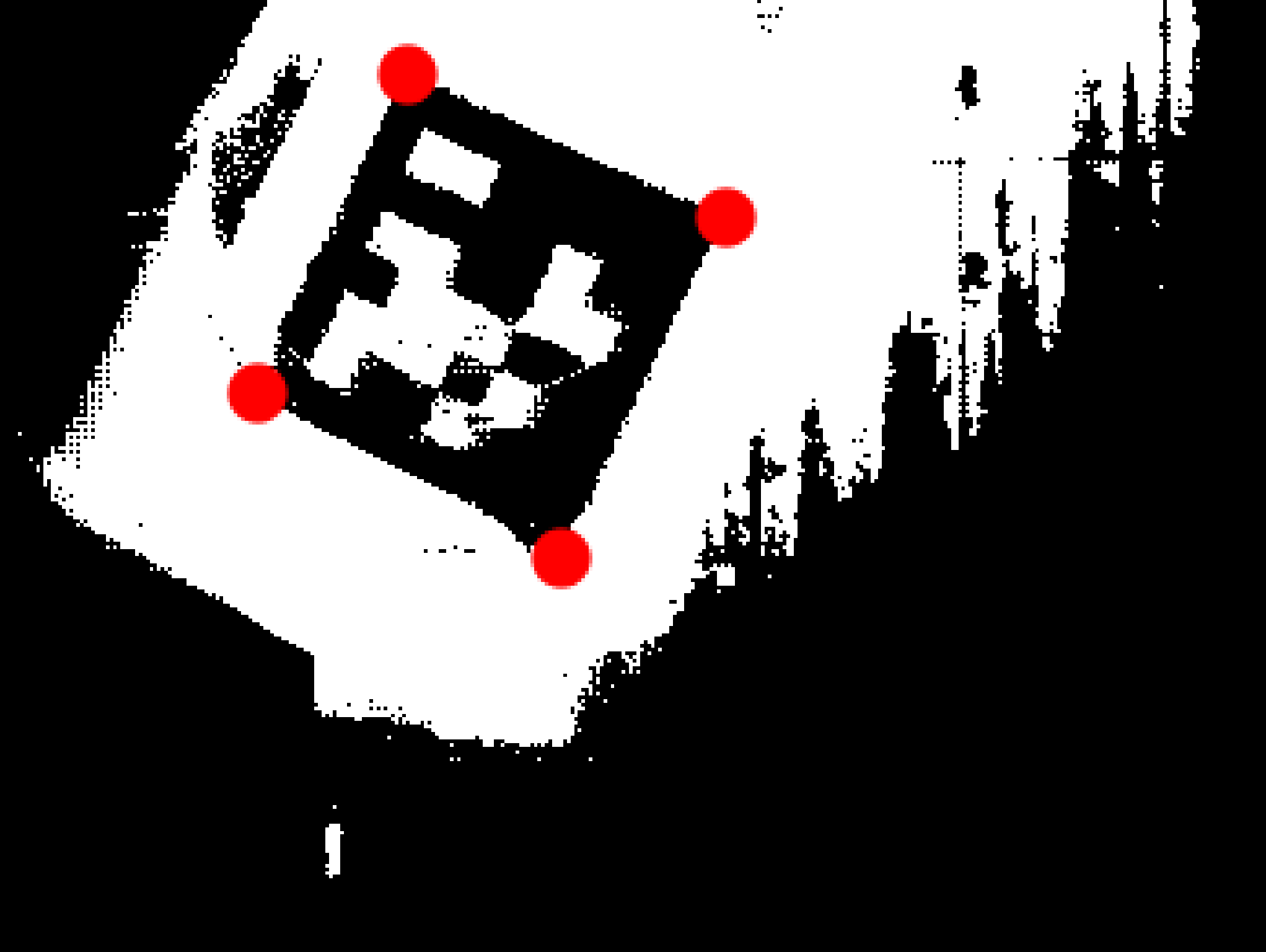}
    \hfill
    \includegraphics[width=0.315\textwidth]{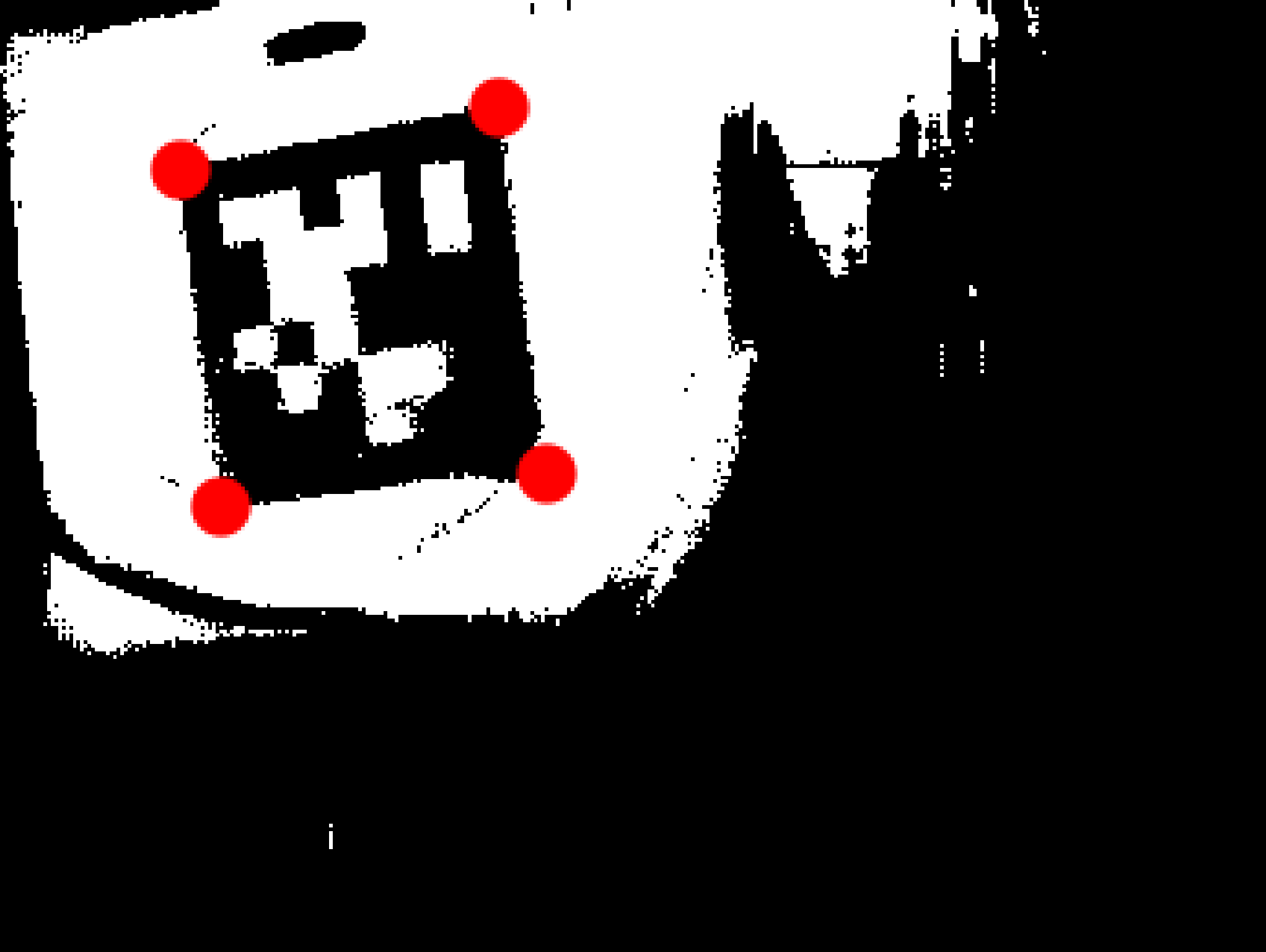}
    \hfill
    \caption{Our video (without AMF filtering)}    \label{fig:apriltag_18}
  \end{subfigure}
  \hfill
  \begin{subfigure}{\linewidth}
    \includegraphics[width=0.315\textwidth]{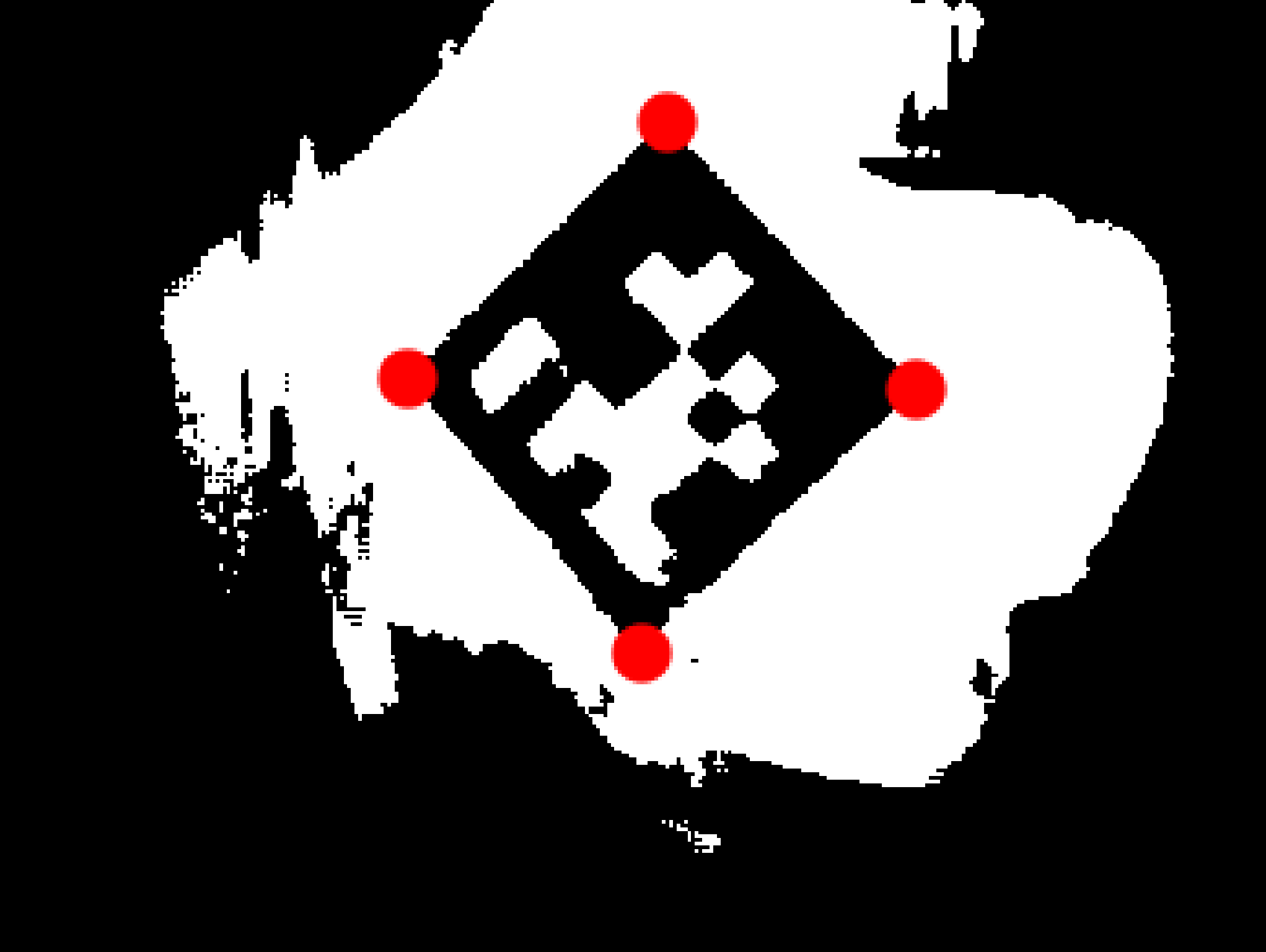}
    \hfill
    \includegraphics[width=0.315\textwidth]{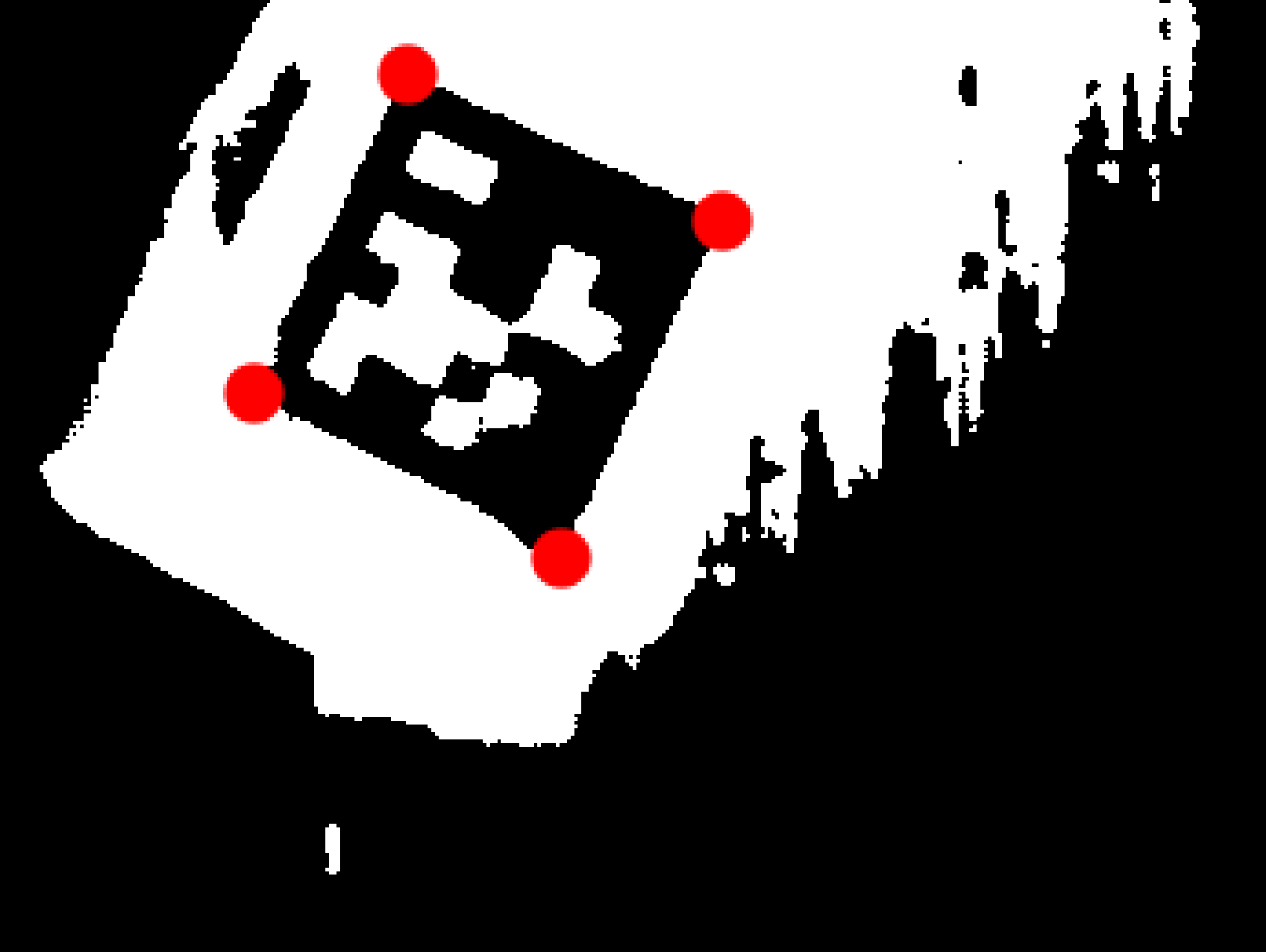}
    \hfill
    \includegraphics[width=0.315\textwidth]{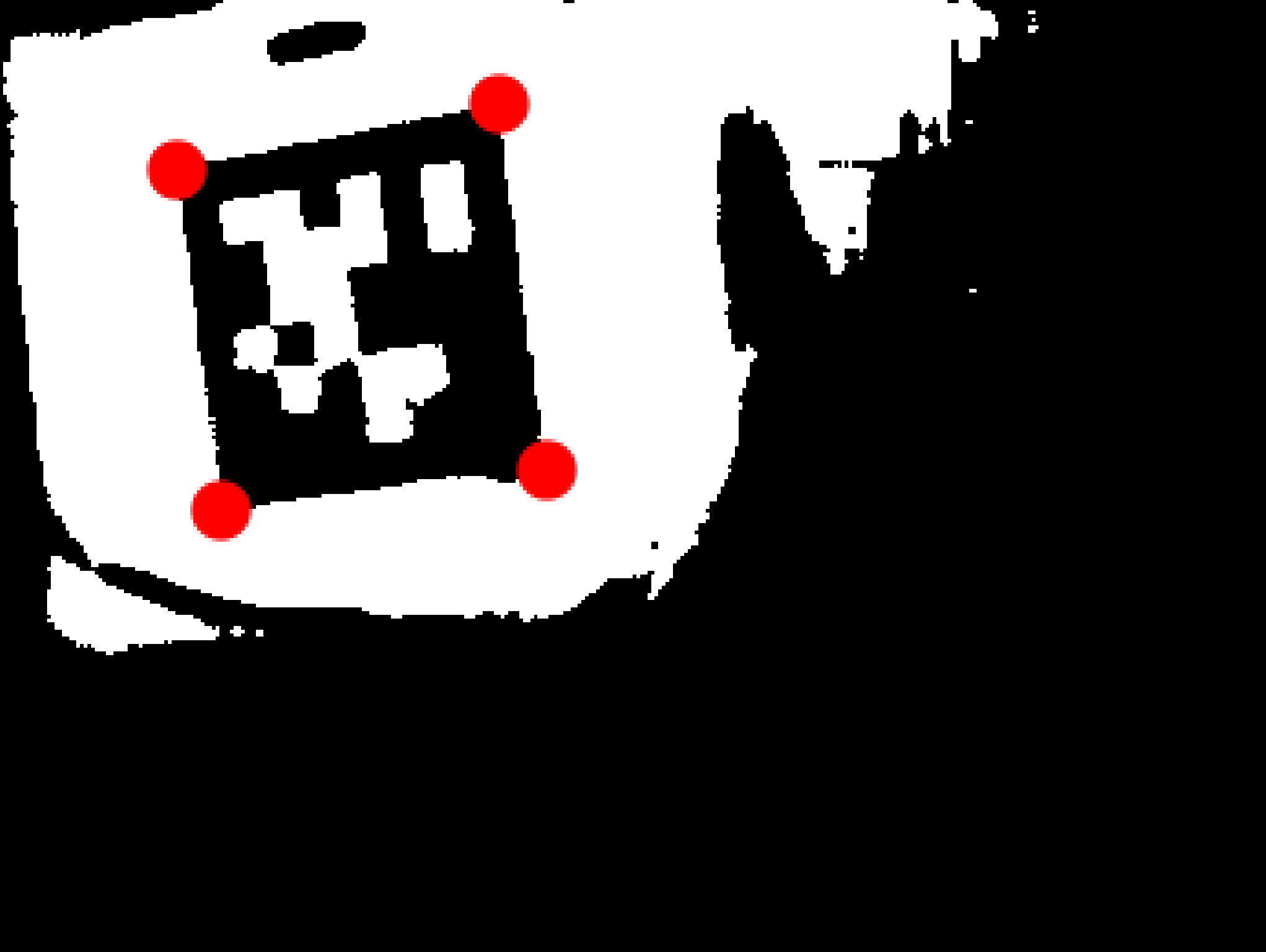}
    \hfill
    \caption{Our video (with AMF filtering)}    \label{fig:apriltag_21}
  \end{subfigure}

  \caption{\lsj{(a) Snapshots of an Apriltag \cite{wang2016apriltag} under high-speed motions. Results in (b), (c), (d), and (e) show that the combination of state-of-the-art video generation methods and image binarization are usually incompatible with each other, resulting in losing track of the target. Results in (e) and (f) demonstrate that our methods can capture the motion and reconstruct sharp binary videos, allowing the tag to be accurately detected (marked by the red dots). Best viewed in color.}}
  \label{fig:rotate_tag}
\end{figure}

\begin{figure}[ht!]
  \centering
  \begin{subfigure}{0.315\linewidth}
    \includegraphics[width=\textwidth]{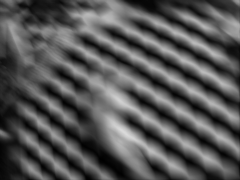}
    \caption{Blur Image}    \label{fig:raw_1581821422508246}
  \end{subfigure}
  \begin{subfigure}{0.315\linewidth}
    \includegraphics[width=\textwidth]{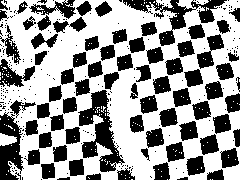}
    \caption{$c=0.25$}    \label{fig:c_25_1581821422508246}
  \end{subfigure}
  \begin{subfigure}{0.315\linewidth}
    \includegraphics[width=\textwidth]{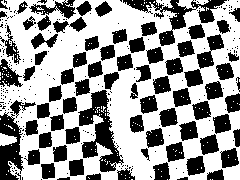}
    \caption{$c=0.5$}    \label{fig:c_50_1581821422508246}
  \end{subfigure}
  \vfill
  \begin{subfigure}{0.315\linewidth}
    \includegraphics[width=\textwidth]{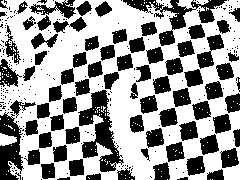}
    \caption{$c=0.75$}    \label{fig:c_75_1581821422508246}
  \end{subfigure}
  \begin{subfigure}{0.315\linewidth}
    \includegraphics[width=\textwidth]{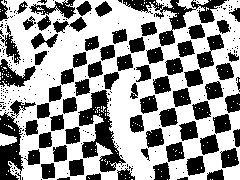}
    \caption{$c=1.0$}    \label{fig:c_100_1581821422508246}
  \end{subfigure}
  \begin{subfigure}{0.315\linewidth}
    \includegraphics[width=\textwidth]{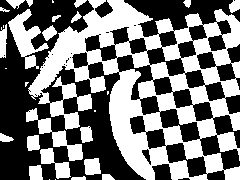}
    \caption{Ground truth}    \label{fig:gt_1581821422508246}
  \end{subfigure}
  
  \caption{Examples of latent binary images generated by our method under various contrast parameters using the HQF dataset.} 
  \label{fig:contrast}
\end{figure}

\begin{table}[t]
\caption{\lsj{Comparisons with state-of-the-art image binarization.}}
\label{table:ib}
\centering
\resizebox{1\linewidth}{!}{
\begin{tabular}{c|lccccc}
\hline
\noalign{\smallskip}
Dataset & Method & Event & Image &  MCC ($\uparrow$) & PSNR ($\uparrow$)  & NRM ($\downarrow$)\\
\noalign{\smallskip}
\hline
\noalign{\smallskip}
 \multirow{8}*{HQF} & Nick \cite{khurshid2009comparison} & & $\surd$  & 0.27       & 3.72   &   0.36 \\ 
 & Adaptive \cite{bradley2007adaptive} & & $\surd$  & 0.21       & 3.48   &  0.39  \\ 
 & Wolf \cite{wolf2004extraction} & & $\surd$         & 0.28       & 4.21    &  0.36  \\ 
  & Wan \cite{mustafa2018binarization}  & & $\surd$         & \underline{0.29}       & \underline{5.24}   & \underline{0.35}   \\ 
 & AE \cite{calvo2019selectional} & & $\surd$  & 0.26      & 4.15  &   0.36  \\ 
 & Dplink-Net \cite{xiong2021dp}    &      & $\surd$ &          0.15    &    2.84           & 0.43  \\
  \cite{stoffregen2020reducing} & Howe \cite{howe2013document}    &      & $\surd$ &          0.06    &   2.41           & 0.47  \\
 & \textbf{Ours}  & $\surd$ & $\surd$   & \textbf{0.54}       & \textbf{6.32} &   \textbf{0.21}   \\ 
\noalign{\smallskip}
\hline
\noalign{\smallskip}
\multirow{8}*{Reblur} & Nick \cite{khurshid2009comparison}  & & $\surd$      & 0.24       & 2.32     & 0.38  \\ 
 & Adaptive \cite{bradley2007adaptive} & & $\surd$ & 0.30       & 2.98   &   0.34  \\ 
 & Wolf \cite{wolf2004extraction}  & & $\surd$       & 0.42       & 4.41      &  0.25\\ 
 & Wan \cite{mustafa2018binarization}  & & $\surd$       & \underline{0.44}       & \underline{6.17}     &   \underline{0.24} \\ 
  & AE \cite{calvo2019selectional} & & $\surd$ &  0.23       & 2.41      &   0.38\\ 
  & Dplink-Net \cite{xiong2021dp}    &      & $\surd$ &          0.22    &    2.27           & 0.39  \\
 \cite{sun2022event} & Howe \cite{howe2013document}    &      & $\surd$ &           0.11    &   1.62           & 0.46  \\
 & \textbf{Ours}  & $\surd$ & $\surd$  & \textbf{0.87}       & \textbf{14.13}    & \textbf{0.06} \\ 
\noalign{\smallskip}
\hline
\noalign{\smallskip}
\multirow{8}*{EBT} & Nick \cite{khurshid2009comparison}  & & $\surd$        & 0.12       & \underline{9.41}    & 0.45  \\ 
 &  Adaptive \cite{bradley2007adaptive} & & $\surd$ & 0.13       & 9.12     & 0.44 \\ 
 & Wolf \cite{wolf2004extraction}   & & $\surd$      & 0.21       & 6.84     &  \underline{0.39}\\ 
 & Wan \cite{mustafa2018binarization}  & & $\surd$         & \underline{0.20}       & 8.82    &  \underline{0.39} \\ 
 & AE \cite{calvo2019selectional} & & $\surd$  & 0.14       & 9.31    &  0.43 \\ 
  & Dplink-Net \cite{xiong2021dp}    &      & $\surd$ &  0.07 &9.31&  0.47 \\
  & Howe \cite{howe2013document}    &      & $\surd$ & 0.06 & 8.19 &  0.47 \\
 & \textbf{Ours} & $\surd$ & $\surd$   & \textbf{0.65}       & \textbf{13.92}   & \textbf{0.19}  \\ 
\noalign{\smallskip}
\hline
\end{tabular}
}
\end{table}



\begin{table}[t]
\caption{\lsj{Comparisons with motion deblurring for the binary image.}}
\label{table:deblur_ib}
\centering
\resizebox{1\linewidth}{!}{
\begin{tabular}{c|lccccc}
\hline
\noalign{\smallskip}
Dataset & Method & Event & Image &  MCC ($\uparrow$) & PSNR ($\uparrow$) & NRM ($\downarrow$)\\
\noalign{\smallskip}
\hline
\noalign{\smallskip}
  \multirow{6}*{HQF} & Jin \cite{jin2018learning} +  Wan \cite{mustafa2018binarization}    &  & $\surd$      & 0.16       & 4.09   &  0.41  \\ 
 & L0-reg \cite{textdeblur2014} + Wan \cite{mustafa2018binarization} &  &  $\surd$ & 0.12       & 3.97     &  0.44 \\ 
 &  EDI \cite{edipami} + Wan \cite{mustafa2018binarization}       & $\surd$ & $\surd$  & 0.37       & 5.43   &  0.31  \\ 
 &  eSL \cite{yu2023learning}+ Wan \cite{mustafa2018binarization}    & $\surd$ & $\surd$         & 0.49       & 6.21     &  0.24 \\ 
 \cite{stoffregen2020reducing}&  LEDVDI \cite{lin2020learning} + Wan \cite{mustafa2018binarization}    & $\surd$ & $\surd$         &   0.53     & 6.23     & 0.23 \\ 
 & \textbf{Ours} & $\surd$  & $\surd$   & \textbf{0.54}       & \textbf{6.28}   &  \textbf{0.21}  \\ 
\noalign{\smallskip}
\hline
\noalign{\smallskip}
\multirow{6}*{Reblur} & Jin \cite{jin2018learning} + Wan \cite{mustafa2018binarization}    &  & $\surd$      & 0.38      & 5.77   &  0.27 \\ 
 & L0-reg \cite{textdeblur2014} + Wan \cite{mustafa2018binarization} &  & $\surd$ & 0.40       & 7.50   &     0.32 \\ 
 & EDI \cite{edipami} + Wan \cite{mustafa2018binarization}       & $\surd$ & $\surd$  & 0.32       & 6.11   &   0.38\\ 
 & eSL \cite{yu2023learning}  + Wan \cite{mustafa2018binarization}  & $\surd$ & $\surd$  & 0.38       & 5.65   &  0.27  \\ 
 \cite{sun2022event} &  LEDVDI \cite{lin2020learning} + Wan \cite{mustafa2018binarization}    & $\surd$ & $\surd$ & 0.66 &  8.69 & 0.13 \\ 
  & \textbf{Ours}  & $\surd$  & $\surd$ & \textbf{0.87}       & \textbf{14.13}  &  \textbf{0.06}   \\ 
\noalign{\smallskip}
\hline
\noalign{\smallskip}
\multirow{6}*{EBT}& Jin \cite{jin2018learning} + Wan \cite{mustafa2018binarization}    &  & $\surd$      & 0.20       & 8.19   &  0.39  \\ 
 & L0-reg \cite{textdeblur2014} + Wan \cite{mustafa2018binarization} &  & $\surd$ & 0.20       & 7.65    & 0.38 \\ 
 & EDI \cite{edipami} + Wan \cite{mustafa2018binarization}       & $\surd$ & $\surd$  & 0.48       & 8.03  &     0.22 \\ 
 & eSL \cite{yu2023learning} + Wan \cite{mustafa2018binarization}    & $\surd$ & $\surd$  & 0.29       & 5.96   &  0.31  \\ 
 &  LEDVDI \cite{lin2020learning} + Wan \cite{mustafa2018binarization}    & $\surd$ & $\surd$    &     0.64        &    12.81 &  0.20\\ 
 & \textbf{Ours}  & $\surd$  & $\surd$ & \textbf{0.69}       & \textbf{13.92} &    \textbf{0.19} \\ 
\noalign{\smallskip}
\hline
\end{tabular}
}
\end{table}

\section{Experiment}
In all of our experiments, unless otherwise specified, the thresholds $\theta_{I}$, and $\theta_{e}$ for binary reconstruction is determined automatically by our optimization method (\cref{sec:thresh}), and the contrast for our method is set to $c=0.35$.

\subsection{Dataset}
We evaluate the proposed method in three different datasets, including two publicly available real-world datasets and our collected event-based bimodal target dataset (EBT). 
\subsubsection{High-quality frame (HQF) dataset}
The HQF dataset \cite{stoffregen2020reducing} contains real-world events and high-quality images captured by a DAVIS240C event camera with resolution 240$\times$180, where the image is captured using well-set exposures to produce sharp images with few blur. We then up-convert the frame rate of captured images and simulate the blurry frames based on the high frame-rate sequences. The blurry frames are obtained by averaging seven sharp images following the common practice of previous works \cite{xu2021motion,zhang2022unifying}. We use sequences containing bimodal objects in the HQF dataset for evaluation, \ie, the \textit{still\_life}.

\subsubsection{REBlur dataset} The REBlur dataset \cite{sun2022event} is captured under indoor lighting using a DAVIS with resolution 340$\times$260 and a slide-rail system mounted on the optical table with accurate control and timing. They first capture real events and blur images, then re-capture sharp images using position and timestamp recorded in the first slide. We use in total 16 sequences named \textit{*\_a01}, \textit{*\_zju}, and \textit{*\_xiaohui}, which contain bimodal objects for evaluation.

\subsubsection{Event-based bimodal target (EBT) dataset}
Existing event-based datasets are most for evaluating the general image reconstruction, lacking common scenes for evaluating binarization tasks. To fully evaluate our method, we collect the event-based bimodal target (EBT) dataset, which contains various represented bimodal objects such as \textit{text}, \textit{sketch}, \textit{visual tag}, \textit{road sign}, \textit{car plate}, and \etc. The EBT dataset has 23 synthetic sequences and 26 natural sequences. The synthetic sequences are simulated under random trajectories to provide quantitative comparisons using high frame-rate latent images in the Event Simulator (ESIM) \cite{Rebecq18corl} (0.35 contrast with Gaussian noise of 0.2 mean and 0.03 standard deviation). The real sequences are captured using a DAVIS346C color event camera\cite{DAVIS} (with resolution 346$\times$240) under different lighting conditions and various motion patterns that naturally include motion blur into the intensity images. We set the camera contrast to 0.35 using the Java tools for Address-Event Representation (jAER) \cite{jaer} following the hardware bias model \cite{nozaki2017temperature}. Please find full dataset details on our \href{https://github.com/eleboss/EBR}{\textbf{project webpage}}.
\subsubsection{Ground truth} We produce the ground truth in two steps using sharp latent images in the above three datasets, where we first estimate the motion-invariant threshold using the latent sharp image \cite{otsu1979threshold} and then check the threshold to make sure each bimodal object is well captured in the ground-truth image. We use the sequences with clear bimodal objects in the Reblur, HQF, and synthetic sequences of the EBT dataset, ensuring no ambiguities about the ground truth. The real sequences of our EBT dataset contain no sharp images. Thus we use it mainly for qualitative evaluations.

\begin{table}[t]
\caption{\lsj{Comparisons on binary video generation.}}
\label{table:video_bin}
\centering
\resizebox{1\linewidth}{!}{
\begin{tabular}{c|lccccc}
\hline
\noalign{\smallskip}
Dataset & Method & Event & Image &  MCC ($\uparrow$) & PSNR ($\uparrow$) & NRM ($\downarrow$)\\
\noalign{\smallskip}
\hline
\noalign{\smallskip}
  \multirow{5}*{HQF } & Jin \cite{jin2018learning} + Wan \cite{mustafa2018binarization}   &  & $\surd$      & 0.27       & 4.65   & 0.36   \\ 
 &  EDI \cite{edipami} + Wan \cite{mustafa2018binarization}       & $\surd$ & $\surd$  & 0.34       & 5.21   &  0.33  \\ 
 &  eSL \cite{yu2023learning} + Wan \cite{mustafa2018binarization}    & $\surd$ & $\surd$         &  0.13       & 4.72     & 0.44 \\ 
  &  LEDVDI \cite{lin2020learning} + Wan \cite{mustafa2018binarization}    & $\surd$ & $\surd$         &    0.58     &   6.92    & 0.18 \\ 
 \cite{stoffregen2020reducing} & \textbf{Ours} & $\surd$  & $\surd$   & \textbf{0.59}       & \textbf{7.17}  &   \textbf{0.16}  \\ 
\noalign{\smallskip}
\hline
\noalign{\smallskip}
\multirow{5}*{EBT}& Jin \cite{jin2018learning} + Wan \cite{mustafa2018binarization}    &  & $\surd$ & 0.27 & 8.55 & 0.35 \\ 
 & EDI \cite{edipami} + Wan \cite{mustafa2018binarization}       & $\surd$ & $\surd$  & 0.42       & 7.39    & 0.23  \\ 
 & eSL \cite{yu2023learning} + Wan \cite{mustafa2018binarization}    & $\surd$ & $\surd$  & 0.26 & 6.65   &  0.34 \\ 
  &  LEDVDI \cite{lin2020learning} + Wan \cite{mustafa2018binarization}    & $\surd$ & $\surd$ &    0.64      & 12.79    & 0.19 \\ 
 & \textbf{Ours}  & $\surd$  & $\surd$ & \textbf{0.80}       & \textbf{17.48}   &  \textbf{0.12} \\ 
\noalign{\smallskip}
\hline
\end{tabular}
}
\end{table}

\subsection{Results of Latent Binary Image Generation}
\lsj{
 We compare our method with state-of-the-art image binarization methods, including methods based on global statistical information \cite{mustafa2018binarization,wolf2004extraction}, local information \cite{khurshid2009comparison,bradley2007adaptive} and learned semantic information \cite{calvo2019selectional,xiong2021dp}. For evaluation, we adopt the Matthews Correlation Coefficient (MCC) \cite{chicco2020advantages}, the Peak Signal-to-Noise Ratio (PSNR) \cite{hore2010image}, and Negative Rate Metric (NRM). We compute three metrics for each binarization method and report the average scores over each dataset.}
 
\lsj{\noindent\textbf{The Matthews Correlation Coefficient (MCC):} The MCC is a widely used metric for evaluating the performance of binary classification tasks, such as image binarization. It takes into account true positives (TP), true negatives (TN), false positives (FP), and false negatives (FN) and provides a measure of the quality of the classification results:
\begin{align*}
\text{MCC}= \frac{\text{TP} \cdot \text{TN}-\text{FP} \cdot \text{FN}}{\sqrt{(\text{TP}+\text{FP}) \cdot(\text{TP}+\text{FN}) \cdot(\text{TN}+\text{FP}) \cdot(\text{TN}+\text{FN})}}.
\end{align*}
MCC ranges from -1 to 1, where 1 represents a perfect classification and -1 represents a complete disagreement between the predicted and actual classes. MCC is suitable for imbalanced datasets and provides an overall assessment of the binarization quality.}

\lsj{\noindent\textbf{Peak Signal-to-noise Ratio (PSNR):} The PSNR is a metric used to evaluate the quality of reconstructed images by measuring the ratio of the maximum possible power of a signal to the power of corrupting noise that affects the fidelity of its representation. The PSNR is defined as:
\begin{equation*}
\text{PSNR}=10 \log _{10}\left({\frac{\text{MAX}^2_I}{\text{MSE}}}\right),
\end{equation*}
where $\text{MAX}^2_{I}$ is the maximum possible pixel value of the image, and $\text{MAX}^2_{I} = 1$ in binarization tasks. the mean squared error MSE is calculated as $\text{MSE} = (\text{FP} + \text{FN})/ (\text{TP} + \text{TN} + \text{FP} + \text{FN})$ for the binarization tasks. Higher PSNR values indicate better preservation of image details and less distortion.}

\lsj{\noindent\textbf{Negative Rate Metric (NRM):} The NRM is a metric specifically designed for evaluating the performance of binarization algorithms to identify the relation between misclassified elements and all other elements in the class. The NRM is defined as: $\text{NRM}=(\text{NR}_\text{FN}+\text{NR}_\text{FP})/2$, where the false negative rate $\text{NR}_\text{FN} = \text{FN}/(\text{TP}+\text{FN})$ and false positive rate $\text{NR}_\text{FP} =  \text{FP}/(\text{TN}+\text{FP})$. A lower NRM suggests that the binarization method performs better in accurately distinguishing bimodal pixels.}


In \cref{fig:ib}, we show qualitative results that our method can produce the sharp binary image and retain fine geometry details of the bimodal objects within the image. In comparison, state-of-the-art image binarization methods fail to generate clear boundaries of the bimodal objects like texts and markers given blurry inputs. As indicated in \cref{table:ib}, our method achieves the best performance on MCC, PSNR, and NRM compared to state-of-the-art methods, which cannot process motion-blurred images, a common challenge for real-world robotic applications.

To further evaluate our method, we apply various state-of-the-art motion deblurring methods \cite{textdeblur2014,yu2023learning,edipami,jin2018learning, lin2020learning} to first recover a sharp intensity image and then feed the recovered image as input to the second best binarization method in \cref{table:ib}, \ie, Mustafa and Kader. \cite{mustafa2018binarization} for binarization. From \cref{fig:deblur_ib}, motion deblurring can remove the blurry effects and produce relatively sharper intensity images. However, such outputs are not compatible with \cite{mustafa2018binarization}. Image-based deblurring methods (Jin \cite{jin2018learning} and L0-reg \cite{textdeblur2014}) cannot accurately deblur the intensity image under complex motions, which leads to distorted intensity information in their output. Thus the following binarization cannot produce accurate binary boundaries with their output. The event-based deblurring methods (eSL \cite{yu2023learning}, EDI \cite{edipami}, and LEDVDI \cite{lin2020learning}) work better than conventional image-based approaches. Still, halo artifacts in their final output could obscure binarization, leading to incorrect binary outputs. Therefore, naively combining deblurring techniques with binarization methods does not necessarily produce satisfactory results. Our method directly leverages the inherent properties of bimodal objects, achieving the best performance, as shown in \cref{table:deblur_ib}.



\begin{table}[t]
\caption{\lsj{Quantitative results in downstream tasks using binary videos.}}
\label{table:app}
\centering
\begin{tabular}{c|cc}
\hline
\noalign{\smallskip}
 Application & Apriltag detection \cite{olson2011apriltag} & Camera calibration \cite{zhang2000flexible}\\
\noalign{\smallskip}
Metric &  Accuracy $\uparrow$ & MRPE $\downarrow$  \\
\noalign{\smallskip}
\hline
\noalign{\smallskip}
Jin \cite{jin2018learning} + Wan \cite{mustafa2018binarization} & 0.275 & 0.82 \\
\noalign{\smallskip}
EDI \cite{edipami} + Wan \cite{mustafa2018binarization} &  0.560 & 1.37  \\
\noalign{\smallskip}
eSL \cite{yu2023learning} + Wan\cite{mustafa2018binarization} & 0.214 & 1.12 \\
\noalign{\smallskip}
LEDVDI \cite{lin2020learning} + Wan \cite{mustafa2018binarization} & 0.748 & 0.45 \\
\noalign{\smallskip}
Ours (w/o filter) & 0.977 &  0.28  \\
\noalign{\smallskip}
Ours (w/ filter)  & \textbf{0.994} &  \textbf{0.19} \\
\noalign{\smallskip}
\hline
\end{tabular}
\end{table}

\subsection{Results of Binary Video Generation}
To verify the effectiveness of our method in generating binary videos, we compare our methods with state-of-the-art video generation methods (Jin \cite{jin2018learning}, EDI \cite{edipami}, eSL \cite{yu2023learning}, LEDVDI \cite{lin2020learning}) by first generating an intensity video and then converting the intensity video to a binary video using image binarization (Wan \cite{mustafa2018binarization}). As the Reblur dataset does not provide intra-frame ground truth, we use the HQF and EBT datasets for evaluation. For the HQF dataset, we generate seven frames for each blurry image by aligning the video frame time with the seven intra-frame ground truth. For the EBT dataset, we also generate seven frames for each blurred image, and then we use the high frame-rate ground truths in EBT to match each video frame for evaluation. We summarized the quantitative and qualitative results in \cref{table:video_bin} and \cref{fig:video_bin_ebt_real,fig:video_bin_ebt_syn,fig:video_bin_hqf}, respectively. From \cref{fig:video_bin_ebt_real,fig:video_bin_ebt_syn,fig:video_bin_hqf}, we see that the previous state-of-the-art methods usually produce halo artefacts to the intensity video. These artefacts confuse the binarization method, resulting in false binary classification and degraded binary video. In contrast, our method does not need any pre-reconstruction. It directly provides high frame-rate binary video, which preserves the bimodal pattern with clear geometry boundaries and obtains the best scores in \cref{table:video_bin}.

\subsubsection*{Downstream applications}
\lsj{
In \cref{fig:rotate_tag} and \cref{table:app}, we show our binary video could benefit downstream applications, including visual tag detection and camera calibration. 
For visual tag tracking, poor binary frames cannot support correct detection. Therefore, we use accuracy as a measure to evaluate the success of detection. The accuracy is calculated as the ratio of correct detects to total detects, \ie, $\text{Accuracy}= (\text{correct detects})/(\text{total detects})$. The calibration task is evaluated using the mean reprojection error (MRPE), which is a commonly used metric in 3D reconstruction tasks to assess the accuracy of an algorithm. The MRPE quantifies the average distance between the re-projected 3D points $\breve{\vec{p}}{i}$ and their corresponding 2D feature points $\vec{p}{i}$. The MRPE is calculated as $\text{MRPE}=1 / N_p \cdot \sum_{i=0}^N \left\|\vec{p}_{i}-\breve{\vec{p}}_{i}\right\|
,$ where $N_p$ is the total number of 2D points. A lower MRPE indicates a higher calibration accuracy.
}

In  \cref{fig:rotate_tag}, the high frame rate binary video increases the sensing rate and allows visual tags under high-speed motions to be accurately tracked. For calibration tasks, motion blur introduced by the handshake motion could also be removed, achieving better performance as shown in \cref{table:app}. The median filtering could be further applied to reduce the dotted noise. In \cref{table:app} and \cref{fig:filtering}, we show the asynchronous median filter could reduce the dotted noise, thus improving the accuracy of tag detection and reducing the error of the camera calibration.

\begin{table}[t]
\caption{Evaluations of different contrast settings ($c = \{0.25,0.5,0.75,1\}$).}
\label{table:contrast}
\centering
\begin{tabular}{c|cccccc}
\hline
\noalign{\smallskip}
Tasks & Dataset & Metric  &  $0.25$ & $0.5$ & $0.75$ & $1.0$\\
\noalign{\smallskip}
\hline
\noalign{\smallskip}
\multirow{9}*{Image} & \multirow{3}*{HQF \cite{stoffregen2020reducing}}     & PSNR ($\uparrow$)   & \textbf{6.31} & 6.28 & 6.30   & 6.30       \\ 
&     & MCC ($\uparrow$)  & \textbf{0.54} & 0.54 & 0.53   & 0.53       \\ 
&     & NRM ($\downarrow$)  & \textbf{0.21} & 0.21 &  0.22  &     0.22   \\

\noalign{\smallskip}
\cline{2-7}
\noalign{\smallskip}
 & \multirow{3}*{Reblur \cite{sun2022event}}     & PSNR ($\uparrow$)  & \textbf{14.13} & 13.91 & 13.95   & 13.99       \\ 
&     & MCC ($\uparrow$)  & \textbf{0.87} & 0.86 & 0.86   & 0.74       \\
&     & NRM ($\downarrow$)  & \textbf{0.06} & 0.07 & 0.07   &   0.07     \\

\noalign{\smallskip}
\cline{2-7}
\noalign{\smallskip}
& \multirow{3}* {EBT}  & PSNR ($\uparrow$)  & 13.65 & 13.92 & \textbf{13.96}  & 13.82       \\ 
&     & MCC ($\uparrow$) & 0.64 & 0.65 & \textbf{0.65}   & 0.64       \\ 
&     & NRM ($\downarrow$)  & 0.19 & \textbf{0.18} &  0.19  &    0.19    \\

\noalign{\smallskip}
\hline
\noalign{\smallskip}
\multirow{6}*{Video} & \multirow{3}*{HQF \cite{stoffregen2020reducing}}     & PSNR ($\uparrow$)  & \textbf{7.18} & 7.17 & 7.12   & 7.14      \\ 
&     & MCC ($\uparrow$) & \textbf{0.59} & 0.59 & 0.58   & 0.58       \\ 
&     & NRM ($\downarrow$)  & \textbf{0.19} & 0.19  & 0.20    &    0.19    \\

\noalign{\smallskip}
\cline{2-7}
\noalign{\smallskip}
& \multirow{3}* {EBT}  & PSNR ($\uparrow$)  & 17.13 & 17.48 & \textbf{17.57}   & 17.36       \\ 
&     & MCC ($\uparrow$)  &  0.78 & 0.80 & \textbf{0.80}   & 0.79       \\ 
&     & NRM ($\downarrow$)  & 0.12  & 0.11 &  \textbf{ 0.11}   &   0.12     \\

\noalign{\smallskip}
\hline
\end{tabular}
\end{table}

\begin{table}[t]
    \centering
   \caption{Runtime evaluation on EBT (Sequence: \textit{exit}, 13 million events, length: 5.7 seconds).}
    \label{table:runtime}
    \centering
    \begin{tabular}{c|ccc}
        \noalign{\smallskip}
        \hline
        \noalign{\smallskip}
        \multirow{2}*{Method} & Processing rate ($\uparrow$)   & Runtime ($\downarrow$) & Real-time \\
         & (Ev/second) & (Second) & factor ($\uparrow$) \\
         \noalign{\smallskip}
        \hline
        \noalign{\smallskip}
        Image - Threshold  & 6.43 $\times 10^6$ & 1.99 & 2.83 \\
        Image - Deblur & 5.18 $\times 10^6$ & 2.54 & 2.28 \\
        \noalign{\smallskip}
        Image - Total & $\mathbf{2.87 \times 10^6}$ & \textbf{4.53} & \textbf{1.26} \\
        \noalign{\smallskip}
        \hline
        \noalign{\smallskip}
        Video (w/o filter) & $\mathbf{5.25 \times 10^6}$ & \textbf{2.44} & \textbf{2.31} \\
        \noalign{\smallskip}
        Video (w/ filter) & $\mathbf{4.02 \times 10^6}$ & \textbf{3.19} & \textbf{1.77} \\
        \noalign{\smallskip}
        \hline
        \noalign{\smallskip}
         eSL \cite{yu2023learning} & 0.0314 $\times 10^6$ & 419.23 & 0.014 \\
        \noalign{\smallskip}
        \hline
        \noalign{\smallskip}
         EDI \cite{edipami} & 0.22 $\times 10^6$ & 59.74 & 0.103 \\
        \noalign{\smallskip}
        \hline
    \end{tabular}
\end{table}

\subsection{Robustness and runtime performance}
\subsubsection{Different contrast}
We evaluate our method under four radically different contrast parameters, \ie, $c=\{0.25, 0.5, 0.75, 1.0\}$, and we summarize quantitative and qualitative results in \cref{table:contrast} and \cref{fig:contrast}. From \cref{table:contrast}, we can see that the performance of our method persists well under different contrast parameters. From \cref{fig:contrast}, we can see that binary images under different contrast are nearly identical. The above results directly show the effectiveness of our method under different contrast parameters, meaning we can significantly relax the need for accurate contrast estimation, thus improving robustness and efficiency.

\subsubsection{Runtime}
We implement our method using C++ and evaluate it on a laptop equipped with Intel i7-10870H@2.2GHz using one CPU core. The EDI \cite{edipami} and eSL \cite{yu2023learning} are also tested using the same setup for fair comparison. LEDVDI \cite{lin2020learning} is excluded as its core modules do not support CPU processing. We employed three standard metrics for evaluation, including the event processing rate, the total runtime, and the real-time factor. The event processing rate indicates the number of events the system processes per second. The real-time factor is calculated by dividing the sequence length by the total processing time. If the real-time factor exceeds 1, it implies that the data generated can be processed within the generation time, indicating that the system can operate at a real-time rate. We summarized the runtime results in \cref{table:runtime}, showing that our proposed method can produce the latent binary image in real-time (with a real-time factor $= 1.26$). Given the latent binary image and the threshold, the generation of binary video is also highly efficient (with a real-time factor $= 1.77$). As the whole pipeline is developed for asynchronous processing, we can also produce a high frame-rate denoised binary video in a highly efficient manner using CPU cores. 
As shown in \cref{table:runtime}, event-based intensity reconstruction methods cannot operate at a real-time rate on CPU-only devices. For example, EDI \cite{edipami} requires complicated contrast optimization of each frame, and eSL \cite{yu2023learning} relies on dense convolutional neural networks.

\section{Conclusion}
We propose a novel method that leverages the synergy between events and images to generate binary videos from motion-blurred videos. Our approach exploits the bimodality in both the events and the image spaces, which naturally relates the motion blurry effect to the formation of blurry images and events. 
This enables our approach to bypass the time-consuming intensity reconstruction process based on events.
We also present an effective fusion method that integrates events and blur images for unsupervised threshold estimation, where our normalization mitigates the need for precise contrast estimation, improving the overall efficiency. Thus, our approach can perform robust and efficient inference of the latent binary image and generate high frame rate binary videos. 
Extensive experiments validate the effectiveness of our proposed method in generating high-quality, high frame-rate binary videos across various motion conditions and demonstrate its potential in downstream tasks such as tag tracking and camera calibration.
\lsj{
\subsubsection*{Limitation} 
The proposed method operates under gentle lighting to avoid saturation in the intensity images, which can result in a non-informative distribution of pixel intensities. 
One possible solution is to incorporate uncertainty into the optimization \cite{saeidi2015uncertain}, and use high certainty areas for estimation.
}












%


\ifCLASSOPTIONcaptionsoff
  \newpage
\fi



\bibliographystyle{IEEEtran}

\bibliography{scibib}


%





\end{document}